\documentclass[lettersize,journal]{IEEEtran}
\usepackage{amsmath,amsfonts}
\usepackage[ruled,vlined]{algorithm2e}
\usepackage{array}
\usepackage[caption=false,font=normalsize,labelfont=sf,textfont=sf]{subfig}
\usepackage{textcomp}
\usepackage{stfloats}
\usepackage{color}
\usepackage{url}
\usepackage{verbatim}
\usepackage{graphicx}
\usepackage{cite}
\usepackage{booktabs} 
\usepackage{ulem}
\usepackage{picins}

\begin{document}

\title{DCUDF2: Improving Efficiency and Accuracy in Extracting Zero Level Sets from Unsigned Distance Fields}

\author{Xuhui Chen, Fugang Yu, Fei Hou,
Wencheng Wang, Zhebin Zhang, and Ying He
\thanks{X. Chen and F. Yu contributed equally to the work. X. Chen, F. Yu, F. Hou and W. Wang are with the Key Laboratory of System Software (CAS) and State Key Laboratory of Computer Science, Institute of Software, Chinese Academy of Sciences, China, and University of Chinese Academy of Sciences. Emails: chenxh$|$yufg$|$houfei$|$whn@ios.ac.cn}
\thanks{Z. Zhang is with the OPPO Research Center, USA. Email: zhebin.zhang@oppo.com}
\thanks{Y. He is with the College of Computing and Data Science, Nanyang Technological University, Singapore. Email: yhe@ntu.edu.sg
 }% <-this % stops a space
\thanks{Corresponding author: F. Hou}
}

\markboth{Journal of \LaTeX\ Class Files,~Vol.~14, No.~8, August~2021}%
{Shell \MakeLowercase{\textit{et al.}}: A Sample Article Using IEEEtran.cls for IEEE Journals}


\maketitle

\begin{abstract}
Unsigned distance fields (UDFs) allow for the representation of models with complex topologies, but extracting accurate zero level sets from these fields poses significant challenges, particularly in preserving topological accuracy and capturing fine geometric details. To overcome these issues, we introduce DCUDF2, an enhancement over DCUDF--the current state-of-the-art method--for extracting zero level sets from UDFs. Our approach utilizes an accuracy-aware loss function, enhanced with self-adaptive weights, to improve geometric quality significantly. We also propose a topology correction strategy that reduces the dependence on hyper-parameter, increasing the robustness of our method. Furthermore, we develop new operations leveraging self-adaptive weights to boost runtime efficiency. Extensive experiments on surface extraction across diverse datasets demonstrate that DCUDF2 outperforms DCUDF and existing methods in both geometric fidelity and topological accuracy. We will make the source code publicly available. 
\end{abstract}

\section{Introduction}
\label{sec:intro}

Signed Distance Fields (SDFs) and Unsigned Distance Fields (UDFs) represent two of  the most popular implicit function techniques for modeling 3D surfaces. SDFs define 3D space through level sets with positive and negative signs to indicate whether a point is inside or outside an object. Its ease of optimization has made SDFs a preferred choice for various applications. However, SDFs are inherently limited to  modeling closed, orientable surfaces due to their reliance on sign information.

In contrast, UDFs can represent a wider variety of 3D models, including open surfaces, non-orientable surfaces, and non-manifolds, by omitting sign information~\cite{Zhou2022CAP-UDF, chibane2020ndf, Long2022NeuralUDFLU}. This flexibility, however, introduces significant challenges in traditional iso-surface extraction algorithms, such as Marching Cubes (MC)~\cite{Lorensen1987MarchingCA} and Dual Contouring (DC)~\cite{Ju2002DualCO}, which depend on sign changes for detecting zero level sets. Moreover, the gradients of UDFs are notably noisy near the target surface, further complicating accurate surface extraction. 

Most existing methods~\cite{guillard2022udf, Zhou2022CAP-UDF, chen2022ndc} attempt to \textit{directly} extract the zero level set from UDFs, a process frequently compromised by inaccuracies and noise in the learned UDFs. To overcome this, DCUDF~\cite{Hou2023DCUDF} employs an \textit{indirect} approach by extracting an iso-surface with an iso-value $r>0$ using the standard Marching Cubes algorithm. This method produces a double covering of the target surface that is always a closed, orientable surface, regardless of the model's topology. As the iso-surface is offset from the zero level set, it inherently contains less noise. DCUDF minimizes a distance loss and a Laplace loss, iteratively refining the double-layered surface to more closely approximate the zero level set. A minimum cut algorithm is then used to separate the two layers. Unlike gradient-based methods~\cite{guillard2022udf, Zhou2022CAP-UDF, Zhang2023SurfaceEF}, DCUDF does not rely on the gradients of UDFs to determine the zero level set, which enhances its robustness. Furthermore, the Laplace term helps to regulate the distribution of vertices on the zero level sets, enabling the production of high-quality triangle meshes. 

While DCUDF offers enhanced robustness compared to gradient-based methods, it also presents two inherent limitations stemming from its design. First, the method tends to over-smooth fine geometric details, diminish the sharpness in regions of high curvature, and require prolonged computation times. Second, the effectiveness of DCUDF critically depends on the correct choice of the iso-value $r$. As illustrated in Figure~\ref{fig:levelset}, the $r$-level set typically exhibits fewer gaps and holes than the actual zero level set, altering the topology. Inappropriate settings of $r$ can result in topological defects, such as holes and gaps. 

\begin{figure*}[!htbp]
    \centering
    \includegraphics[height=2.1in]{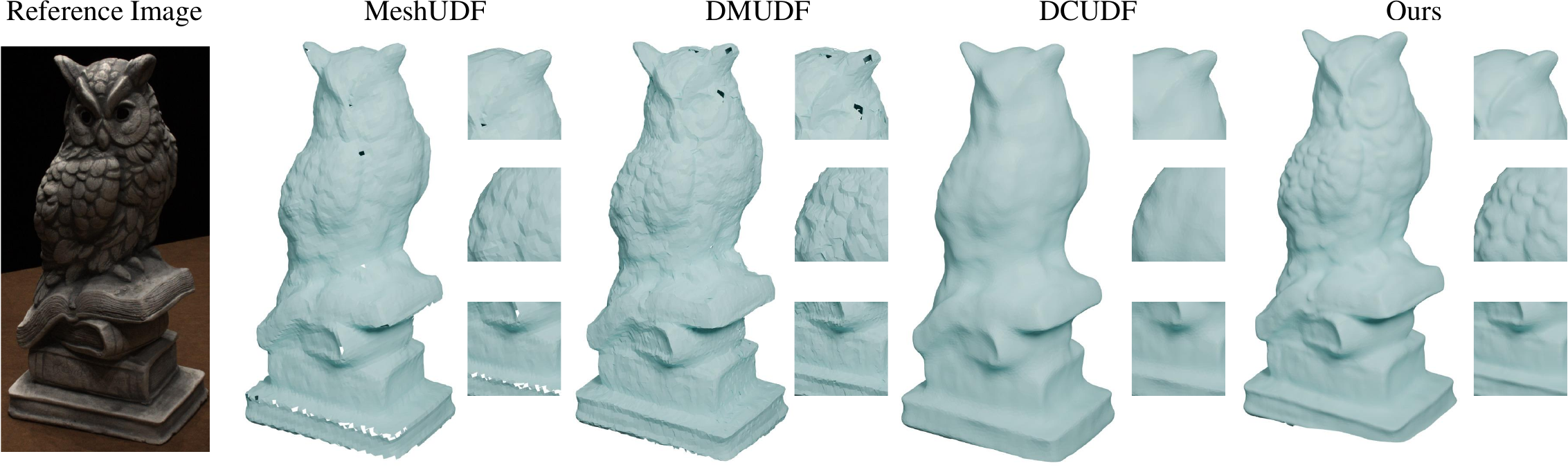}\\
    \caption{DCUDF, a leading method for extracting zero level sets from UDFs, often suffers from over-smoothing that compromises geometric details. Our enhanced approach improves upon DCUDF by more effectively preserving geometric fidelity and topological accuracy, while also enhancing runtime performance. It outperforms existing methods in maintaining detailed geometry and accurate topology. }
    \label{fig:teaser}
\end{figure*}

This paper introduces significant enhancements to overcome the limitations of DCUDF. We propose an accuracy-aware loss function tailored to local geometric variations. This function employs self-adaptive weights to ensure that each part of the model, especially areas of high curvature, receives the necessary attention to effectively mitigate the smoothing (see Figure~\ref{fig:teaser}). Moreover, we tackle the topological challenges arising from DCUDF's dependence on the iso-value $r$ with a dynamic topology correction strategy that adjusts the topology during the optimization process. In addition, we augment DCUDF by introducing activation masks, which progressively reduce the number of variables involved in the optimization, and optimization direction correction to prevent the process from  getting stuck at local minima. 

Through extensive experiments, we validate that these enhancements significantly improve the DCUDF framework. Our results demonstrate enhanced precision, improved geometric fidelity and superior topological accuracy. These improvements effectively address the inherent limitations of the original DCUDF approach. Comparative analyses with leading methods further confirm the enhanced accuracy and robustness of our approach.

\section{Related Works}
\label{sec:related}

Modeling 3D surfaces with signed or unsigned distance fields is a pivotal topic in computer graphics and 3D vision. SDFs are preferred for surface reconstruction due to their reduced sensitivity to noise and outliers~\cite{Park2019DeepSDFLC, Michalkiewicz2019ImplicitSR, yariv2020multiview, wang2021neus, Mildenhall2020NeRF, Gropp2020ImplicitGR, Wang2022IDF, Wang2022NeuS2FL, Fu2022GeoNeusGN, Yu2022MonoSDFEM}. 
Traditionally, these fields are computed directly from point clouds~\cite{Kazhdan2006PoissonSR, Hou2022IterativePS}, utilizing MC~\cite{Lorensen1987MarchingCA} or DC~\cite{Ju2002DualCO} to extract surfaces. More recently, the integration of neural networks to form neural implicit functions has expanded the input possibilities beyond point clouds~\cite{Park2019DeepSDFLC, Michalkiewicz2019ImplicitSR, Gropp2020ImplicitGR, Wang2022IDF } to include images and videos~\cite{yariv2020multiview, wang2021neus, Wang2022NeuS2FL, Fu2022GeoNeusGN, Yu2022MonoSDFEM}.

However, the use of SDFs is constrained to watertight models due to the necessity of inside/outside labels that denote the sign of distance values. To address this limitation, several approaches adapt SDFs for open models~\cite{Meng2023, Liu2023GhostOT,Chen20223PSDFTS}, although they often encounter accuracy issues at boundaries and are incapable of managing complex surfaces inner structures.

In contrast, UDFs eliminates the need for sign information, thereby facilitating the representation of models with complex topologies~\cite{Liu2023NeUDFLN, Long2022NeuralUDFLU, Ren2022GeoUDFSR, chibane2020ndf, Zhou2023LearningAM, Zhou2022CAP-UDF, chen2022ndc}. Despite their potential, the practical application of UDFs is hampered by the lack of effective techniques for extracting zero level sets, primarily due to the absence of sign information which guides traditional extraction algorithms.

Several methods aim to mitigate this by employing UDF gradients to identify zero crossing, such as MeshUDF~\cite{guillard2022udf} and CAP-UDF~\cite{Zhou2022CAP-UDF}. Nevertheless, the instability of UDFs near the target surface can lead to incorrect gradient directions and distance values, thus misleading the extraction. 
NDC~\cite{chen2022ndc} leverages a pre-trained model to enhance the Dual Contouring with UDFs, yet it struggles with generalization, often failing to produce accurate results for new, unseen models. 

Given these challenges, alternative methods such as DMUDF~\cite{Zhang2023SurfaceEF} and DCUDF~\cite{Hou2023DCUDF} adopt strategies that exploit the stability of off-surface level sets. 
DMUDF~\cite{Zhang2023SurfaceEF} projects points onto the zero level set, which can be used for solving the quadratic error function used in Dual Contouring.  Although DMUDF includes a filter to reduce noisy points, it still suffers from low accuracy and tends to yield meshes with topological defects. DCUDF~\cite{Hou2023DCUDF}, on the other hand, extracts the mesh of a non-zero level set and iteratively optimizes it to approximate the zero level set. While DCUDF exhibits robustness against noise and inaccuracies of the input UDFs, its optimization tends to generate over-smoothed meshes.

\section{Method}
\label{sec:method}
Our method is designed to enhance the accuracy and efficiency of DCUDF. In Section~\ref{subsec:preliminaries}, we present preliminaries of DCUDF, and examine the reasons for its tendencies for over-smoothing and dependency on the parameter $r$. This discussion sets the state for introducing our adaptive weight learning strategy and the accuracy-aware loss function in Section~\ref{subsec:adaptiveweights}. Subsequent sections~\ref{subsec:mask} and~\ref{subsec:orientation} detail activation masks and optimization direction correction for improving runtime efficiency and preventing the process getting stuck at local minima. After that,  Section~\ref{subsec:topology} presents our topology correction strategy, which addresses topological inconsistencies that arise from improper selection of the iso-value $r$. We provide the pseudocode for DCUDF2 in Algorithm~\ref{alg:DCUDF2}, and illustrate the algorithmic pipeline using the Bunny model in Figure~\ref{fig:dynamic_weights}.

\begin{figure}[!htbp]
    \centering
    \includegraphics[height=2.0in]{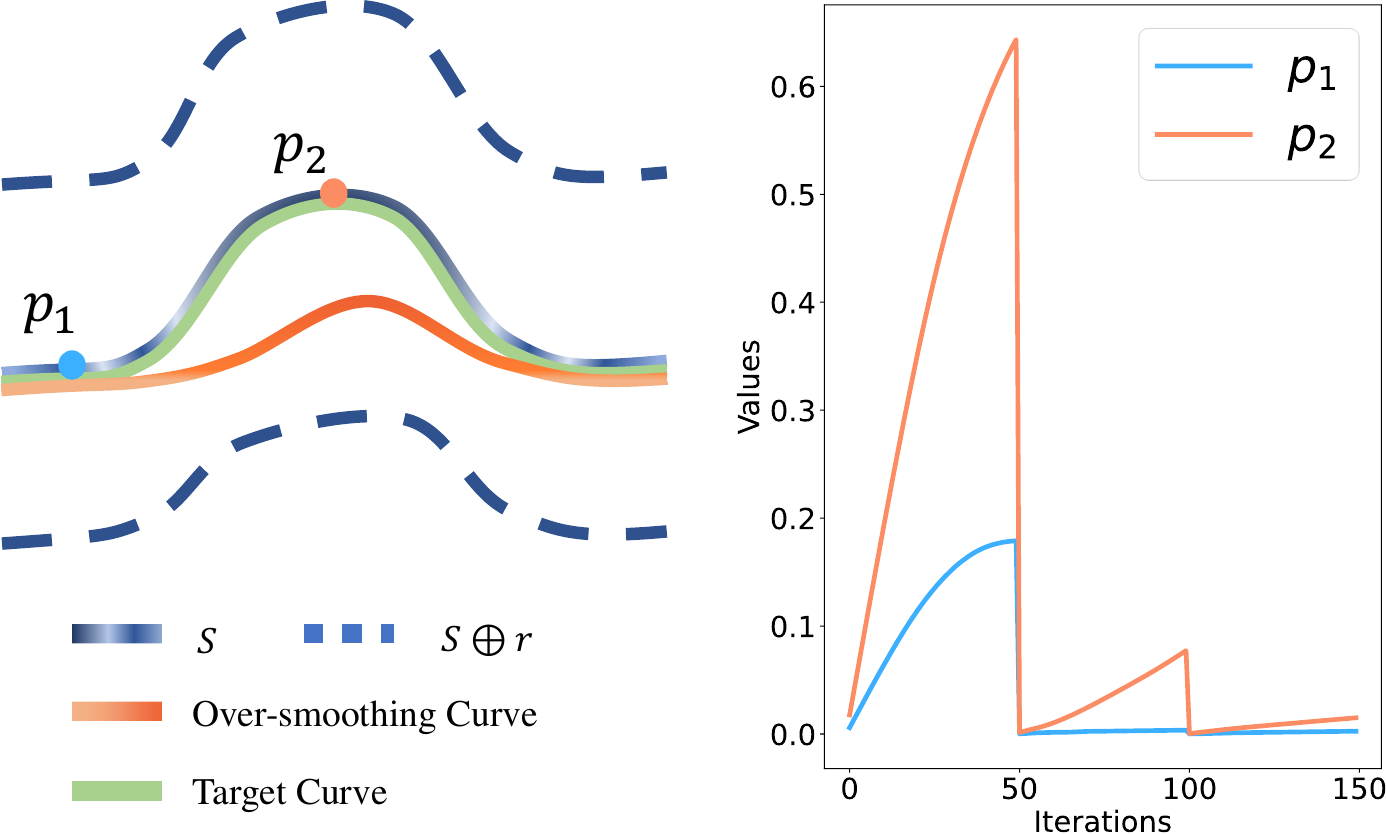}\\
    \caption{Conceptual illustration of the over-smoothing side effect in DCUDF, particularly pronounced when the target surface has rich geometric details and  a large iso-value $r$ is used. In this illustration, $S$ represents the target zero level set, while $M = S \oplus r$ denotes the $r$-level set, forming a double covering of $S$. Due to the large iso-value $r$, the initial $r$-level set starts at a considerable distance from $S$. During the iterative shrinking of the region between the two layers, the opposing effects of distance loss and Laplacian loss frequently result in a significant gap between DCUDF's extracted level set (orange curve) and the actual zero level set, especially in regions of high curvature. DCUDF2 addresses this by incrementally decreasing the effect of the Laplacian loss in areas of high curvature, thereby increasing the influence of the distance loss. The plots on the right display the self-adaptive weights $w_\text{sa}$ for two representative points—one in a low-curvature area and the other in a high-curvature area. This adjustment allows the optimization process to effectively reduce the gap, drawing the orange curve closer to the true zero level set. }
    \label{fig:local_optimize}
\end{figure}

\subsection{Preliminaries}
\label{subsec:preliminaries}

\subsubsection{Two-stage Optimization}

To extract a mesh corresponding to the zero level set $\mathcal{S}$ from an unsigned distance field $f$, DCUDF first generates a double-layer mesh using the Marching Cubes algorithm at an iso-value $r$ (where $r>0)$. This operation produces a dilated version of the target mesh. The resulting mesh, denoted by $\mathcal{M}$, is then transformed by a continuous covering map $\pi$, which projects $\mathcal{M}$ onto $\mathcal{S}$. 
The objective is to optimize $\mathcal{M}$ to closely approximate $\mathcal{S}$ while maintaining high geometric fidelity.
After the optimization converges, a post-processing step isolates the final single-layer mesh by detaching the double-layered structure. 

To ease optimization, DCUDF employs a two-stage strategy. In the initial stage, the focus is on stabilizing the mesh's structure while aligning it closely with the zero level set. This is achieved through the minimization of the following loss function:
\begin{equation}
\label{eqn:step1}
\min_{\pi}\sum_{p_i\in\mathcal{M}\cup\mathcal{C}} f\big(\pi(p_i)\big)
    +\lambda_1\sum_{p_i\in\mathcal{M}} w(p_i)\|\bigtriangleup \pi(p_i)\|^2,
\end{equation}
where $\mathcal{C}$ denotes the set of triangle centroids and $w(p_i)$ is a position-dependent weight.     

The first term, $f(\pi(p_i))$, ensures that the projected point $\pi(p_i)$ seeks the local minimum of the input UDF, effectively guiding the mesh towards the zero level set. The second term, denoted by
$$\bigtriangleup \pi(p_i)=\pi(p_i)-
    \frac{1}{|\mathcal{N}(p_i)|}\sum_{p_j\in\mathcal{N}(p_i)}\pi(p_j)$$  
where $\mathcal{N}(p_i)$ represents the 1-ring neighbors of $p_i$ in the mesh $\mathcal{M}$, serves as a
Laplacian regularization. This term stabilizes the geometric structure and promotes a uniform distribution of the mesh vertices during the optimization process.

In the second-stage optimization, the Laplacian loss is eliminated to enhance geometric fidelity. This change allows the mesh $\mathcal{M}$ to more closely approximate the actual zero level set $\mathcal{S}$, especially in regions of high curvatures. By removing this constraint, the mesh can adjust more freely, resulting in a more accurate depiction of local geometry.

\subsubsection{The Choice of $r$}
The choice of the iso-value $r$ is a critical factor in DCUDF, as it fundamentally affects the topological accuracy of the resulting mesh. When DCUDF employs the Marching Cubes algorithm to initially extract a double-layer mesh $\mathcal{M}$, the topology of $\mathcal{M}$ is set and remains unchanged during the subsequent optimization process. Therefore, the selection of $r$ is vital for ensuring that the final mesh accurately reflects the topology of the desired zero level set, $\mathcal{S}$.

In general, the choice of the iso-value $r$ is influenced by several factors: the geometry of the model, the level of noise in the input UDF, and the resolution of the Marching Cubes used for mesh generation. Selecting a very small $r$ can produce an initial mesh $\mathcal{M}$ that closely matches the topology of the desired zero level set $\mathcal{S}$. However, this proximity can lead to instability in the mesh, making it susceptible to errors, particularly the formation of unwanted holes. Conversely, opting for a larger $r$ generally results in a more stable $r$-level set, but with a higher risk of deviating from the true topology of $\mathcal{S}$, as illustrated in Figure~\ref{fig:levelset}. Thus, finding an optimal $r$ is crucial to balancing mesh stability with topological accuracy.

In DCUDF, the iso-value $r$ is determined within the following bounds:
\begin{equation}
\label{eqn:r_chosen}
    \max\left\{\frac{1}{2k}, d_{\max}\right\} \leq r \leq \frac{\eta_{\min}}{2}
\end{equation}
where $d_{\max}$ represents the maximum UDF value at points on the zero level set (indicative of noise level), $\eta_{\min}$ is the smallest gap size within the target surface, and $k$ denotes the Marching Cubes resolution. 

When the recommended range for $r$ is narrow, selecting an appropriate value without compromising the final mesh quality can be challenging. In response, our method reduces the heavy dependence on the precise selection of $r$ by incorporating a topological correction strategy that dynamically updates the topology of $\mathcal{M}$ during the optimization process. This enhancement affords users greater flexibility in choosing $r$. However, we still advise adhering to the lower bound of the range, $r\geq \max\{\frac{1}{2k}, d_{\max}\}$, to ensure a stable $r$-level set for initialization.

\subsubsection{Pros and Cons of the Laplacian Loss}
The optimization goal of DCUDF is to minimize the geometric discrepancies between the initial double layer mesh $\mathcal{M}$ and the zero level set $\mathcal{S}$. This is primarily achieved by reducing the distance field values across $\mathcal{M}$, thereby drawing it closer to $\mathcal{S}$.

However, relying solely on minimizing the distance field can introduce issues such face flipping and self-intersections, mainly due to the noise present in the input UDF. To mitigate these issues, DCUDF incorporates a Laplacian regularization, which enhances the stability of the optimization process by smoothing the mesh. Although this addition helps maintain a stable iteration process, it leads to a trade-off: 1) Over-smoothing: The Laplacian loss tends to overly smooth regions rich in detail, which can diminish the sharpness and accuracy of the reconstructed surface features. 2) Noise robustness: Reducing the influence of the Laplacian loss enhances detail preservation but at the cost of increased sensitivity to noise, which can destabilize the optimization process. 

Balancing the weight between the distance field loss and the Laplacian loss is challenging. The optimization process, therefore, needs to carefully manage these opposing forces to optimize both geometric fidelity and topological accuracy.

\begin{figure*}[!htbp]
    \centering
    \makebox[0.9in]{Initialization}
    \makebox[0.9in]{50}
    \makebox[0.9in]{100}
    \makebox[0.9in]{150}
    \makebox[0.9in]{200}
    \makebox[0.9in]{400}
    \makebox[0.9in]{Output}\\
    \includegraphics[width=0.9in]{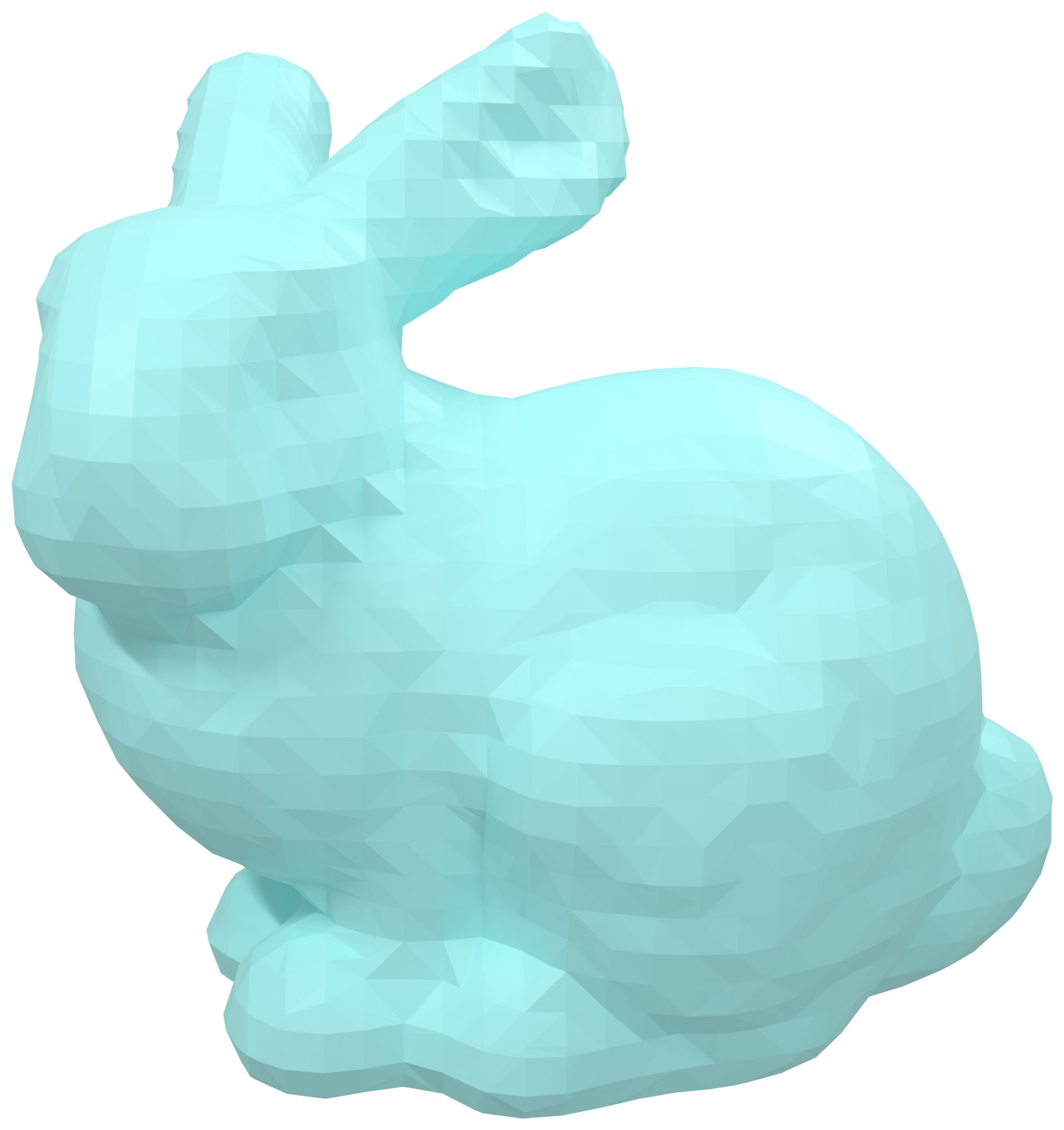 }
    \includegraphics[width=0.9in]{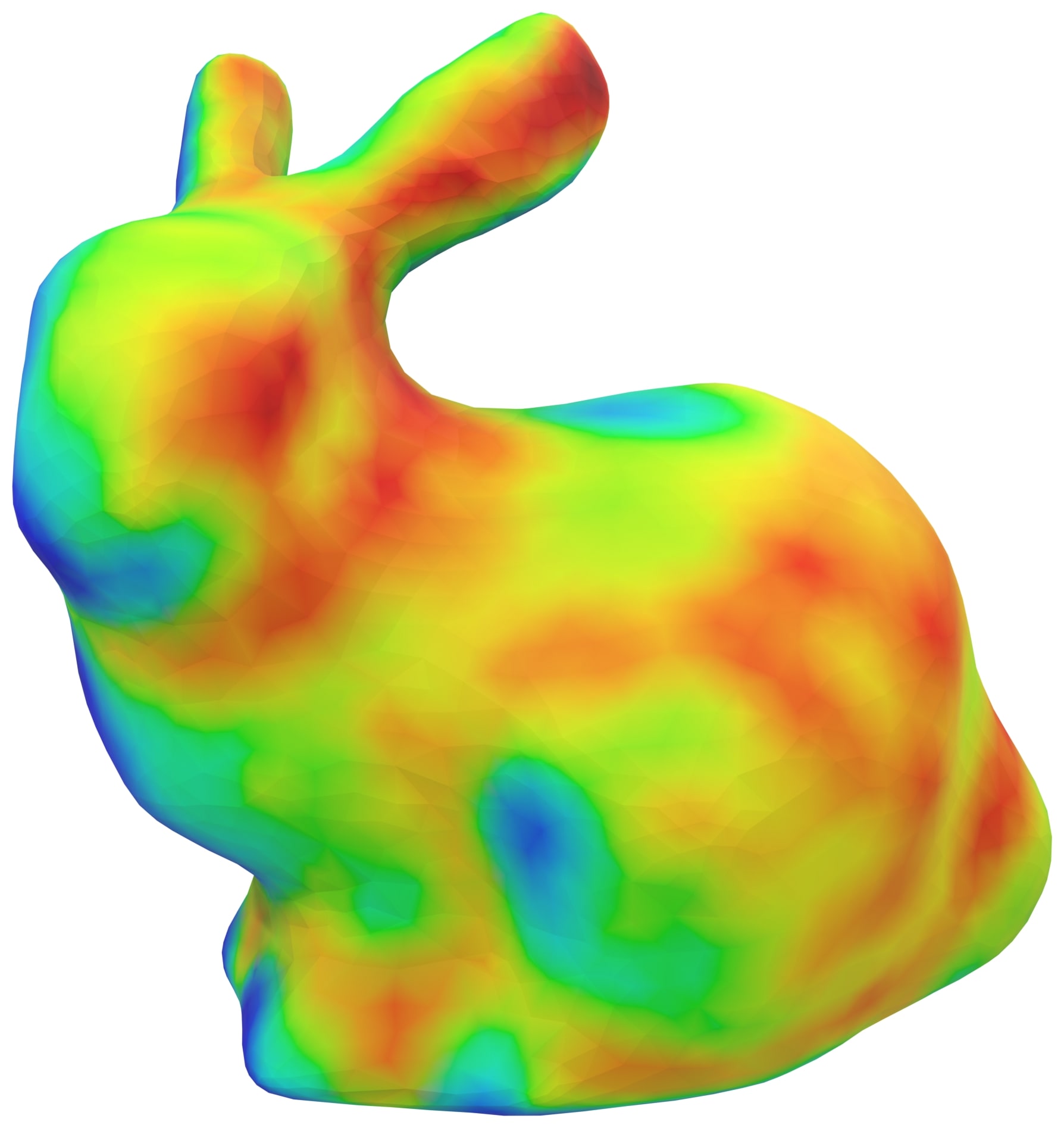 }
    \includegraphics[width=0.9in]{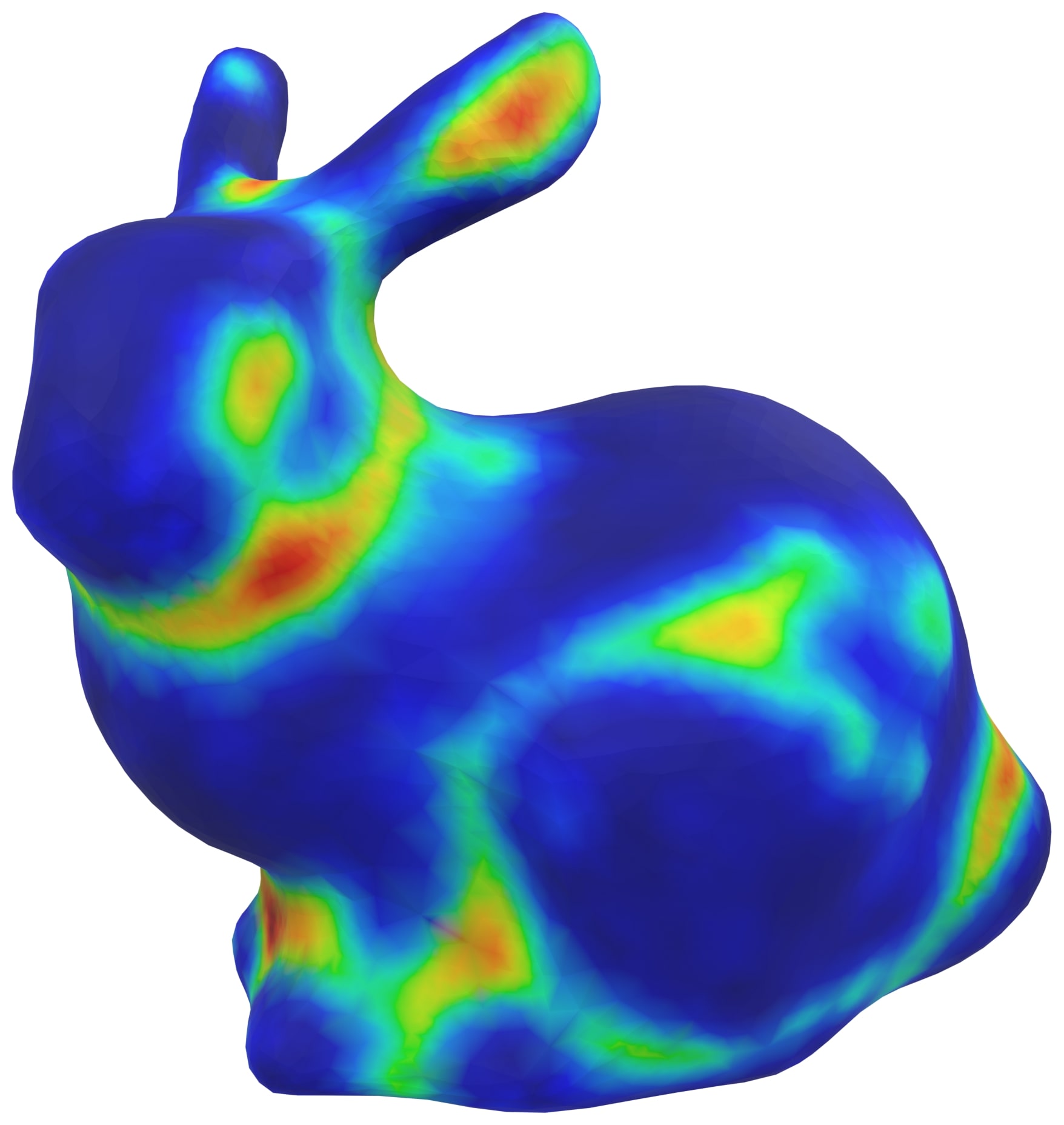 }
    \includegraphics[width=0.9in]{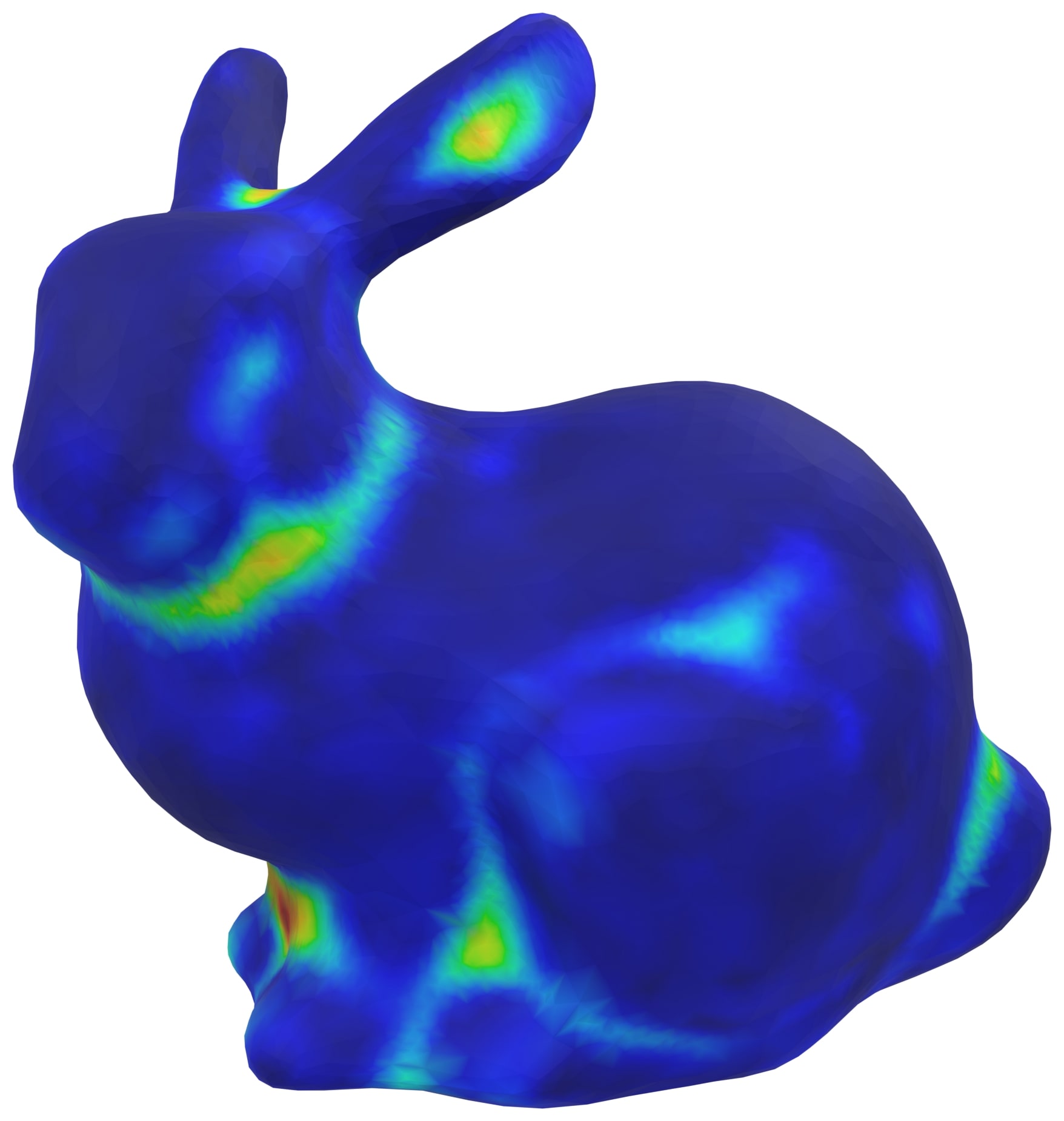 }
    \includegraphics[width=0.9in]{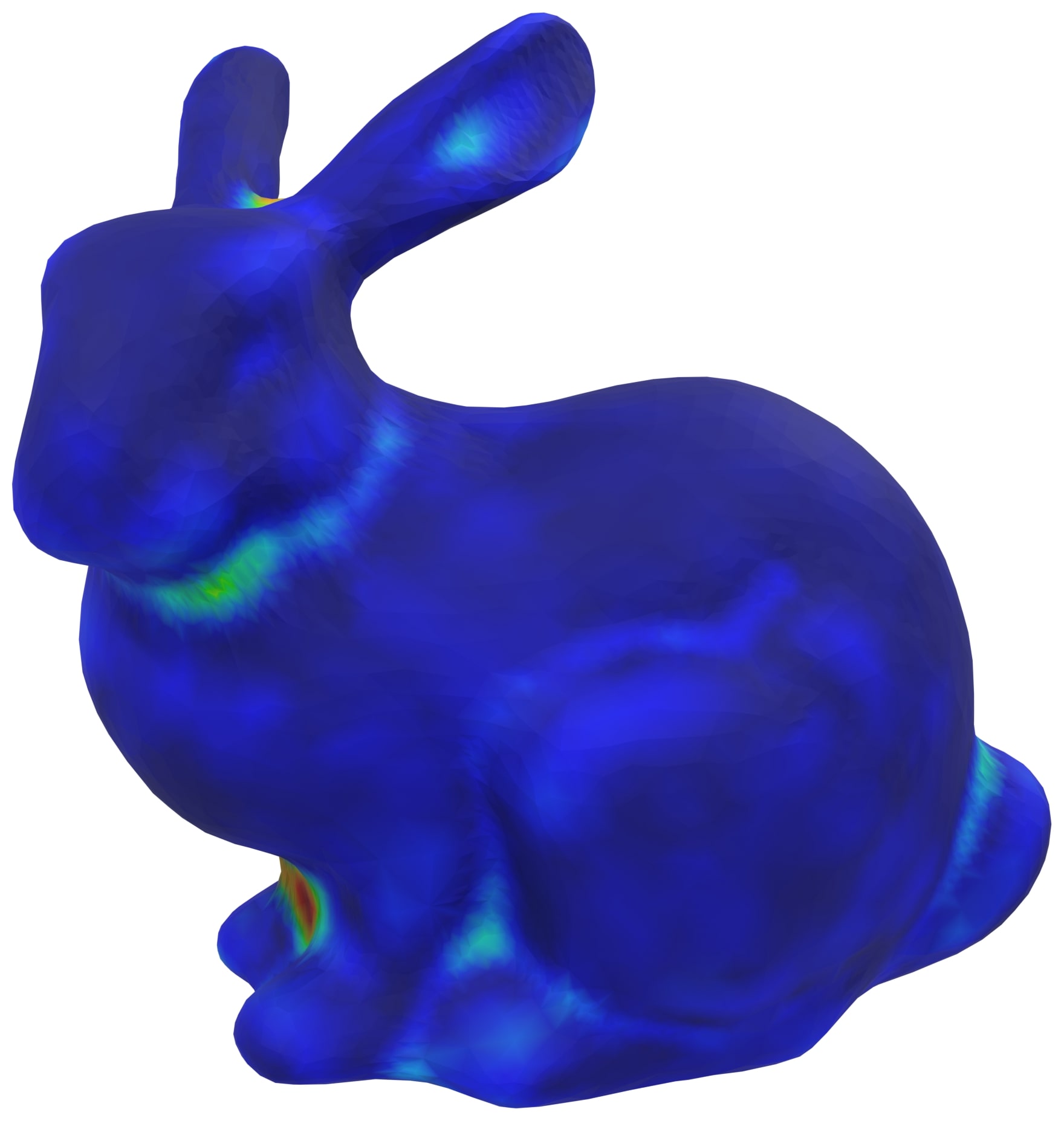 }
    \includegraphics[width=0.9in]{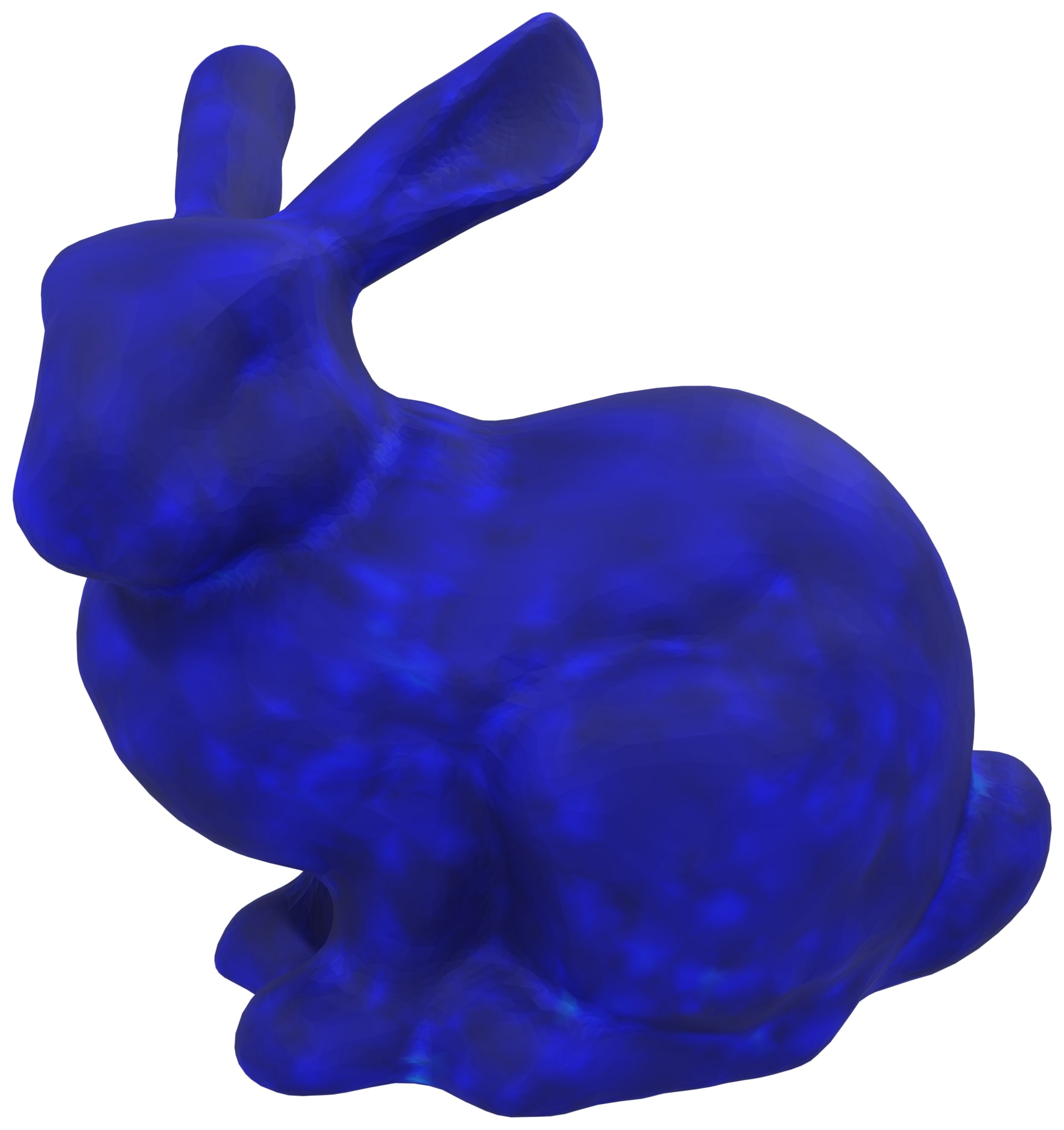 }
    \includegraphics[width=0.9in]{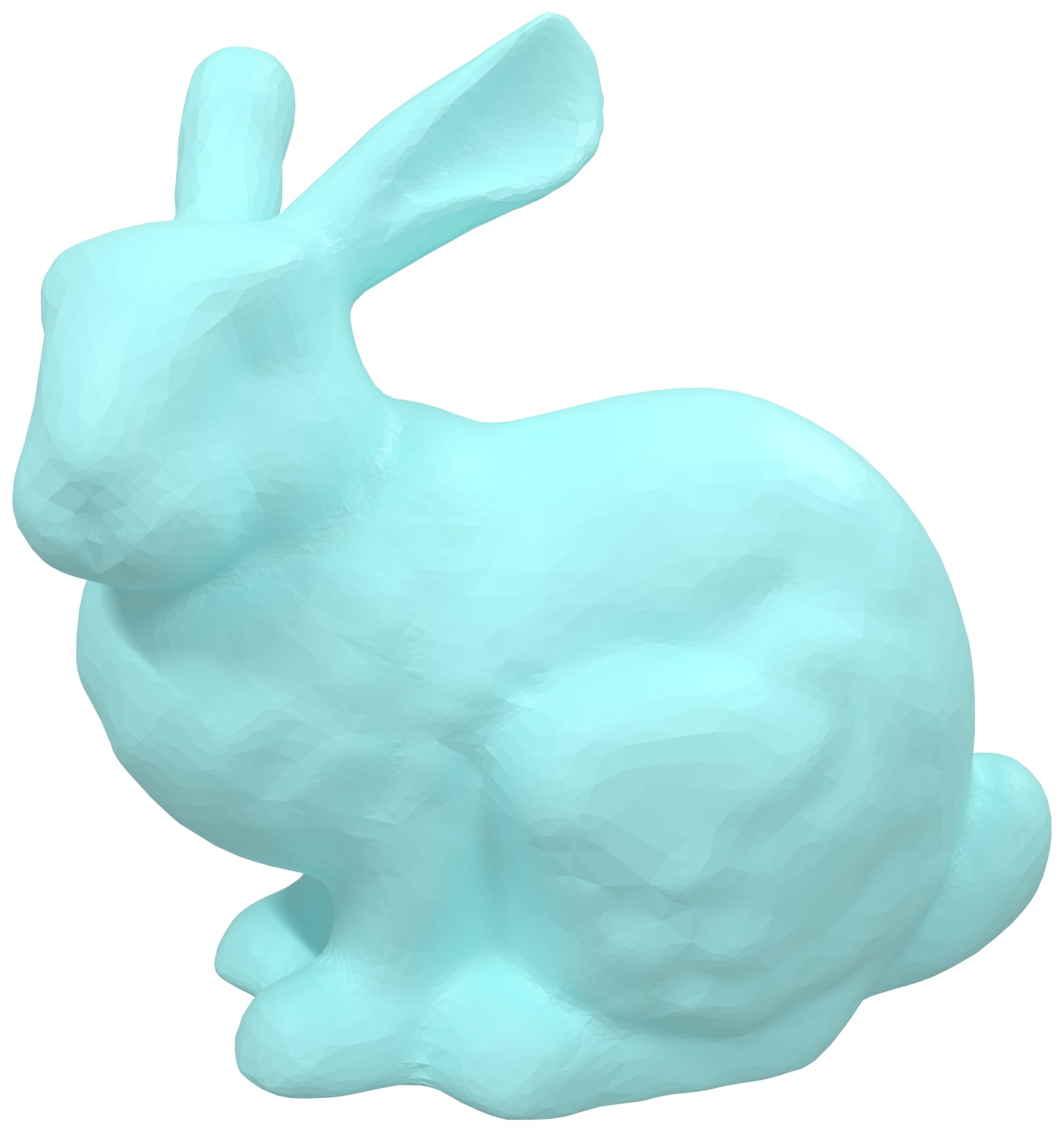 }\\
    \makebox[0.9in]{Weight}\\
    \includegraphics[width=0.9in]{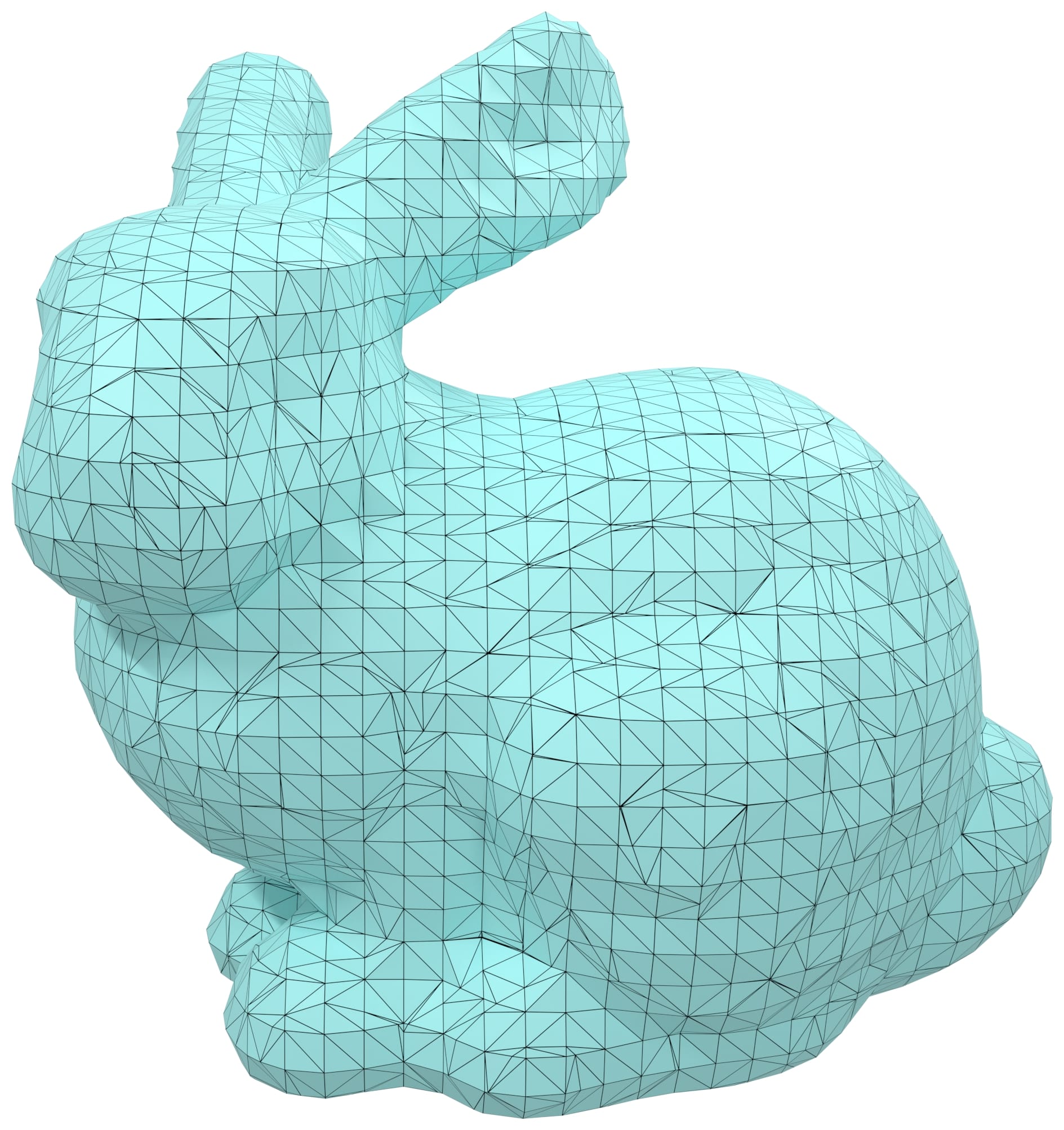 }
    \includegraphics[width=0.9in]{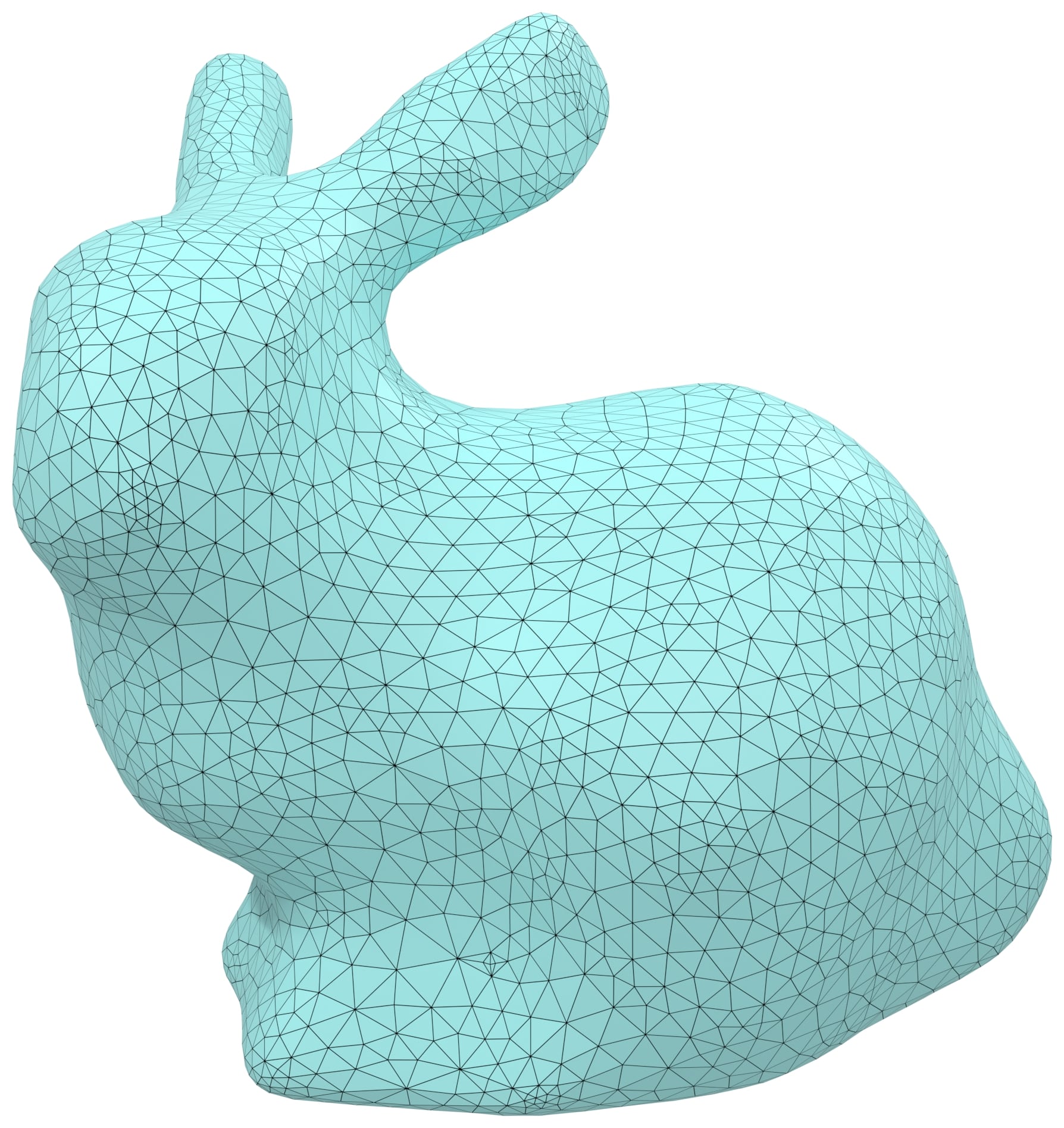 }
    \includegraphics[width=0.9in]{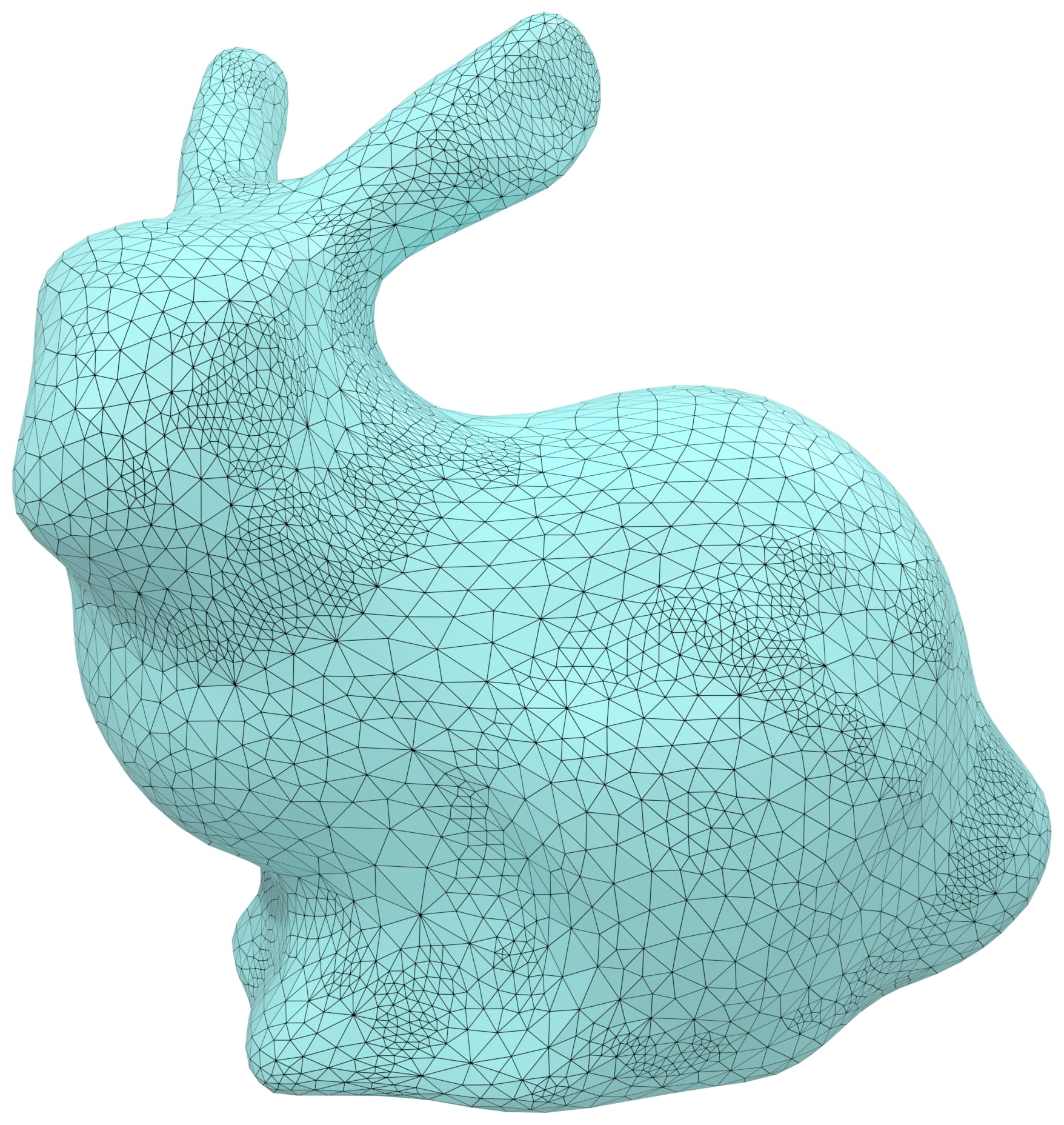 }
    \includegraphics[width=0.9in]{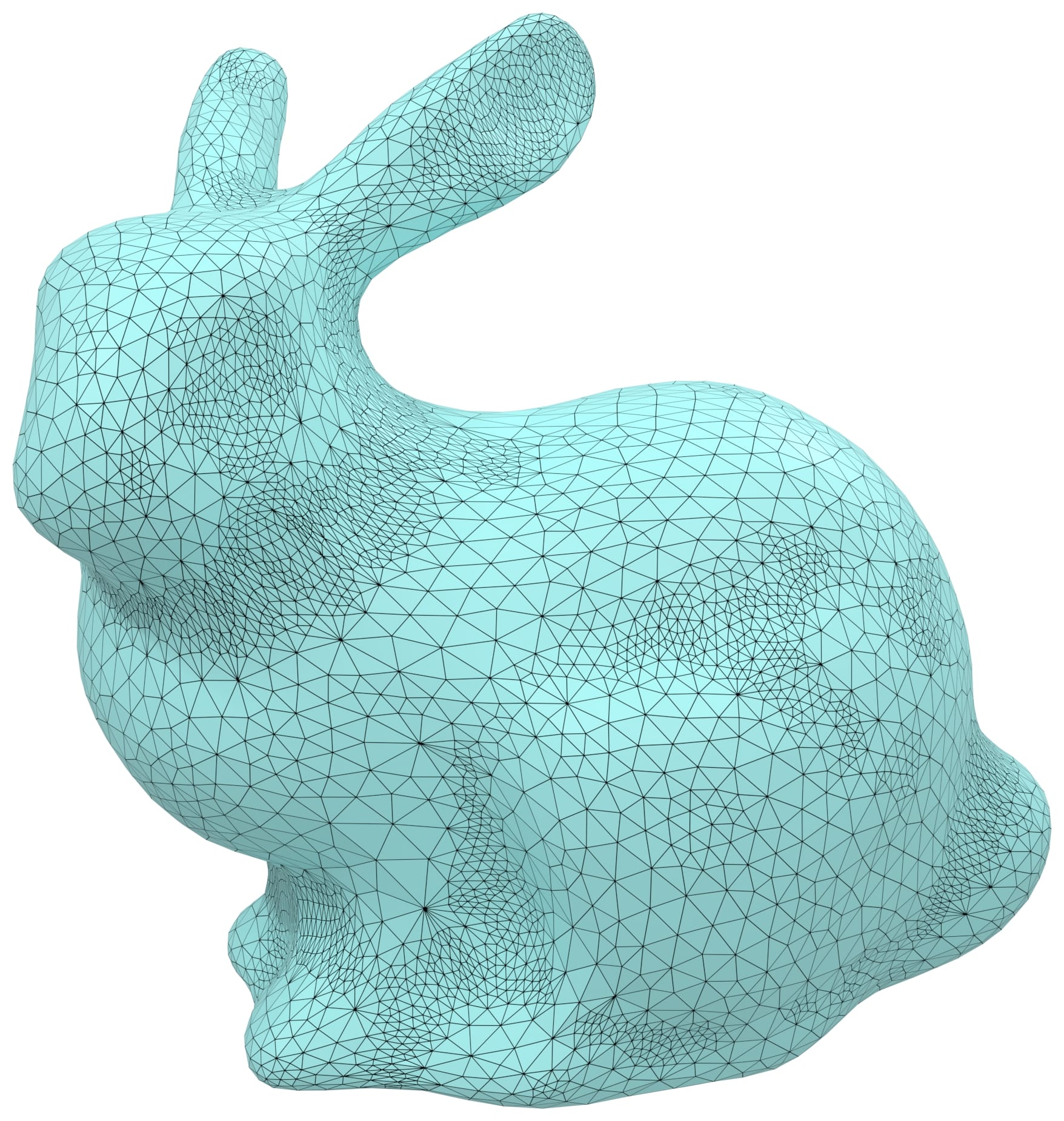 }
    \includegraphics[width=0.9in]{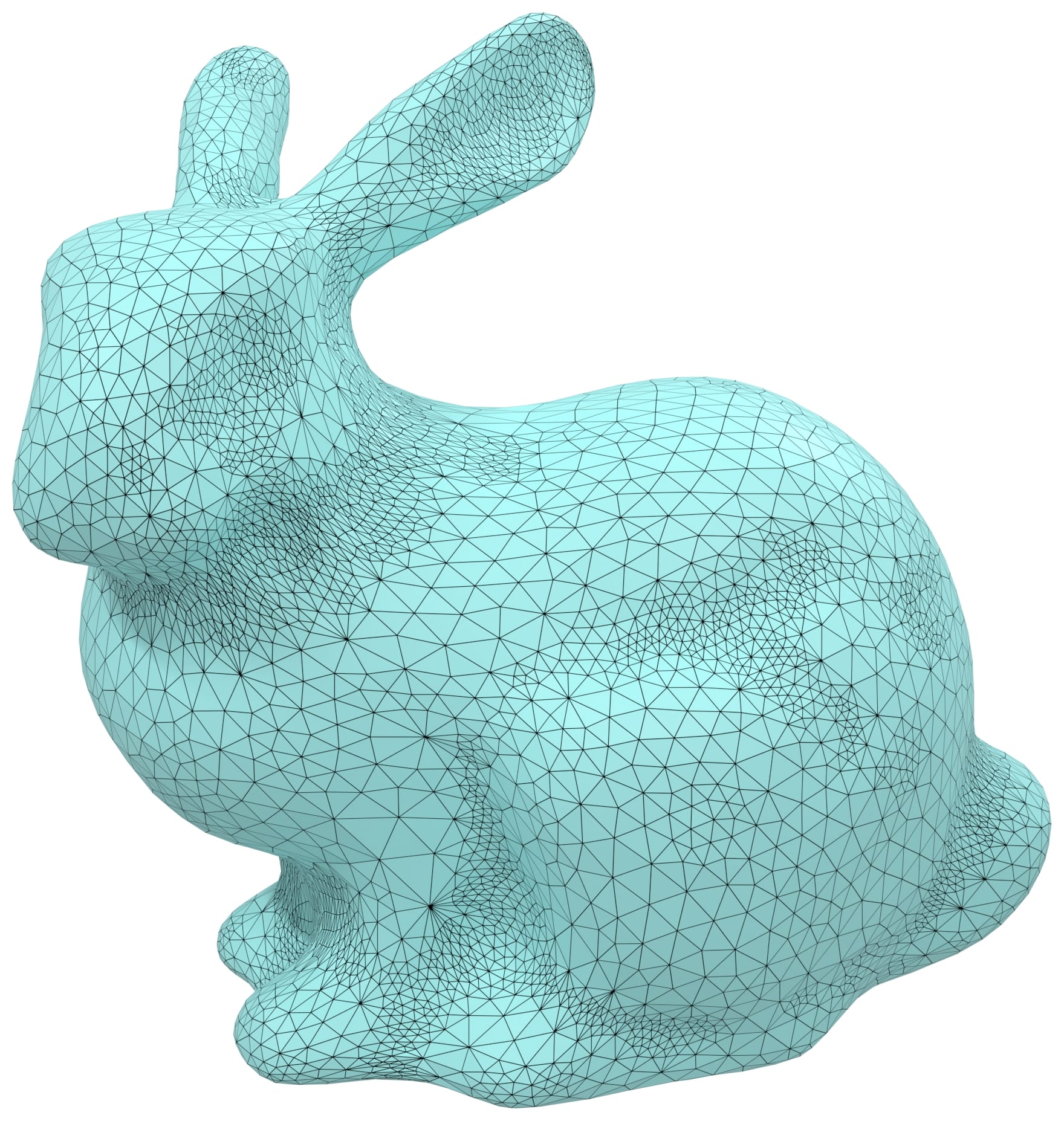 }
    \includegraphics[width=0.9in]{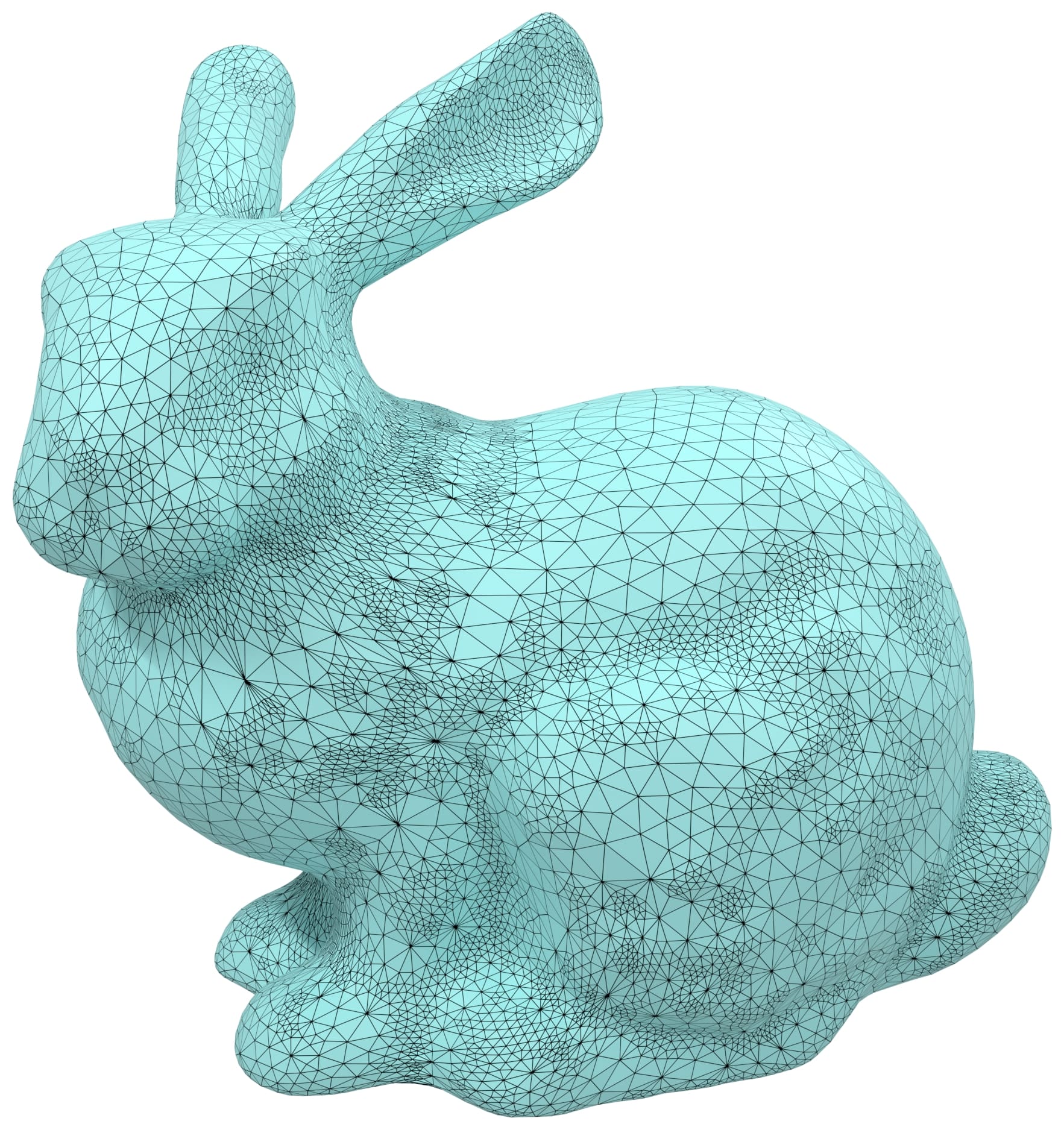 }
    \includegraphics[width=0.9in]{./images/Dynamic_weights/new/wire_400 }\\
    \makebox[0.9in]{Wireframe}\\
    \includegraphics[width=0.9in]{./images/Dynamic_weights/new/ours_mesh0 }
    \includegraphics[width=0.9in]{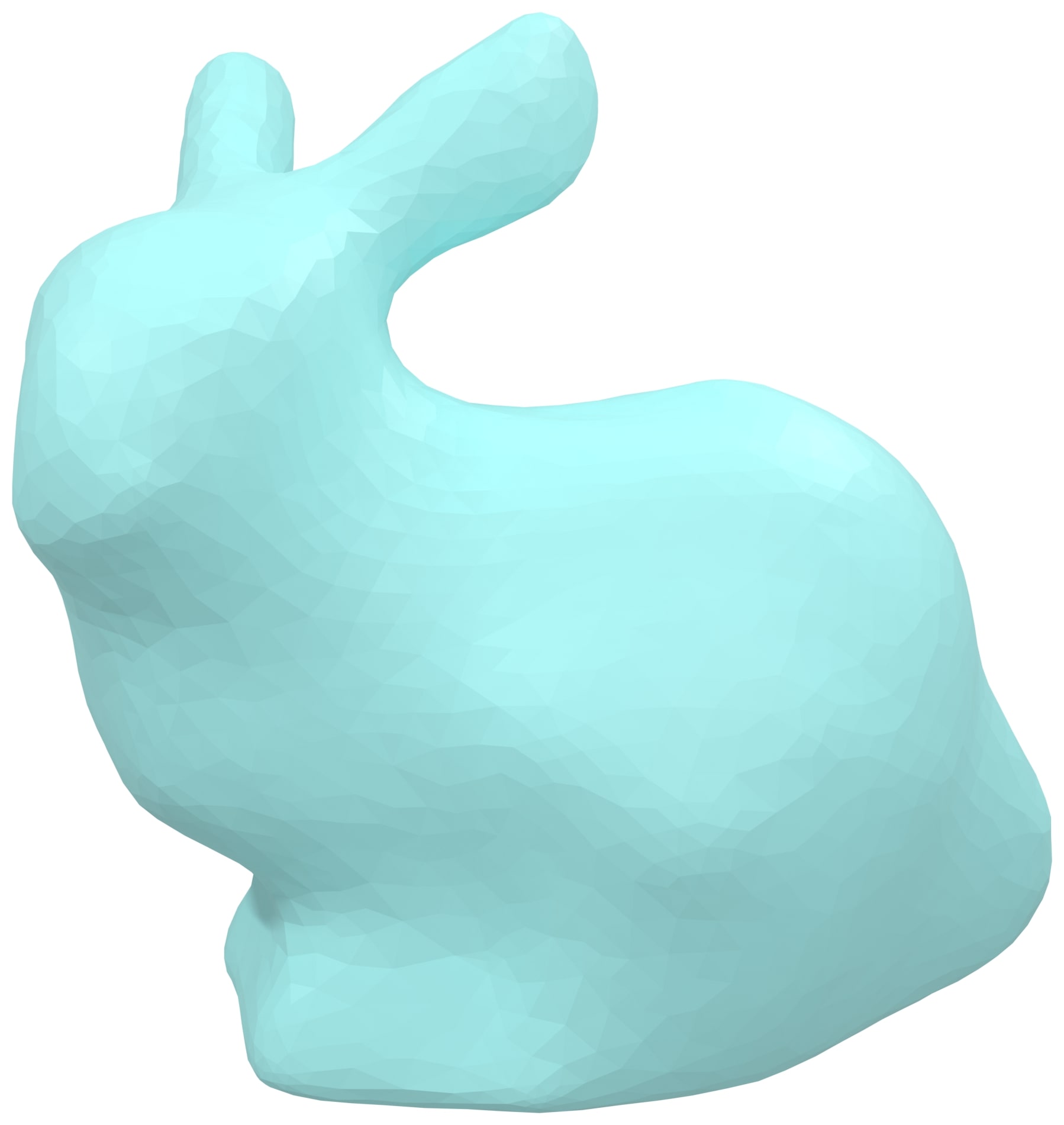 }
    \includegraphics[width=0.9in]{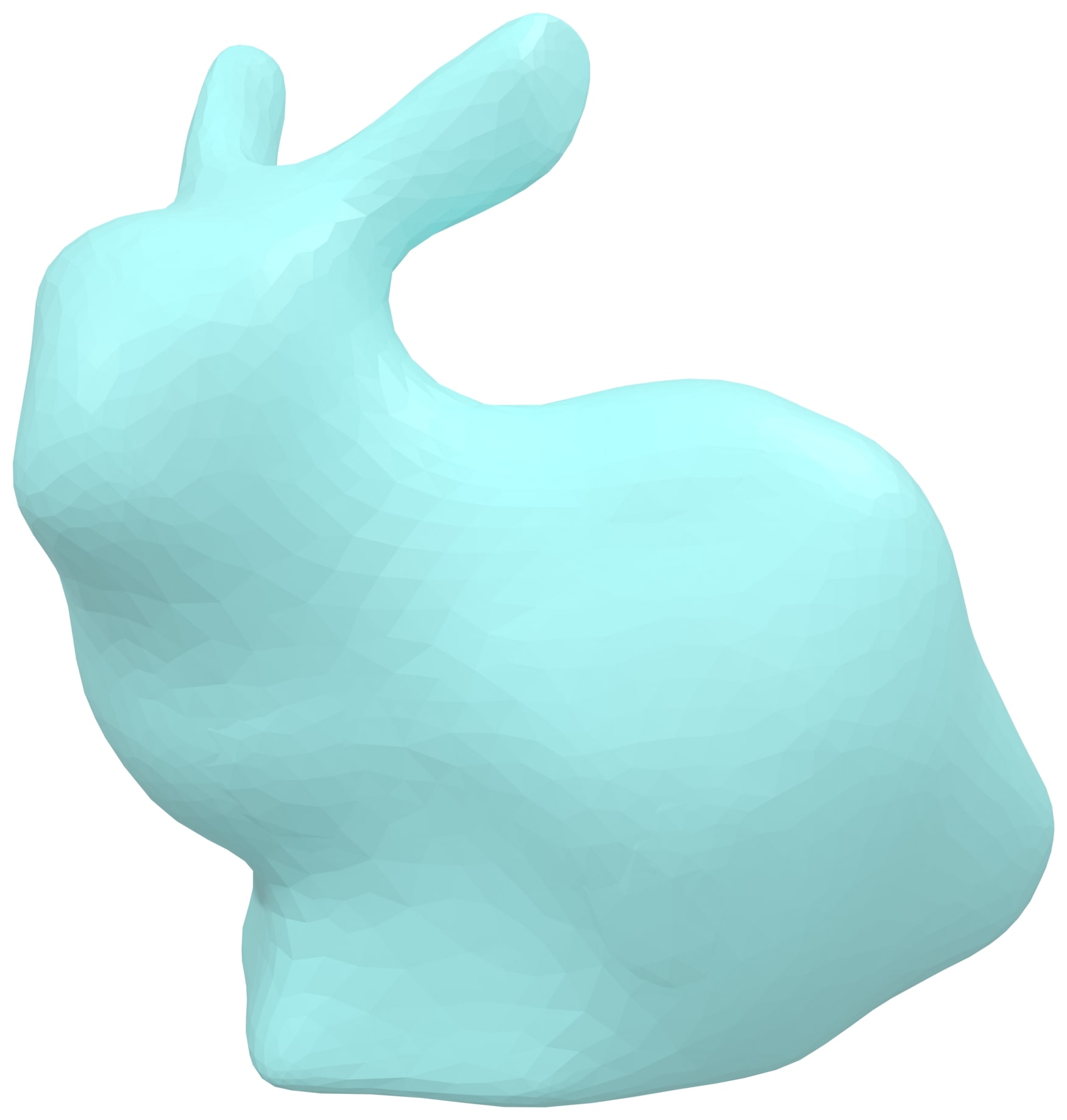 }
    \includegraphics[width=0.9in]{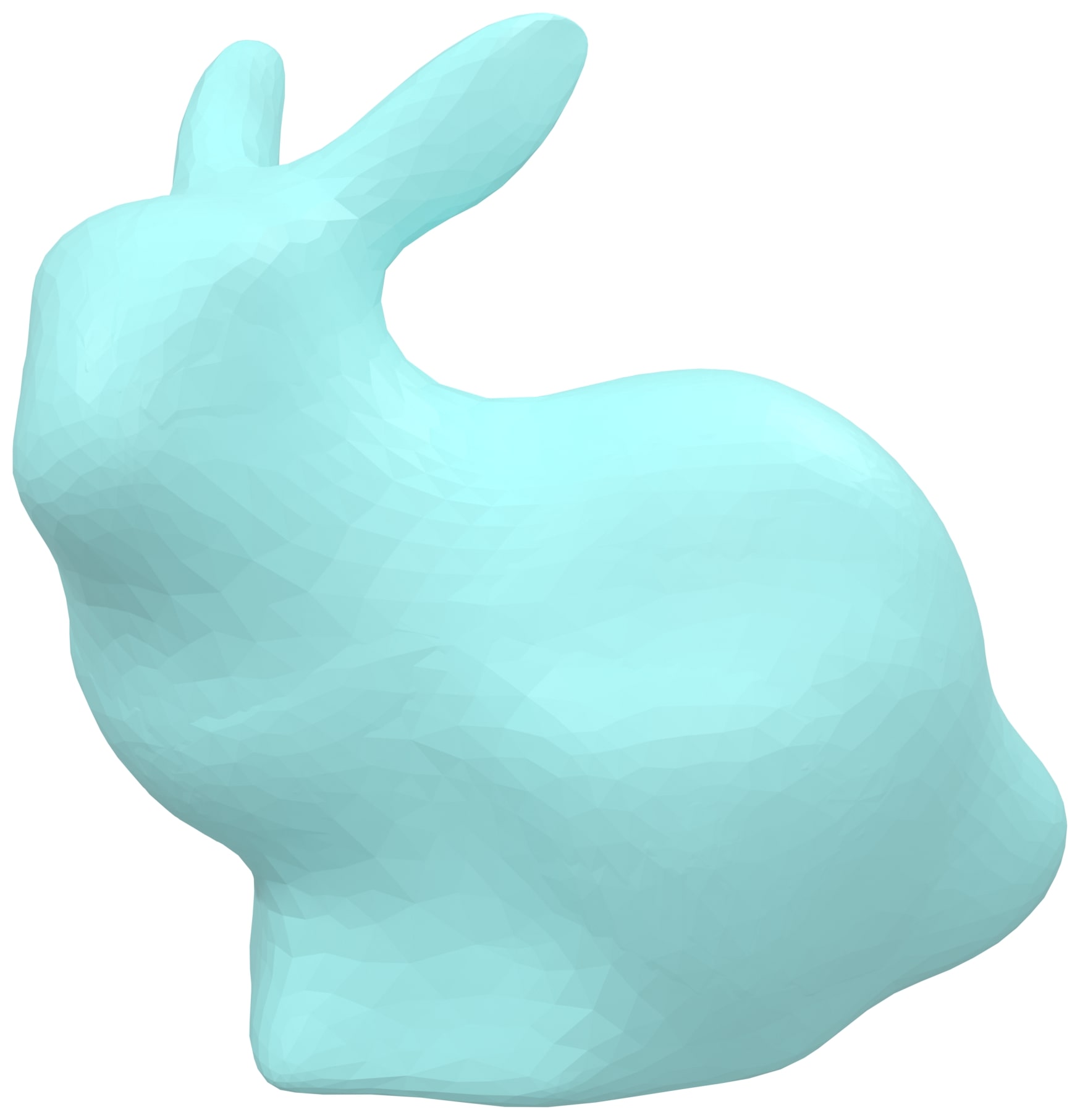 }
    \includegraphics[width=0.9in]{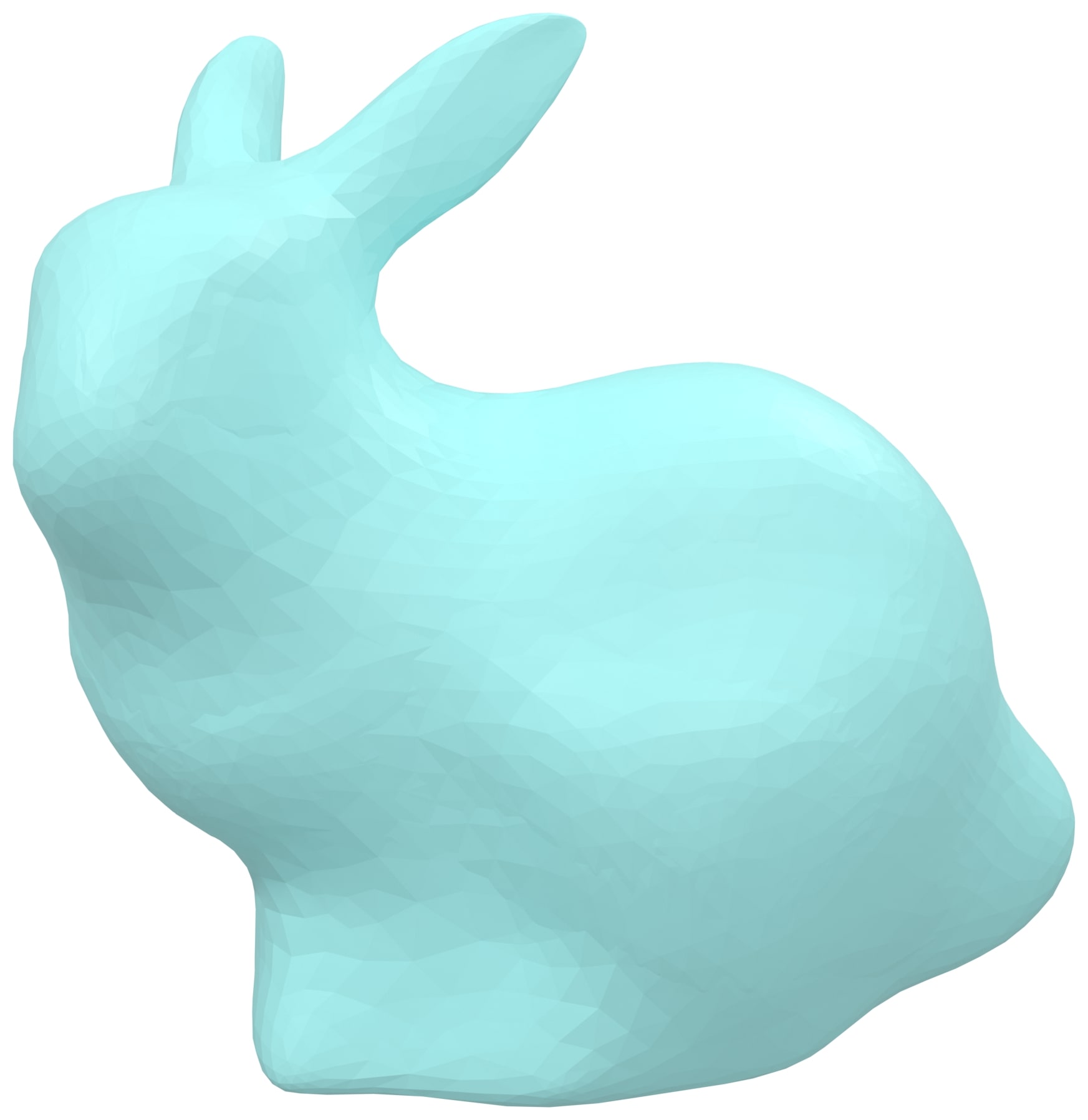 }
    \includegraphics[width=0.9in]{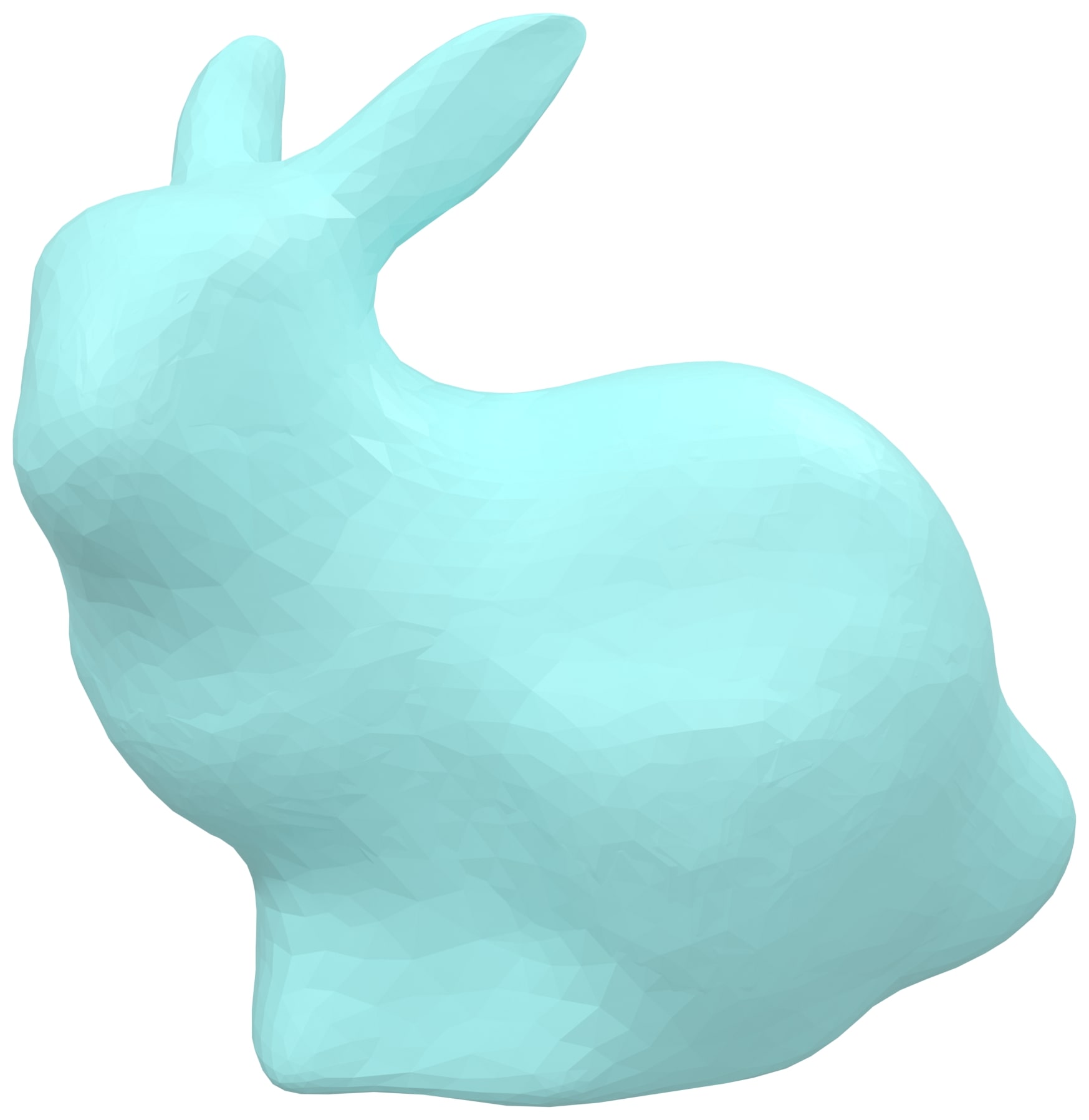 }
    \includegraphics[width=0.9in]{./images/Dynamic_weights/new/dcudf_400 }\\
    \makebox[0.9in]{DCUDF}\\
    \caption{Visual results during the minimization of the accuracy-aware loss function. Initially, uniformly high weights are applied across the entire low-resolution model. As optimization progresses, these weights are dynamically adjusted to focus increasingly on regions requiring finer fitting and refinement. The top row displays the weight values using a heat color map, where warm colors indicate higher weights and cool colors represent lower weights. The middle row showcases the meshes with wireframes, highlighting the introduction of new vertices that provide additional degrees of freedom necessary for better adapting to regions with geometric details. The bottom row illustrates the iterative results of DCUDF, which, using uniform weights, fails to capture the finer geometric details.}
    \label{fig:dynamic_weights}
\end{figure*}

\newcommand\mycommfont[1]{\footnotesize\ttfamily{#1}}
\SetCommentSty{mycommfont}

\begin{algorithm}
    
    \SetAlgoLined
    \SetKwInOut{Input}{input}
    \SetKwInOut{Output}{output}
    \SetKwRepeat{Do}{do}{while}
    \Input{A differentiable UDF $f$, the offset parameter $r>0$, and the weight thresholds $w_s$, $w_t$}
    \Output{A triangular mesh representing the zero-crossing iso-surface $\mathcal{M}$ of the UDF}
    Extract the iso-surface $\mathcal{M}$ of iso-value $r$ by marching cubes on $f$\;
    \tcc{Project $\mathcal{M}$ onto the zero level set by minimizing the accuracy-aware loss function}
    optimizer = VectorAdamOptimizer($\mathcal{M}$)\;
    \For{$iter\gets1$ \KwTo $\text{MaxIteration}$}{
    
    \If {$iter \mod s_1 ==0$}
    {
        Cluster regions where $w_{sa}>w_s$\;
        \ForEach{region $\in$ regions}{
          \If{Region contains two sets of closed curves}
            {
                \tcc{region is a columnar region}
                Remove region and fix holes\;
            }
        }
        
        Mask low weight regions ($w_{sa}>w_t$)\;
    }
    \If {$iter \mod s_2 ==0$}
    {
        Subdivide faces where $w_{sa}>w_s$\;
    }

    loss = $F(\mathcal{M})$ (\ref{eqn:step1})\;
    
    Update $w_{sa}$ according to (\ref{eqn:accumulation})\;
    loss.backward()\;
    Correct optimization direction (\ref{eqn:optimize_correct})\;
    optimizer.step()\;
    }
    \Return{the projected double-layered mesh $\mathcal{M}$}\;

    \caption{DCUDF2}
    \label{alg:DCUDF2}
\end{algorithm}

\subsection{Self-adaptive Weights and Accuracy-aware Loss Function}
\label{subsec:adaptiveweights}

A primary cause of the over-smoothing observed in DCUDF is the conflicting dynamics between the distance field loss and the Laplacian loss. This effect becomes  particularly evident in regions of high curvature, as illustrated in Figure~\ref{fig:local_optimize}. When the optimization process reaches a local optimum, these two losses find an equilibrium that results in a significant gap between the optimized mesh $\mathcal{M}$ and the target zero level set $\mathcal{S}$, thereby failing to capture the fine details of the geometry. 

To counteract this tendency towards over-smoothing, we propose an enhanced loss function equipped with self-adaptive weights. These weights are calculated based on the \textit{accumulated} loss for each point $p_i$ throughout the optimization process, increasing the loss function's sensitivity to deviations from the target geometry. We define the self-adaptive weight $w_{\text{sa}}$ for point $p_i$ under the $j$-th iteration as follows:
\begin{equation}
\label{eqn:accumulation}
    w_{\text{sa}}(p_i, j) = 
    \frac{\sum_{t=1}^{j-1}f(\pi^{(t)}(p_i))}
    {\frac{1}{|\mathcal{M}\cup\mathcal{C}|}\sum_{p_i\in\mathcal{M}\cup\mathcal{C}}
    \sum_{t=1}^{j-1}f(\pi^{(t)}(p_i))},
\end{equation}
where $\pi^{(t)}(p_i)$ denotes the position of $p_i$ in the $t$-th iteration. The numerator represents the accumulated loss for point $p_i$ over the last $j-1$ iterations, indicating how much $p_i$ has derived from the zero level set throughout the entire optimization process. The denominator normalizes this value by the average accumulated loss across all mesh vertices and triangle centroids, ensuring that the weight adaptation is scaled appropriately across the entire mesh. Points with higher accumulated losses over the $(j-1)$ iterations are given greater weights, prioritizing their adjustment in subsequent optimization steps to better capture the fine details of the geometry. Conversely, points with lower accumulated losses receive less emphasis, maintaining stability in already well-fitted regions.

With these self-adaptive weights, we significantly enhance the impact of the distance field loss in regions where the mesh has previously shown poor fit. This targeted adjustment allows the mesh to more effectively prioritize its optimization efforts toward minimizing the distance field loss where it is most needed. Consequently, the modified loss function is expressed as: 
\begin{equation}
\label{eqn:step1_dynamic}
\min_{\pi}\sum_{p_i\in\mathcal{M}\cup\mathcal{C}} w_{\text{sa}} f\big(\pi(p_i)\big)+\lambda_1\sum_{p_i\in\mathcal{M}} w(p_i)\Bigl\|\bigtriangleup \pi(p_i)\Bigr\|^2,
\end{equation}
where $w_{\text{sa}}(p_i)$
 dynamically adjusts the weight for each point $p_i$
  based on its accumulated loss, enhancing the focus on reducing the distance field error in critical areas. The Laplacian term continues to enforce smoothness across the mesh, ensuring that while the focus is on fitting the zero level set, the overall mesh structure remains smooth and the vertices are well-distributed.

\textbf{Remark:} In our implementation, we reset the self-adaptive weights $w_{\text{sa}}$ as defined in Equation~(\ref{eqn:step1_dynamic}) every $N$ iterations. The purpose for this reset is to re-calibrate the weight calculations to adapt to the evolving geometry of the mesh, which prevents the accumulation of errors and ensures that the optimization remains sensitive to the most current deviations from the target geometry. We empirically set $N=50$ in all experiments conducted for this paper.

\subsection{Editing Topology}
\label{subsec:topology}

\begin{figure}[!htbp]
    \centering
    \hspace{6pt}
    \includegraphics[height=0.95in]{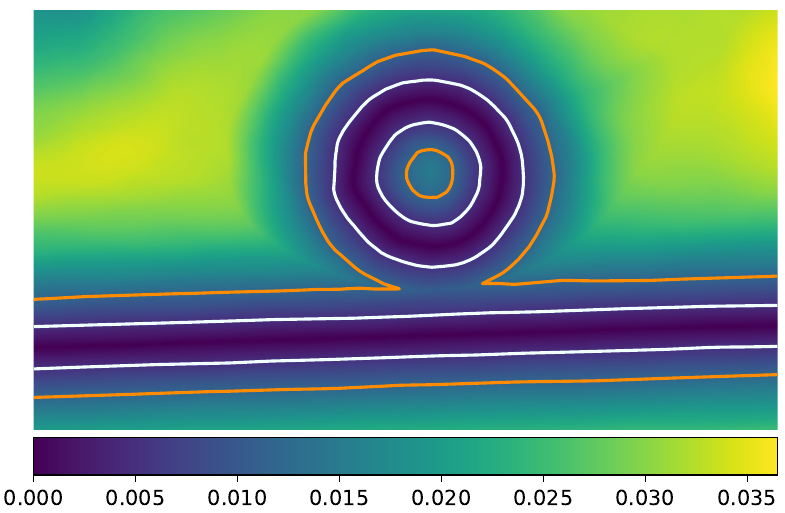}
    \hspace{2pt}
    \includegraphics[height=0.9105in]{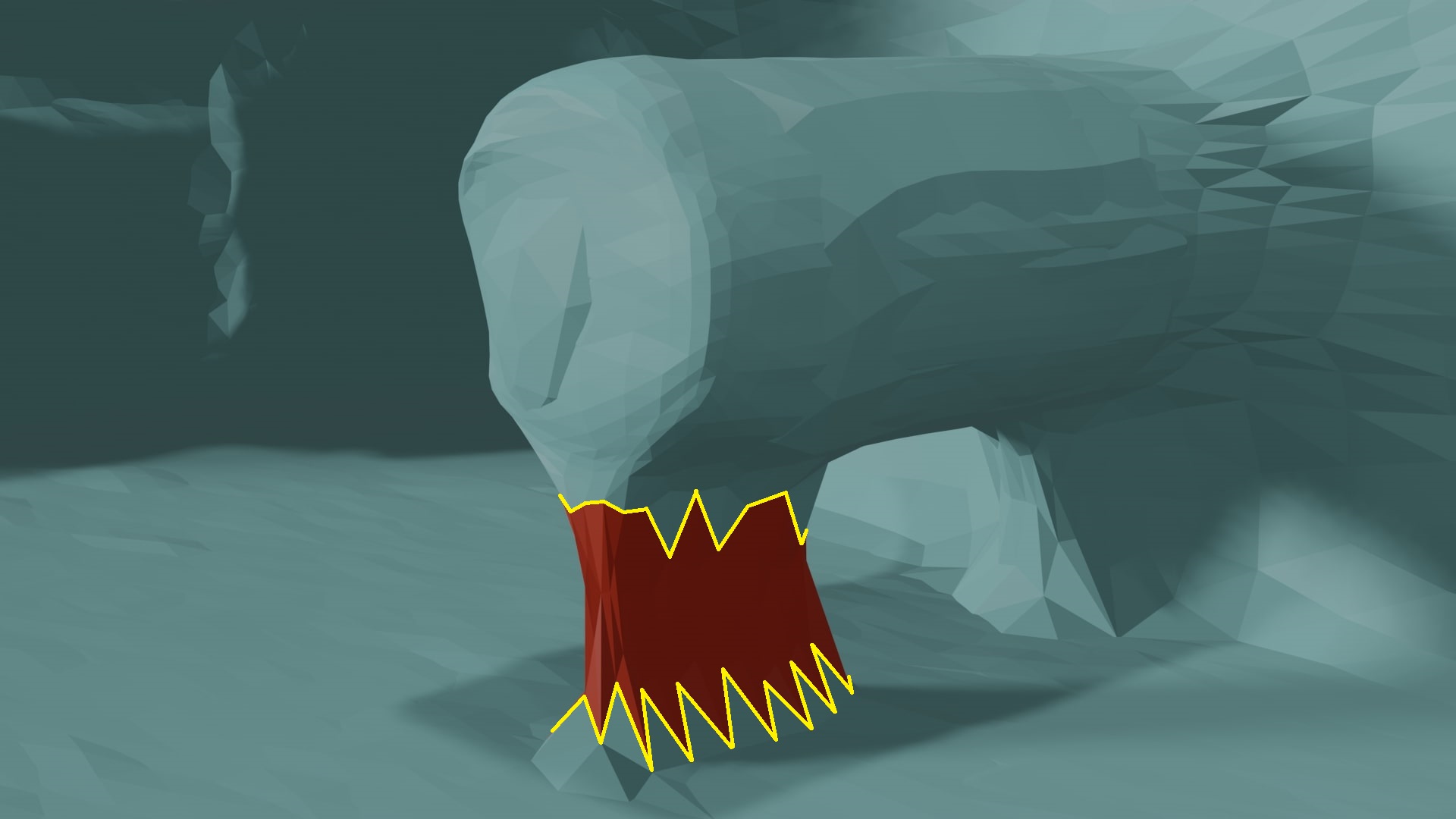}\\
    \makebox[1.55in]{(a)}
    \makebox[1.55in]{(b)}\\
    \includegraphics[height=0.9105in]{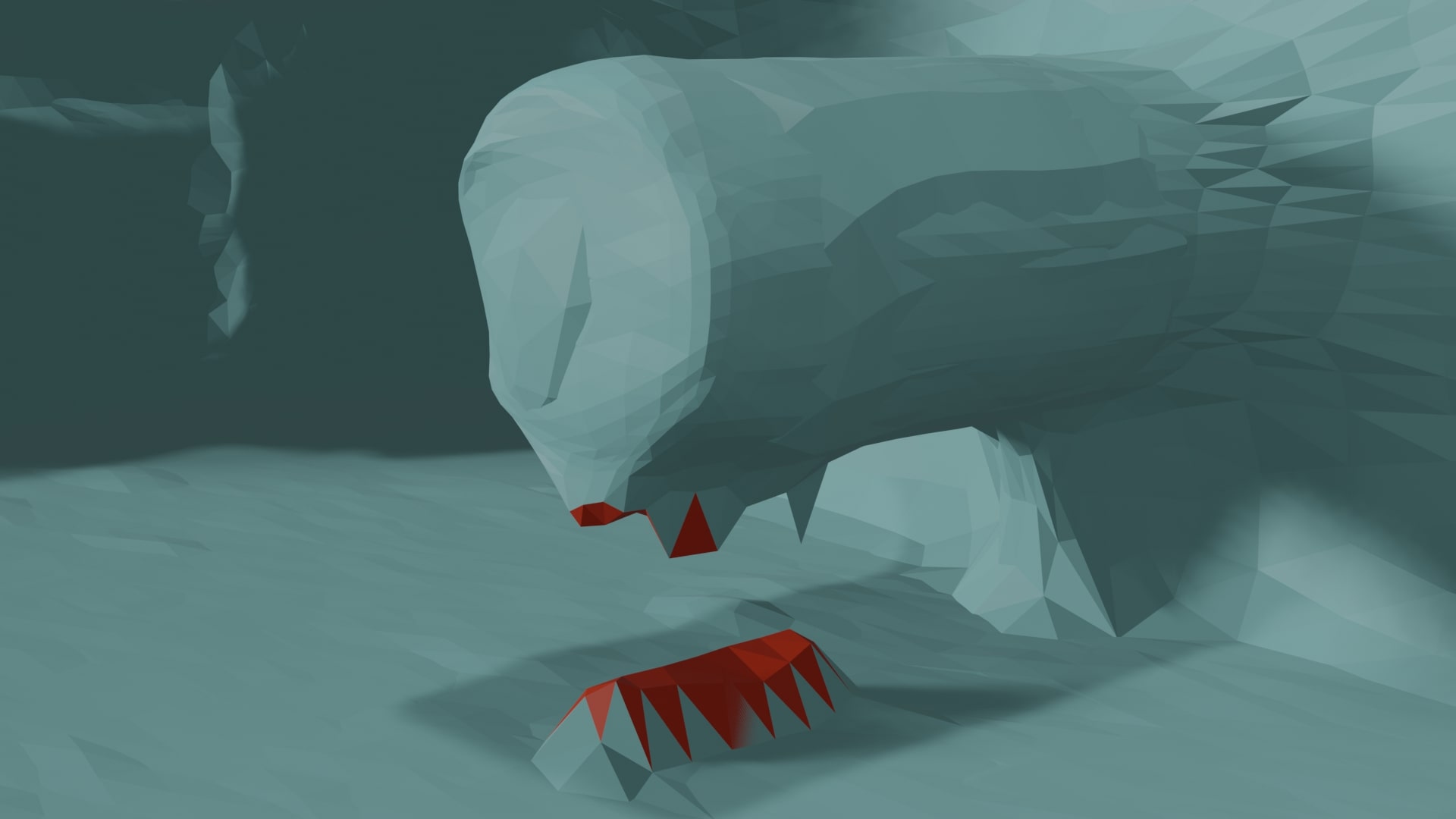}
    \includegraphics[height=0.9105in]{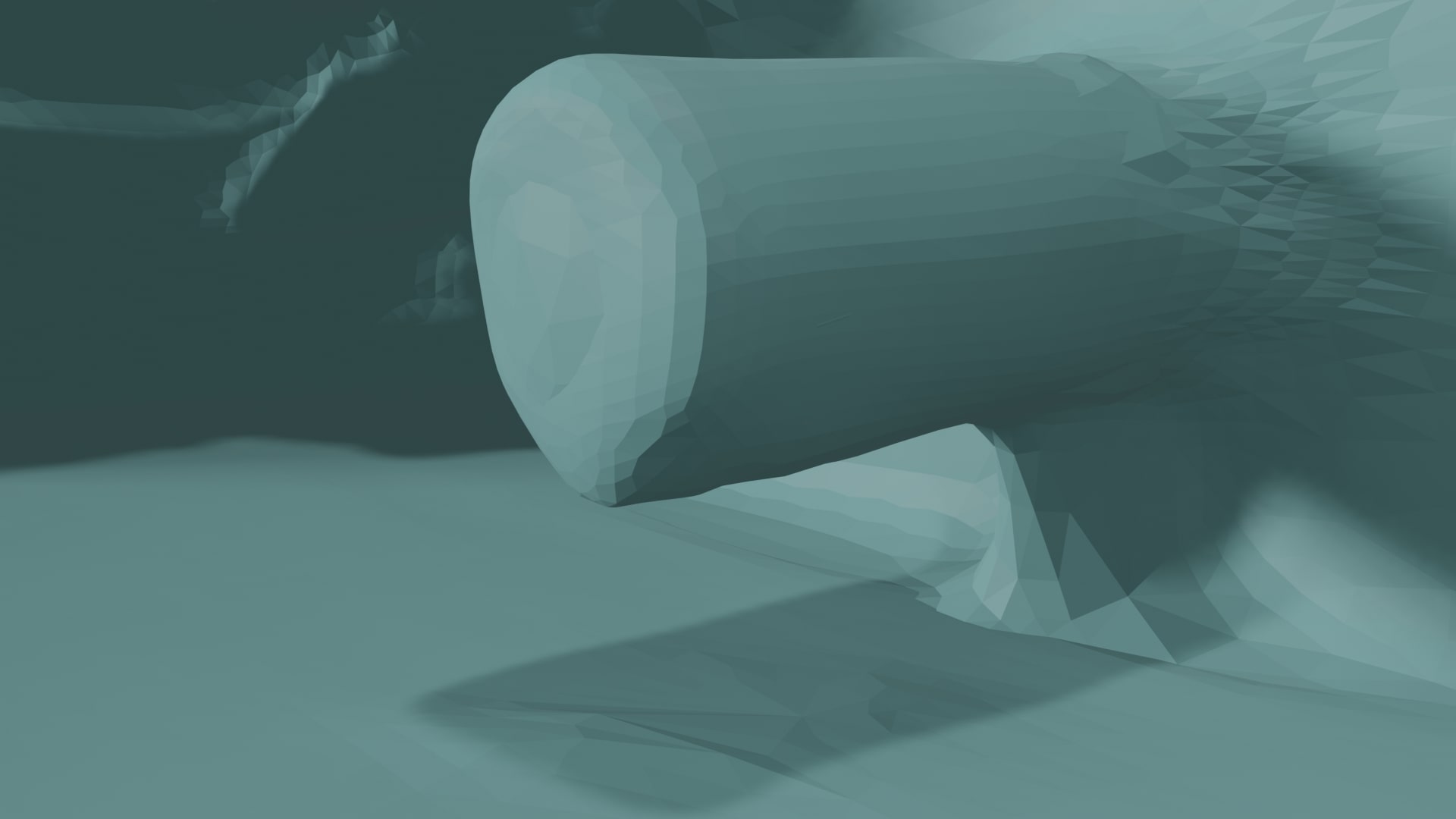}\\
    
    \makebox[1.55in]{(c)}
    \makebox[1.55in]{(d)}\\
    \caption{Topology correction can reduce dependency on the optimal setting of the  iso-value $r$. In (a), we provide a UDF visualization  of a cross-section involving a cylinder and a flat face, which are close to each other. A suitable small $r$, represented by the white iso-curves, accurately separates the cylinder from the plane, yielding an initial mesh $\mathcal{M}$ that preserves the topology of the actual zero level set $\mathcal{S}$. In contrast, an excessively large $r$, shown with orange iso-curves, leads to poor initialization by merging the cylinder and the plane, as illustrated in (b). By clustering the high weight region ($w_{sa}>w_s$), shown with red, we identifies it as cylinder if there are two closed boundary, as shown with yellow line in (b). } Our topology correction algorithm identifies and rectifies this topological error, successfully restoring the correct topology from an initially flawed configuration (c). This enables the subsequent optimization process to achieve a high-quality result with accurate topology  (d).
    \label{fig:levelset}
\end{figure}

Supporting topological editing in optimization allows a more flexible constraint to input hyperparameters.
As mentioned in DCUDF, when the value of $r$ is too large, the topology of the $r$ level set is not identical to $S$. Some small holes are filled, and at the same time, unnecessary columnar structures may be generated in the initial double-layer mesh, as shown in Figure~\ref{fig:levelset}. In DCUDF, a proper $r$ requires repeated attempts to determine and may not exist. In our work, we detect topological inconsistencies using an algorithm and correct the topology. This makes our method more robust to $r$.

We perform this algorithm every $s_1$ iterations. Since removing highly weighted columnar regions introduces holes, we need to handle them in a more precise manner. Firstly, we obtain clusters by performing a clustering operation on the highly weighted faces based on their topological connections. For each cluster, we determine the region as a columnar region if the edges of the cluster are two sets of closed curves. Subsequently, we delete the corresponding columnar region and perform a holes filling operation~\cite{fillholoesLiepa2023} based on the detected closed curves to obtain the final result. As shown in Figure~\ref{fig:levelset}, our method demonstrates efficacy in the removal of columnar regions, thereby enhancing the robustness of the input $r$.

Detail regions need a high resolution mesh to capture details while a high resolution mesh needs a long time to optimize. As detail regions are often high weight regions, we perform a subdivision operation~\cite{amresh2003adaptive} on the high weight regions every $s_2$ iterations to increase the number of faces in the detail-rich regions to better capture the local details. We define regions with $w_{acc} > w_s$ as high weight region to perform subdivision. As shown in Figure~\ref{fig:subdivision}, this help us generate a detail-rich mesh with a low resolution initialization.

\subsection{Activation Masks}
\label{subsec:mask}
The time required for each iteration is directly proportional to the number of vertices being optimized. To enhance efficiency, reducing the active vertices is essential. Our method employs self-adaptive weights to determine how well each part of the mesh fits the desired geometry. As illustrated in  Figure~\ref{fig:dynamic_weights}, regions with lower weights,  indicating satisfactory alignment, need note be updated in subsequent steps. We utilize an activation mask to manage this process, initially including all vertices. At every $\beta_m$ iterations, the mask is adjusted to exclude vertices whose self-adaptive weight $w_{\text{sa}}$ is below a pre-defined threshold $w_t$. This selective activation substantially reduces the computational cost, with our tests showing a reduction of up to 90\% vertices after 100 iterations, thereby significantly improving the optimization efficiency. 

\subsection{Optimization Direction Correction}
\label{subsec:orientation}

\begin{figure}[!htbp]
    \centering   \includegraphics[height=0.95in]{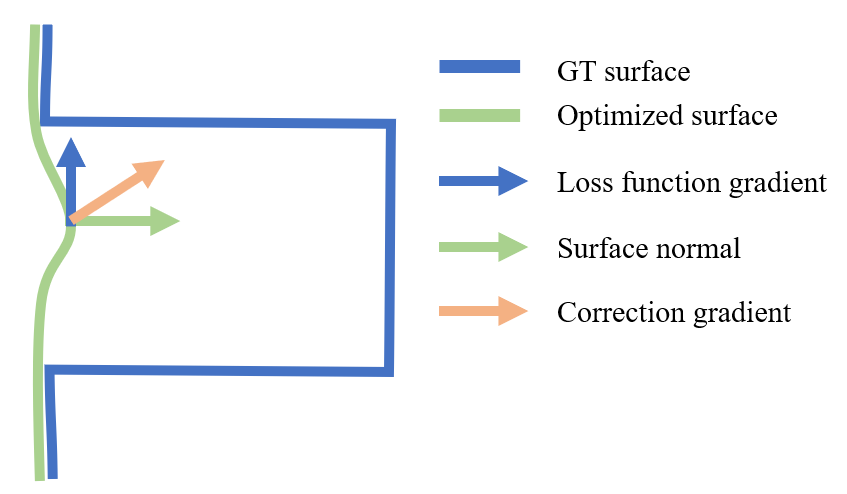}    \includegraphics[height=0.95in]{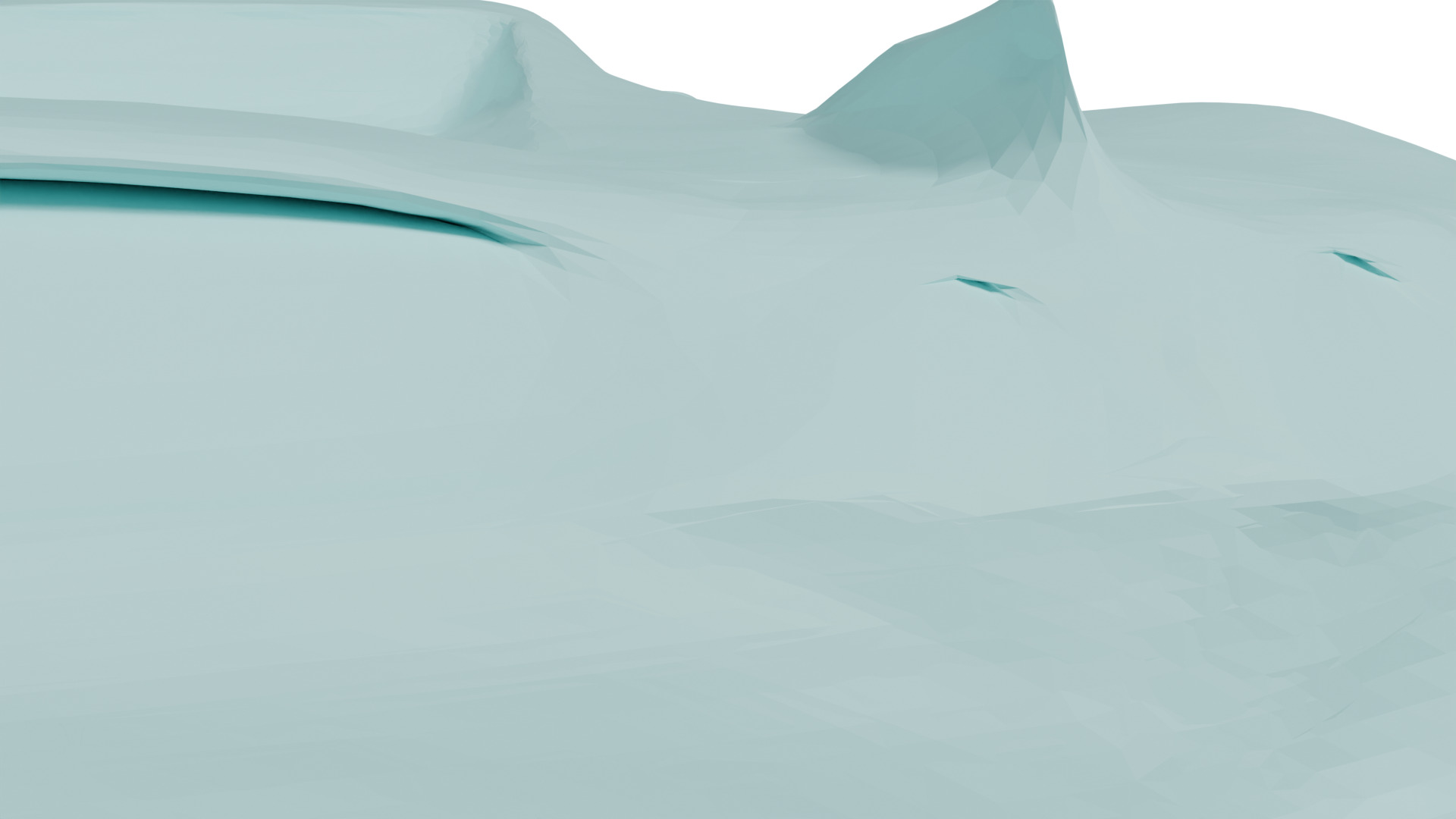}\\
    \makebox[1.55in]{(a)}
    \makebox[1.55in]{(b)}\\    \includegraphics[height=0.95in]{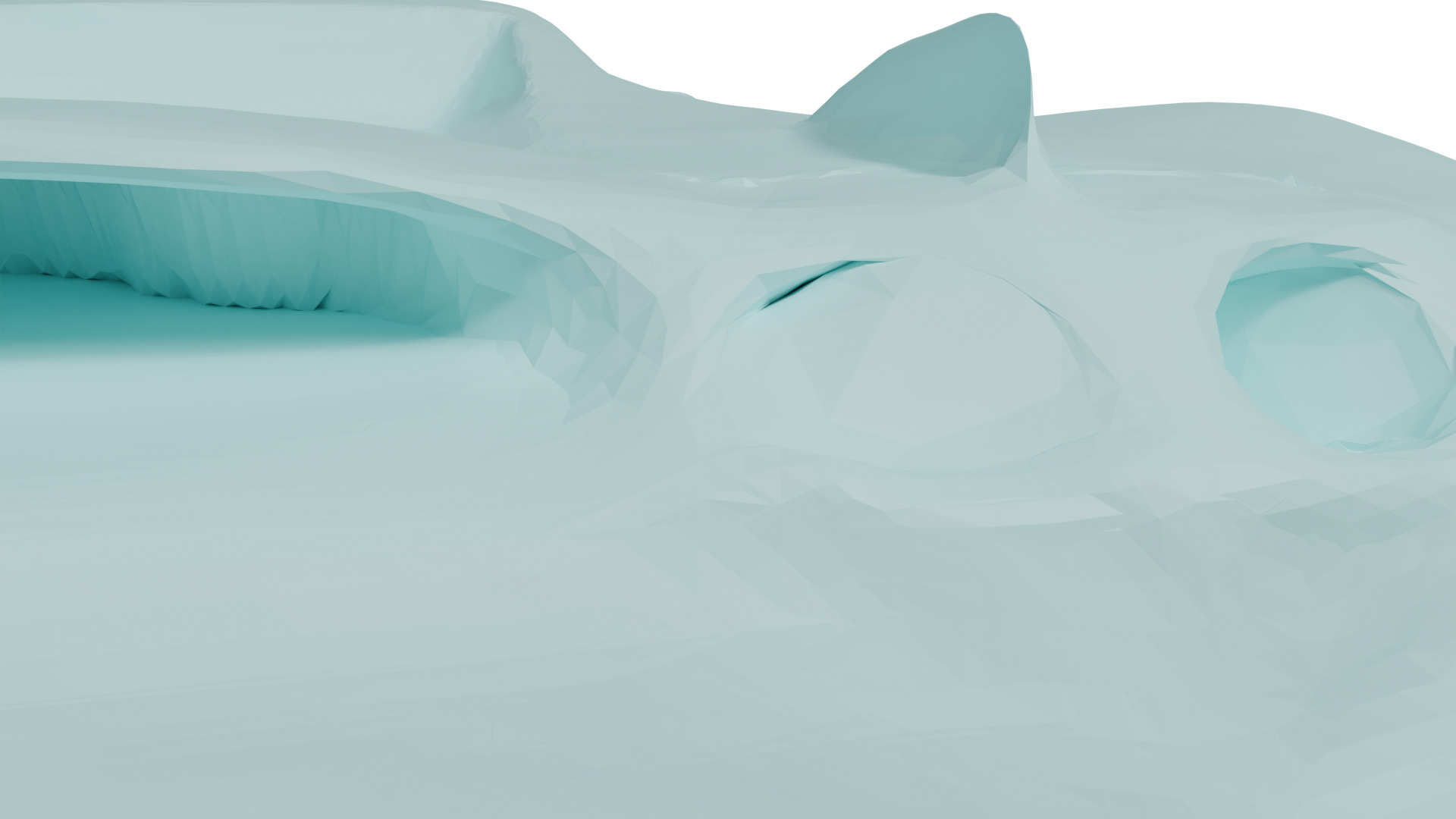}    \includegraphics[height=0.95in]{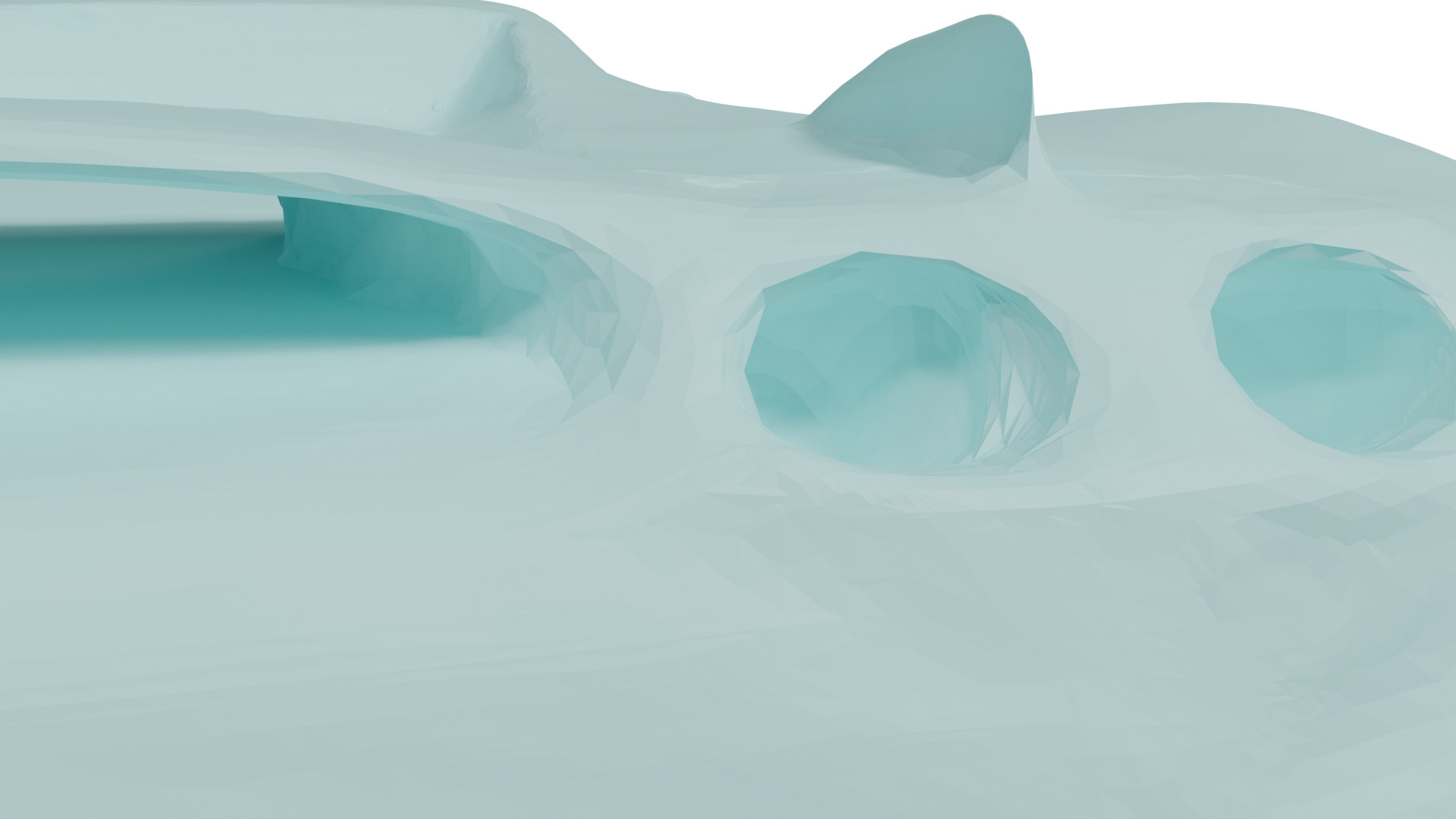}\\
    \makebox[1.55in]{(c)}
    \makebox[1.55in]{(d)}\\
    \caption{Optimization correction in U-shaped region. Optimizing the $r$ level set towards zero level set may fall into a local optimal solution in a U-shaped region. Although our self-adaptive weights can push the surface a little inward from (b) to (c). As shown in (a), it is difficult to optimize further as the loss gradient is almost perpendicular to the expected optimization direction. By correcting the optimization direction with the mix of gradient and surface normal, we can reach a good performance in (d).}
    \label{fig:U-shape}
\end{figure}

Optimization challenges are common in concave regions, where the process might get stuck. Figure~\ref{fig:U-shape} illustrates this issue with a U-shaped area, demonstrating the difficulty in effectively pushing the mesh $\mathcal{M}$ into the concave valley. In such cases, the unfitted points, still distant from the valley's base, have loss gradients that are nearly perpendicular to the desired direction of optimization. Influenced by the Laplacian loss, these points do not advance toward the valley but rather remain static, resulting in a suboptimal local equilibrium. 

To overcome local optima in concave areas, we utilize the mesh normal, which aligns with the desired direction of optimization, to correct the optimization direction as follows:
\begin{equation}
\label{eqn:optimize_correct}
    g_c(p_i) = w_{n}(p_i) \|\mathbf{g}_r(p_i)\| \mathbf{n}(p_i) + (1-w_{n}(p_i))\mathbf{g}_r(p_i)
\end{equation}
and
\begin{equation}
\label{eqn:correct_weight_reg}
   w_{n}(p_i) = \frac{w_{\mathrm{1\text{-}ring}}(p_i)-\min\limits_{p_i\in\mathcal{M}}(w_{\mathrm{1\text{-}ring}}(p_i))}{\max\limits_{p_i\in\mathcal{M}}(w_{\mathrm{1\text{-}ring}}(p_i))-\min\limits_{p_i\in\mathcal{M}}(w_{\mathrm{1\text{-}ring}}(p_i))}
\end{equation}
where 
\begin{equation}
\label{eqn:correct_weight}
   w_{\mathrm{1\text{-}ring}}(p_i) = \sum_{p_j\in\mathcal{F}(p_i)}w_{acc}(p_j),
\end{equation}
and $\mathbf{g}_r$ is the origin loss gradient, $\mathcal{F}(p_i)$ denotes the 1-ring neighboring face centroids $p_i\in\mathcal{C}$.
As the U-shaped region is one class of poorly fitted regions, we mainly modify the high weight region and maintain the origin gradient in the low weight region. As shown in Figure~\ref{fig:U-shape}, we optimize the mesh further inward into a concave region.

\section{Experiments and Comparisons}

\subsection{Experiment Setup}
Our experiments were conducted on a workstation equipped with an Intel Xeon E5-2630 v4 CPU, 64GB RAM, and an NVIDIA TITAN V GPU with 12GB of memory. Computations involving the covering map $\pi$ were performed on the GPU using Pytorch, while computations for the dilated double cover and topology  editing were handled on the CPU. Following DCUDF, We employed the vector Adam solver~\cite{Ling2022}--a geometry optimization tailored variant of the standard Adam optimizer~\cite{Kingma2015}--to minimize the loss function.

All input models were scaled uniformly to fit within a unit box to normalize parameter scales.We set the Laplacian weights $\lambda_1=1800$ of Equation~(\ref{eqn:step1_dynamic}). We set $w_t = 2$ to generate the optimization mask used in Sec.~\ref{subsec:mask}. We set $w_s =2$ for most of our experiments to perform subdivision and it is enough to subdivide the properer region with $w_s = 3$ in the Shapenet-Car dataset \cite{Chang2015ShapeNetAI}. We use 150 iterations with a learning rate of 0.0005 by default to optimize the surface $S$. We extract the initial mesh with $r=0.005$ in $128^3$ resolution by default and set $r = 0.01$ for noisy data like the DeepFashion3D dataset ~\cite{Park2019DeepSDFLC}. 
We set $s_1 = 100$ and $s_2 = 50$ for topology editing.

\subsection{Hyper-parameter Settings}
DCUDF2 employs two critical hyper-parameters for topology correction: $w_s$ and $w_t$. The optimal settings for these parameters vary by dataset. As shown in Figure~\ref{fig:rate}, for cloth data, setting both $w_s$ and $w_t$ to 2 targets local regions effectively. For car models in the ShapeNet-Car dataset, a setting of $w_s=3$ enhances runtime efficiency without excessive mesh subdivision, whereas $w_s=2$ tends to over-subdivide. The influence of these settings on accuracy is quantitatively presented in Table~\ref{tab:w}.

In addition, there are two model-independent parameters, $s_1$ and $s_2$, for the subdivision process. Typically, we only need to perform one time to slice the columns, so we set $s_1=100$ when we use 150 iterations to optimize the mesh. In practice, setting $s_2=50$ yields good results. If the mesh is initialized at a lower resolution and a higher iso-value $r$, more iterations and subdivisions are necessary to achieve the desired results. We show an example in Figure~\ref{fig:dynamic_weights}.

\subsection{Baselines \& Quantitative Measures}

We compared our method against three leading approaches in zero level set extraction from UDFs: 
\begin{itemize}
    \item MeshUDF~\cite{guillard2022udf}: A variant of the Marching Cubes algorithm, which utilizes gradient directions to determine edge intersections. 
    \item  DMUDF~\cite{Zhang2023SurfaceEF}: A variant of Dual Contouring, which projects points from a non-zero level set to the zero level set, facilitating mesh extraction via a customized DC algorithm.
    \item DCUDF~\cite{Hou2023DCUDF}: An optimization-based approach that extracts and refines a non-zero level set to approximate the target surface\footnote{DCUDF's post-processing step for obtaining a single-layer mesh, which involves computing a minimum cut by trial-and-error searches for source and sink, exhibits instability at lower resolutions (e.g., $128^3$). Therefore, we chose not to utilize this step in DCUDF. }. 
\end{itemize}

We assessed the quality of the extracted meshes for each method using both geometrical and topological metrics. Geometrically, we utilized the Chamfer distance (CD) to quantify the dissimilarity between the generated meshes and the ground truth. To calculate CD, we sampled 10K points from the ground truth meshes or point clouds.

For topological evaluation, we utilized metrics that highlight non-manifold configurations, genus and boundaries: 
\begin{itemize}
    \item Percentage of Non-manifold Configurations: Non-manifold  vertices are key topological anomalies, occurring where the local topology consists of multiple topological disks or half-disks converging at a single point. Like non-manifold vertices, non-manifold edges occur when several surfaces intersect along an edge. These configurations can undermine the mesh's stability during further processing, since simply removing them yield undesirable gaps in the mesh.
    \item Genus and Boundaries: These metrics assess the mesh's overall topological structure, offering insights into the global topology. Note that genus is typically computed using the Euler characteristic for manifold meshes. 
\end{itemize}

These metrics provide insights into both the geometric fidelity and the topological structure of the generated meshes, crucial for assessing their suitability for downstream applications.

\subsection{Evaluations \& Comparisons}

\begin{figure*}[!htbp]
    \centering
    \makebox[1.2in]{GT}
    \makebox[1.2in]{MeshUDF}
    \makebox[1.2in]{DMUDF}
    \makebox[1.2in]{DCUDF}
    \makebox[1.2in]{DCUDF2}\\
    \includegraphics[width=1.2in]{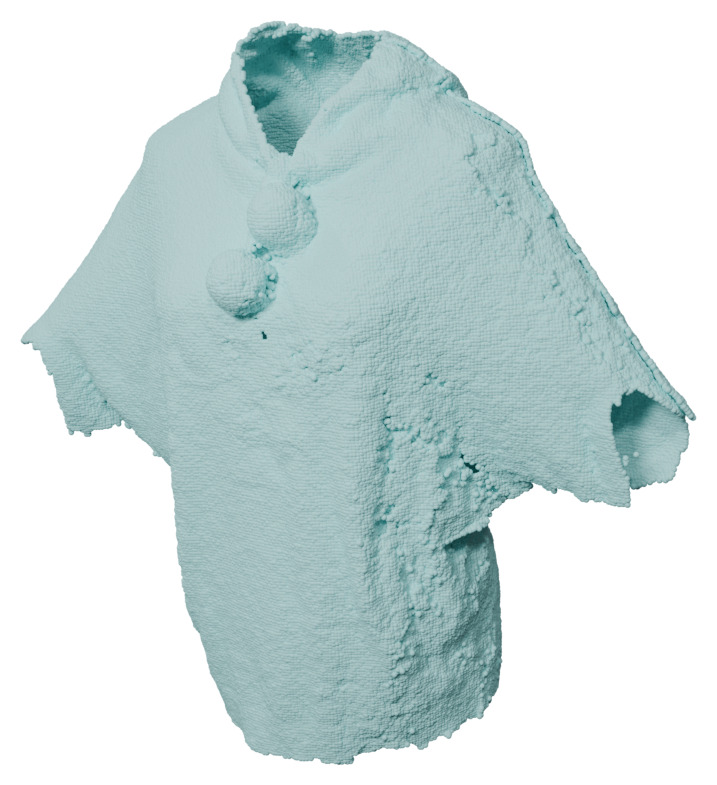 }
    \includegraphics[width=1.2in]{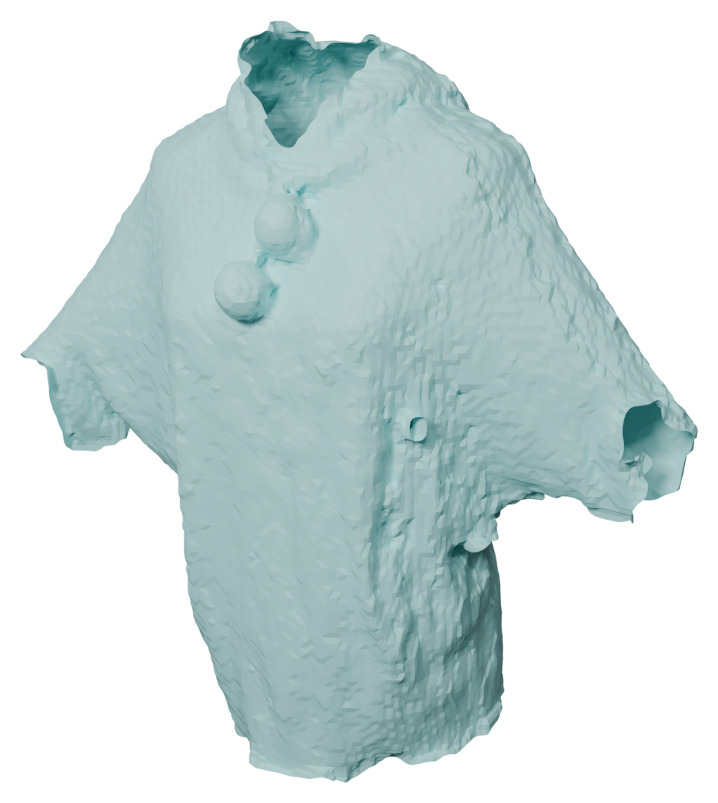 }
    \includegraphics[width=1.2in]{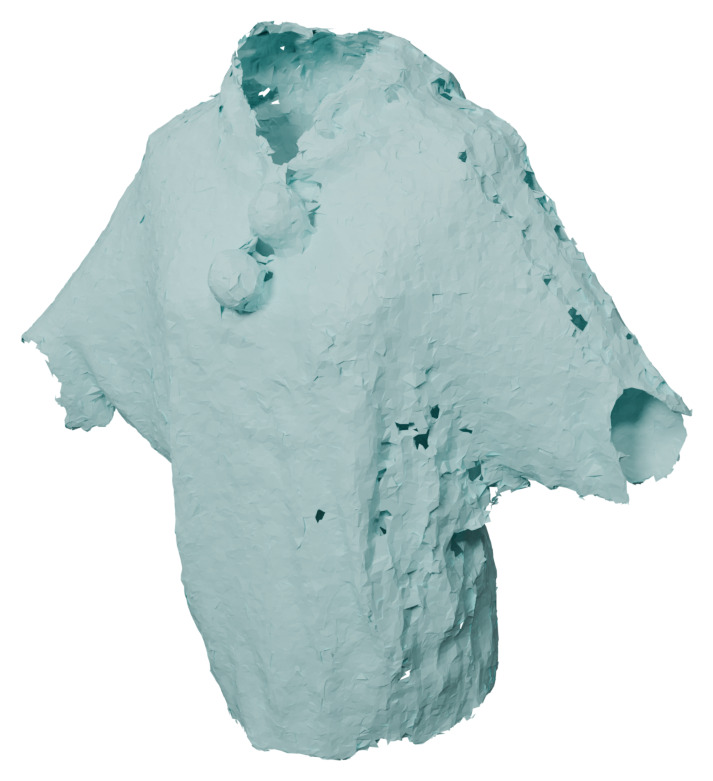 }
    \includegraphics[width=1.2in]{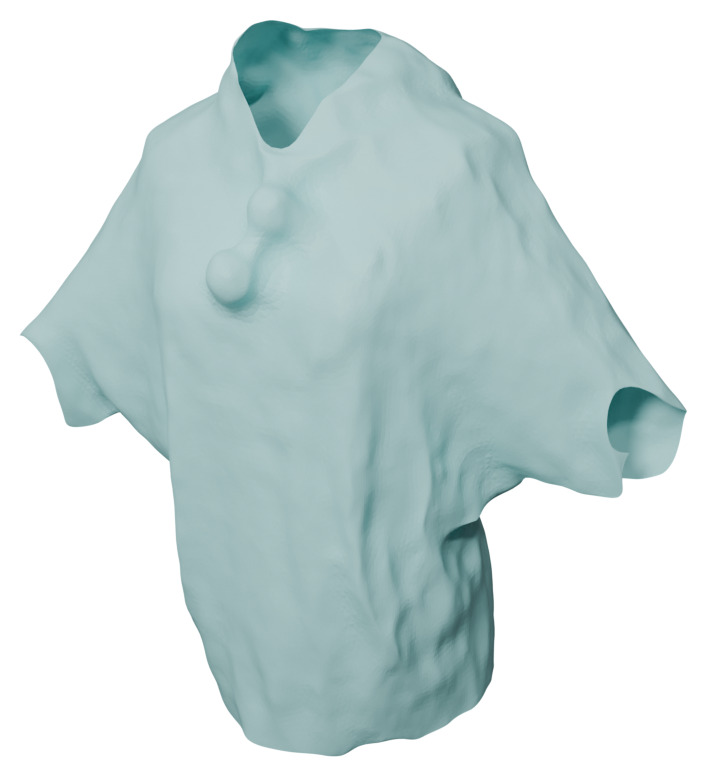 }
    \includegraphics[width=1.2in]{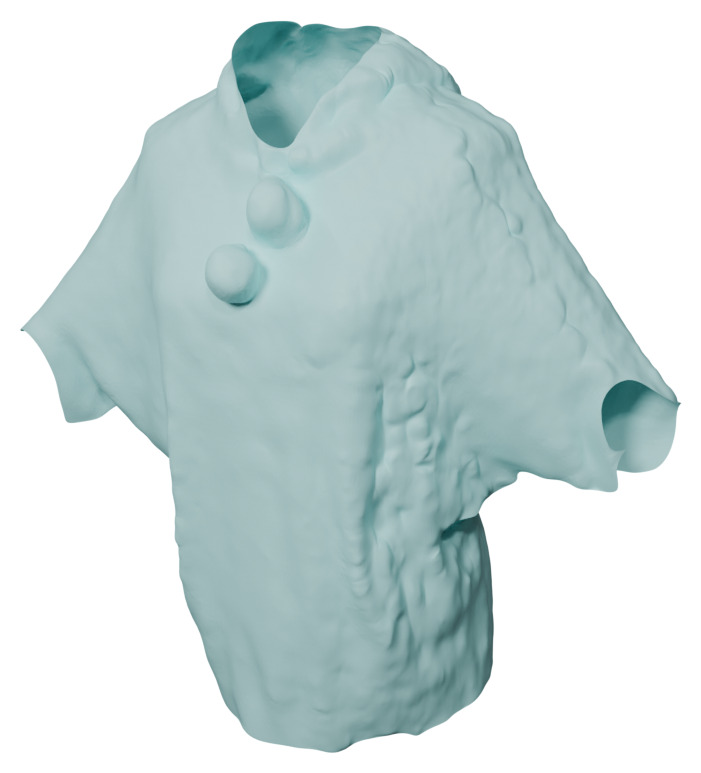 }\\
    \includegraphics[width=1.2in]{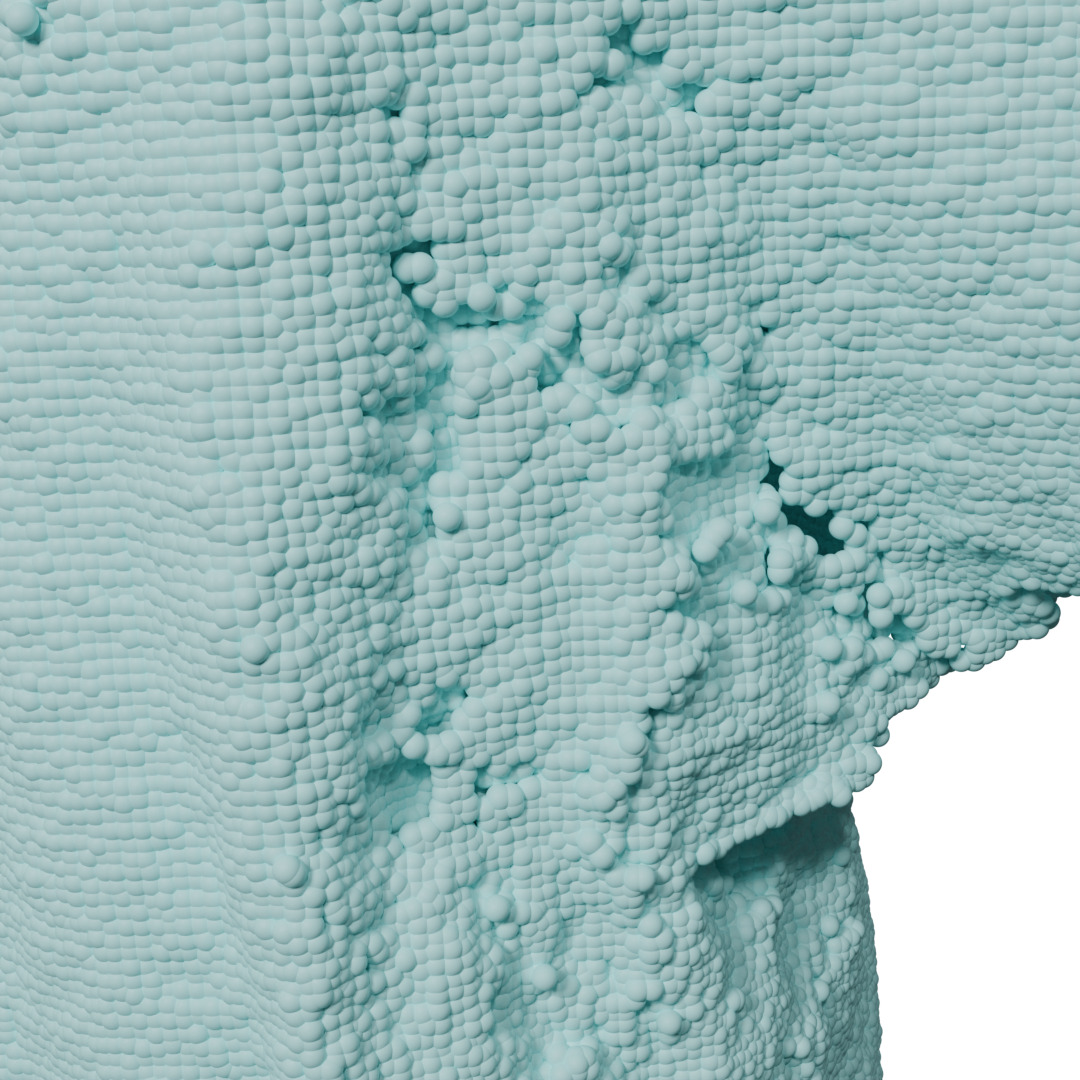 }
    \includegraphics[width=1.2in]{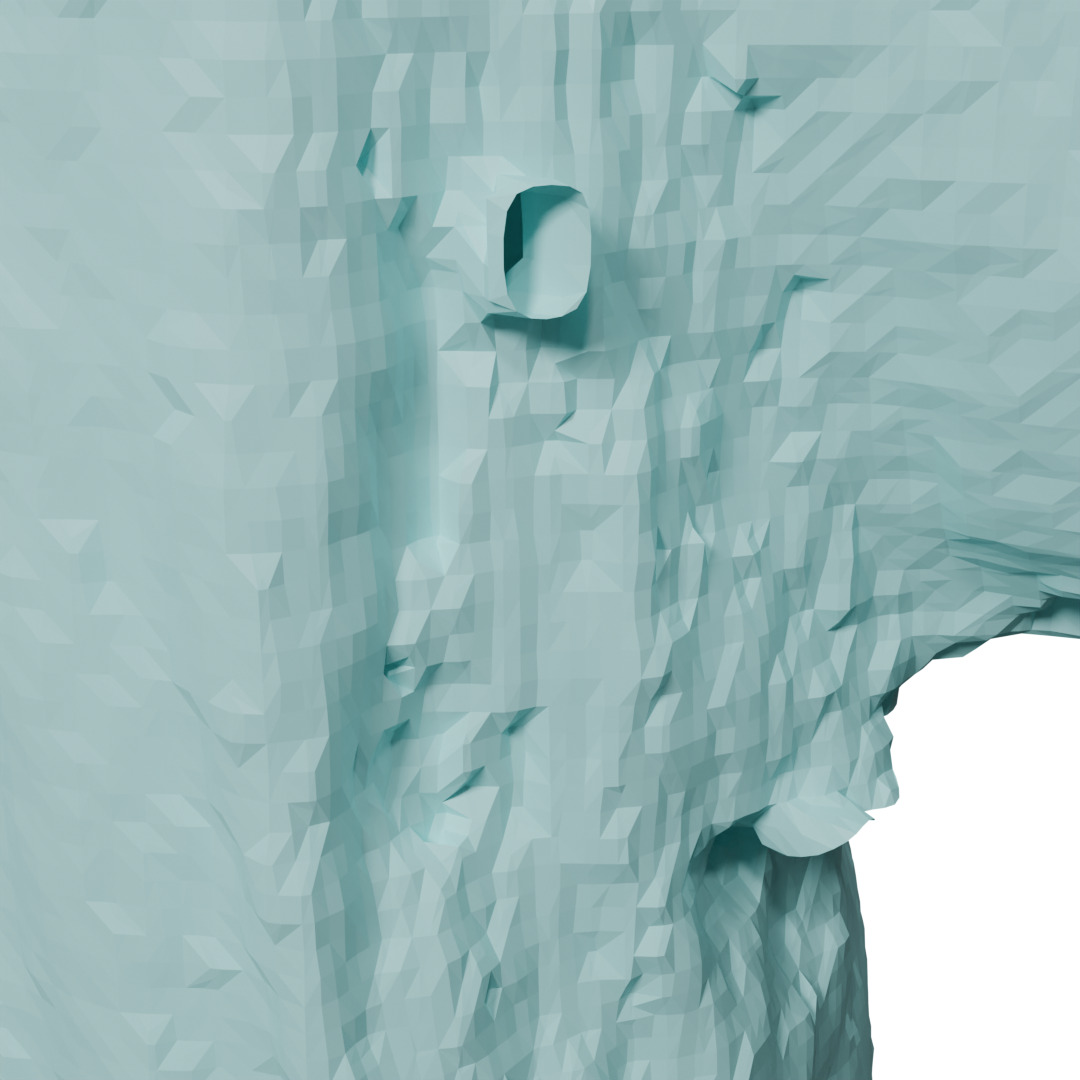 }
    \includegraphics[width=1.2in]{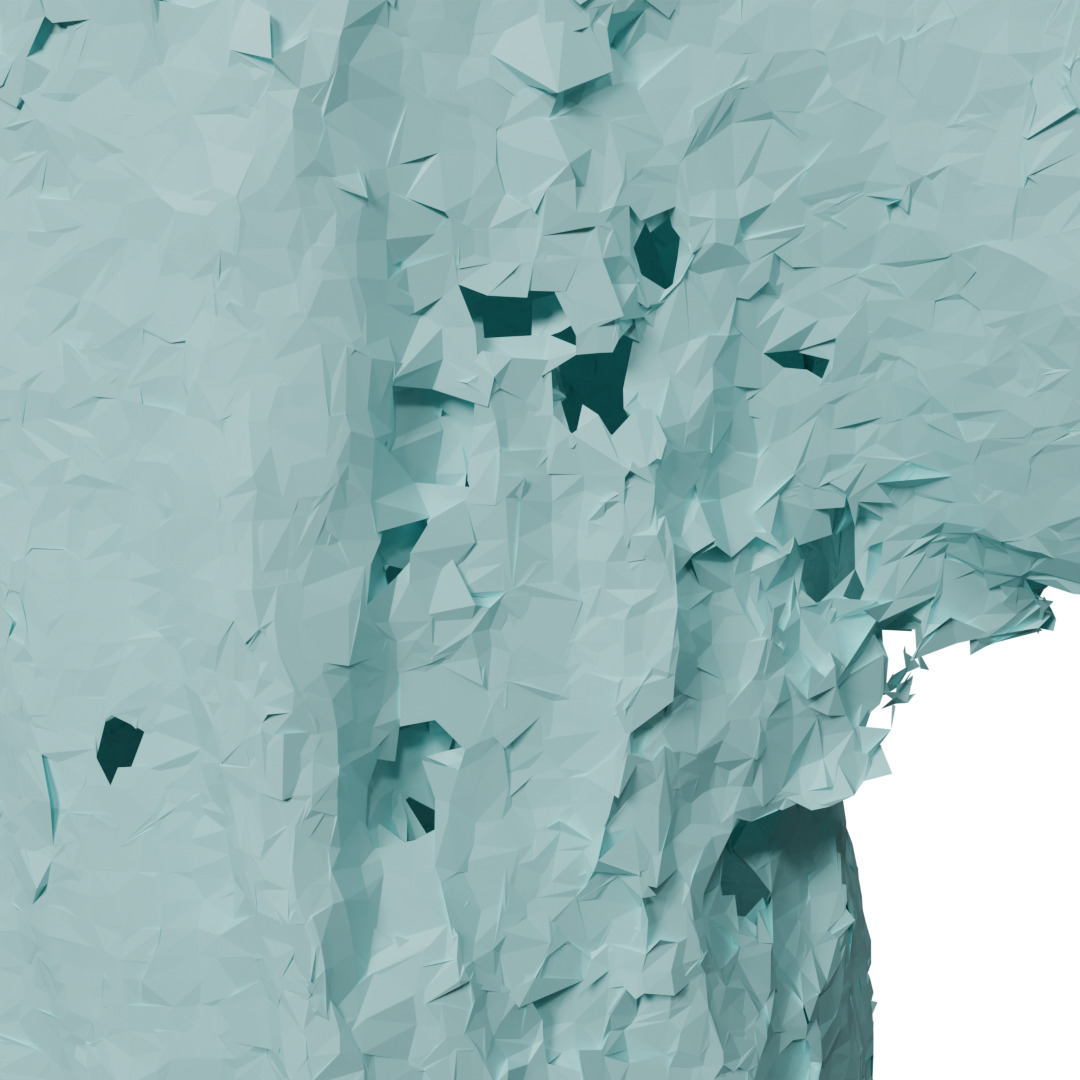 }
    \includegraphics[width=1.2in]{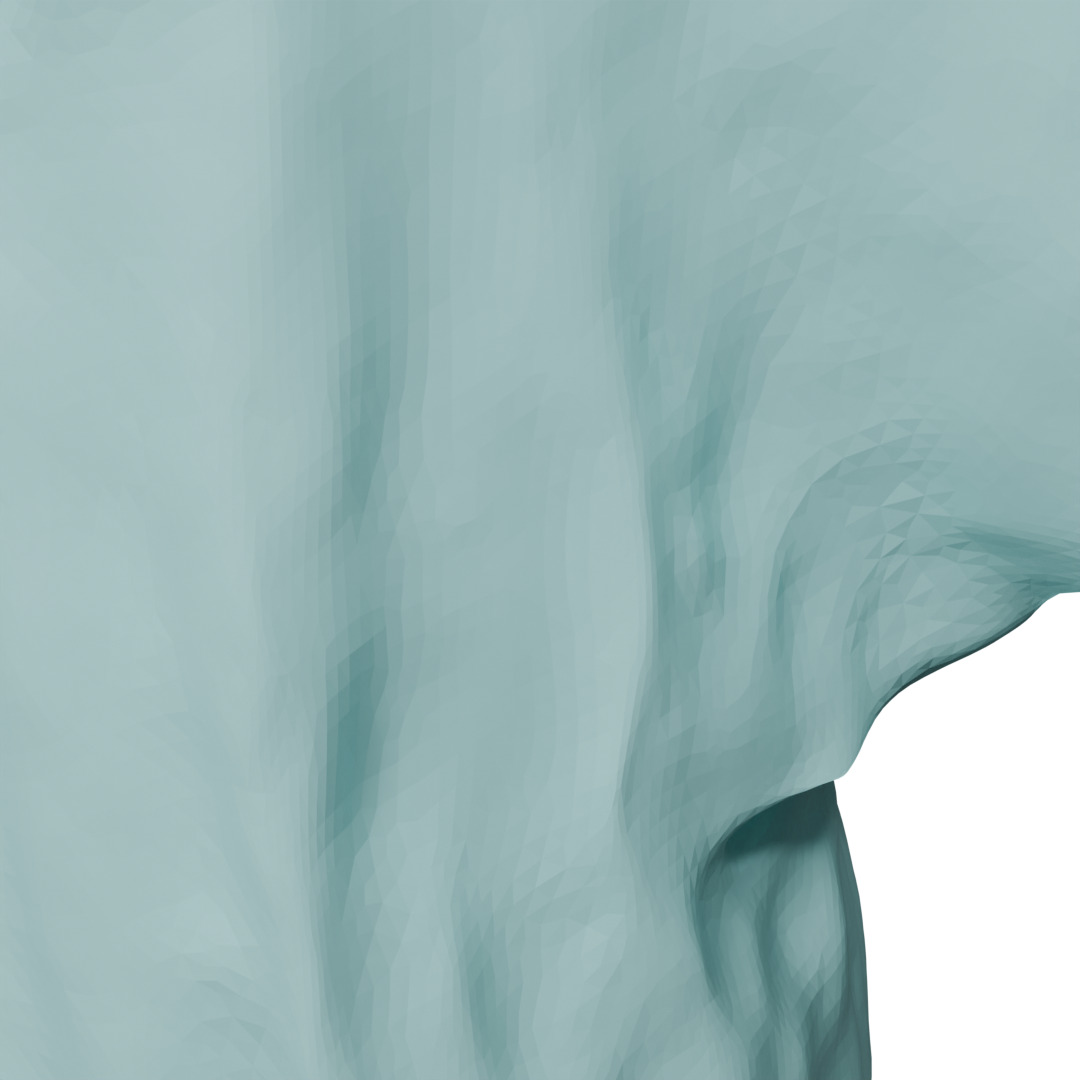 }
    \includegraphics[width=1.2in]{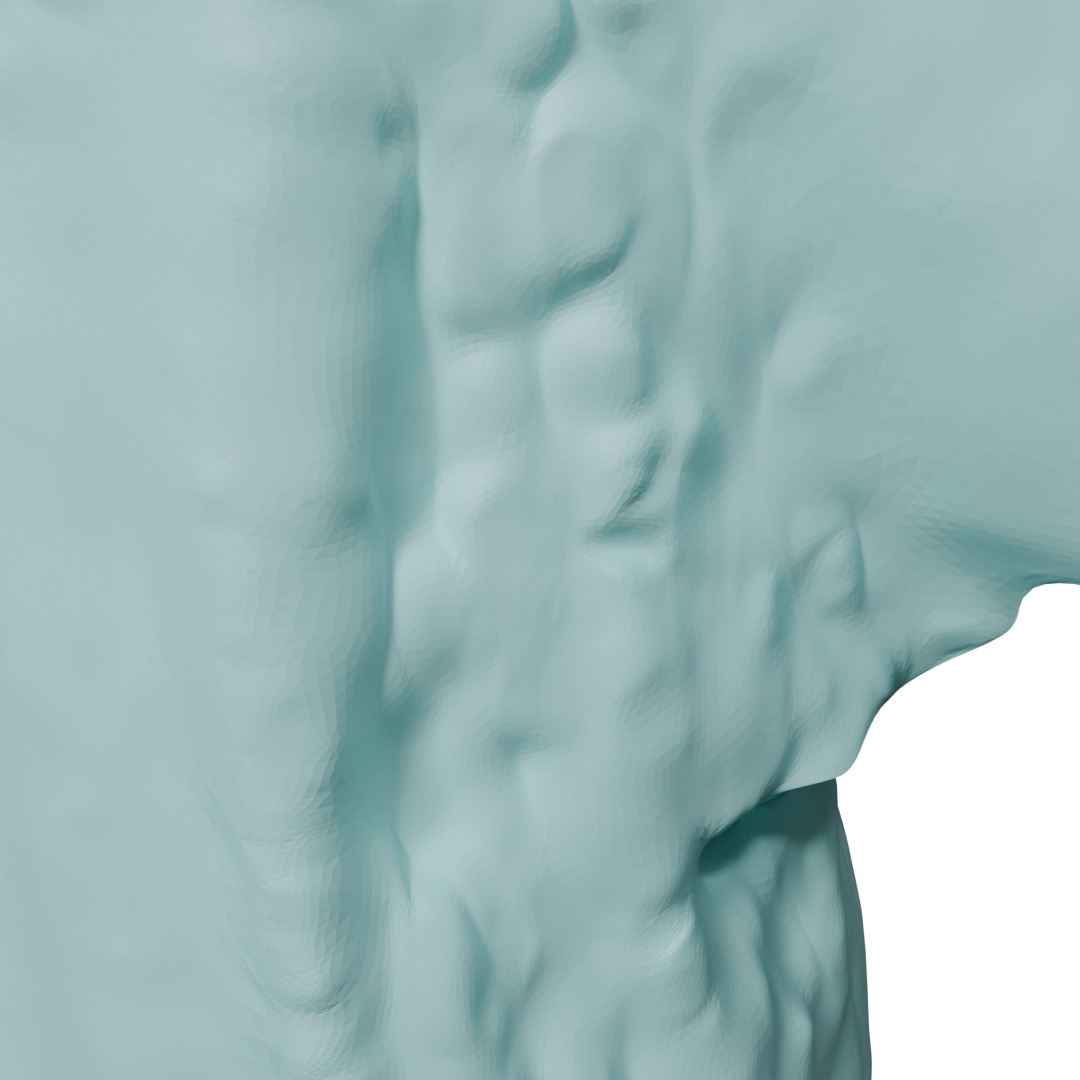 }\\
    \includegraphics[width=1.2in]{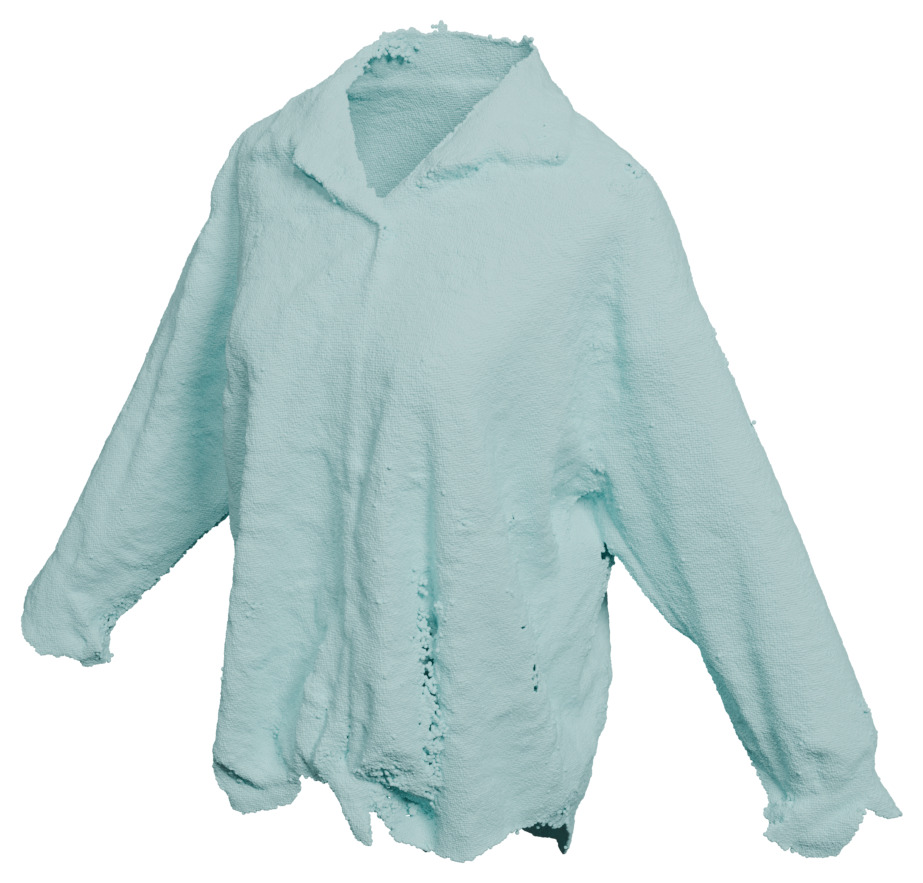 }
    \includegraphics[width=1.2in]{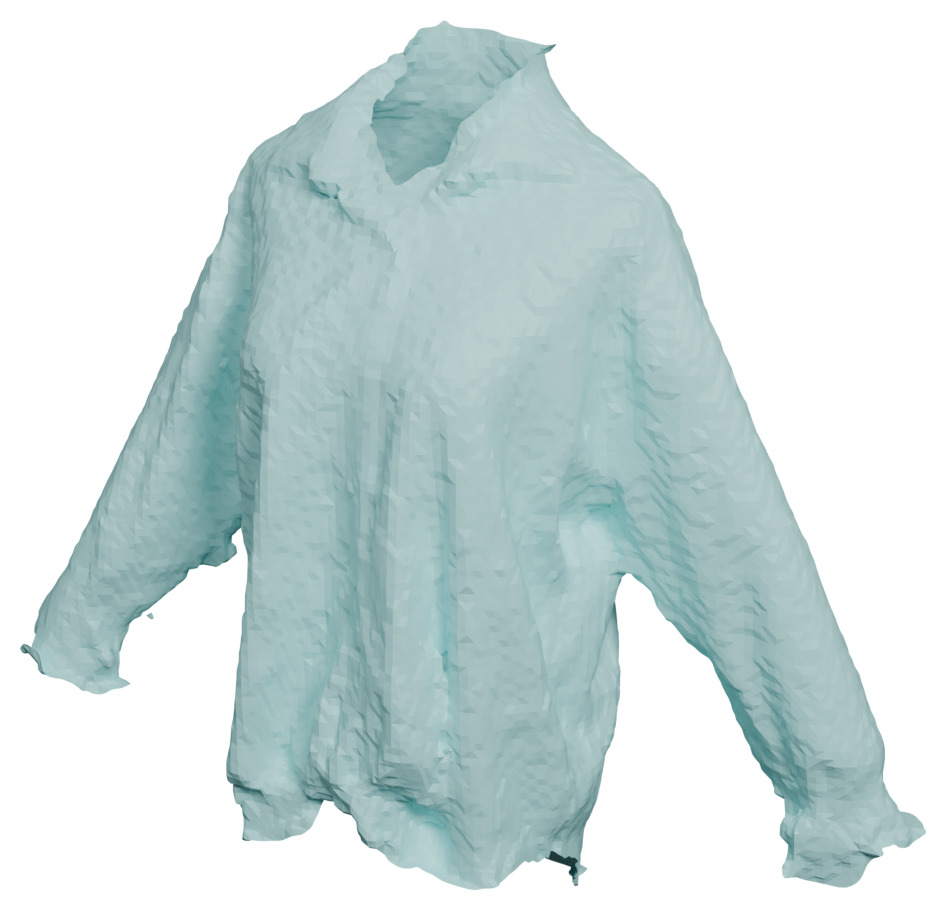 }
    \includegraphics[width=1.2in]{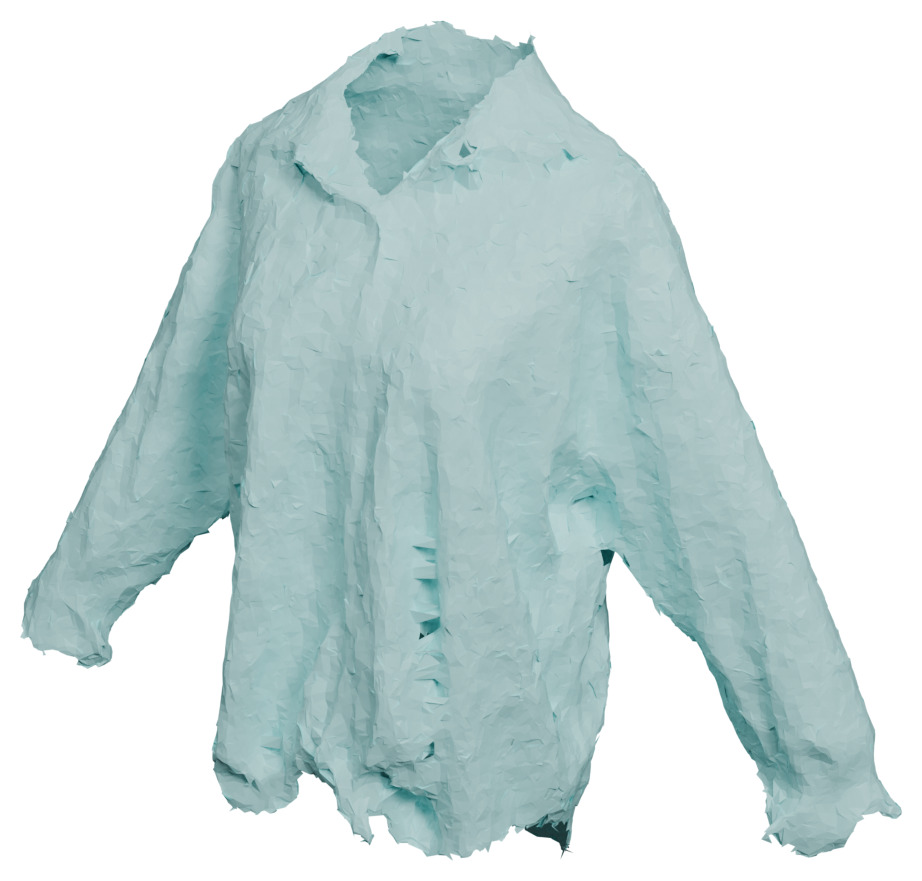 }
    \includegraphics[width=1.2in]{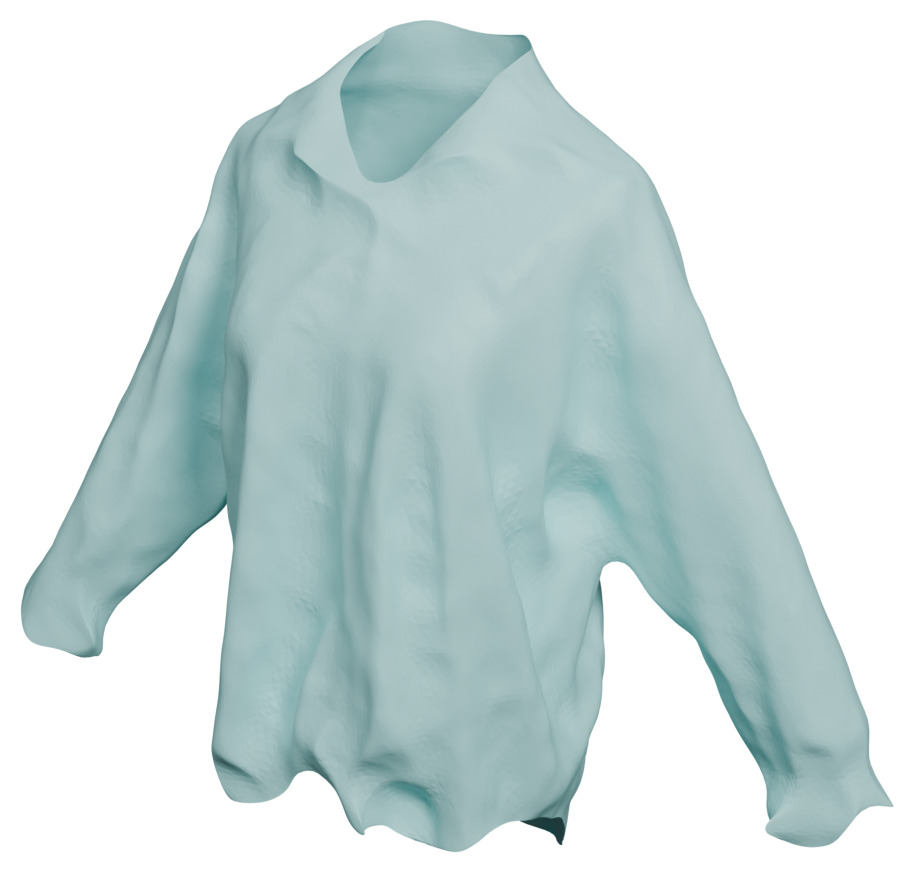 }
    \includegraphics[width=1.2in]{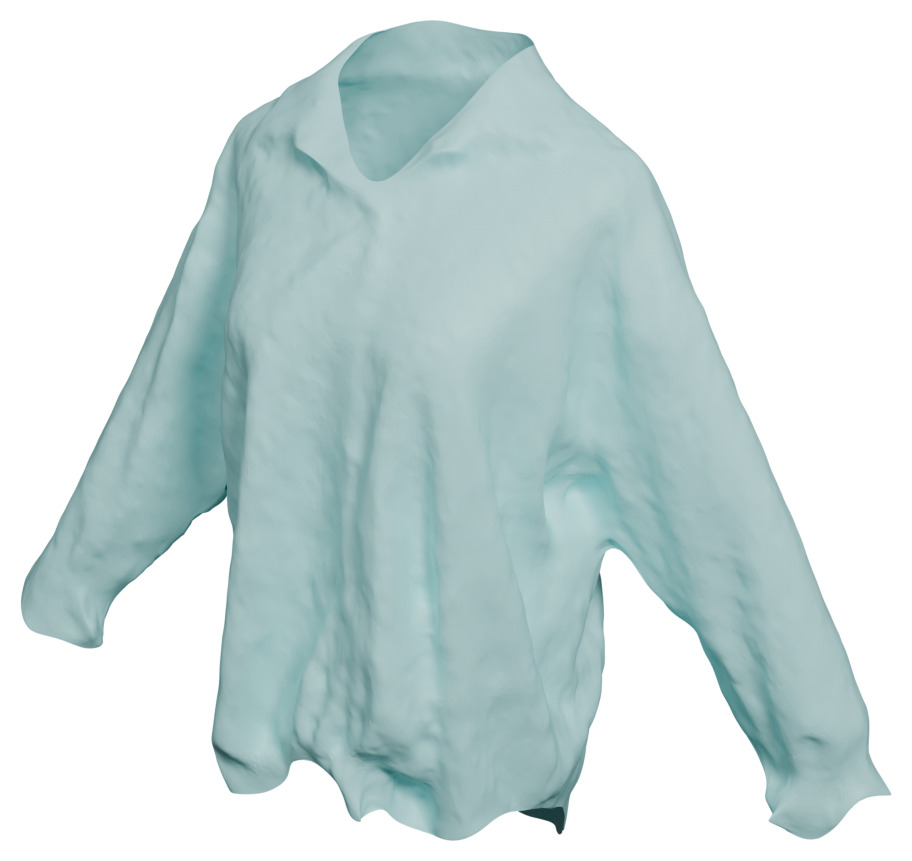 }\\
    \includegraphics[width=1.2in]{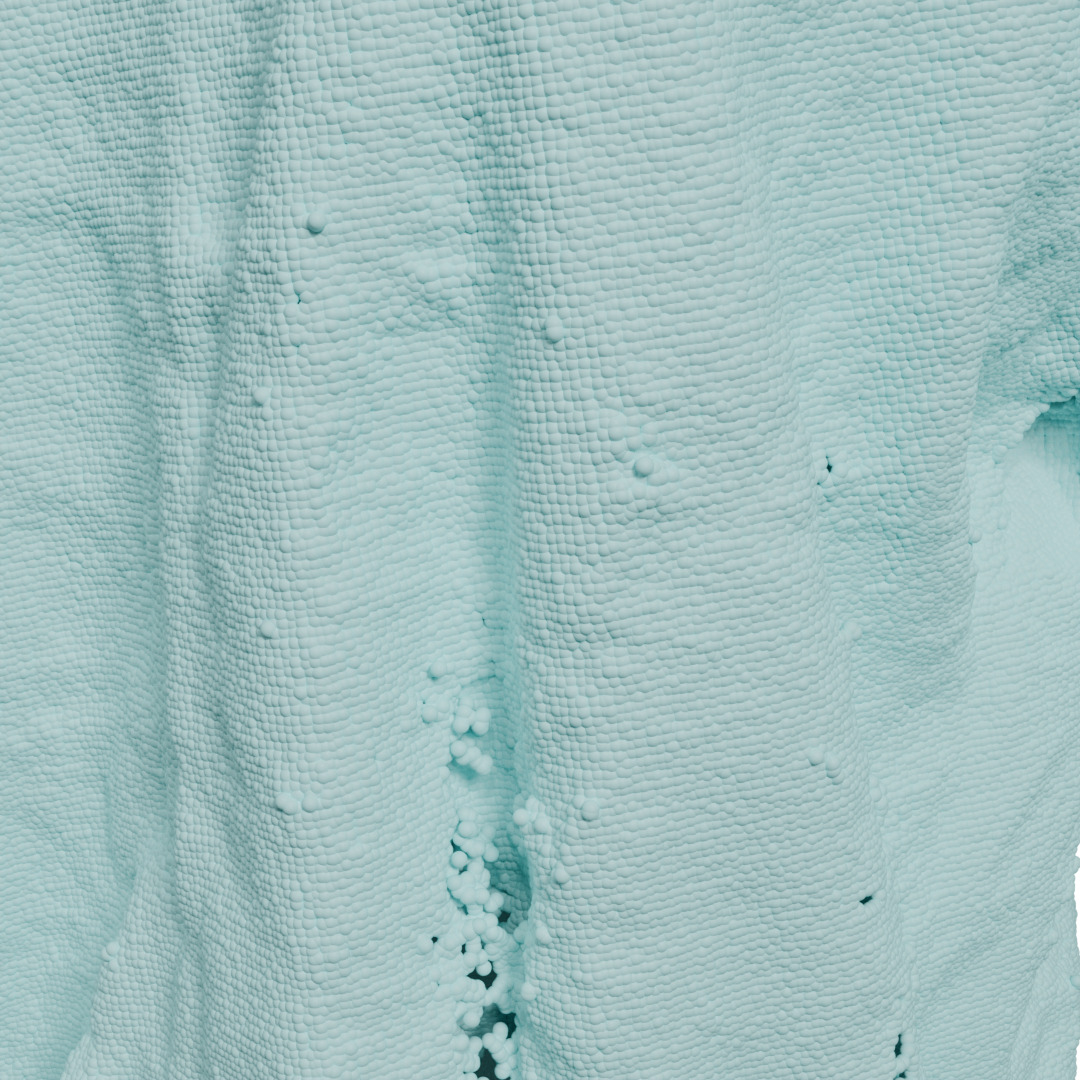 }
    \includegraphics[width=1.2in]{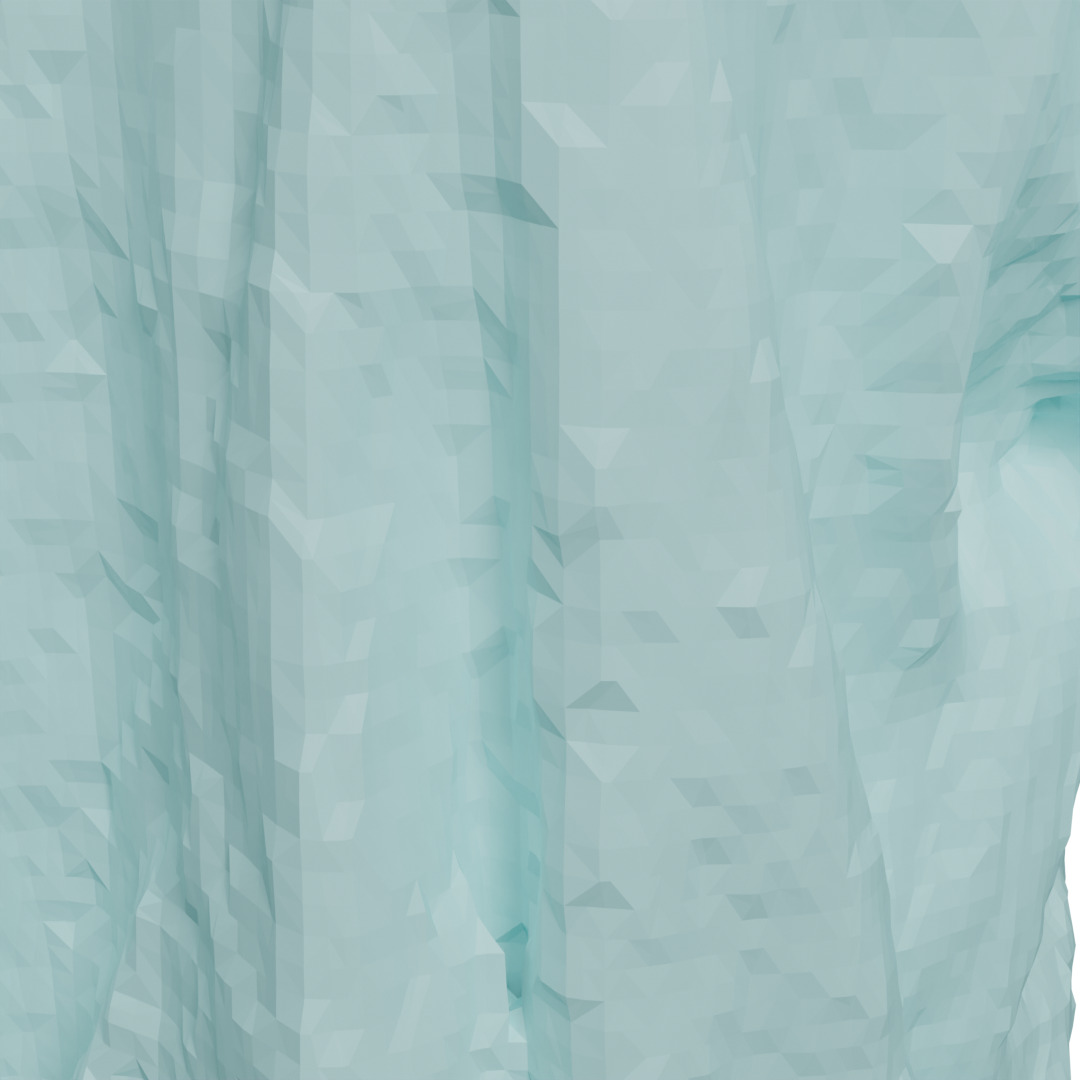 }
    \includegraphics[width=1.2in]{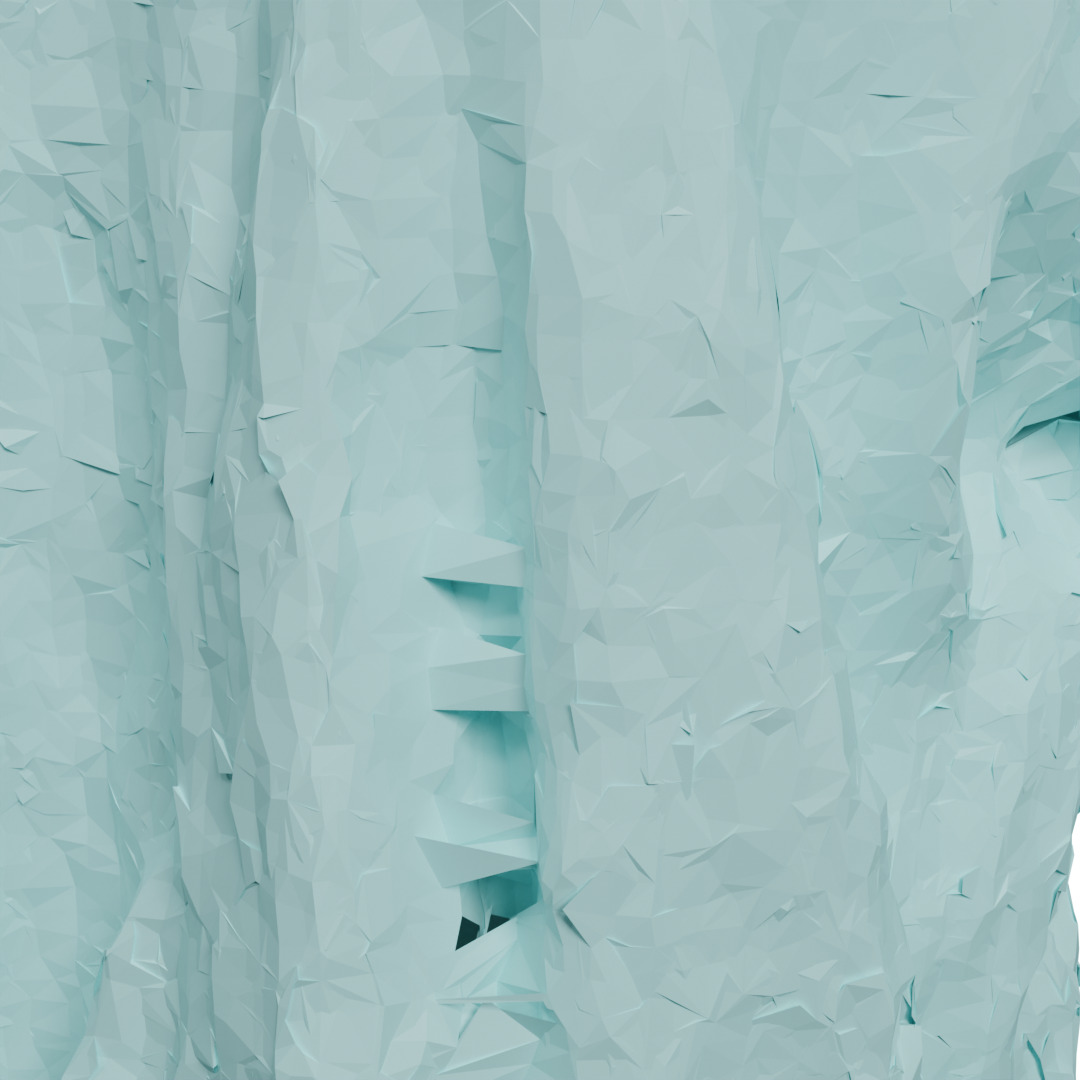 }
    \includegraphics[width=1.2in]{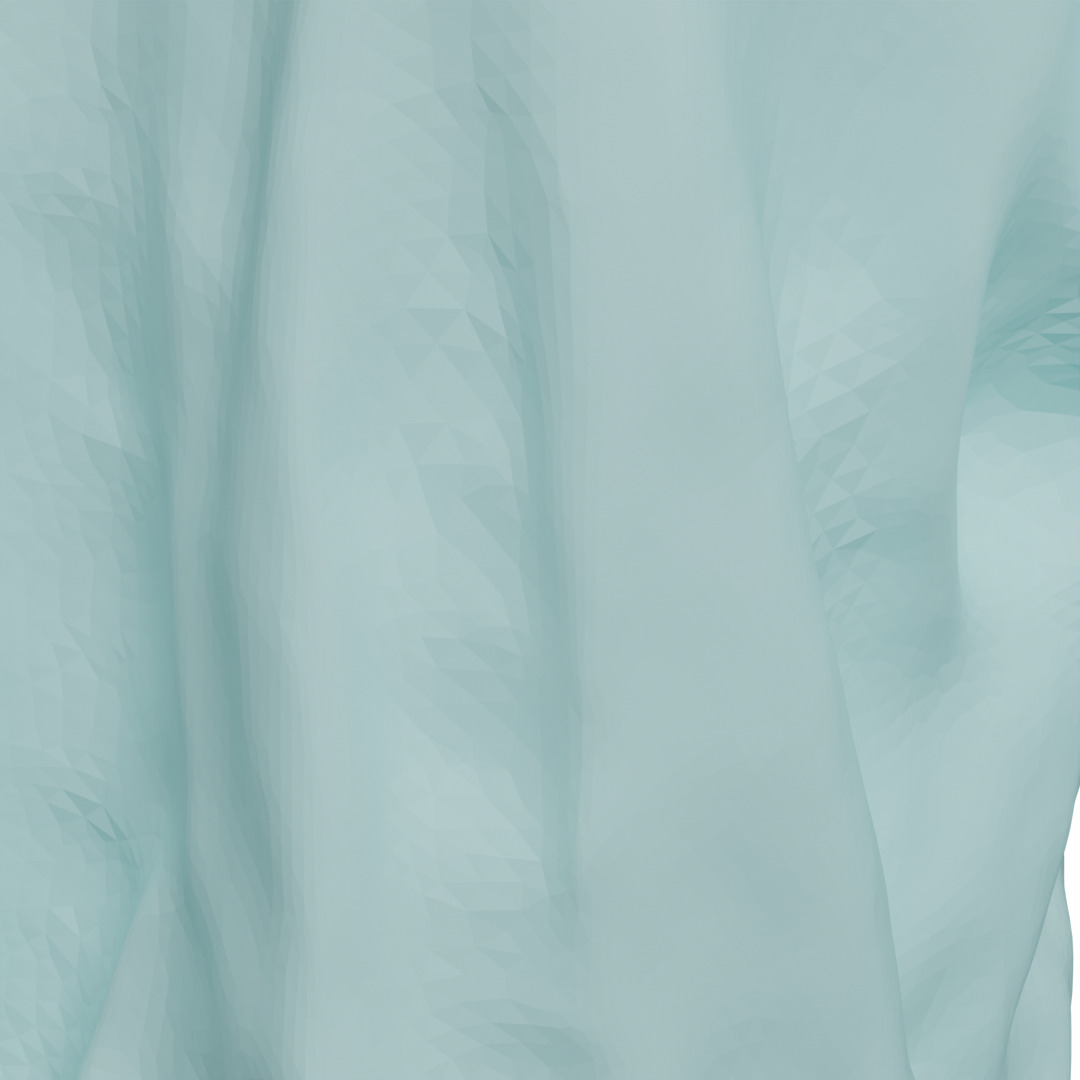 }
    \includegraphics[width=1.2in]{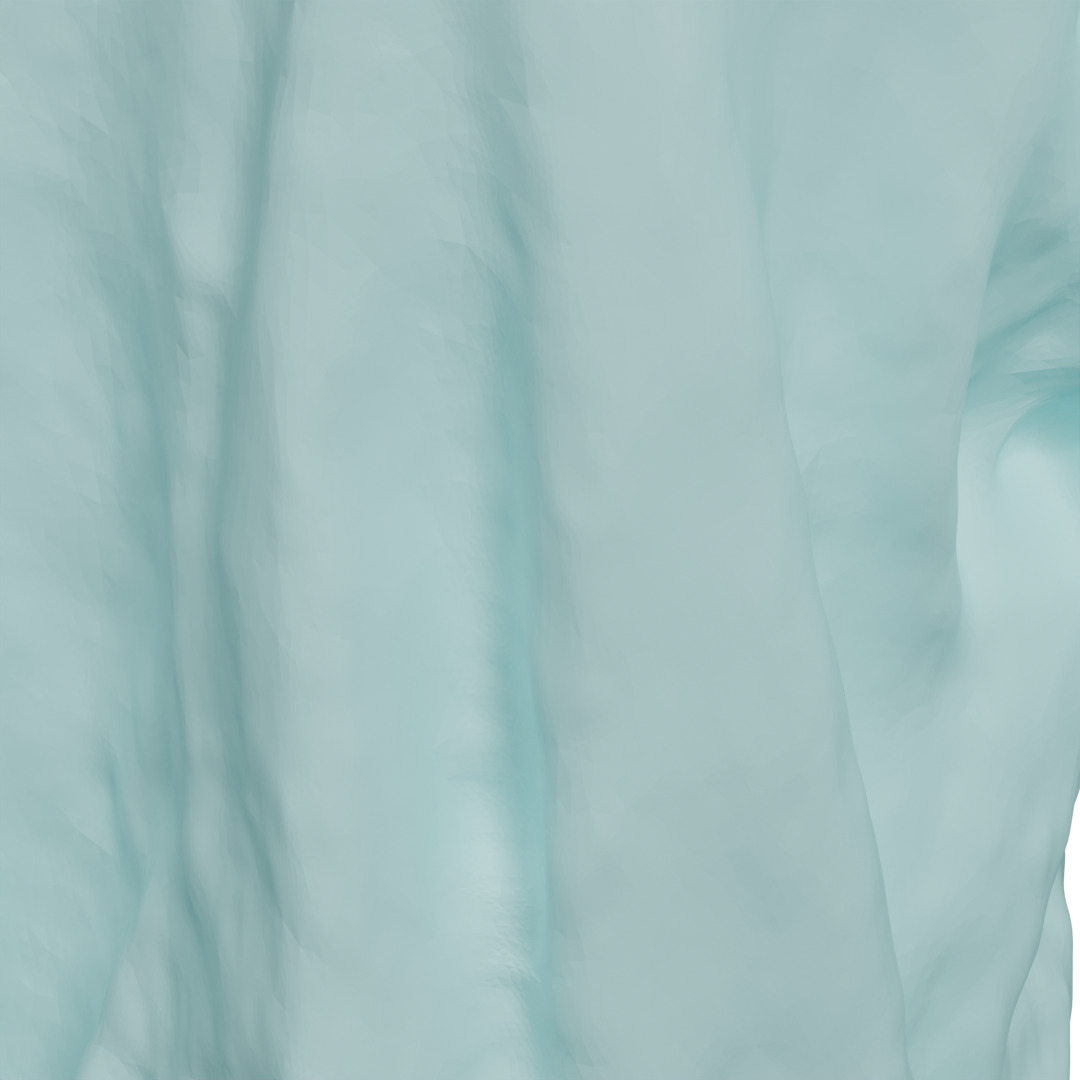 }\\
  
    \includegraphics[width=1.2in]{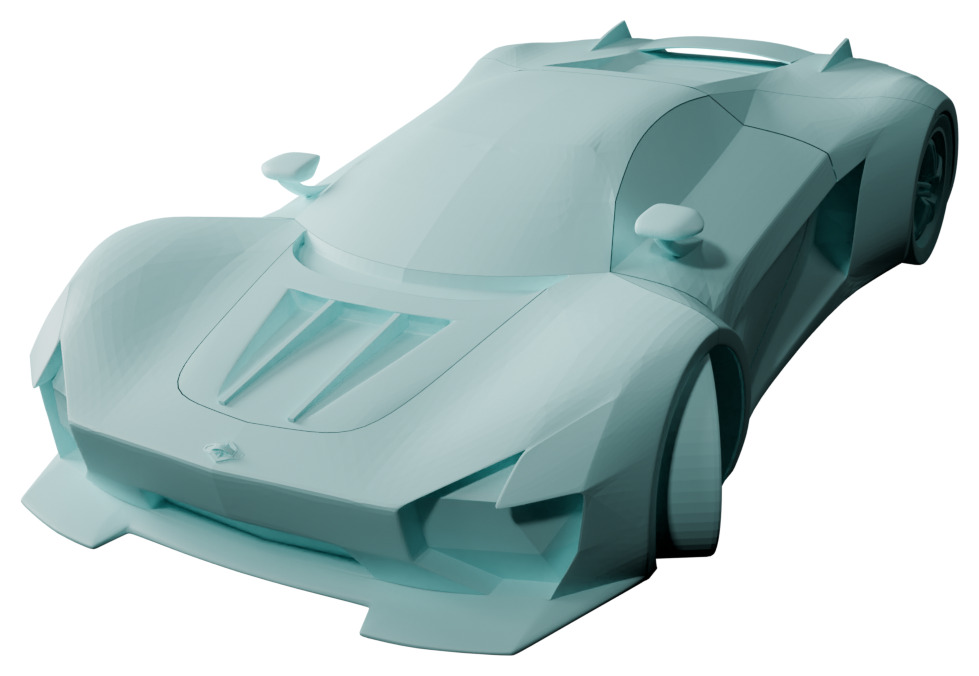 }
    \includegraphics[width=1.2in]
    {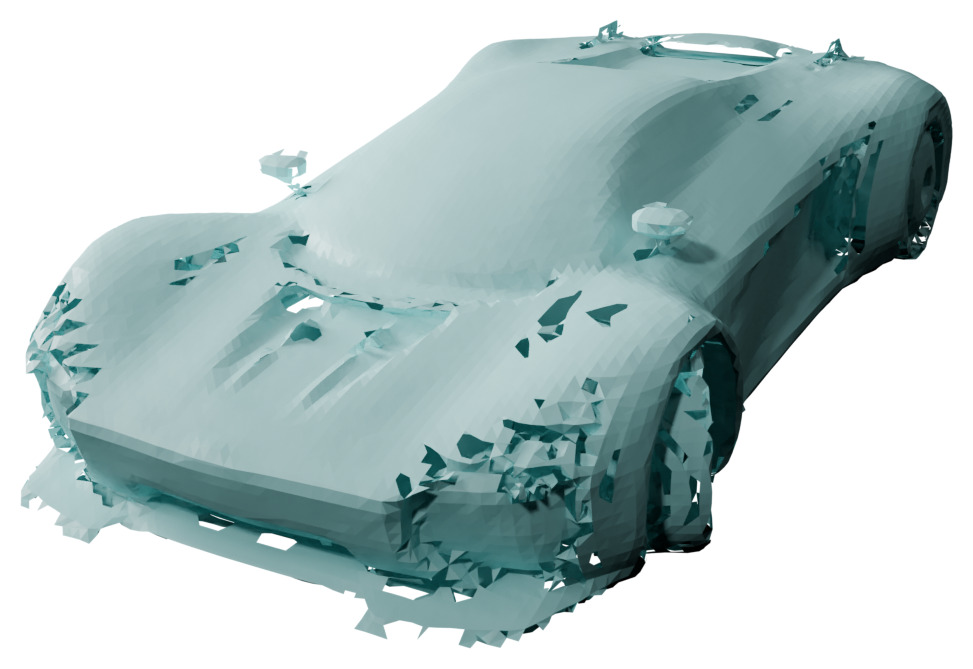 }
    \includegraphics[width=1.2in]{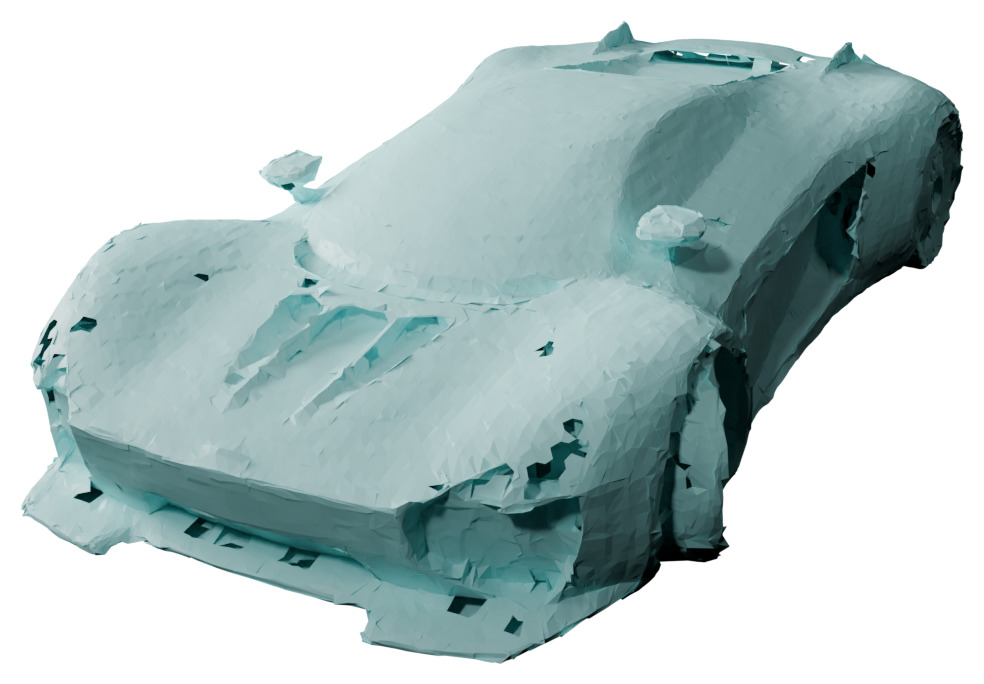}
    \includegraphics[width=1.2in]{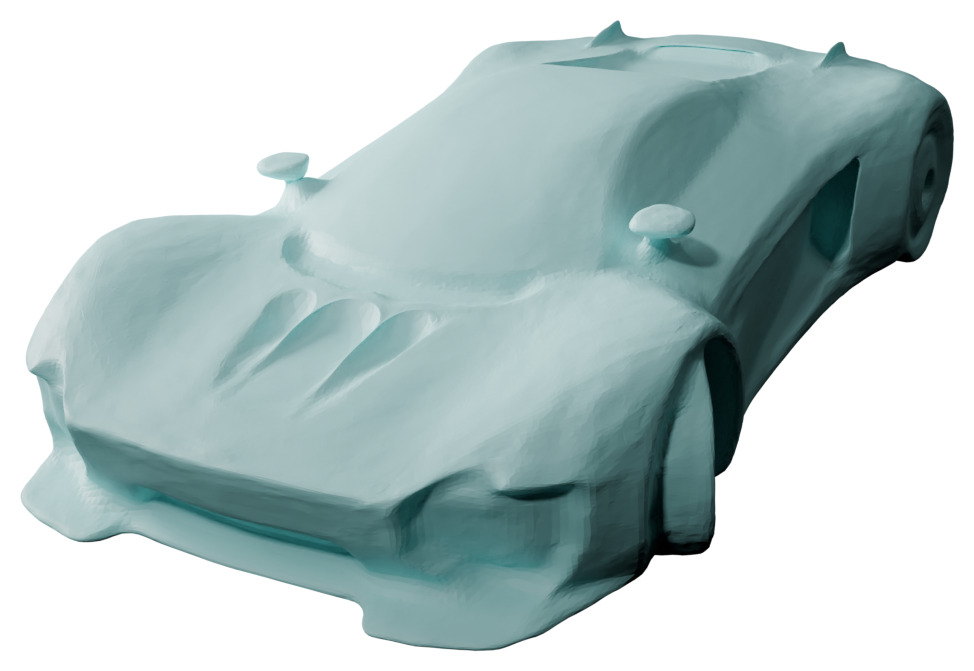}
    \includegraphics[width=1.2in]{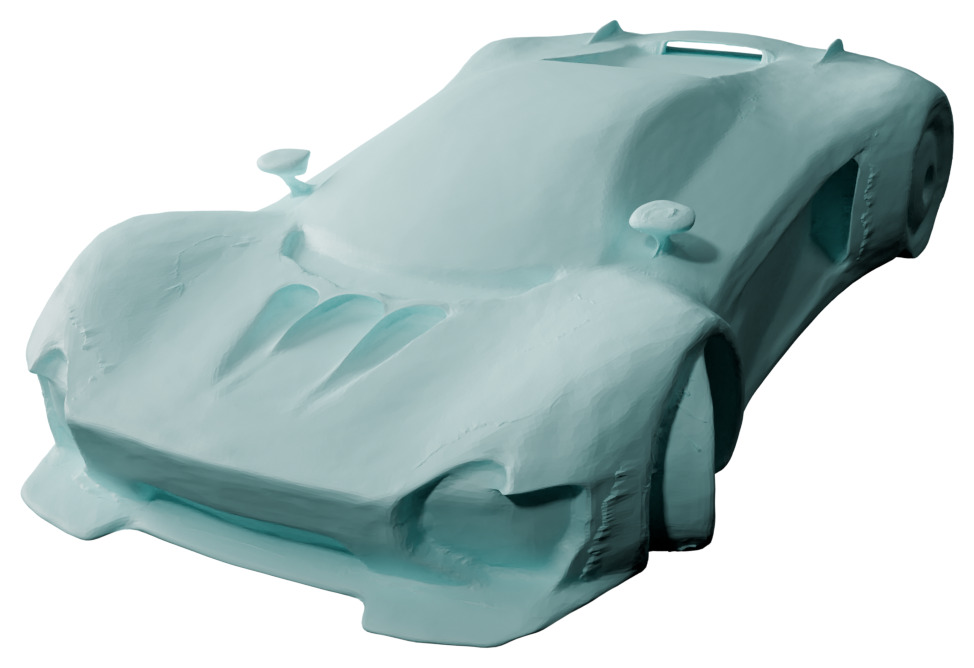 }\\
    \includegraphics[width=1.2in]{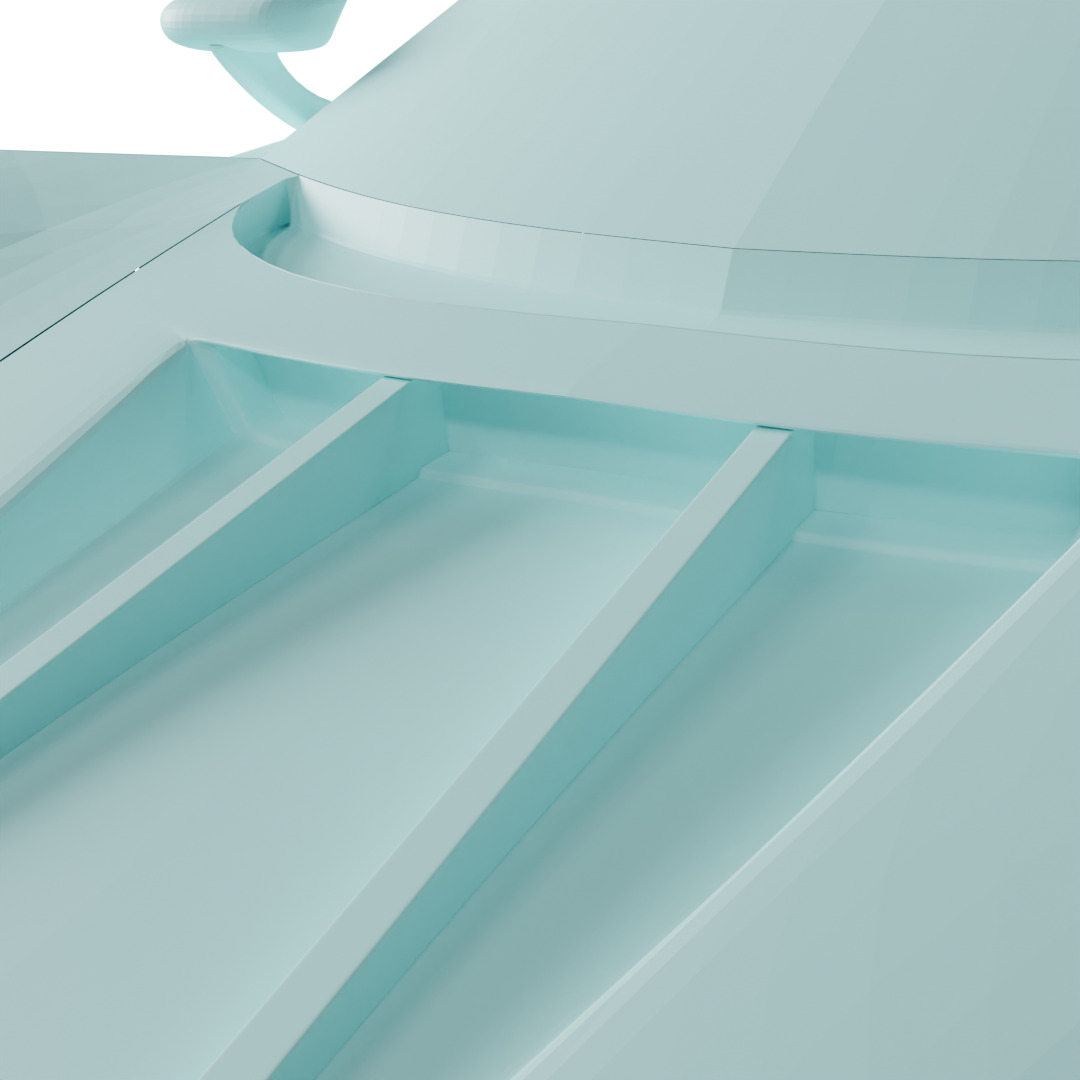 }
    \includegraphics[width=1.2in]
    {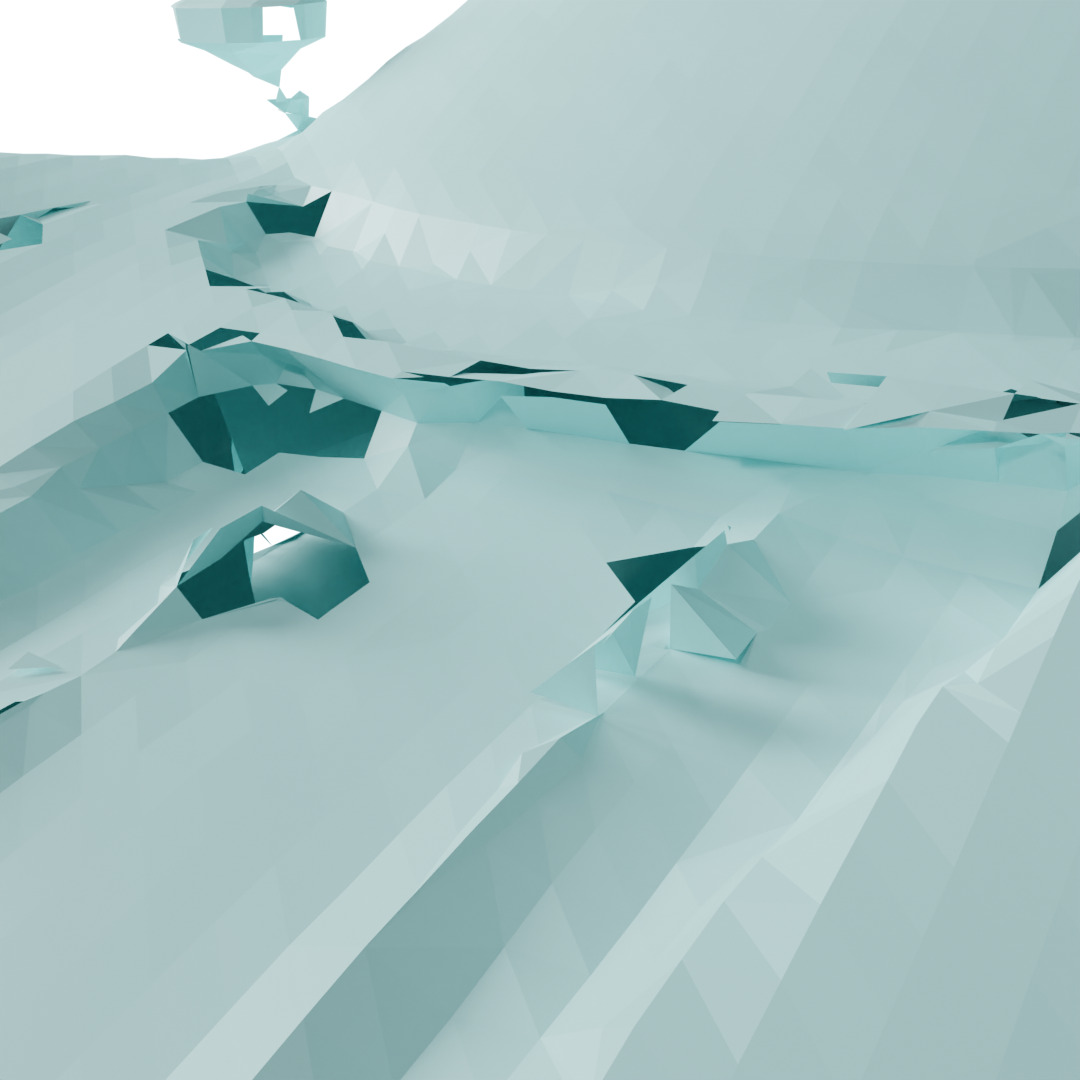}
    \includegraphics[width=1.2in]{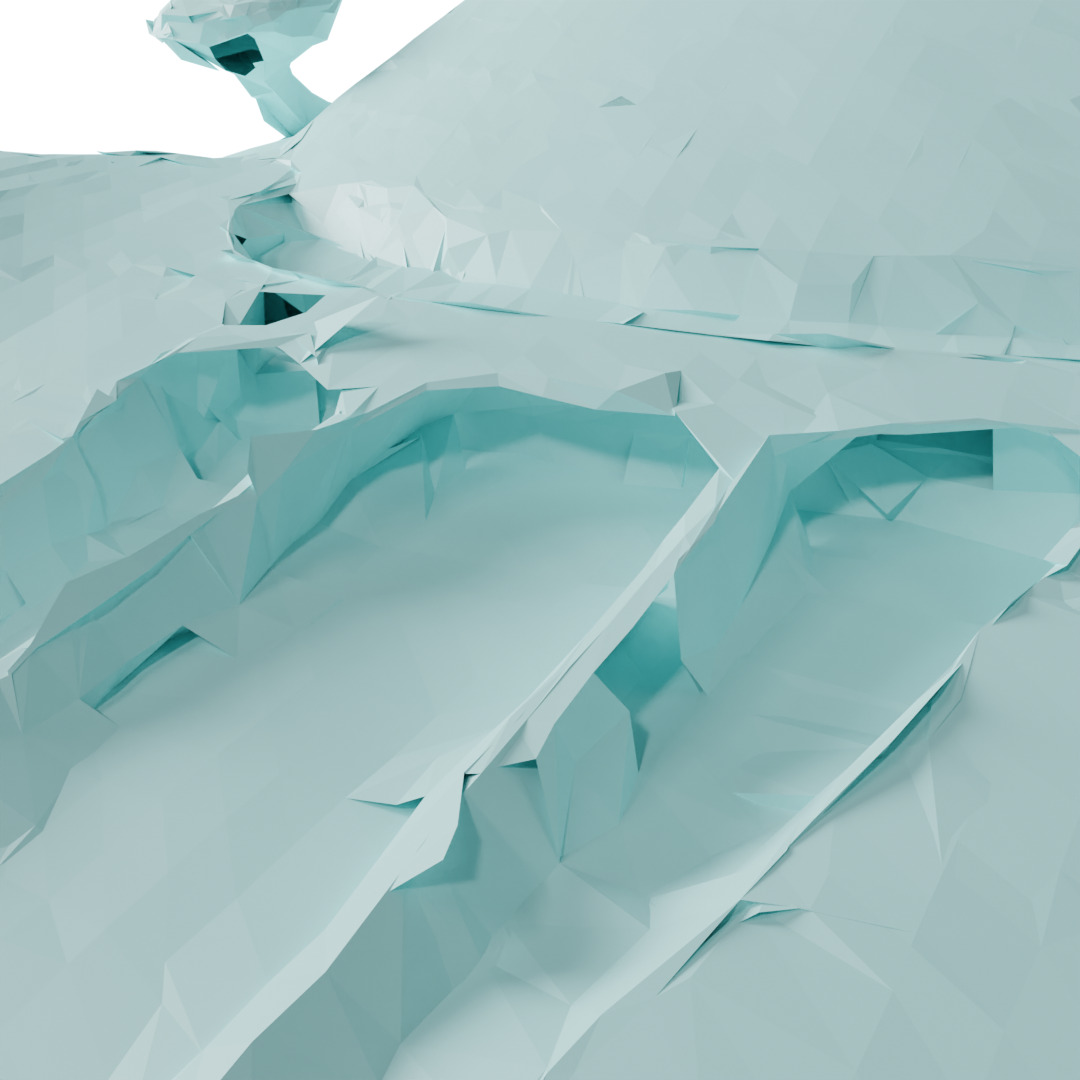}
    \includegraphics[width=1.2in]{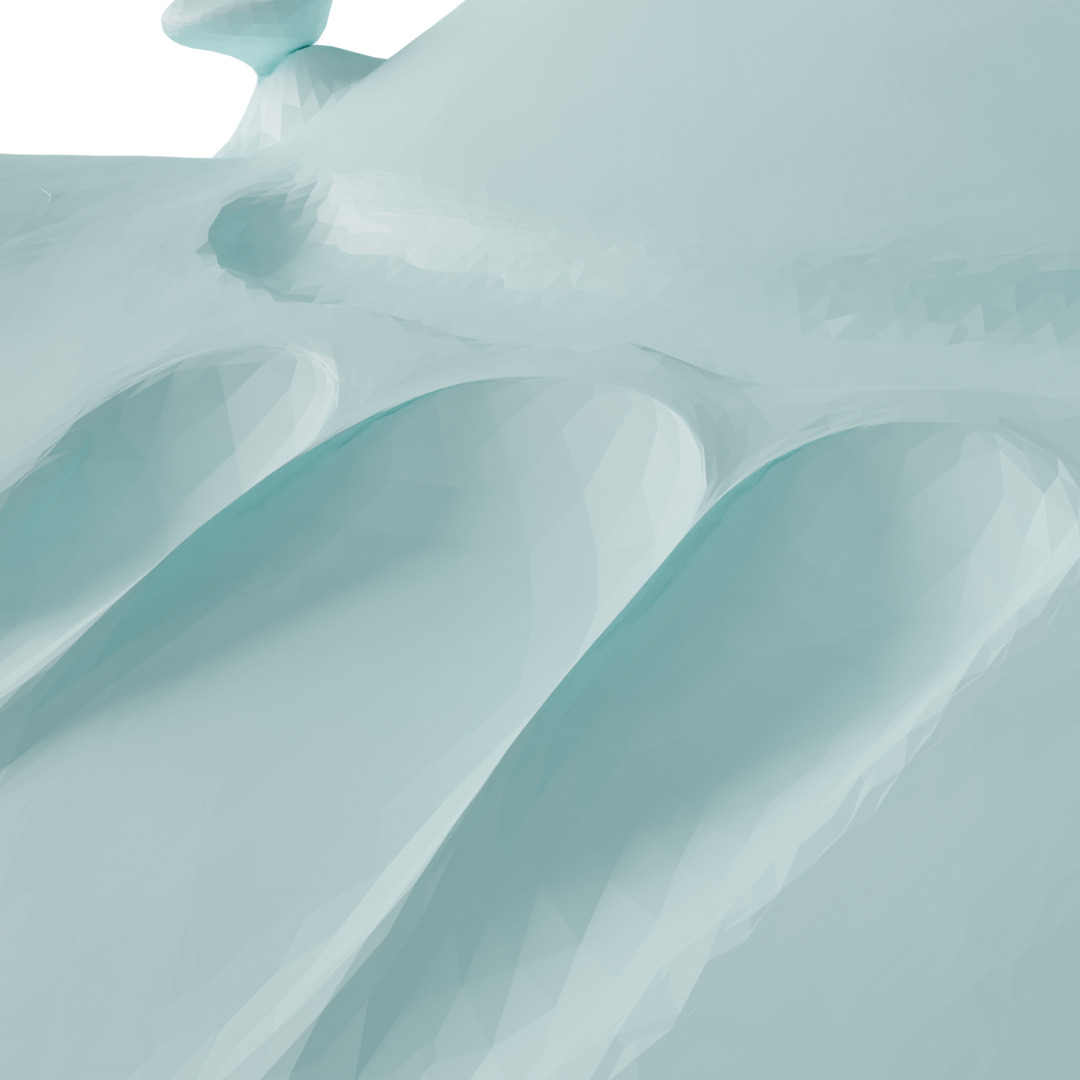}
    \includegraphics[width=1.2in]{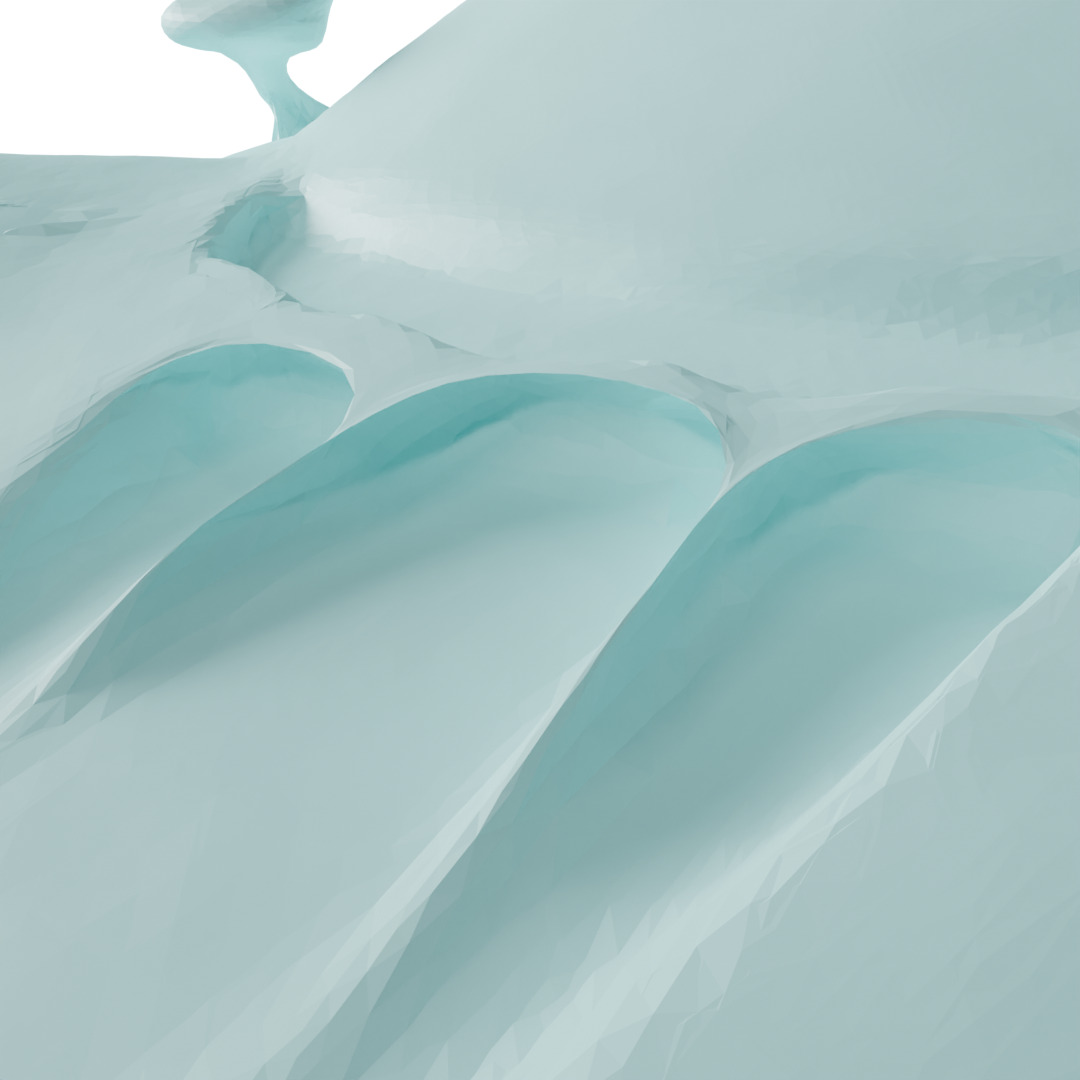}\\
    \includegraphics[width=1.2in]{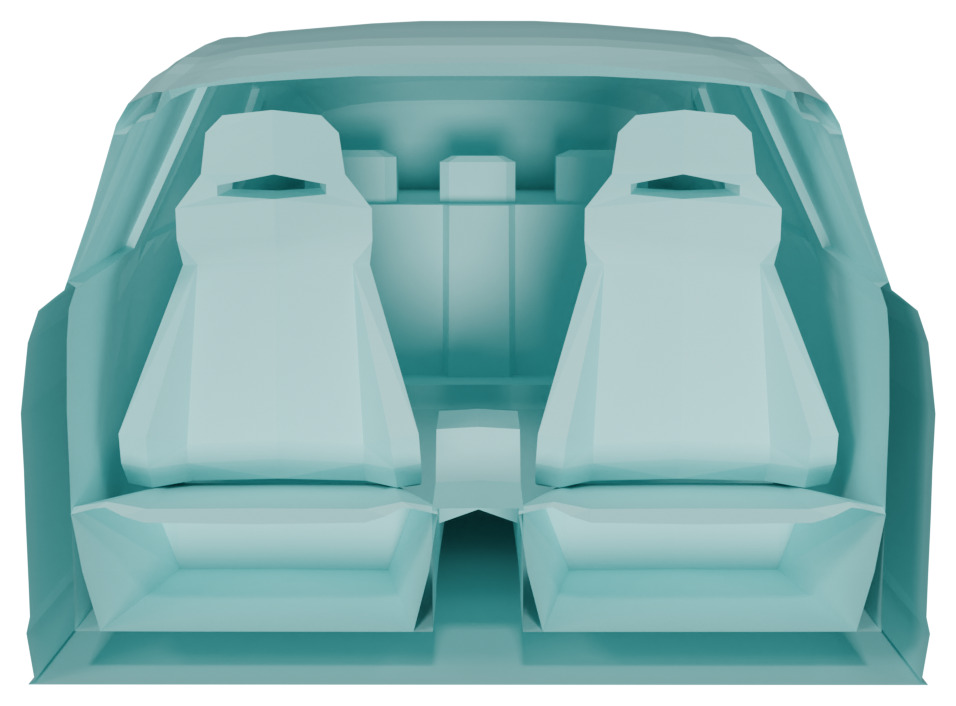}
    \includegraphics[width=1.2in]
    {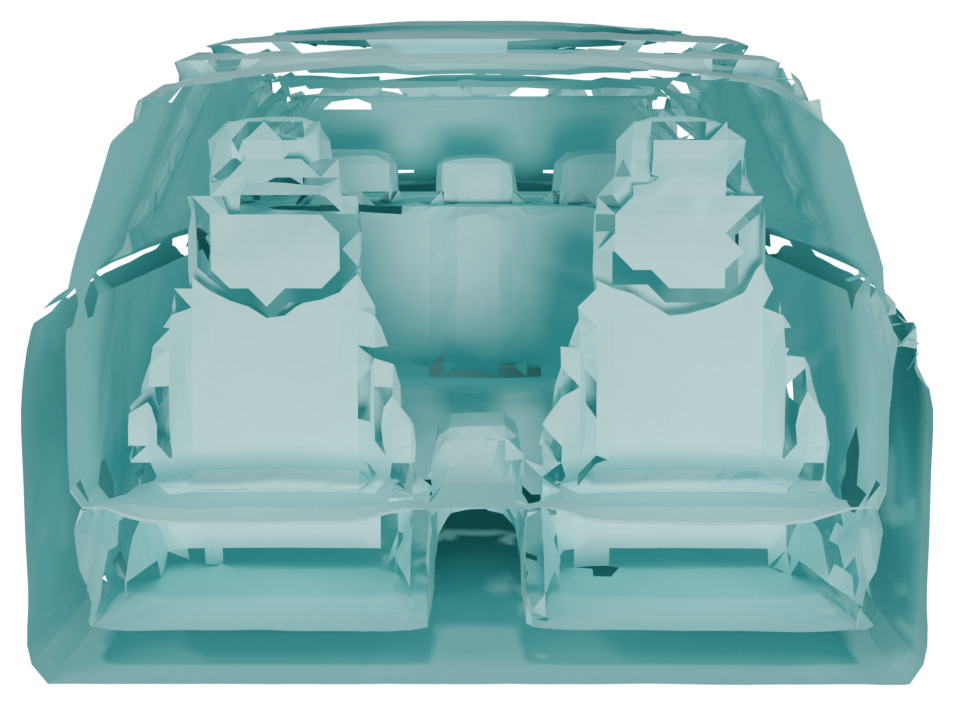}
    \includegraphics[width=1.2in]{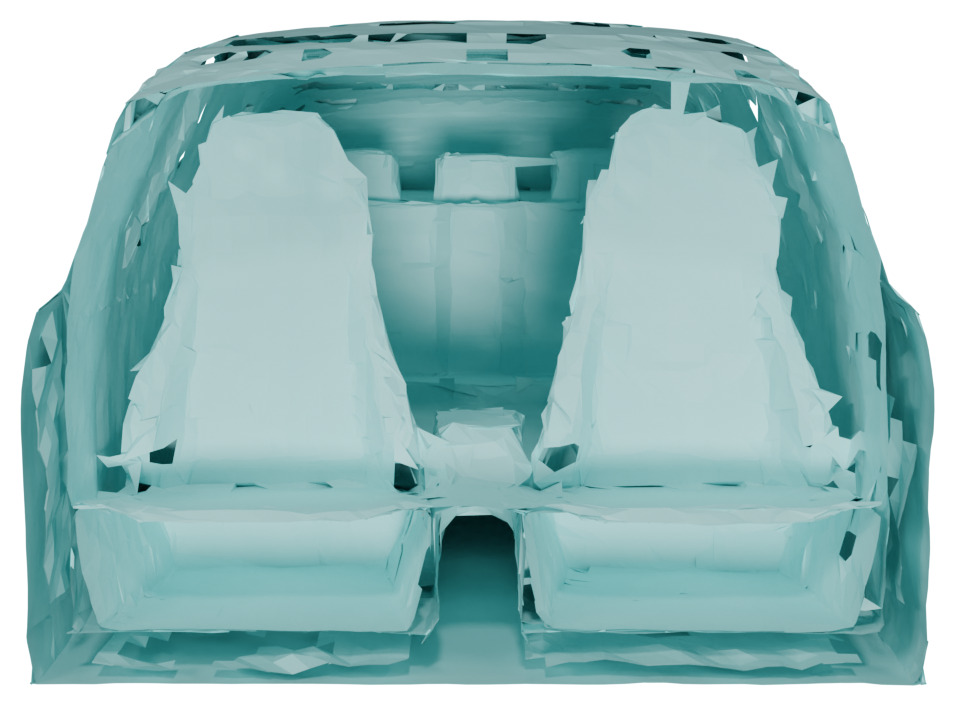}
    \includegraphics[width=1.2in]{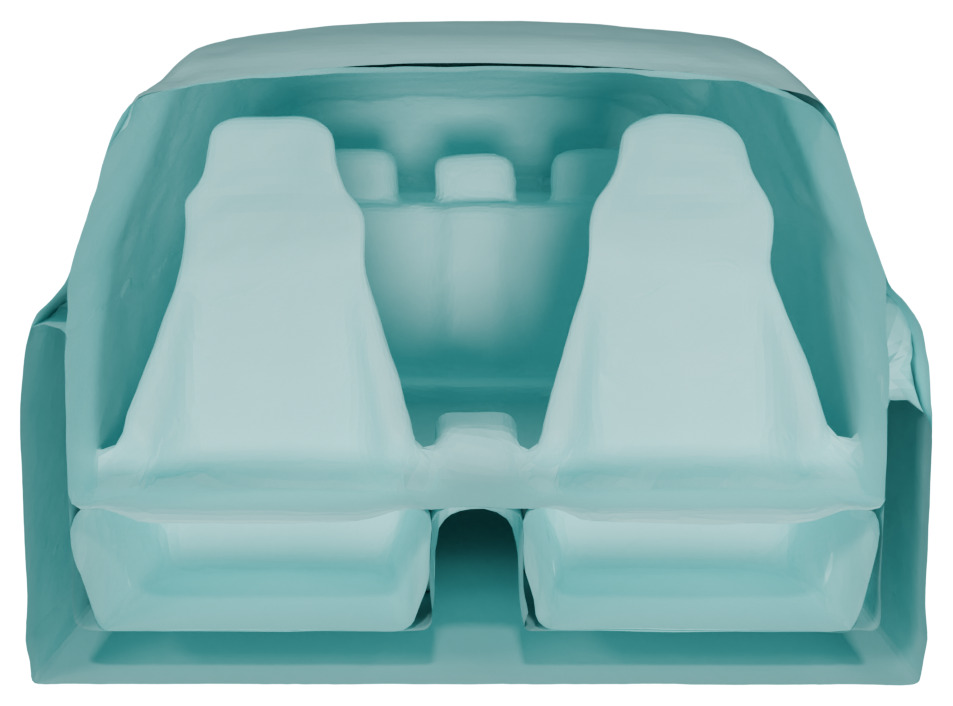}
    \includegraphics[width=1.2in]{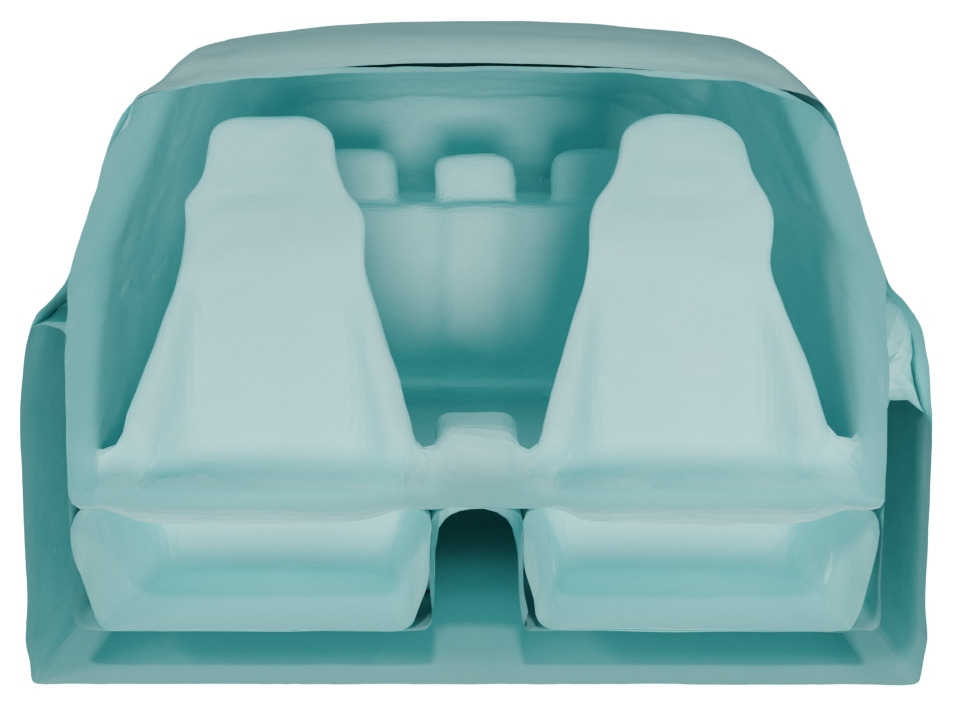}\\
    \includegraphics[width=1.2in]{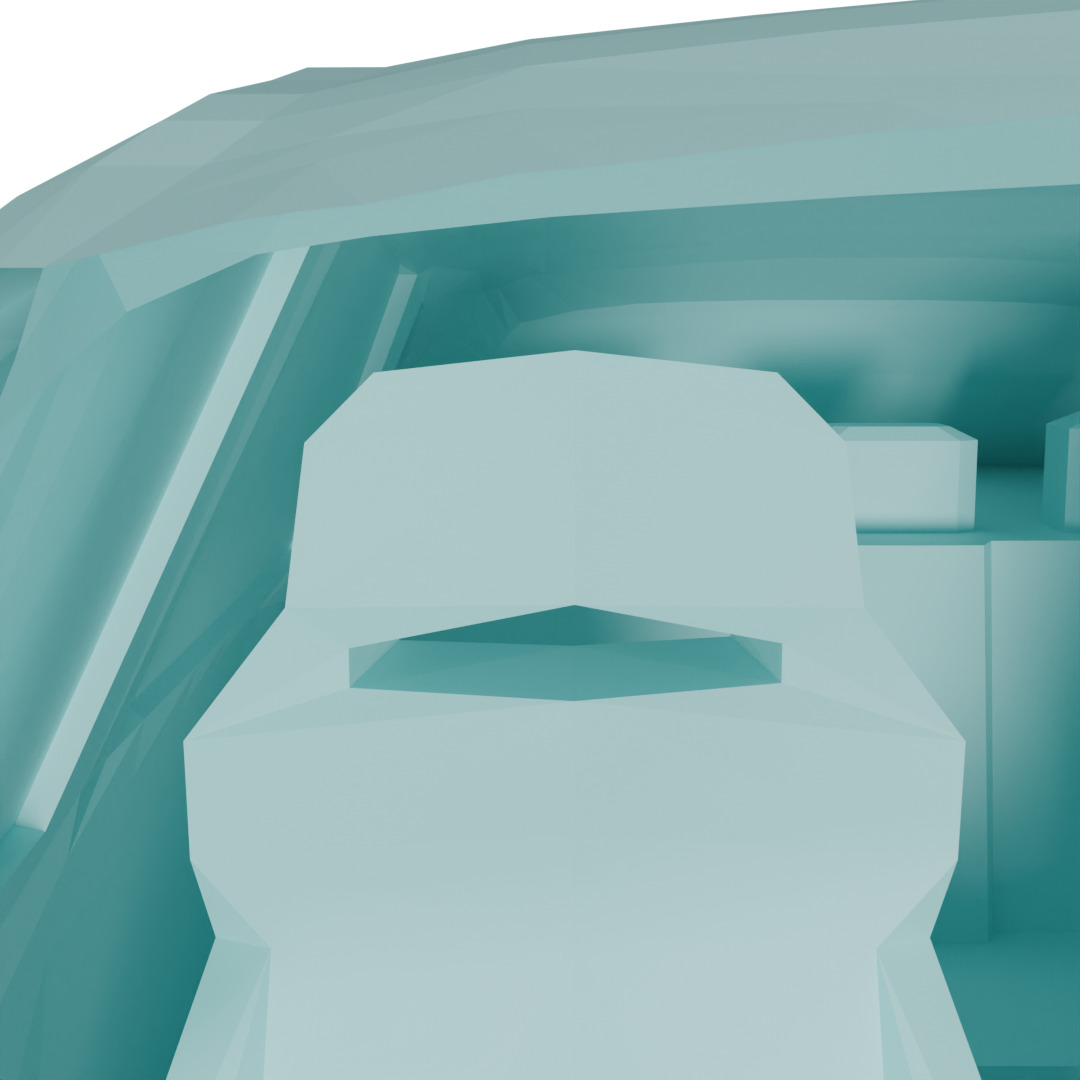}
    \includegraphics[width=1.2in]
    {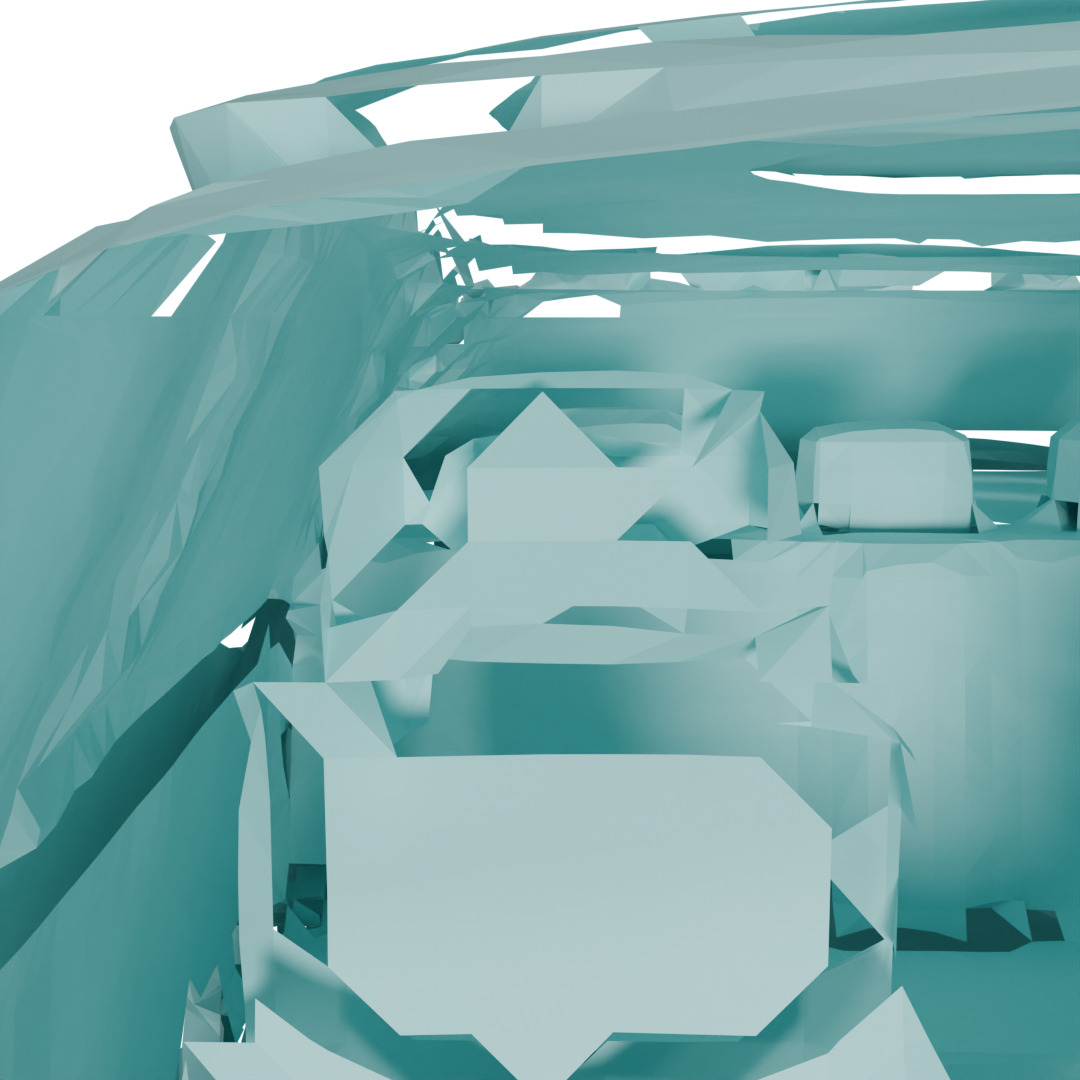}
    \includegraphics[width=1.2in]{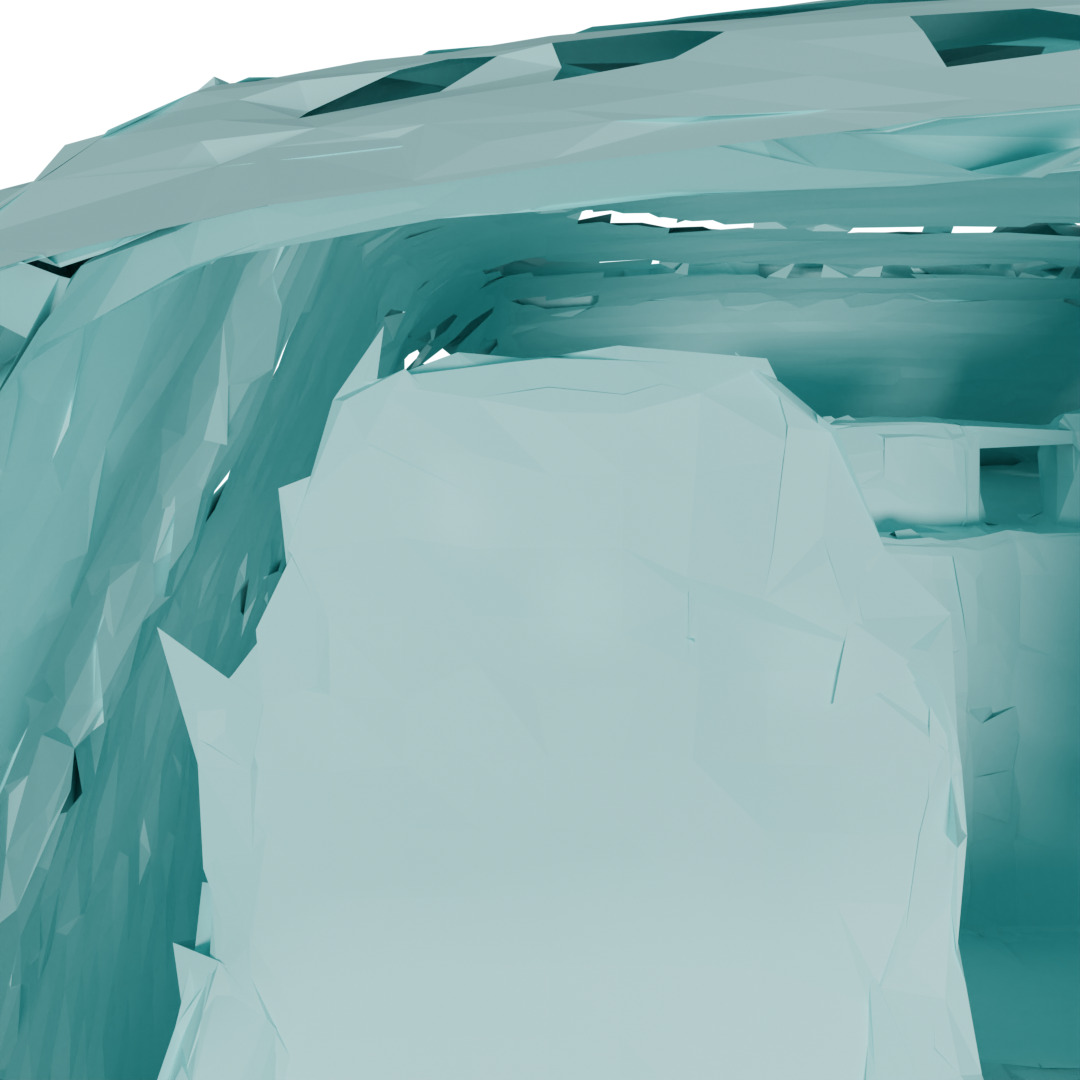}
    \includegraphics[width=1.2in]{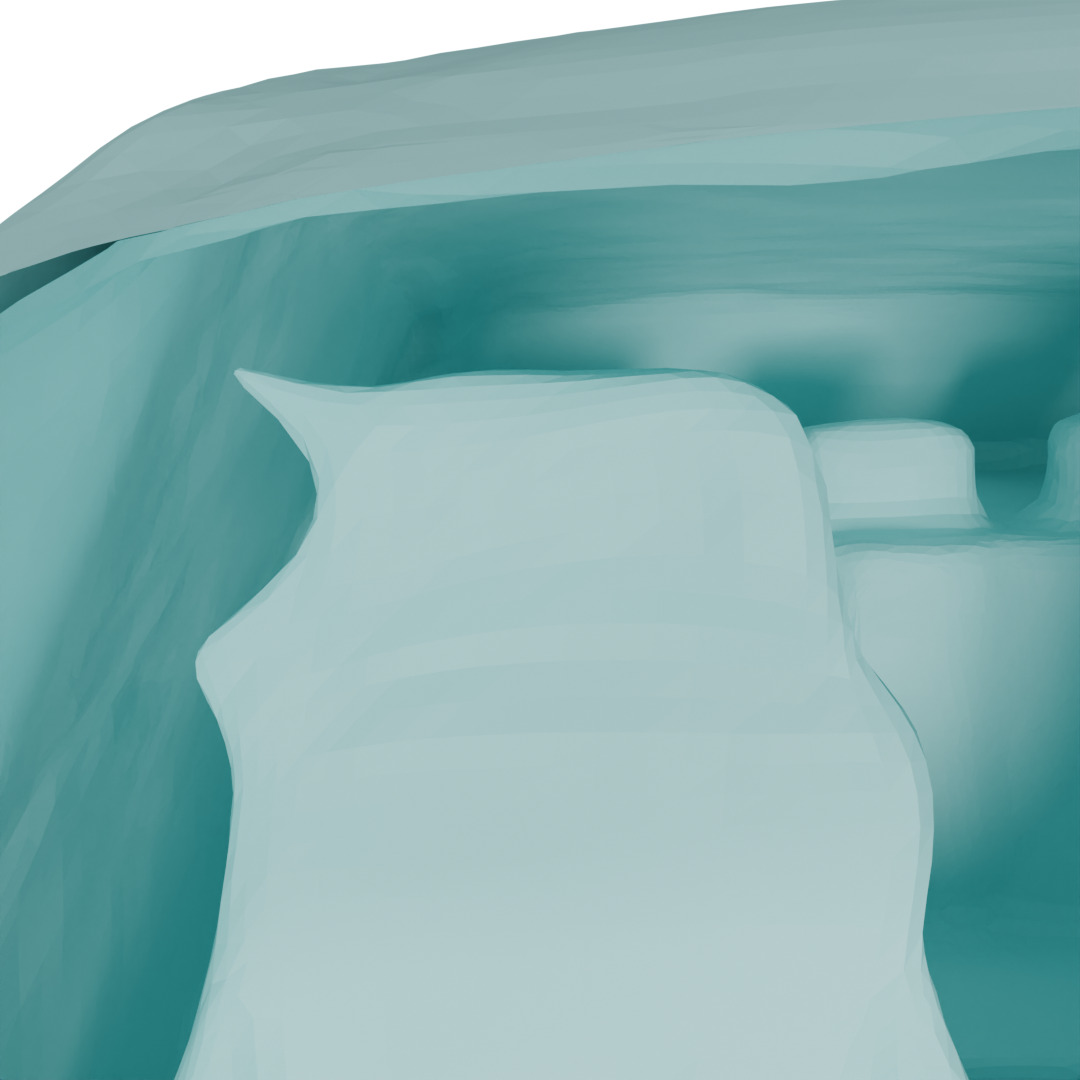}
    \includegraphics[width=1.2in]{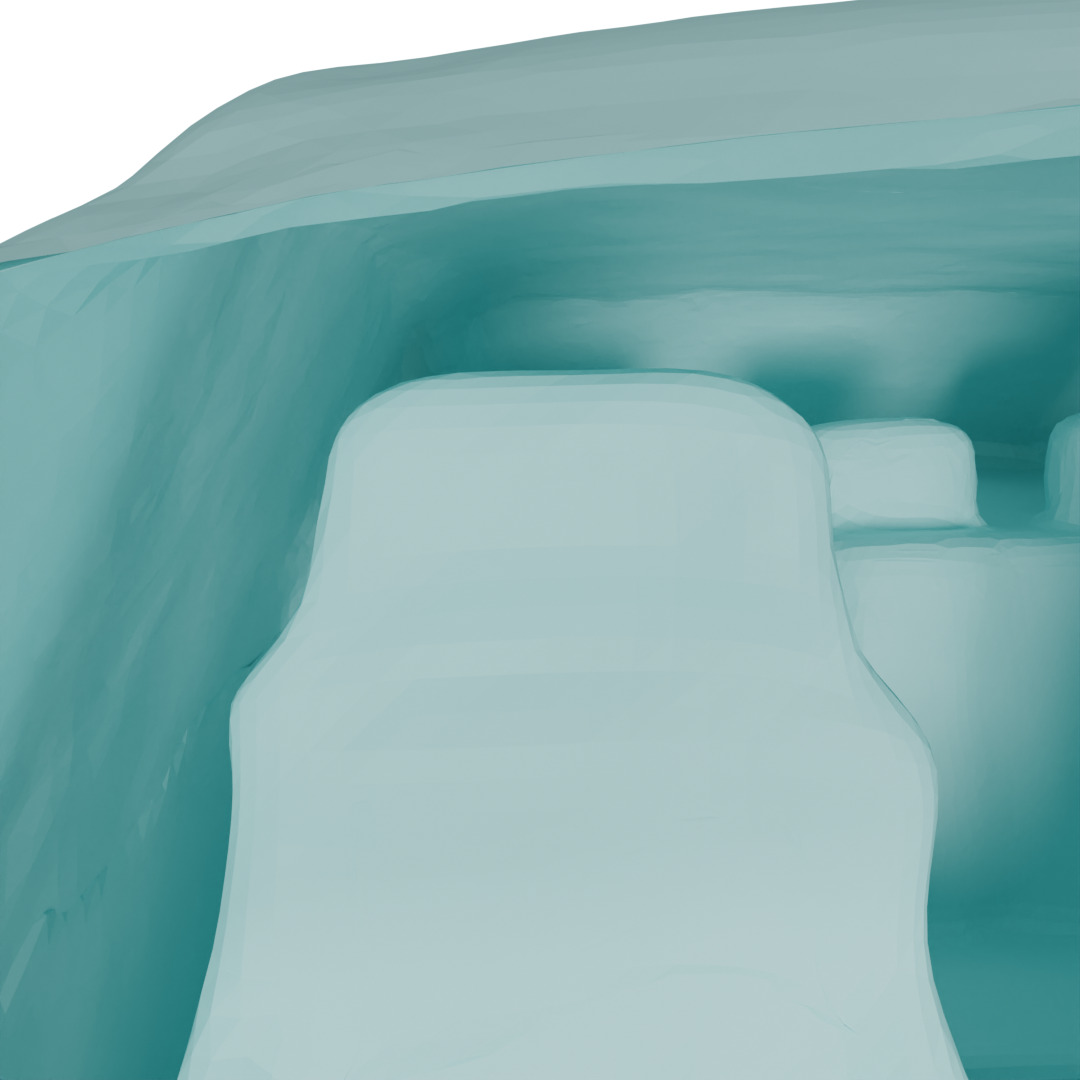}\\
    \caption{Visual results on the DeepFashion3D dataset featuring open surfaces, and the Shapenet-Car dataset showcasing both exterior views and interior cross-sections. The input UDFs were learned using  simpleMLP in DCUDF for the DeepFashion3D dataset and CAPUDF for the Shapenet-Car dataset. Both DCUDF and DCUDF2 use the same iso-value $r$, set at $0.01$ for the clothing models and $0.005$ for the car models. }
    \label{fig:shapenet_car}
\end{figure*}

\paragraph{\textbf{Evaluation on Open Models with Noise}} The DeepFashion3D dataset~\cite{zhu2020deep} includes 598 scanned clothing items and provides point clouds generated from images. We evaluated our method against DMUDF and MeshUDF on this dataset, focusing on performance in noisy distance fields. Table~\ref{tab:deepfashion3d} shows that our method outperforms others in terms of accuracy. MeshUDF suffers from artifacts such as unwanted triangular facets, while DMUDF is adversely affected by noise, leading to topological errors and broken structures due to inaccurate point projections. Furthermore, both methods frequently generate non-manifold configurations. In contrast, both DCUDF and our improved DCUDF2 consistently achieve high topological accuracy and effectively capture the textures of clothing, as evidenced by superior Chamfer distance scores. See Figure~\ref{fig:shapenet_car} for visual results. 

\begin{table}[!htbp]
  \caption{Quantitative comparison on the DeepFashion3dD dataset. UDFs were learned by simpleMLP as in DCUDF~\cite{Hou2023DCUDF} with mesh extractions performed at a resolution of $128^3$. The iso-value $r$, set at 0.01, was consistently applied in both DCUDF and our method. 
  }
  \label{tab:deepfashion3d}
  \setlength\tabcolsep{2pt}
  \centering
  \begin{tabular}{l|c|c|c|c}
    \hline
                              & MeshUDF     & DMUDF &    DCUDF &    Ours   \\ \hline\hline

      CD (average, $10^{-3}$)       &  1.073      &   0.868    &  0.876  & 0.853  \\  
   avg. non-manifold vertices   &  0.745      &   45.923    &   0       & 0       \\
    avg. non-manifold edges     &  0.336      &   56393.371  &   0       & 0       \\
    avg. genus                  &  22.667     &   1633.603   &   10.234  & 10.234  \\
    avg. boundaries             &  25.017     &   2902.516   &   0.039   & 0.039   \\
\hline
  \bottomrule
  \end{tabular}
\end{table}

\paragraph{\textbf{Evaluation on Models with Complex Inner Structures}} 
We selected car models from the ShapeNet dataset~\cite{Chang2015ShapeNetAI} to evaluate our method's performance on complex structures. This subset is referred to as ShapeNet-Car. As reported in Table~\ref{tab:shapnetcar}, DMUDF struggles with numerous non-manifold structures and many small, disconnected faces strongly affect the accuracy of their method. This issue often arises when points are incorrectly projected across narrow spaces, resulting in floating, skinny triangles. Similarly, MeshUDF tends to produce many boundaries or holes. In contrast, our method not only achieves the CD values but also significantly reduces topological defects. Moreover, DCUDF2 enhances the visual quality of sharp features on car models, providing superior aesthetic quality. See Figure~\ref{fig:shapenet_car}.

\begin{table}[tb]
  \caption{Quantitative comparison on the ShapeNet-Cars dataset. UDFs were learned using CAPUDF~\cite{Zhou2022CAP-UDF} with zero level set extraction at a resolution of $128^3$. The parameter $r$ for both DCUDF and ours was set to 0.005. 
  }
  \label{tab:shapnetcar}
  \centering
  \setlength\tabcolsep{2pt}
  \begin{tabular}{l|c|c|c|c}
    \hline
                              & MeshUDF     & DMUDF &    DCUDF &    Ours   \\ \hline\hline
    
  CD (average, $10^{-3}$)       &  3.582       &   4.360      &   3.447   & 3.316   \\  
   avg. non-manifold vertices   &  1449.904    &   102.183    &   0       & 0       \\
    avg. non-manifold edges     &  98.462      &   58829.440  &   0       & 0       \\
    avg. genus                  &  284.158     &   1630.502   &   38.025  & 23.695  \\
    avg. boundaries             &  854.915     &   3371.993   &   0.261   & 1.518   \\
\hline
  \bottomrule
  \end{tabular}
\end{table}

\paragraph{\textbf{Evaluation on UDFs Learned from Multi-view images}}
Following NeUDF~\cite{Liu2023NeUDFLN}, we evaluate our method on the multi-view reconstruction task using the DTU datset~\cite{jensen2014large}. NeUDF has been tested across 15 DTU scenes. Due to space limitations, we present selected results here, with additional outcomes available in the supplementary material. As shown in Figure~\ref{fig:dtu}, our method achieves the best visual results at a resolution of $128^3$, effectively capturing model details and preserving topological integrity. However, given that the point clouds from the DTU dataset include noise, we recommend a higher resolution for optimal numerical accuracy.

\begin{figure}[!htbp]
    \centering
    \makebox[1.1in]{MeshUDF}
    \makebox[1.1in]{DMUDF}
    \makebox[1.1in]{Ours}\\
    \includegraphics[width=1.1in]{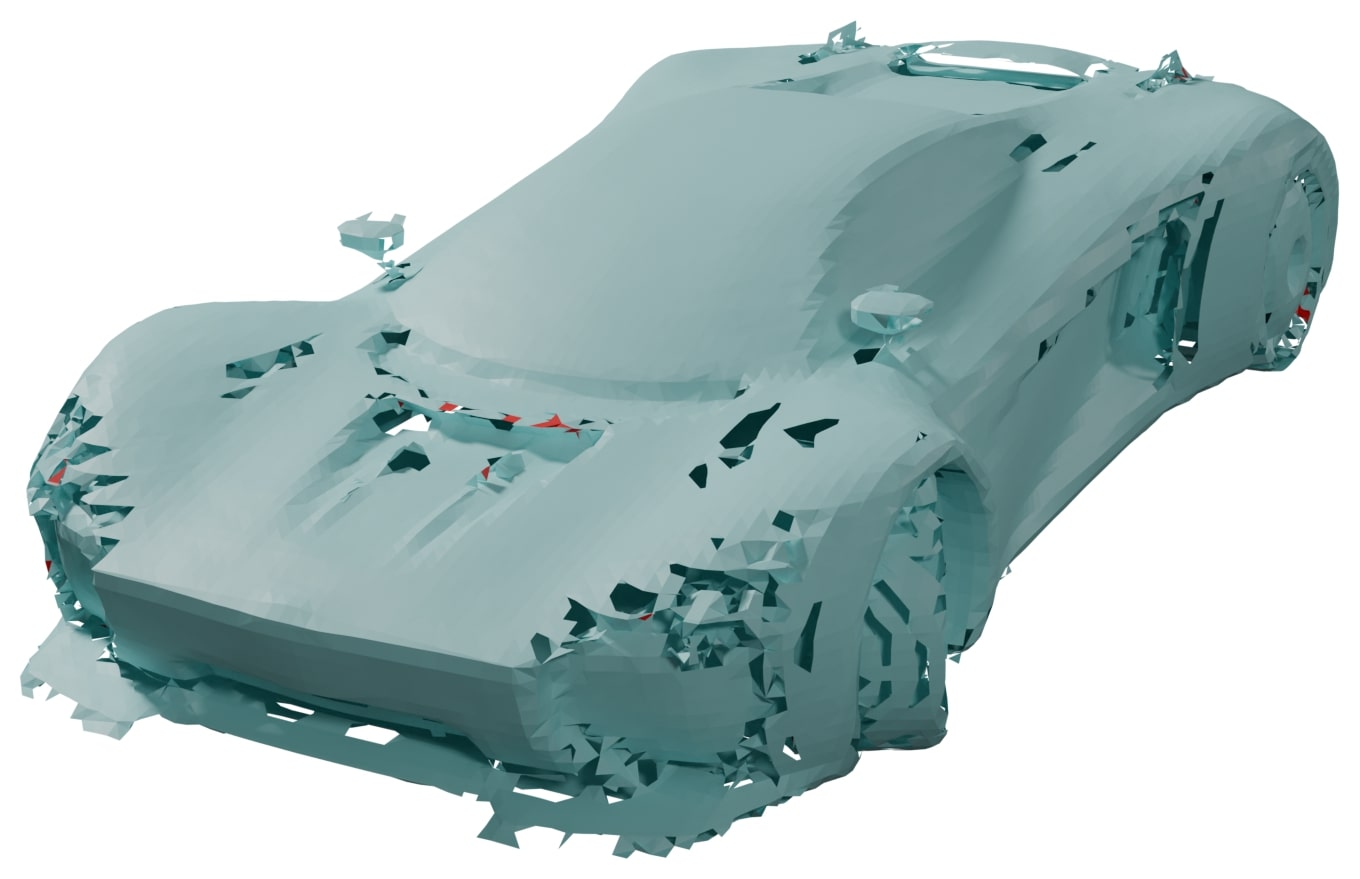 }
    \includegraphics[width=1.1in]{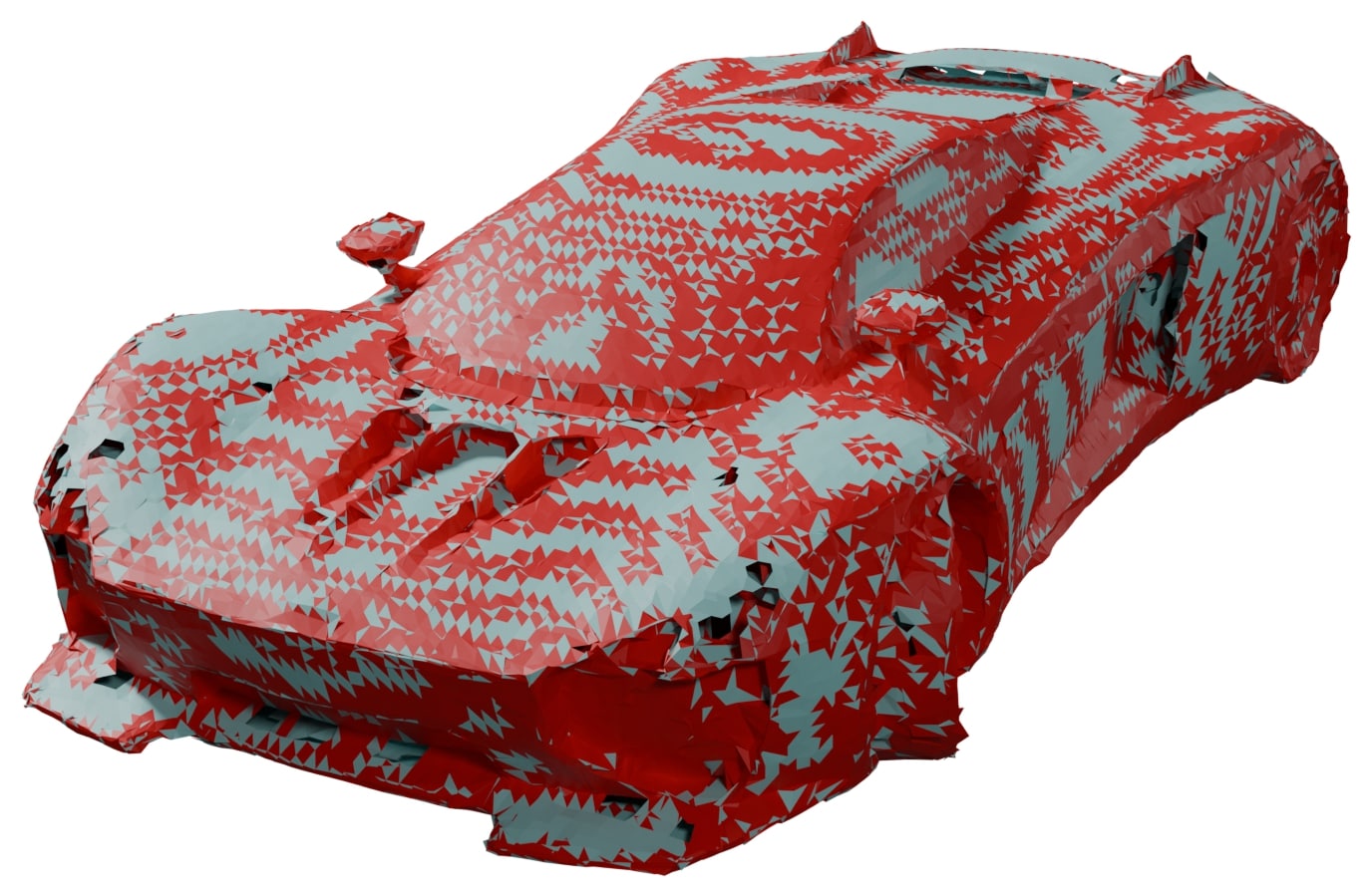 }
    \includegraphics[width=1.1in]{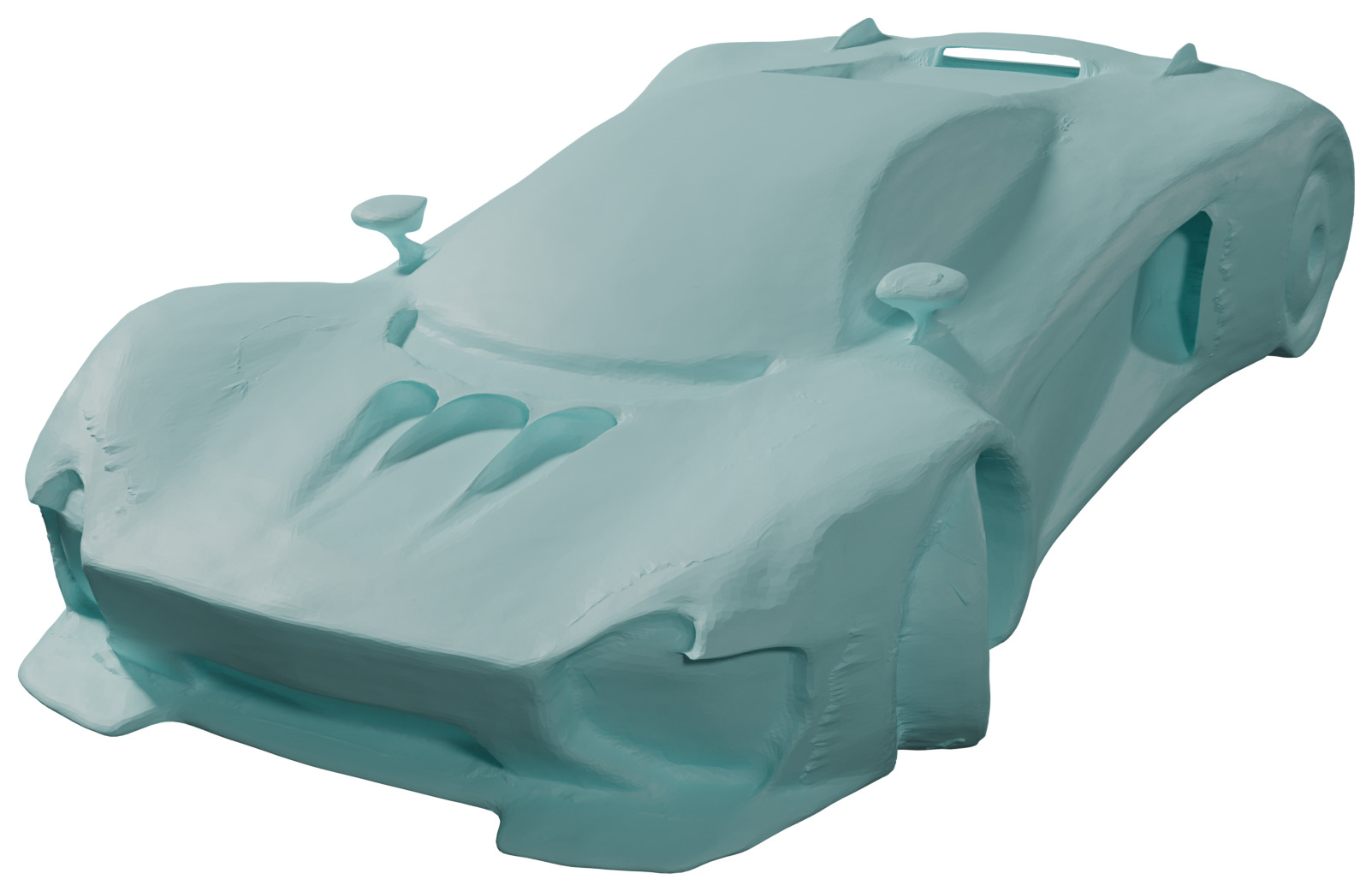 }\\
    \includegraphics[width=1.1in]{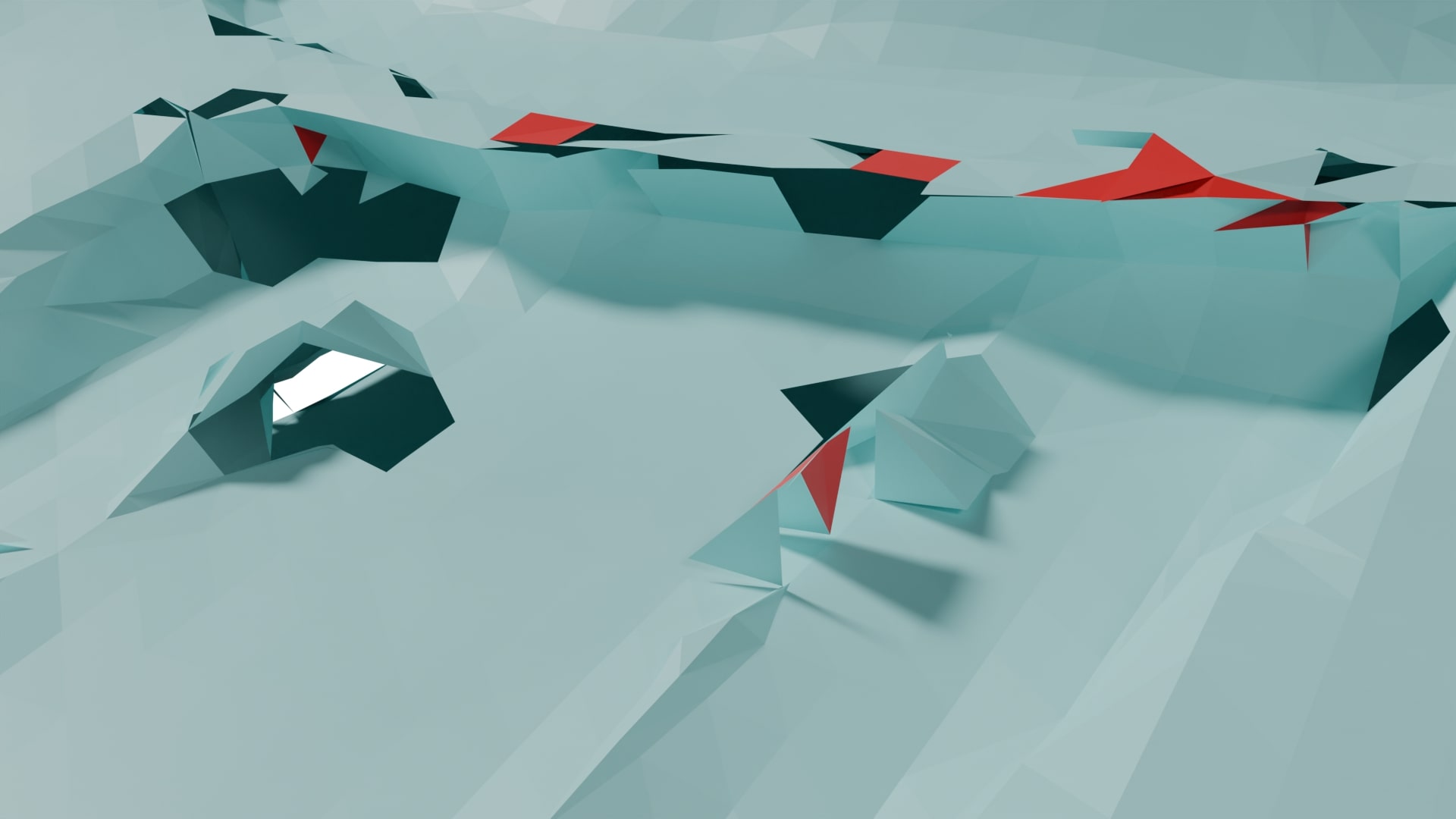 }
    \includegraphics[width=1.1in]{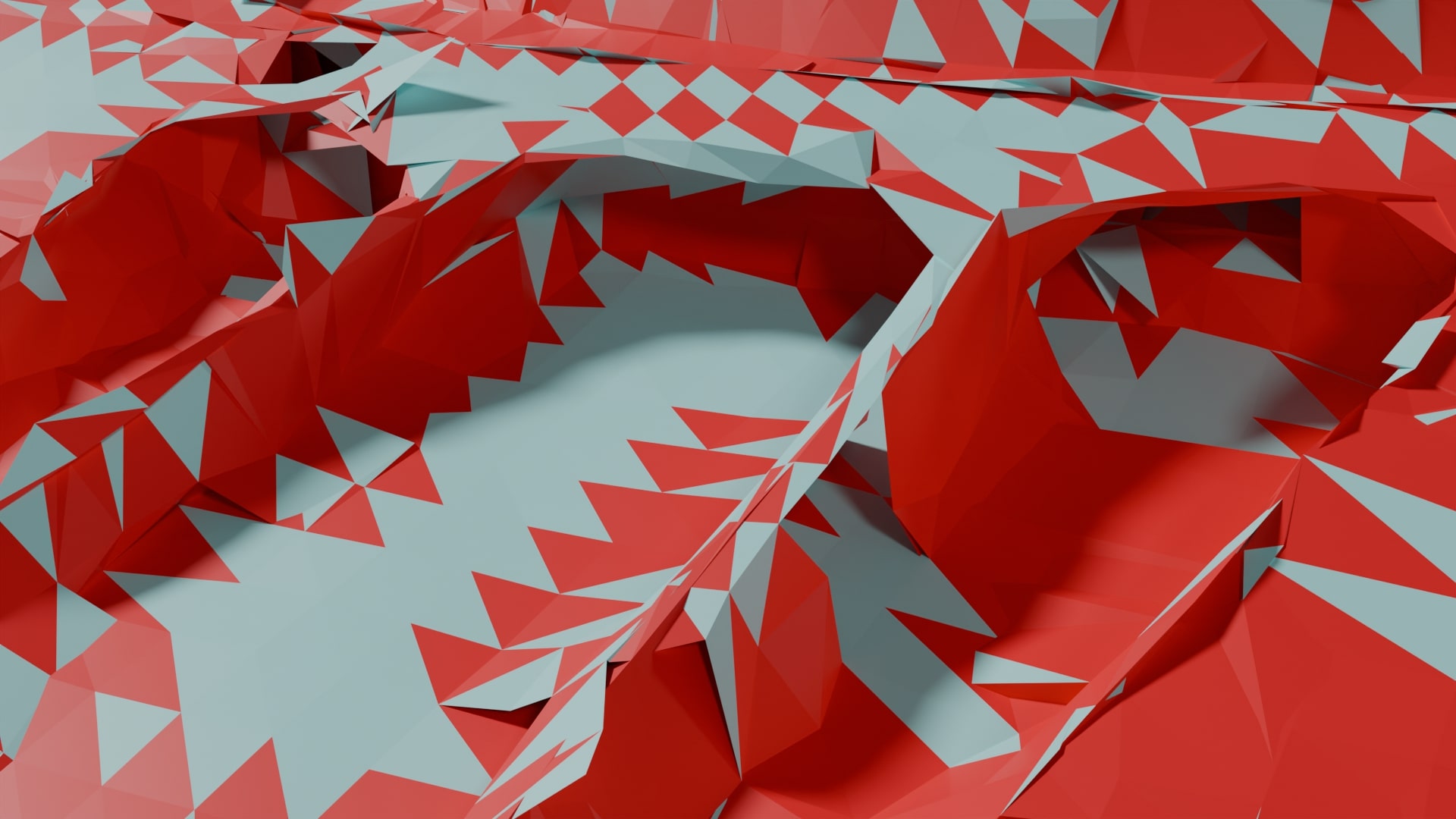 }
    \includegraphics[width=1.1in]{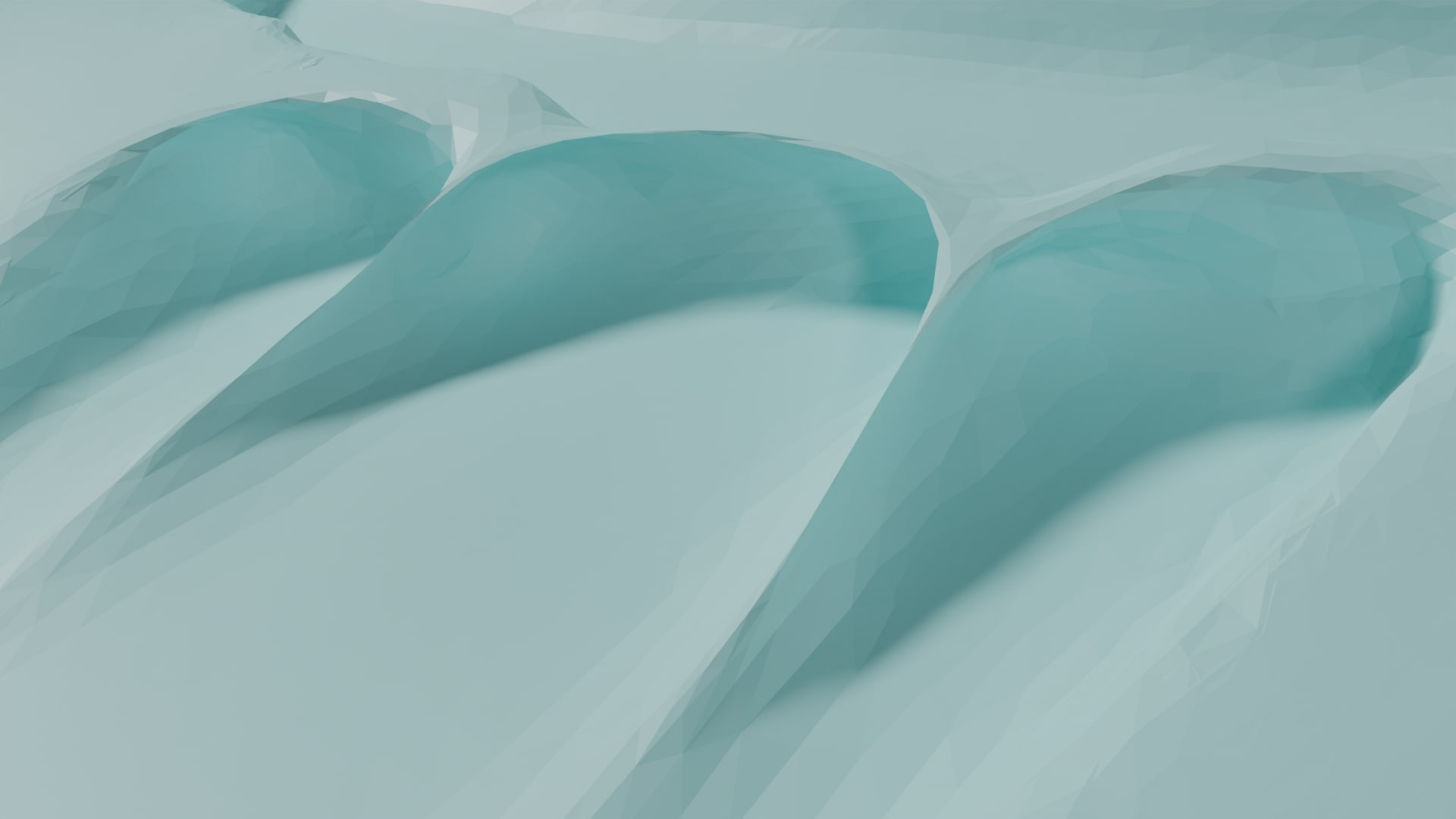 }\\
    \caption{Visualization of non-manifold structures highlighted in red. Both MeshUDF and DMUDF frequently produce non-manifold vertices and edges. In contrast, our method consistently avoids non-manifold issues, ensuring our results are ideally suited for downstream applications that require manifold inputs.}
    \label{fig:non-manifold}
\end{figure}

\paragraph{\textbf{Eliminating Non-manifold Configurations}} DMUDF tends to produce a large amount of non-manifold edges, which can complicate further processing. While MeshUDF reduces the frequency of non-manifold structures, it does not eliminate them entirely. Our approach, however, completely avoids the creation of non-manifold structures, as shown in Figure~\ref{fig:non-manifold}. This capability makes our method particularly valuable for downstream applications that require manifold configurations. See  for visual results. 

\begin{figure}[!htbp]
    \centering
    \includegraphics[height=2.6in]{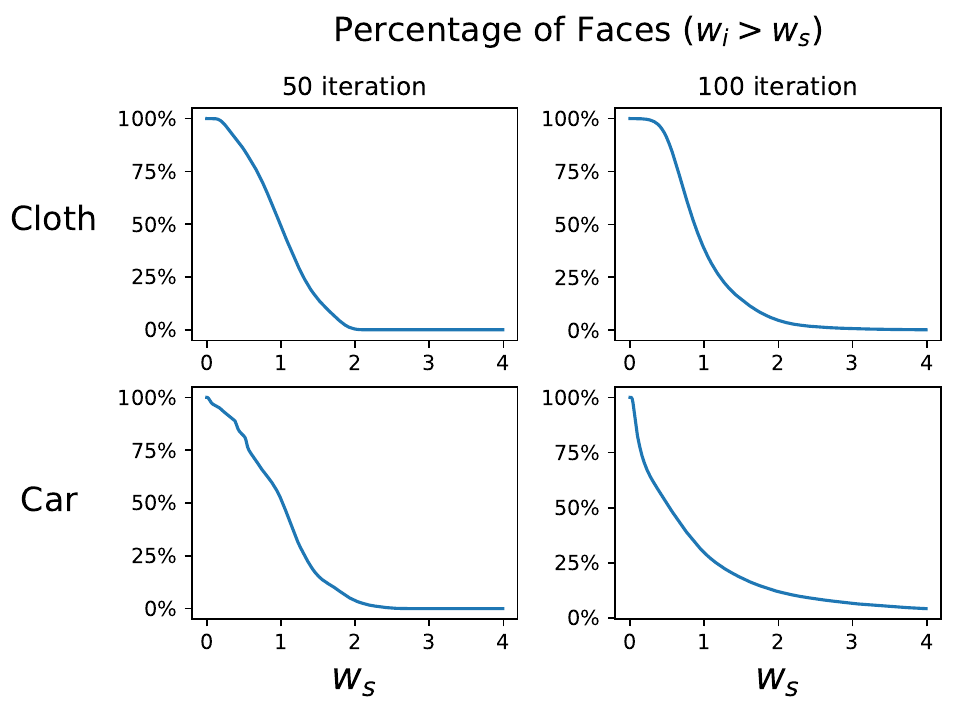}\\
    \caption{We show the percentages of faces whose weights are larger than $w_s$. The vertical axes are the percentages and horizontal axes are the weights $w_s$.}
    \label{fig:rate}
\end{figure}
\begin{figure}[tb]
    \centering
   
    \makebox[0.8in]{$128^3$}
    \makebox[0.8in]{$256^3$}
    \makebox[0.8in]{$256^3$}
    \makebox[0.8in]{$512^3$}\\
    \makebox[0.8in]{$r=0.005$}
    \makebox[0.8in]{$r=0.015$}
    \makebox[0.8in]{$r=0.005$}
    \makebox[0.8in]{$r=0.005$}\\
    \includegraphics[width=0.8in]{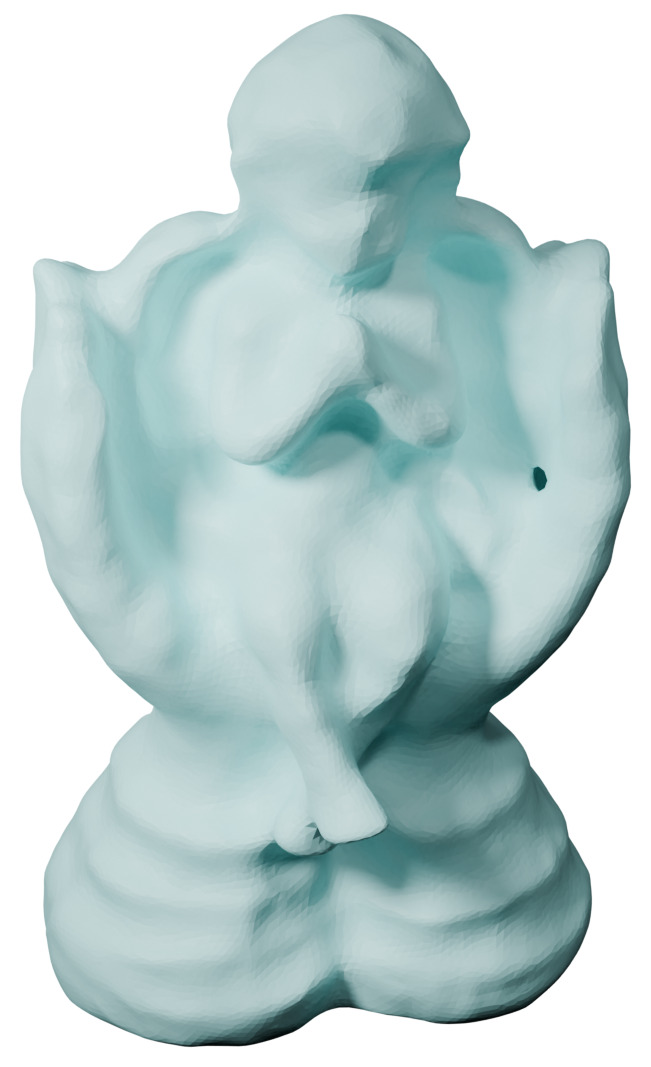 }
    \includegraphics[width=0.8in]{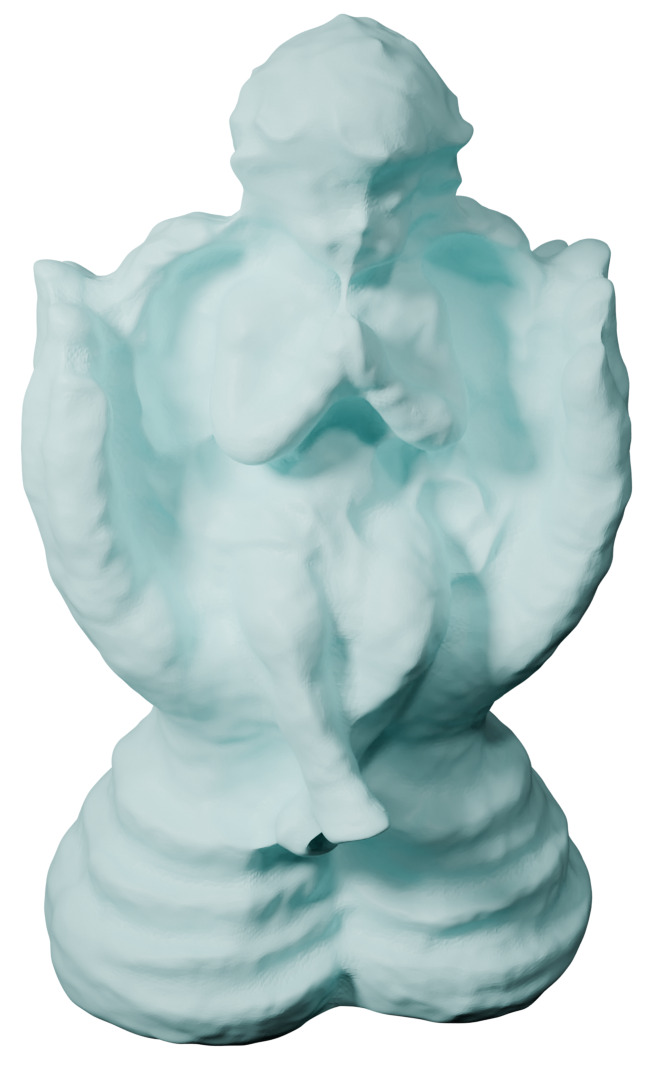 }
    \includegraphics[width=0.8in]{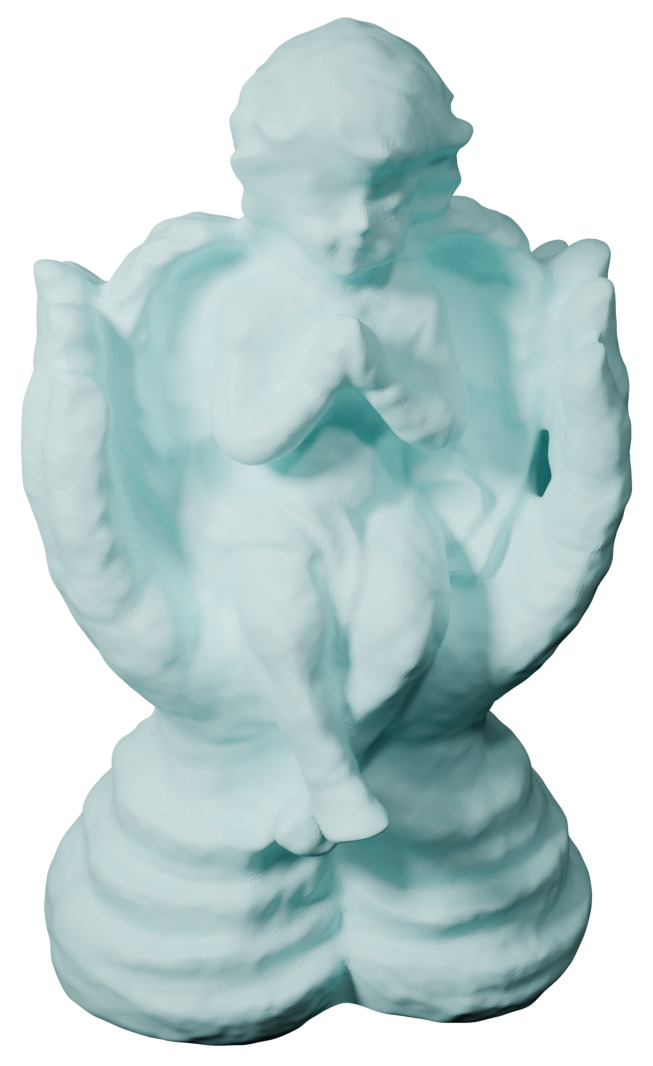 }
    \includegraphics[width=0.8in]{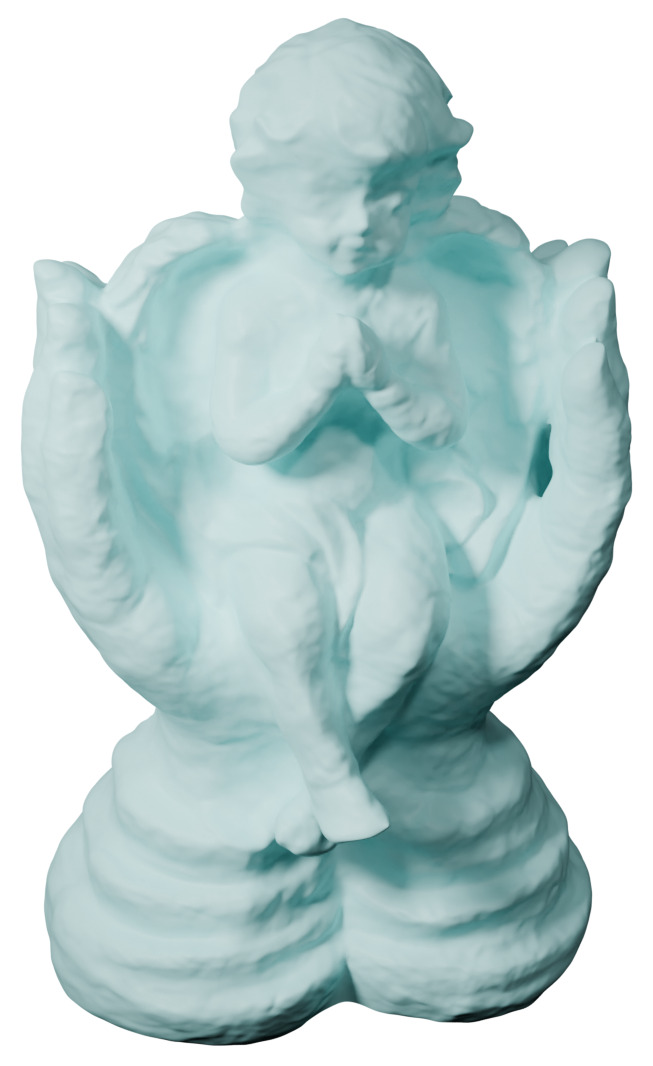 }\\
    \includegraphics[width=0.8in]{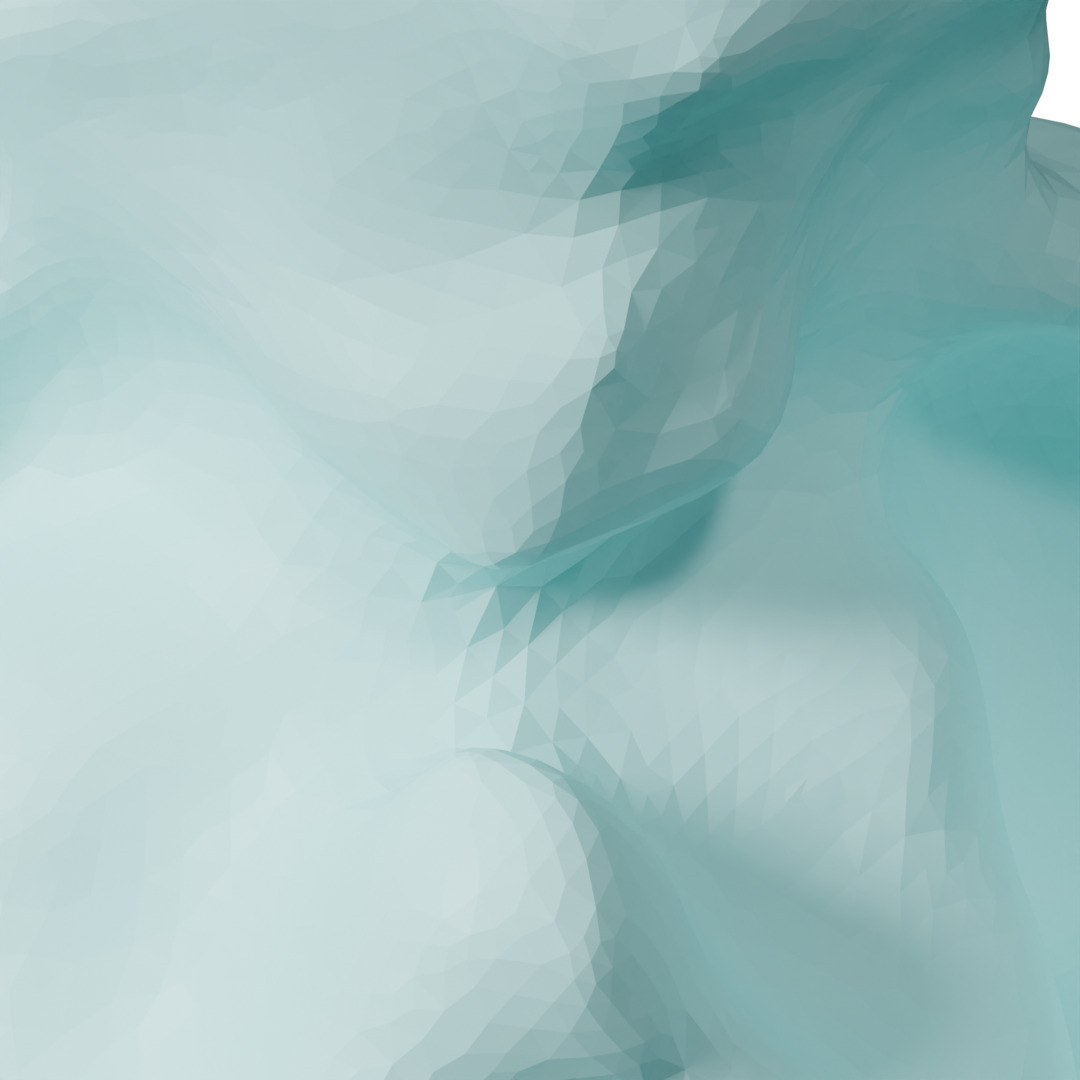 }
    \includegraphics[width=0.8in]{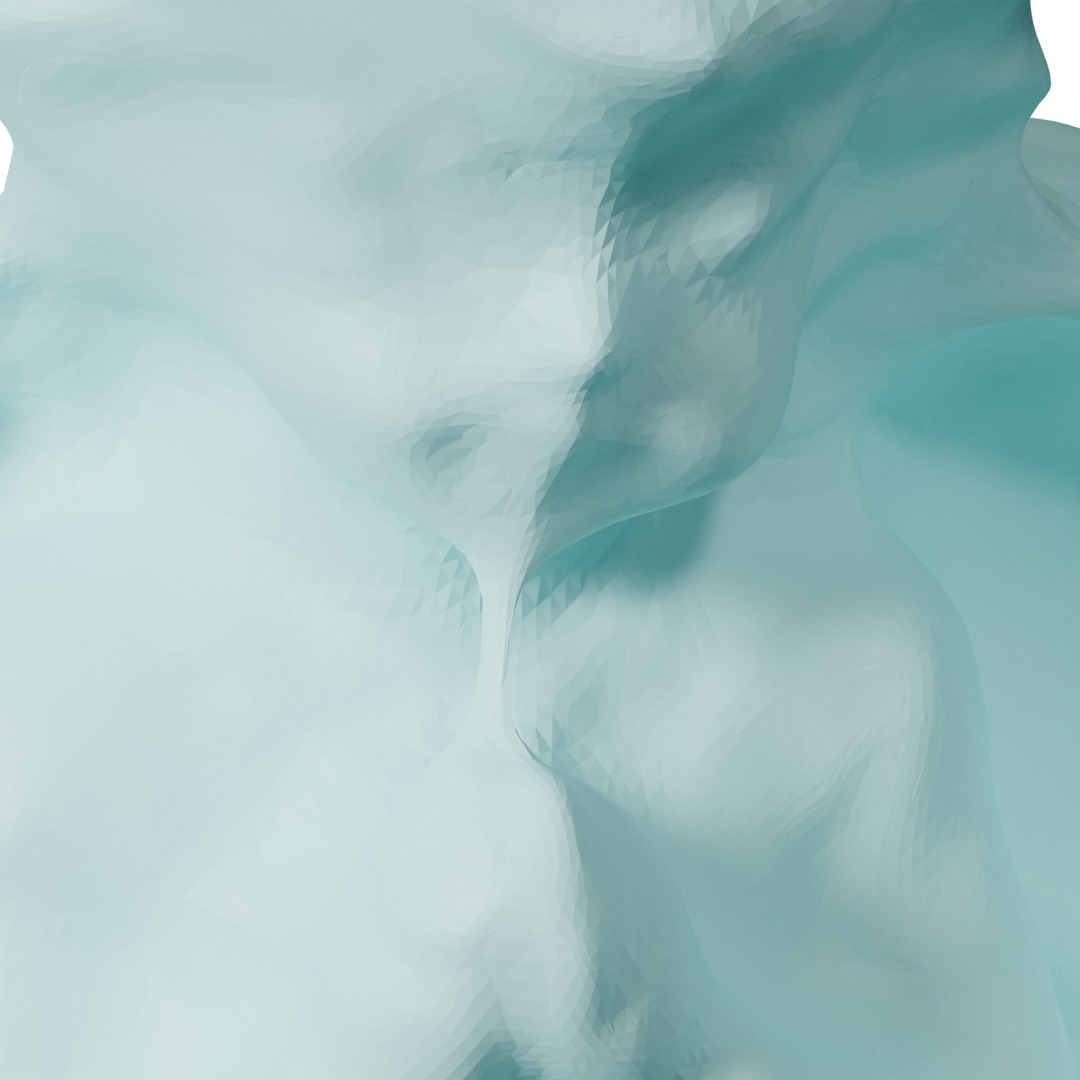 }
    \includegraphics[width=0.8in]{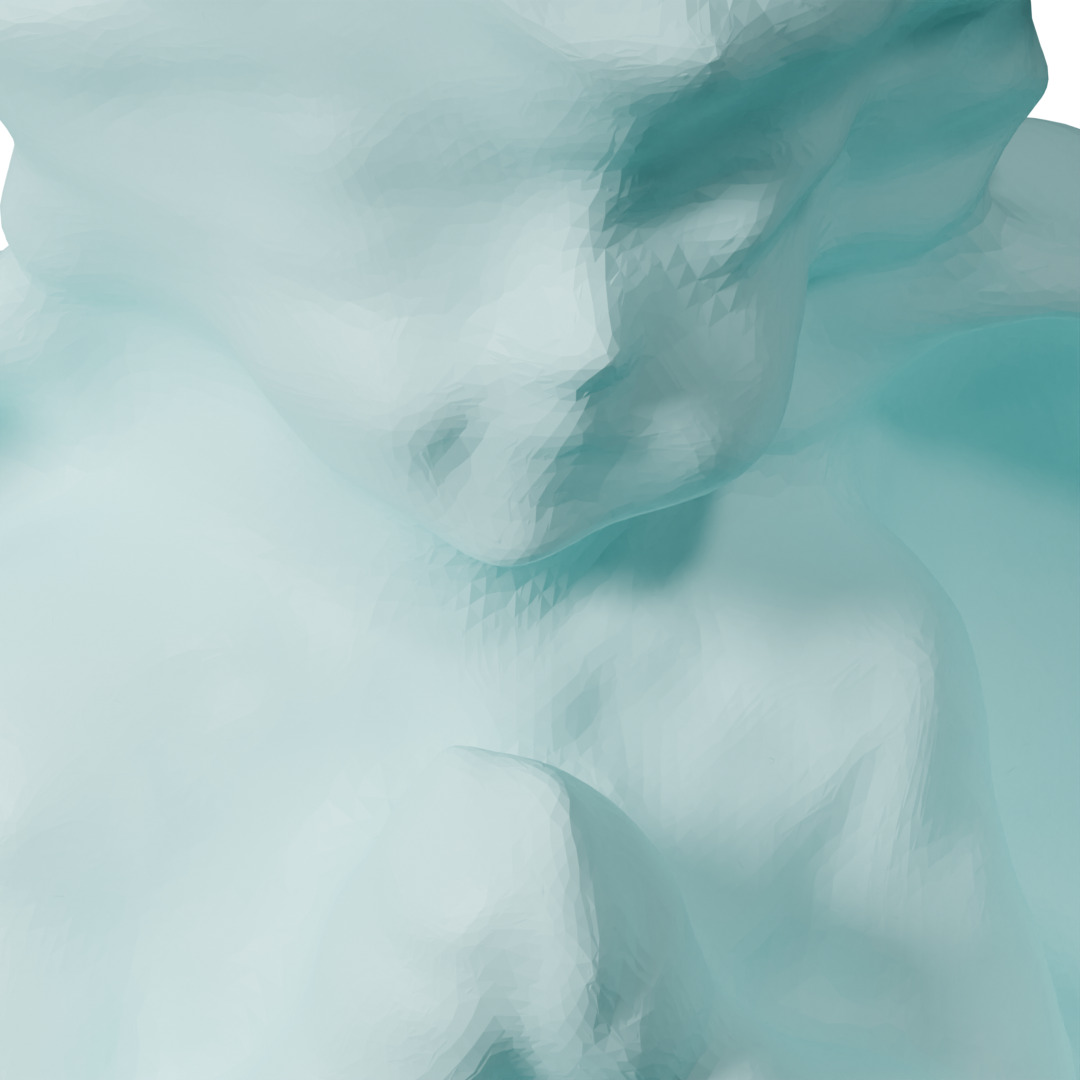 }
    \includegraphics[width=0.8in]{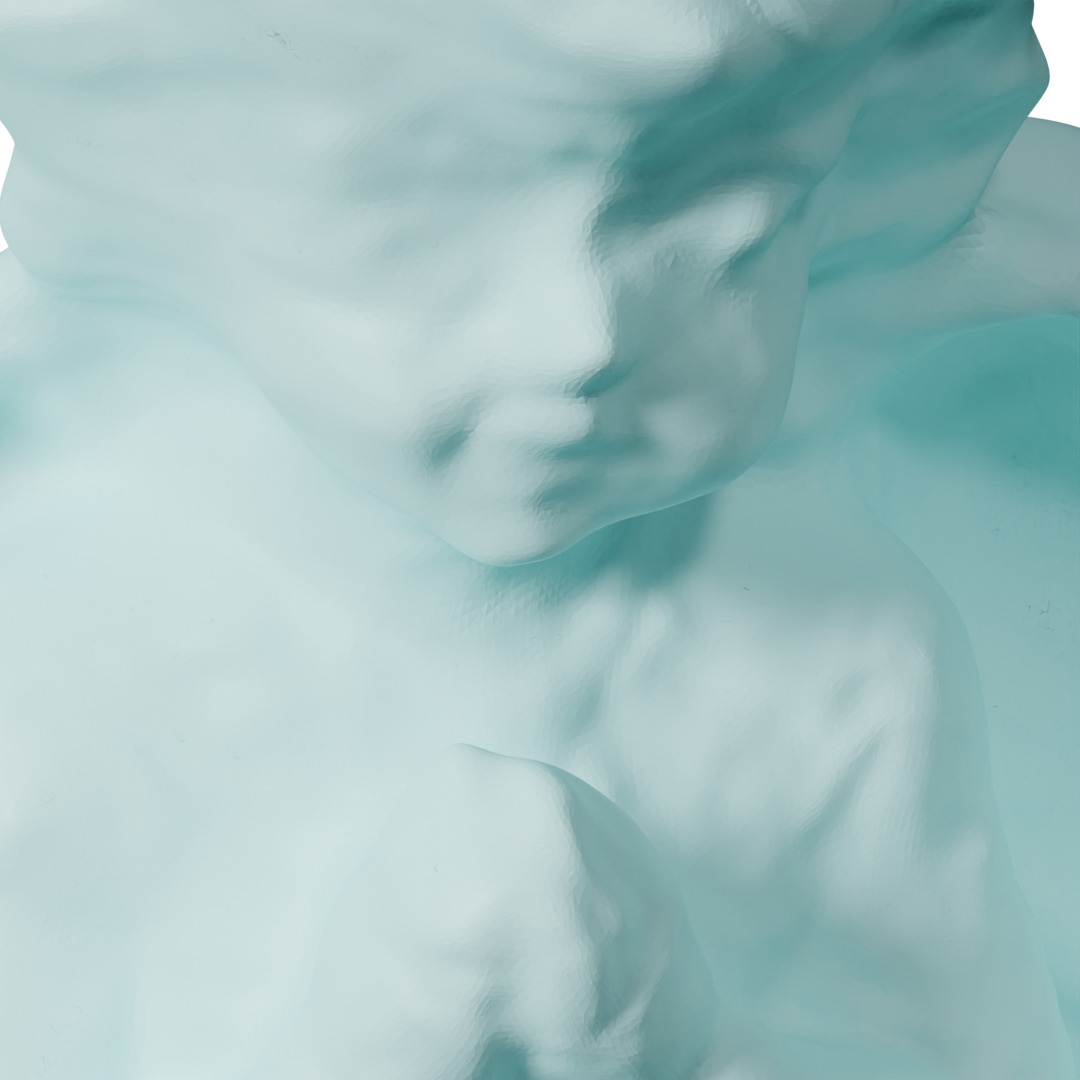 }\\
    \makebox[0.8in]{0.796}
    \makebox[0.8in]{0.757}
    \makebox[0.8in]{0.733}
    \makebox[0.8in]{0.727}\\
     \makebox[1.2in]{DCUDF}\\

    \includegraphics[width=0.8in]{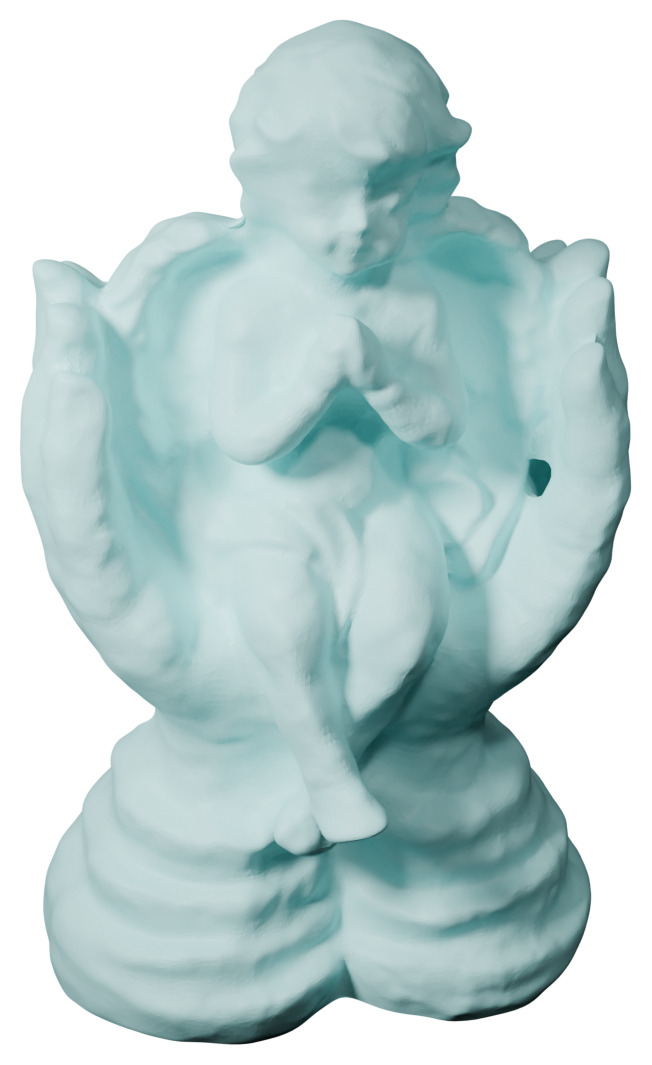 }
    \includegraphics[width=0.8in]{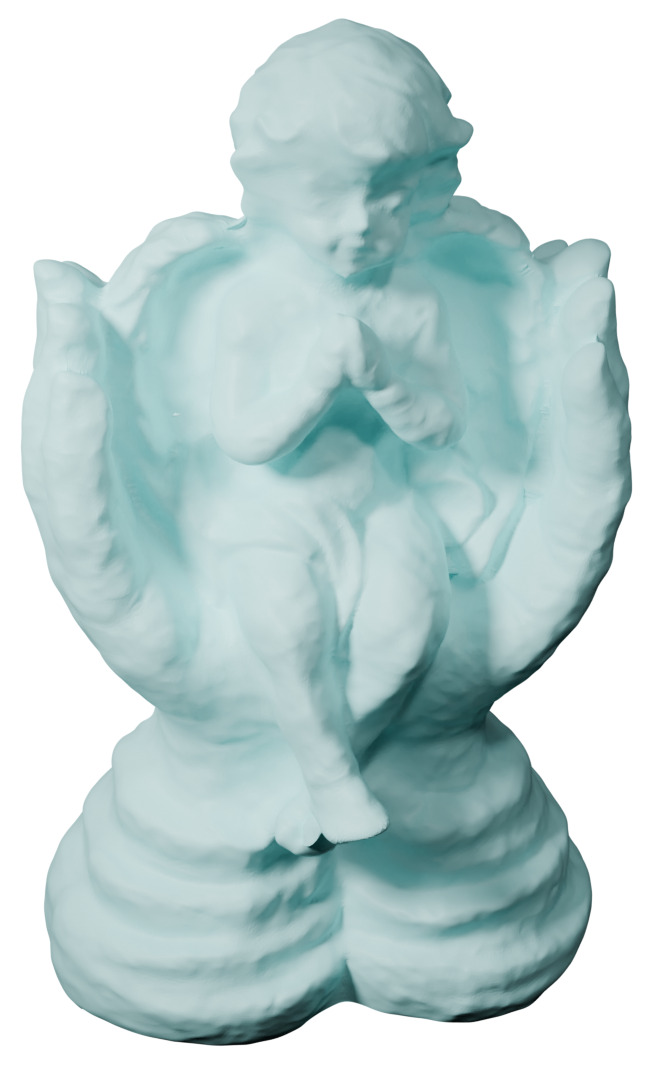 }
    \includegraphics[width=0.8in]{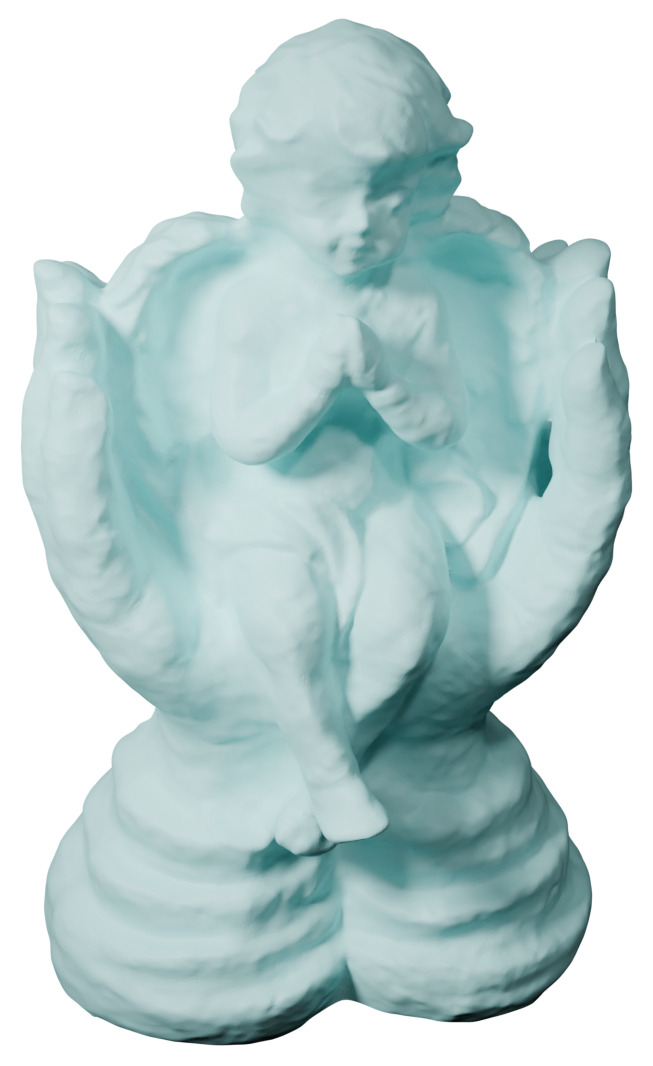 }
    \includegraphics[width=0.8in]{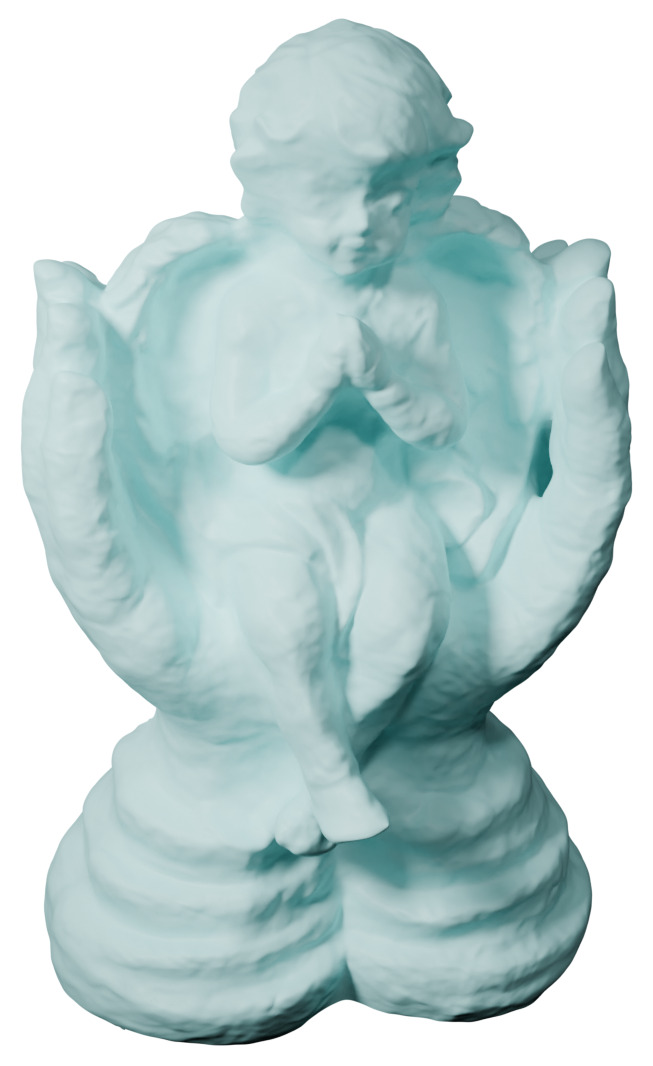 }\\
    \includegraphics[width=0.8in]{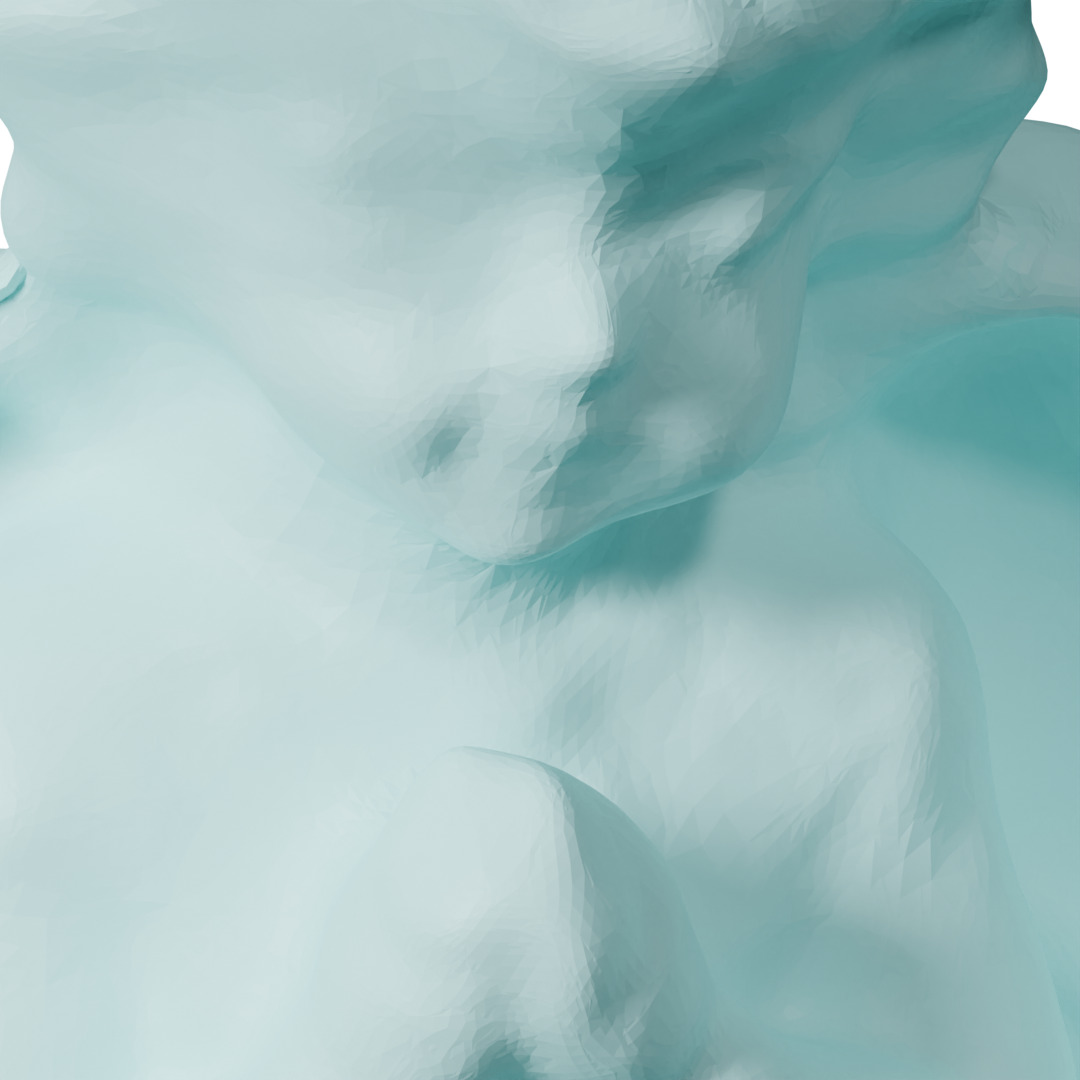 }
    \includegraphics[width=0.8in]{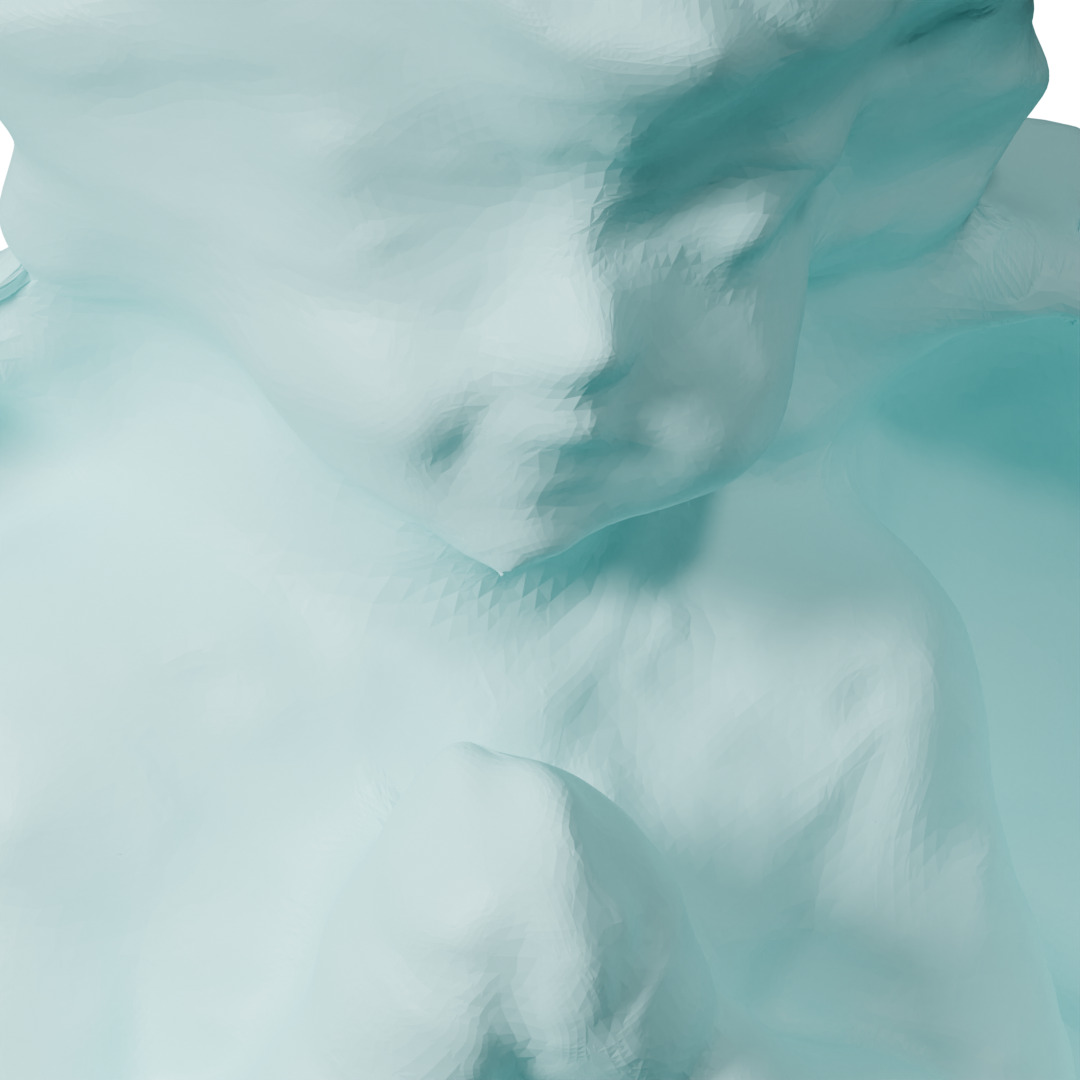 }
    \includegraphics[width=0.8in]{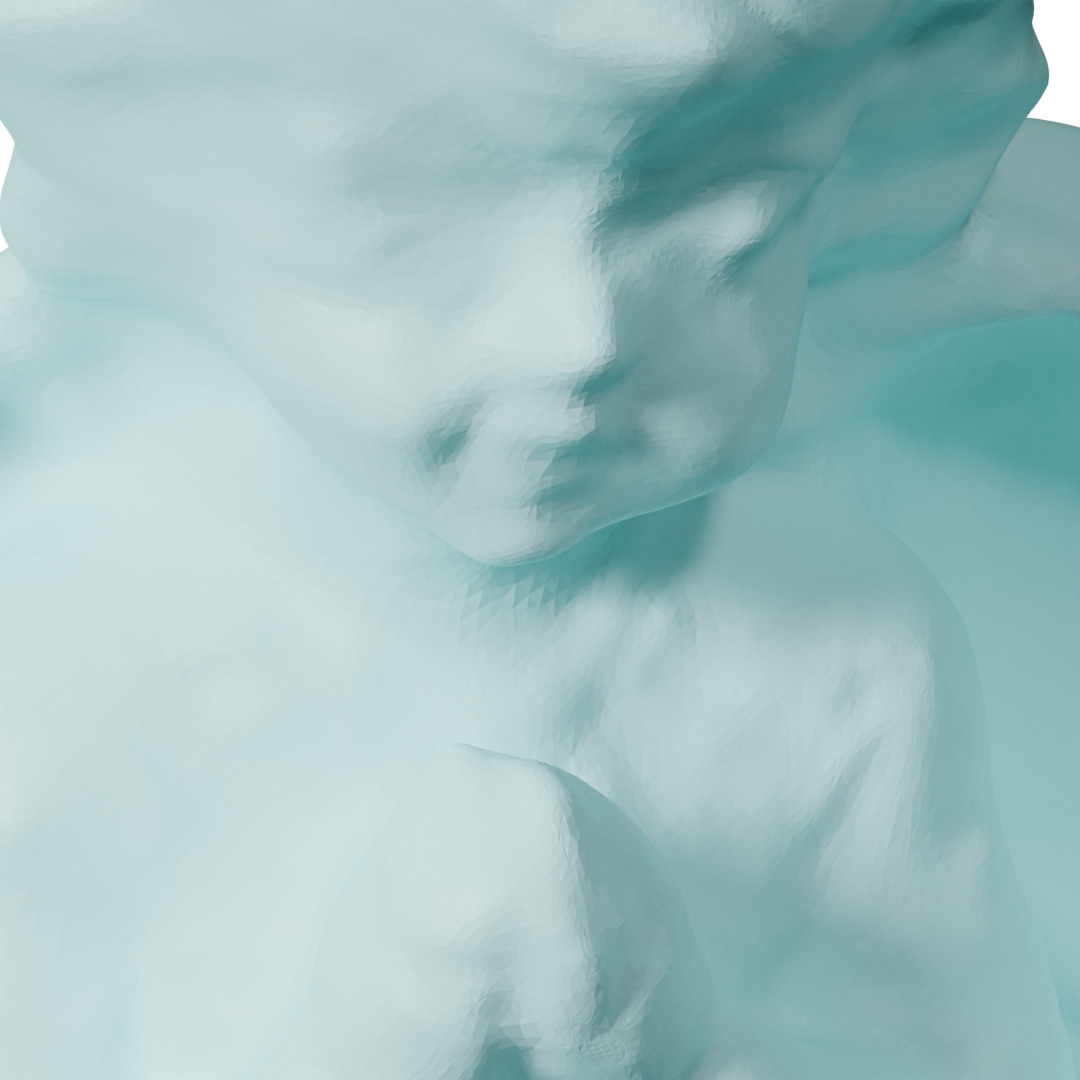 }
    \includegraphics[width=0.8in]{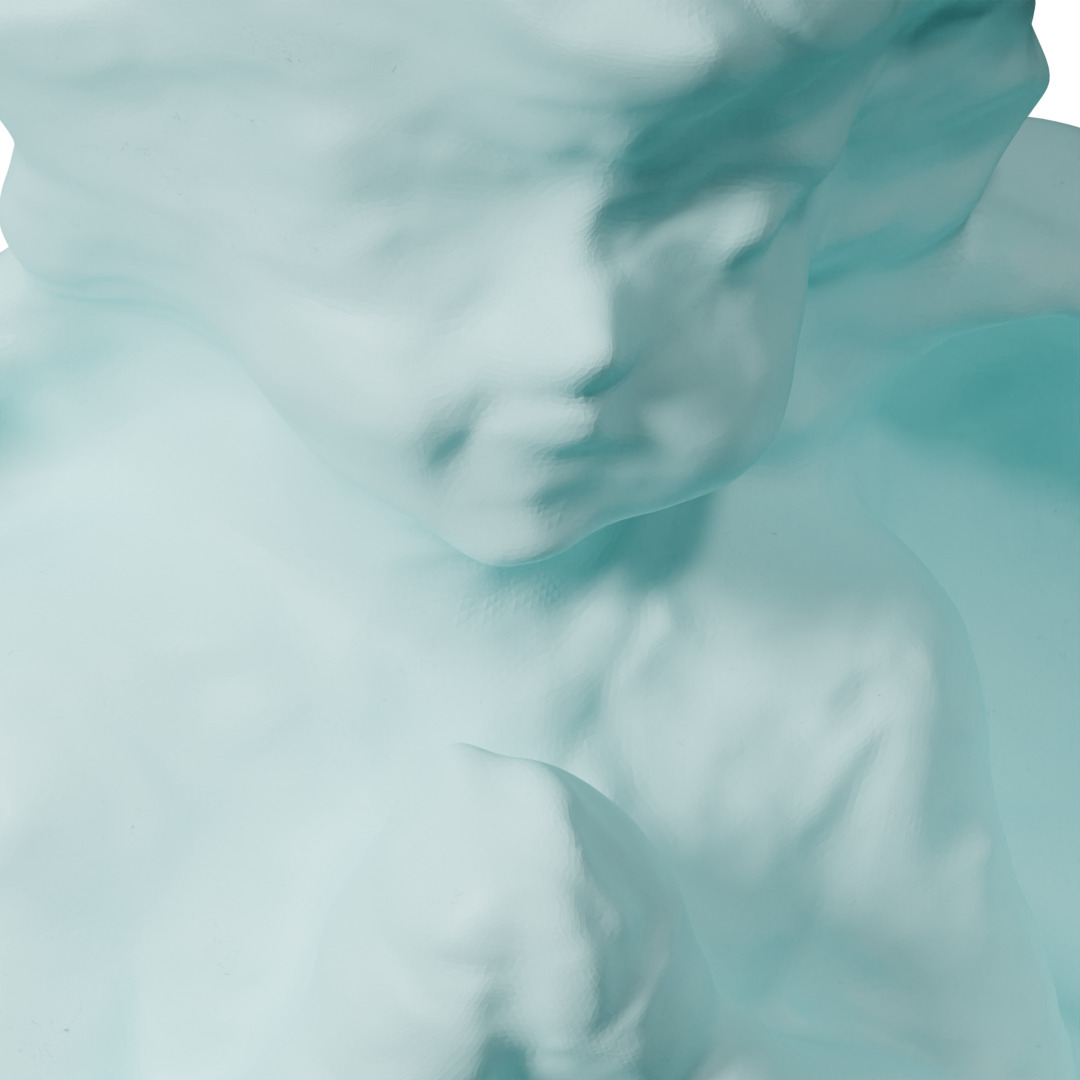 }\\
    \makebox[0.8in]{0.735}
    \makebox[0.8in]{0.730}
    \makebox[0.8in]{0.724}
    \makebox[0.8in]{0.724}
\\
 \makebox[1.2in]{DCUDF2}\\

    \caption{Comparison of our method with DCUDF across different settings of MC resolutions and iso-values $r$. Visually, our results at a lower resolution of $256^3$ match the quality of DCUDF results at a higher resolution of $512^3$. Additionally, our outputs with higher iso-values $r$ demonstrate superior quality due to our topology correction strategy. The Chamfer distance (scaled by $10^3$) is provided below each figure for quantitative comparison.}
    \label{fig:multi-resolution}
\end{figure}

\paragraph{\textbf{Comparison with DCUDF under Various Parameter Settings}}
We show the performance of our method and DCUDF at various resolutions and iso-values $r$ using a DTU scene as shown in Figure~\ref{fig:multi-resolution}. We calculate CD with the evaluation code provided by NeUDF~\cite{Liu2023NeUDFLN}. With a high MC resolution $512^3$, we observe that both DCUDF and DCUDF2 yield results of similar quality, however, DCUDF takes almost 3 times than ours. While our method can use a low resolution mesh with fewer vertices to extract a high-quality mesh from UDF.

\begin{figure}[!htbp]
    \centering
    \includegraphics[width=0.8in]{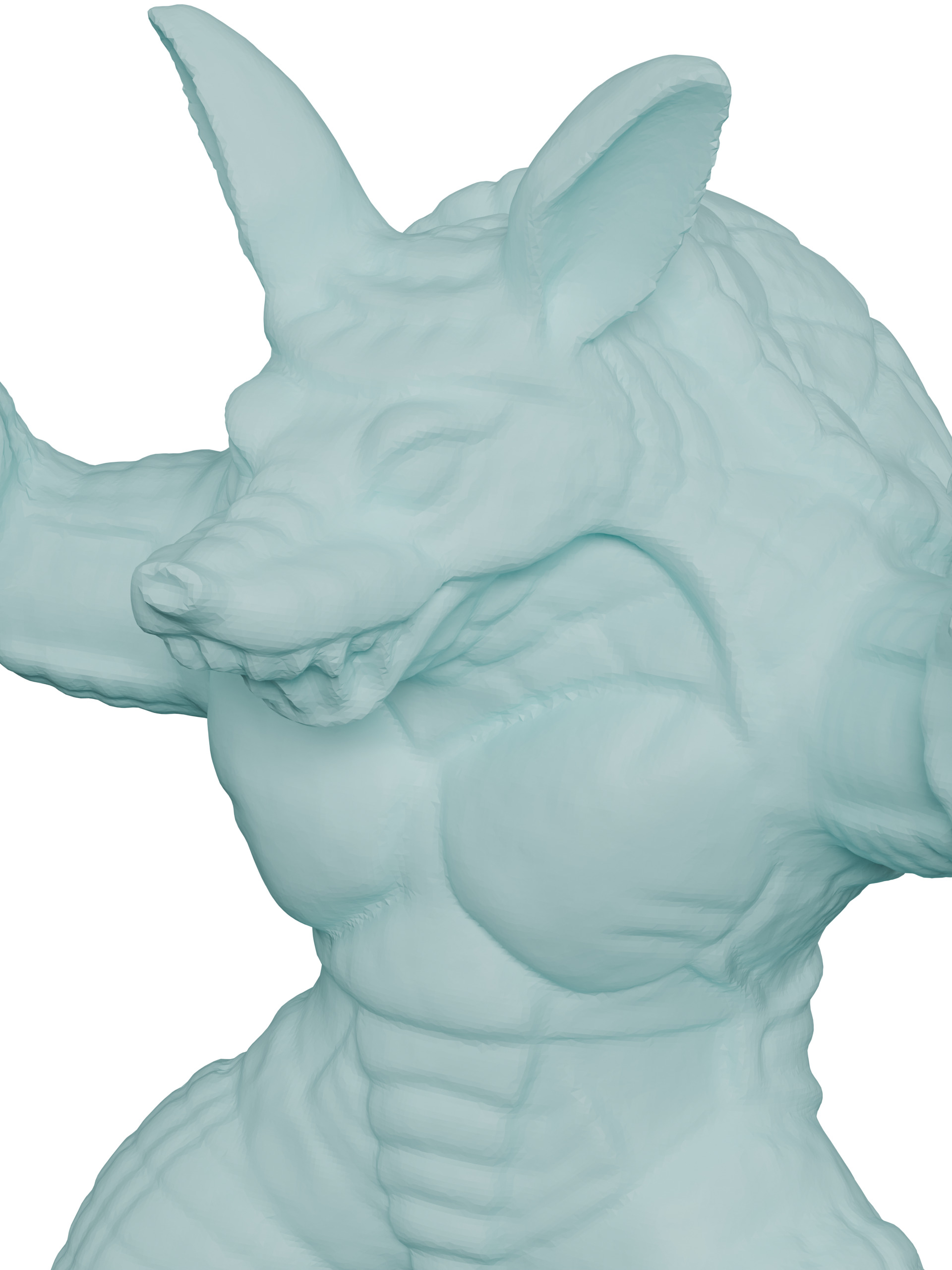 }
    \includegraphics[width=0.8in]{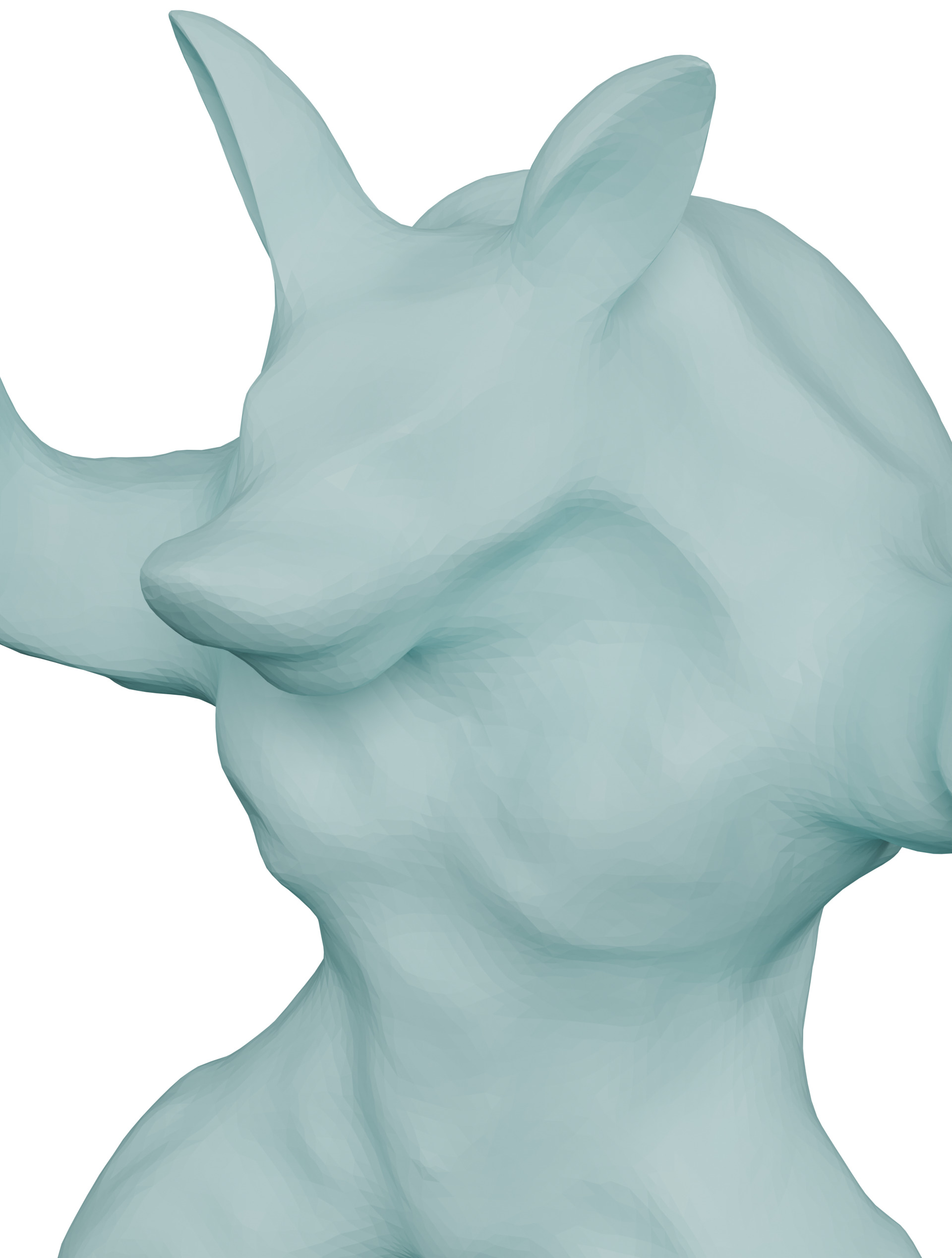 }
    \includegraphics[width=0.8in]{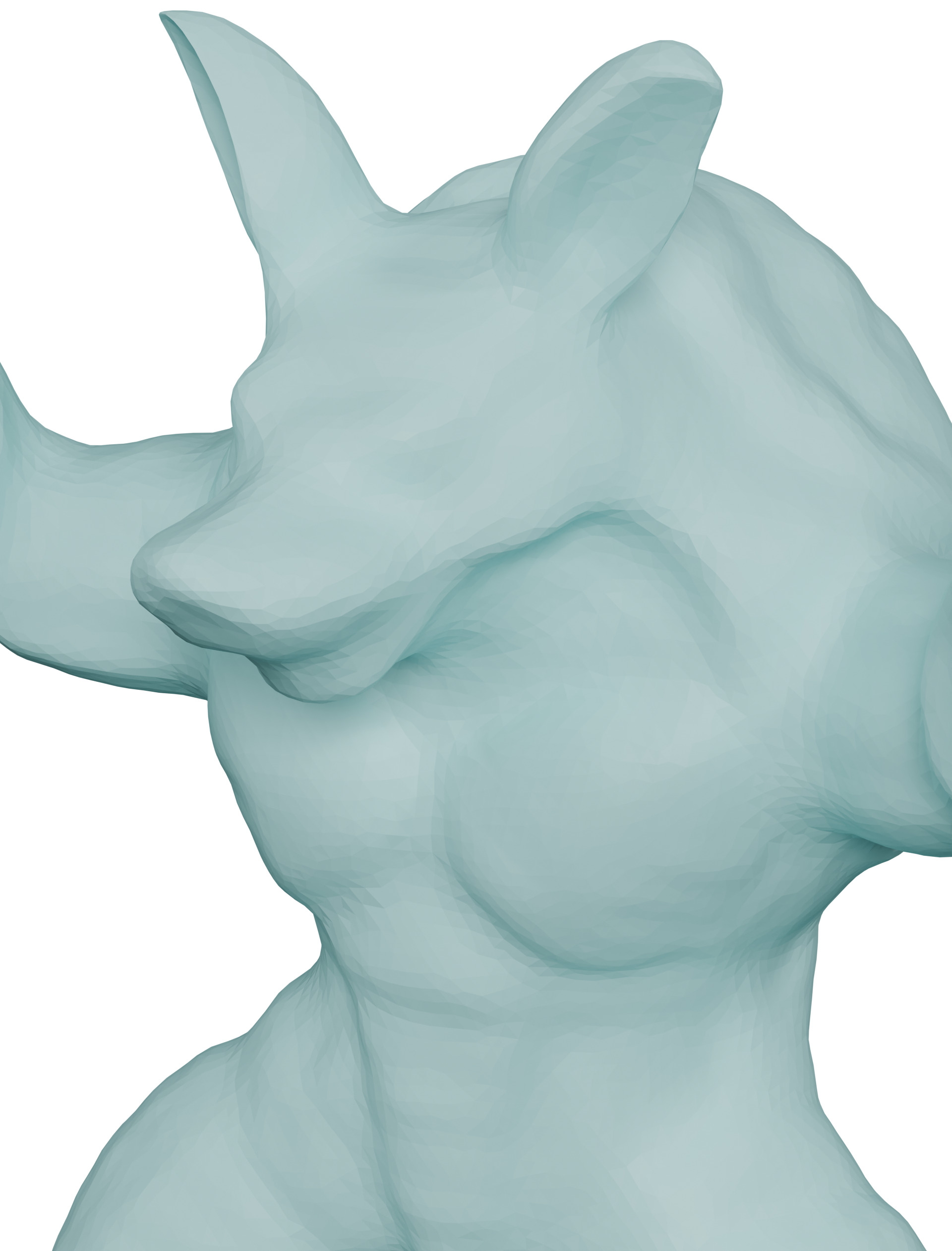}
    \includegraphics[width=0.8in]{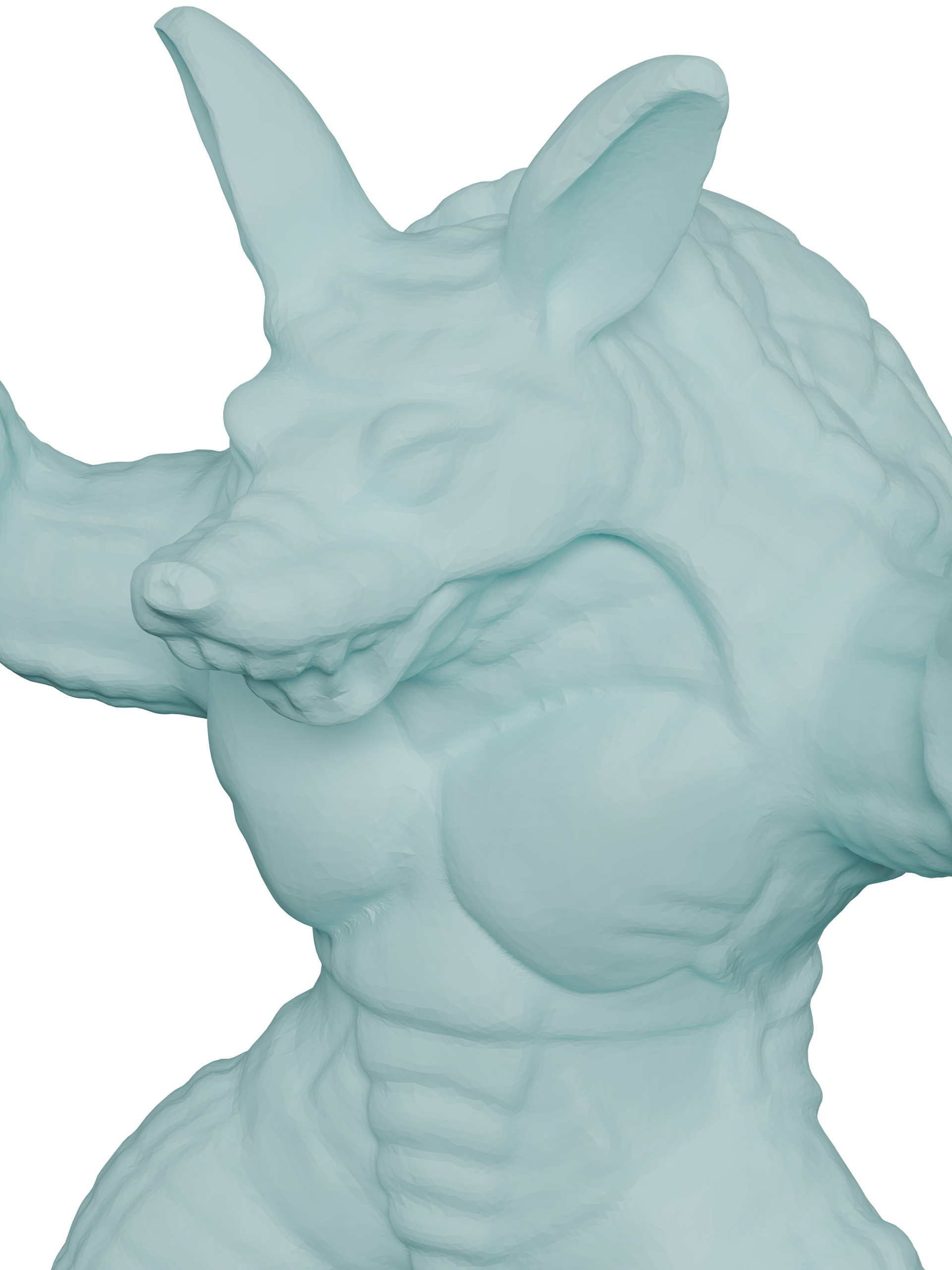 }\\
    \vspace{0.05in}
    \includegraphics[width=0.8in]{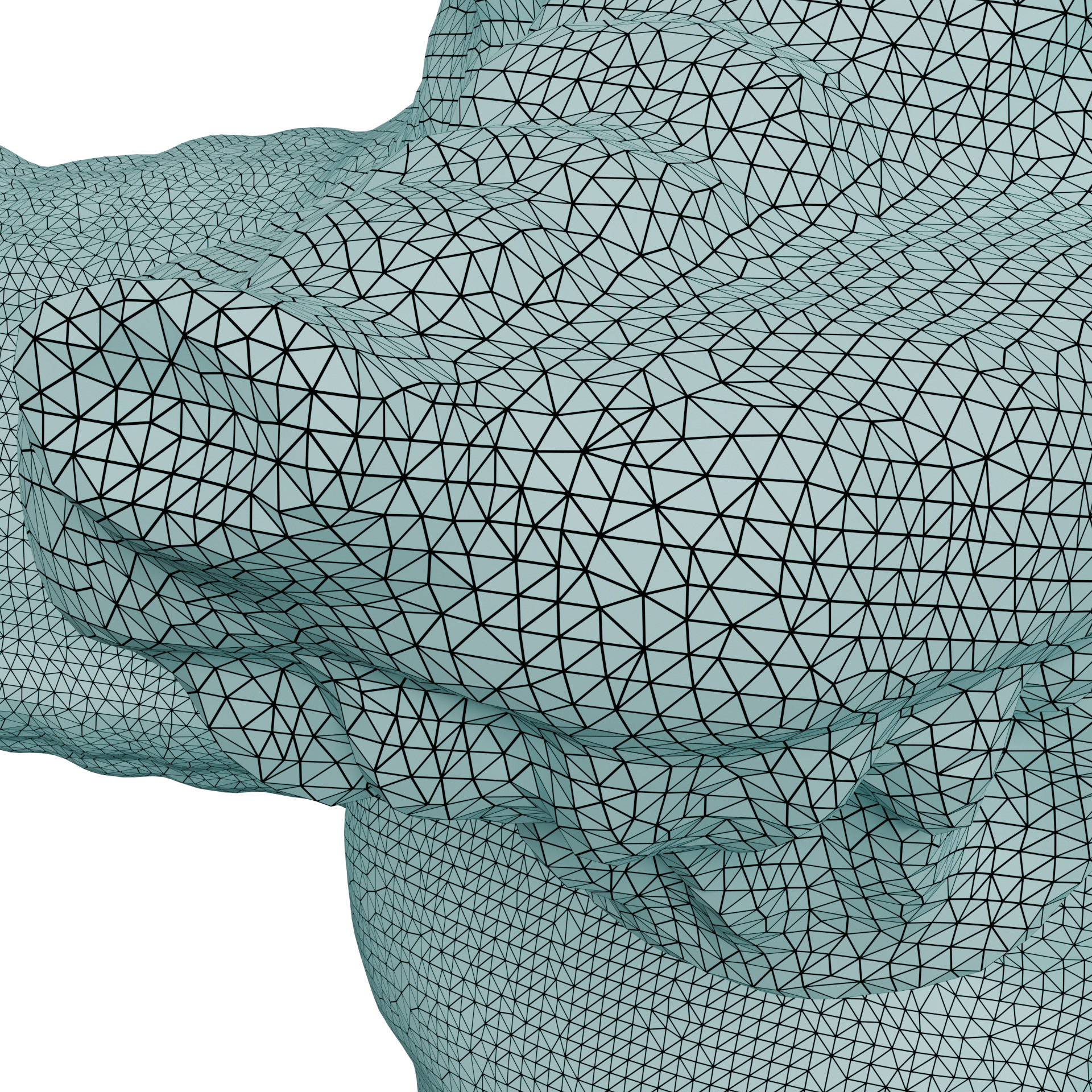 }
    \includegraphics[width=0.8in]{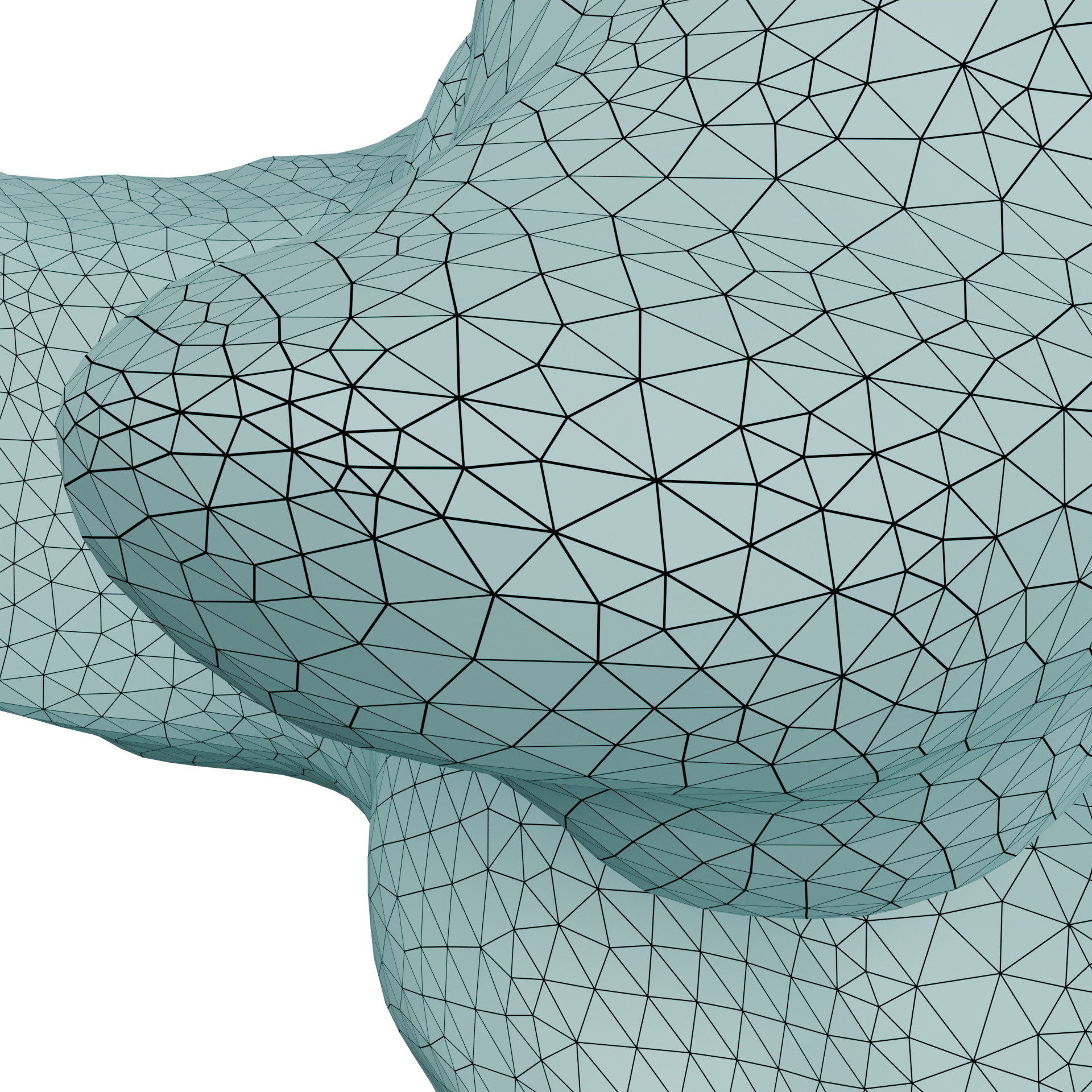 }
    \includegraphics[width=0.8in]{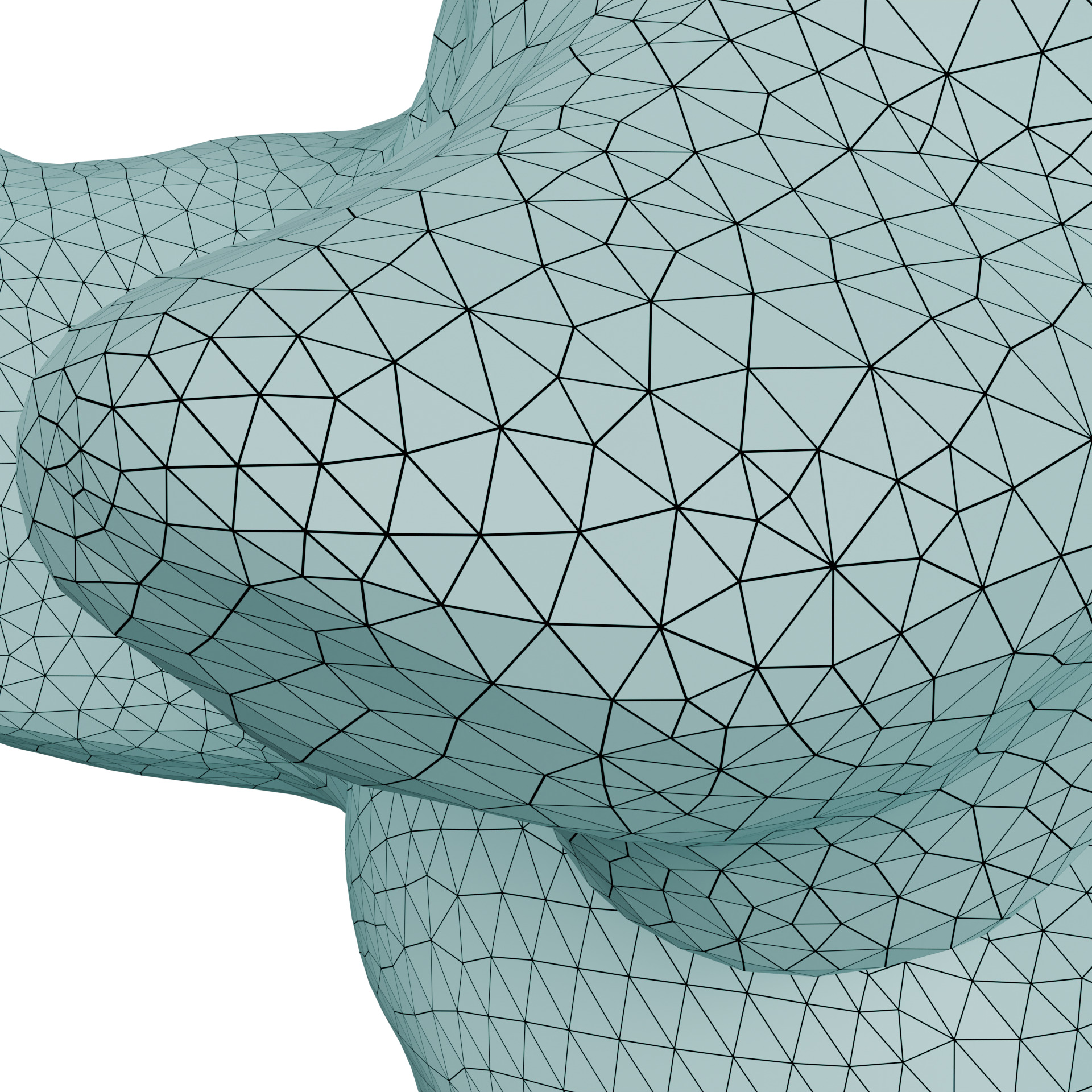}
    \includegraphics[width=0.8in]{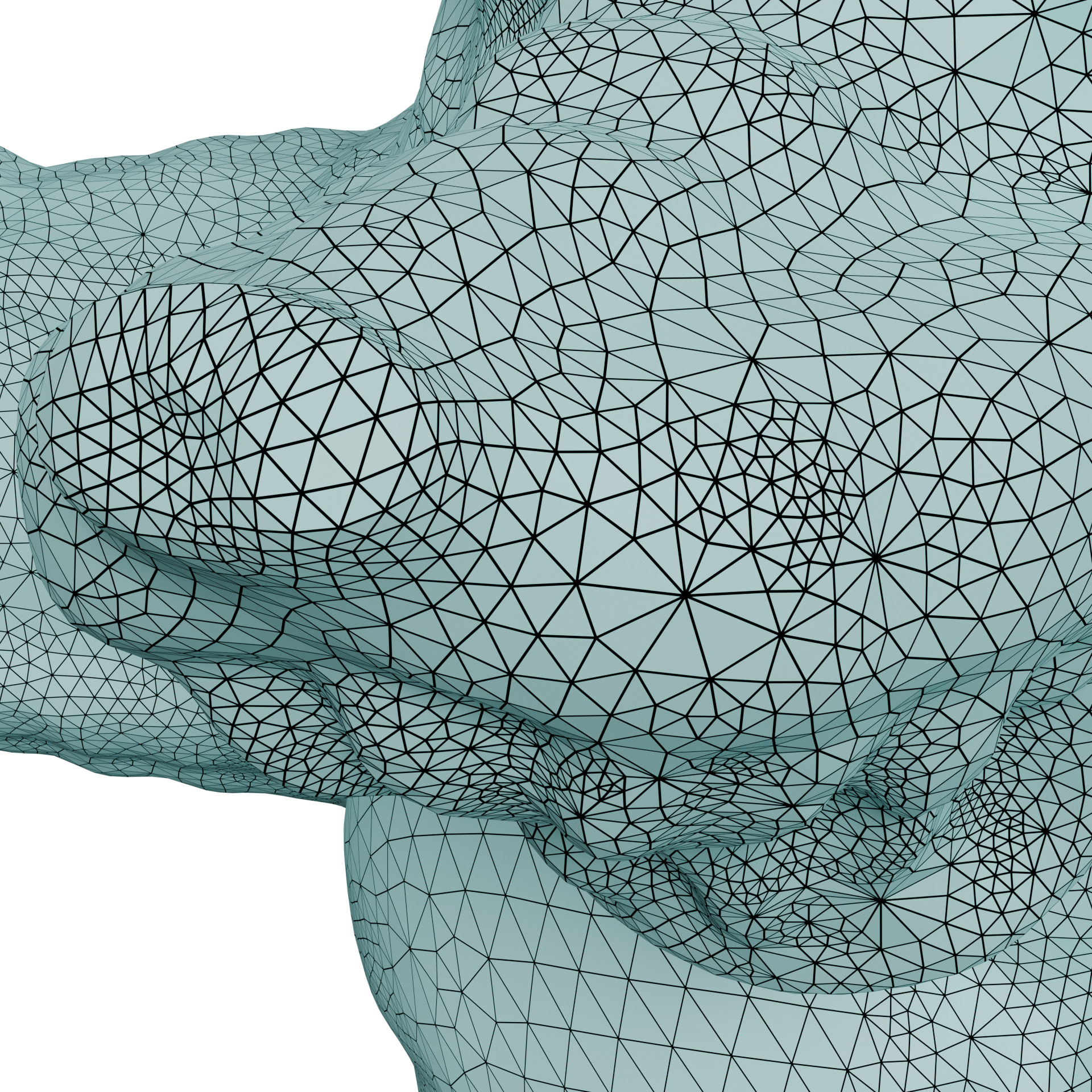 }\\
        \makebox[0.8in]{Ground truth}
    \makebox[0.8in]{DCUDF}
    \makebox[0.8in]{Ours w/o sub.}
    \makebox[0.8in]{Ours w/ sub.}\\

    \caption{Ablation on adaptive mesh subdivision. The mesh is subdivided adaptively in concave/convex regions leading to high accuracy.}
    \label{fig:subdivision}
\end{figure}

\begin{figure}[!ht]
    \centering
    \makebox[0.8in]{Ref. image}
    \makebox[0.8in]{Weights $w_{\text{sa}}$}
    \makebox[0.8in]{Output}
    \makebox[0.8in]{Wireframe}\\
    \includegraphics[width=0.45\textwidth]{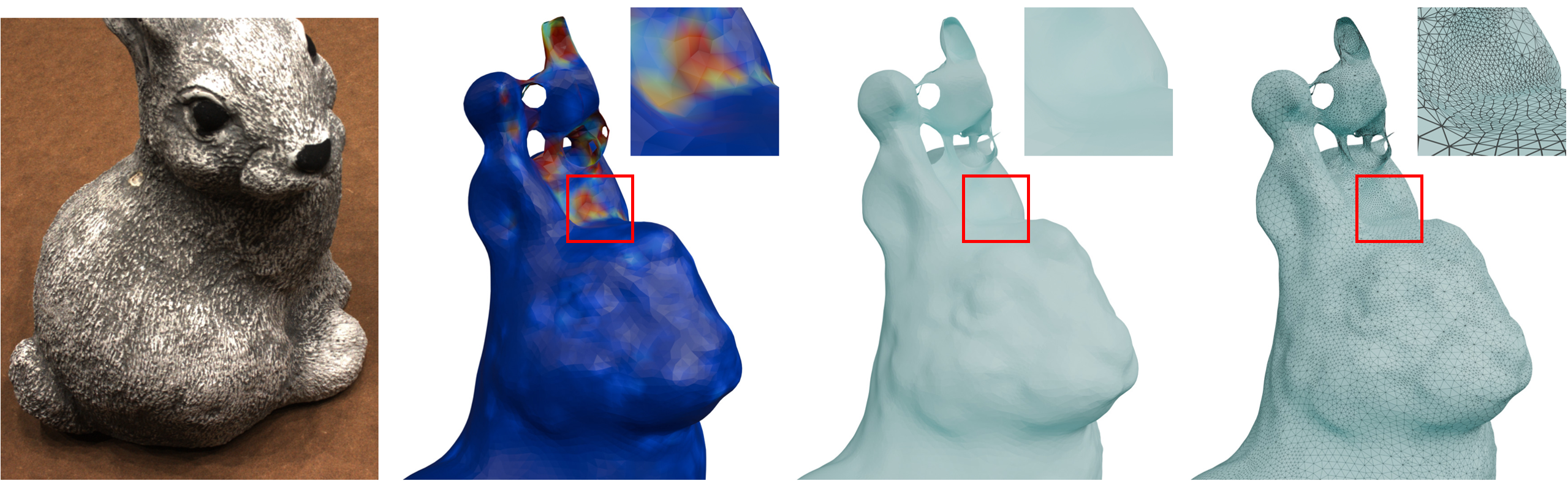}\\
    \caption{Failure case for the Rabbit model from the DTU dataset due to low-quality input UDF, particularly around the ears. The UDF inaccuracies stem from insufficient image coverage of the ear tips, causing high fitting errors during optimization. Focusing on regions with significant distance errors, DCUDF2 overemphasizes the ears by continuously subdividing and assigning high weights there, at the expense of other areas. Consequently, the overall quality of the extracted mesh suffers.}
    \label{fig:bad_fit}
\end{figure}

\begin{figure*}[!htbp]
    \centering
    \begin{scriptsize}
    \makebox[1.2in]{Reference image}
    \makebox[1.2in]{MeshUDF}
    \makebox[1.2in]{DMUDF}
    \makebox[1.2in]{DCUDF}
    \makebox[1.2in]{DCUDF2}\\
    \end{scriptsize}
    \includegraphics[width=1.2in]{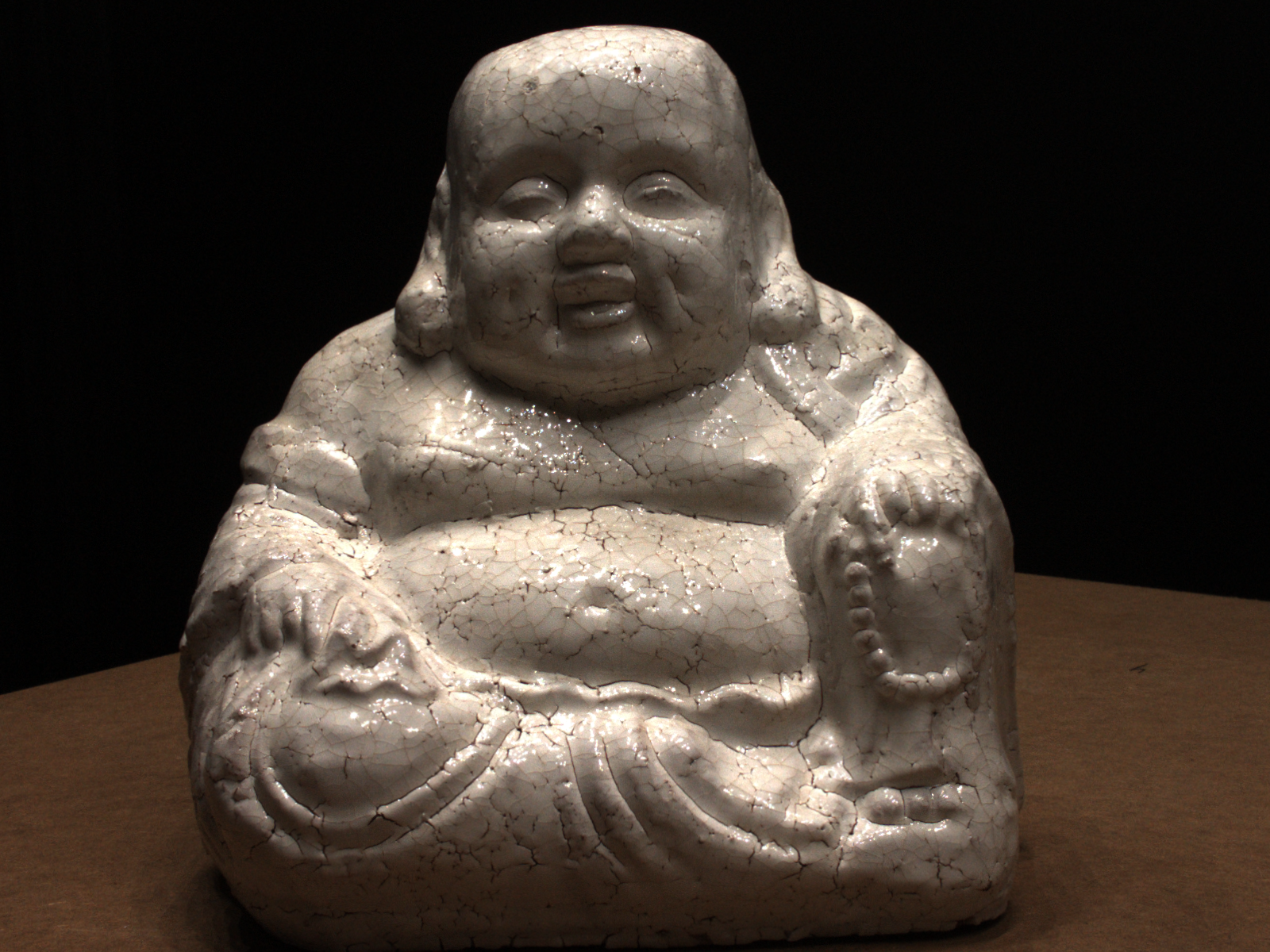 }
    \includegraphics[width=1.2in]{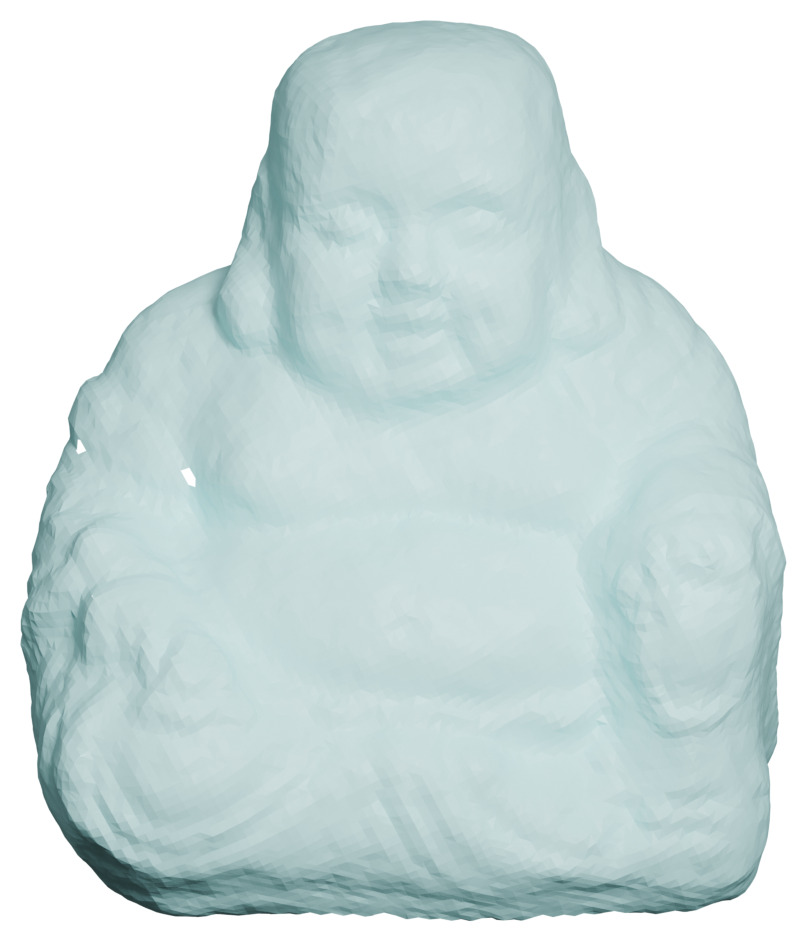 }
    \includegraphics[width=1.2in]{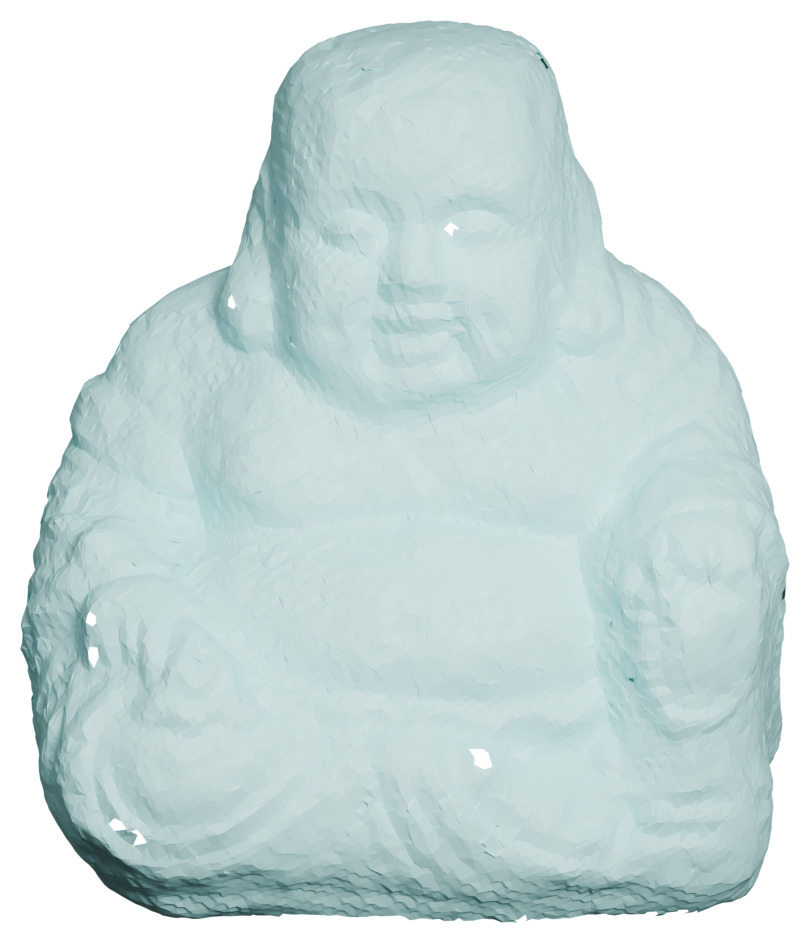 }
    \includegraphics[width=1.2in]{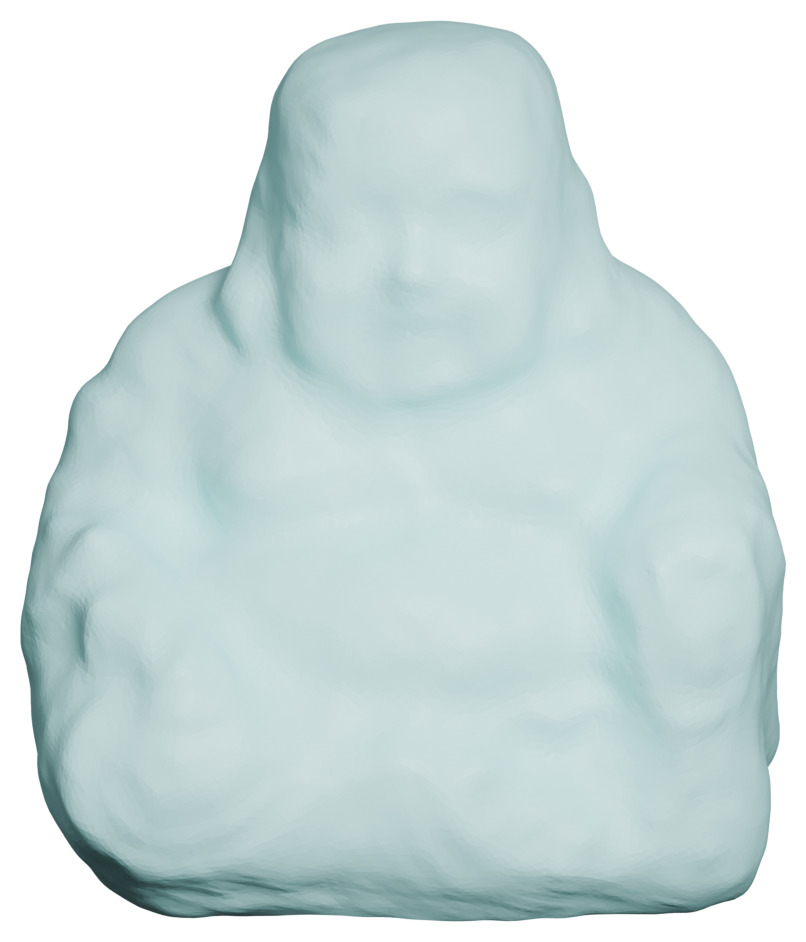 }
    \includegraphics[width=1.2in]{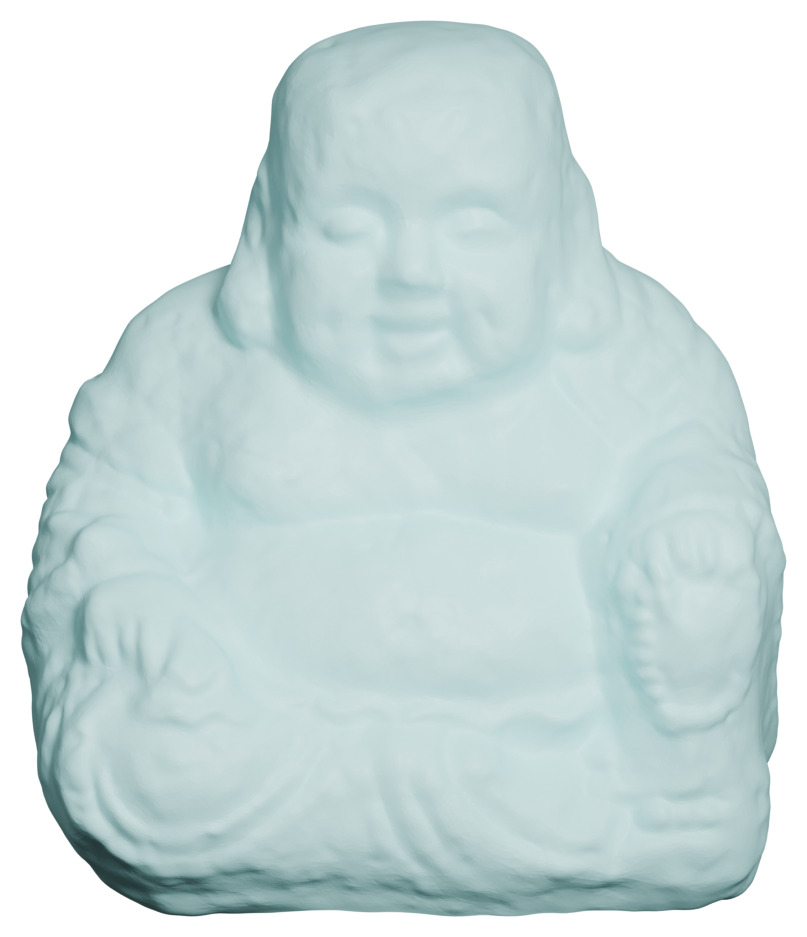 }\\
    \includegraphics[width=1.2in]{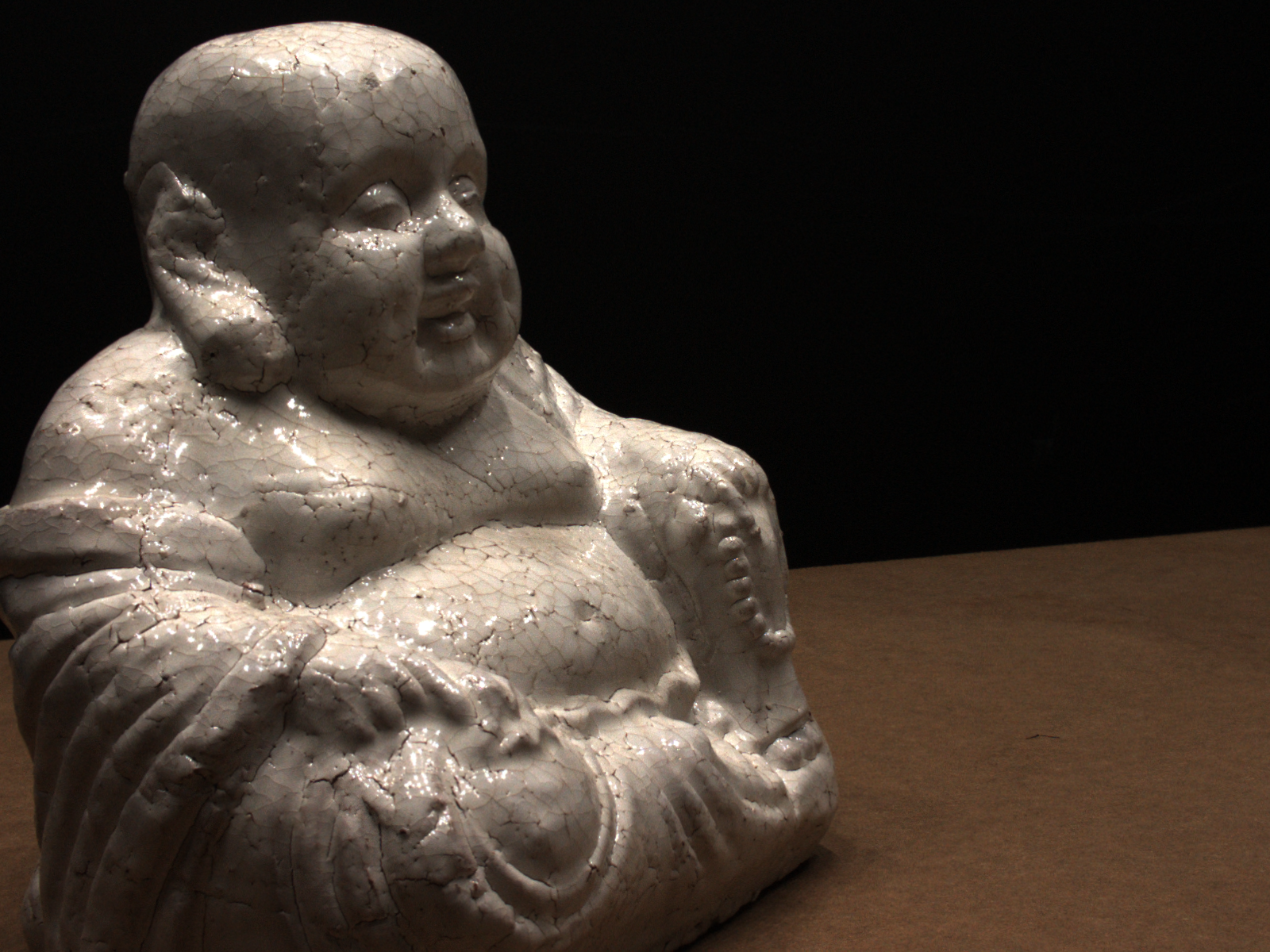 }
    \includegraphics[width=1.2in]{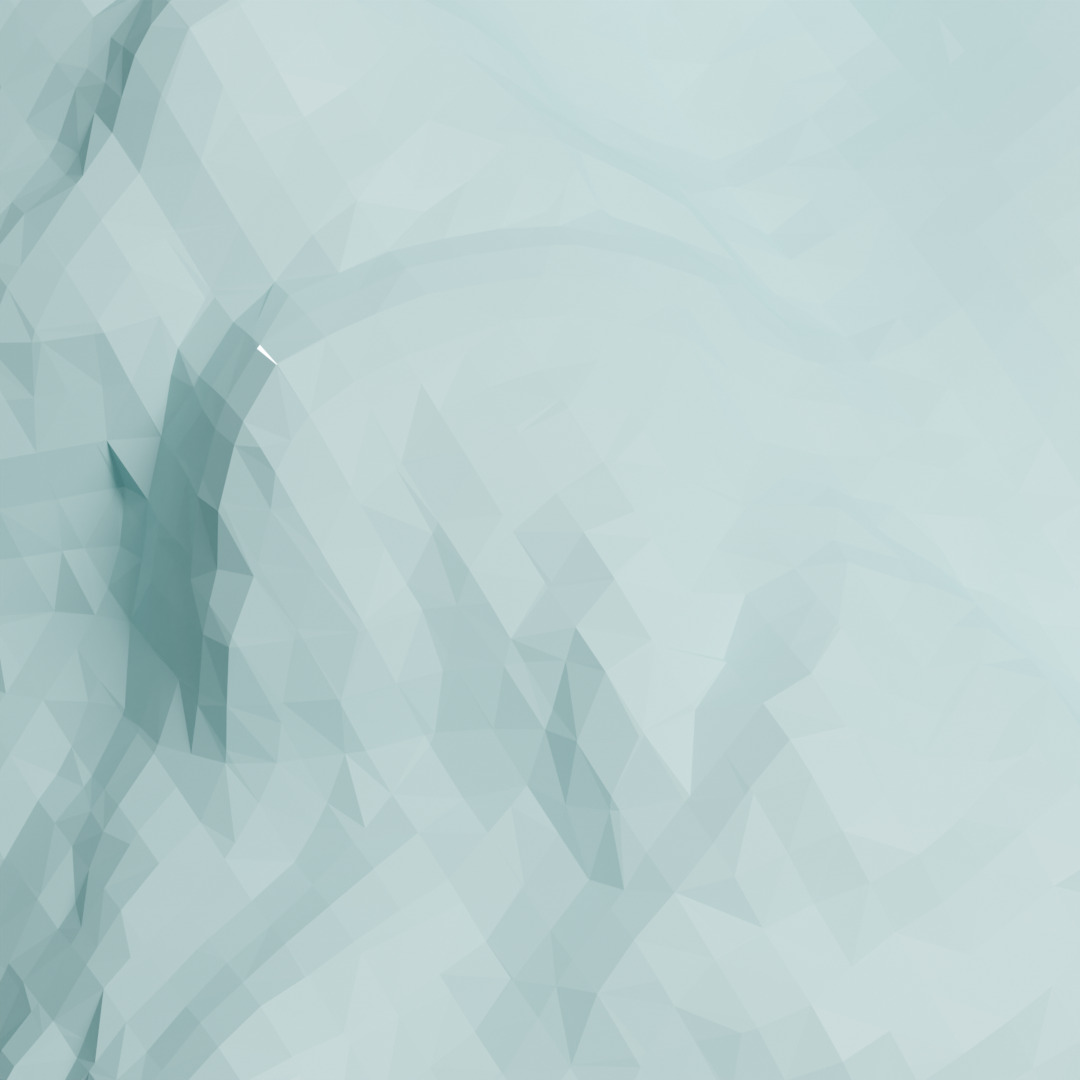 }
    \includegraphics[width=1.2in]{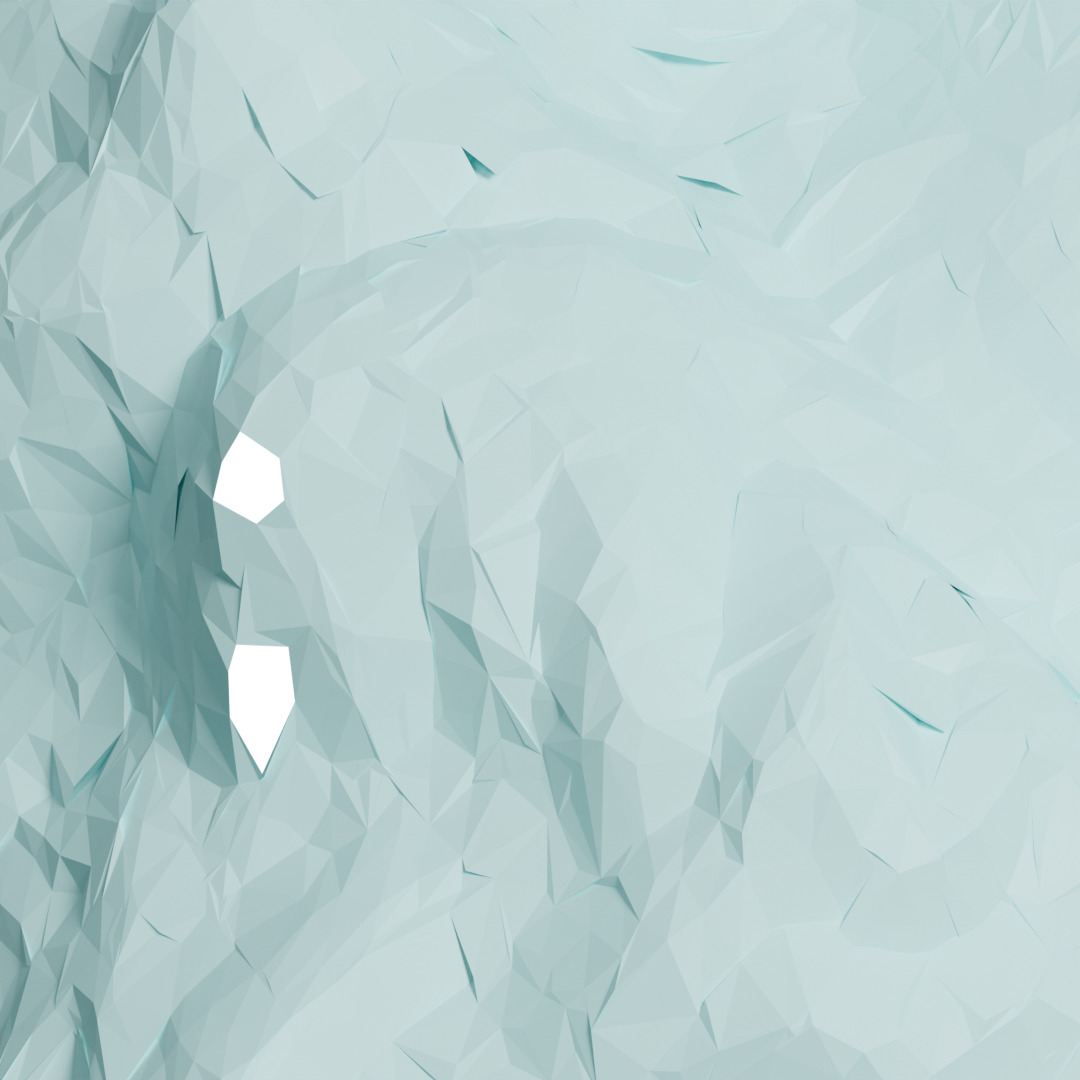 }
    \includegraphics[width=1.2in]{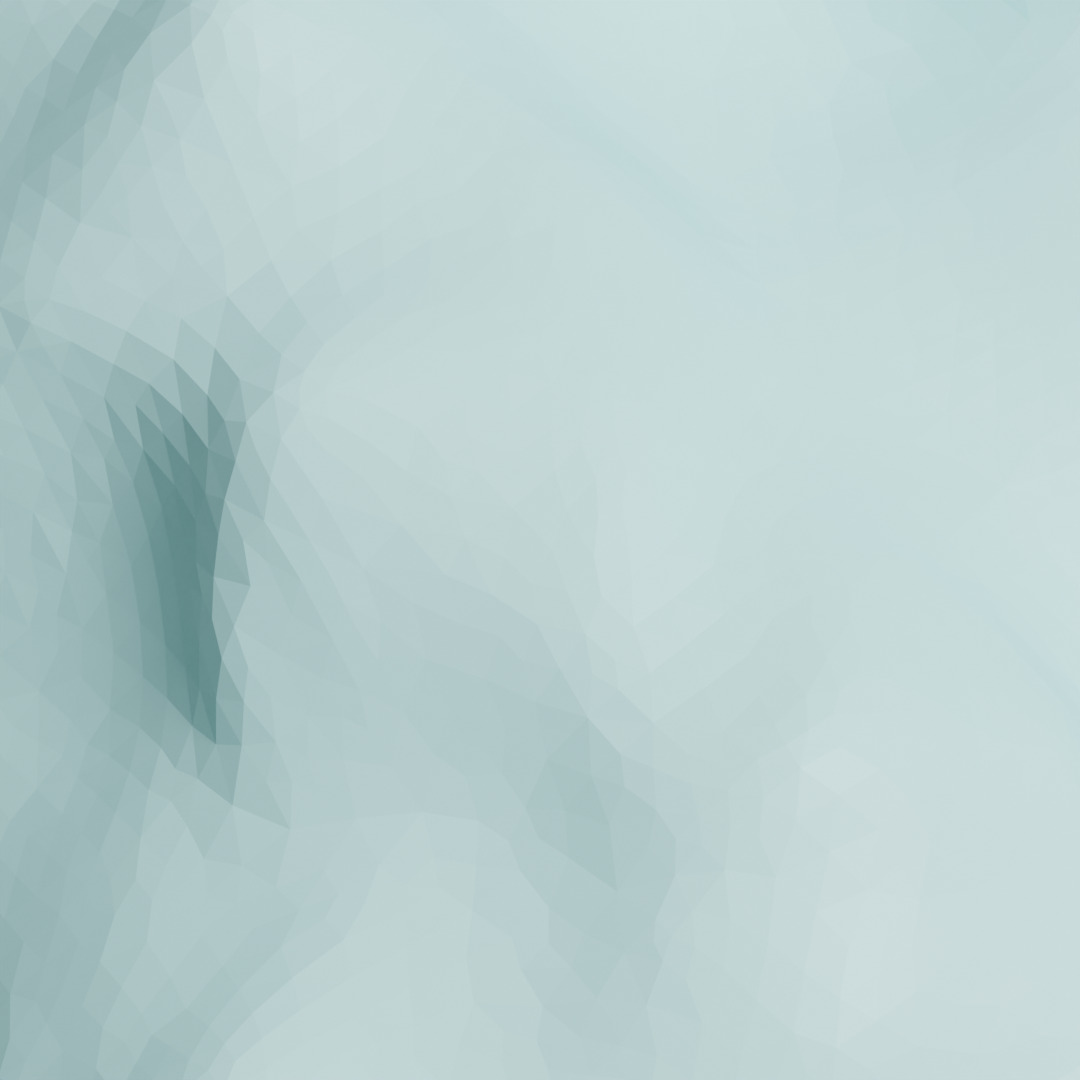 }
    \includegraphics[width=1.2in]{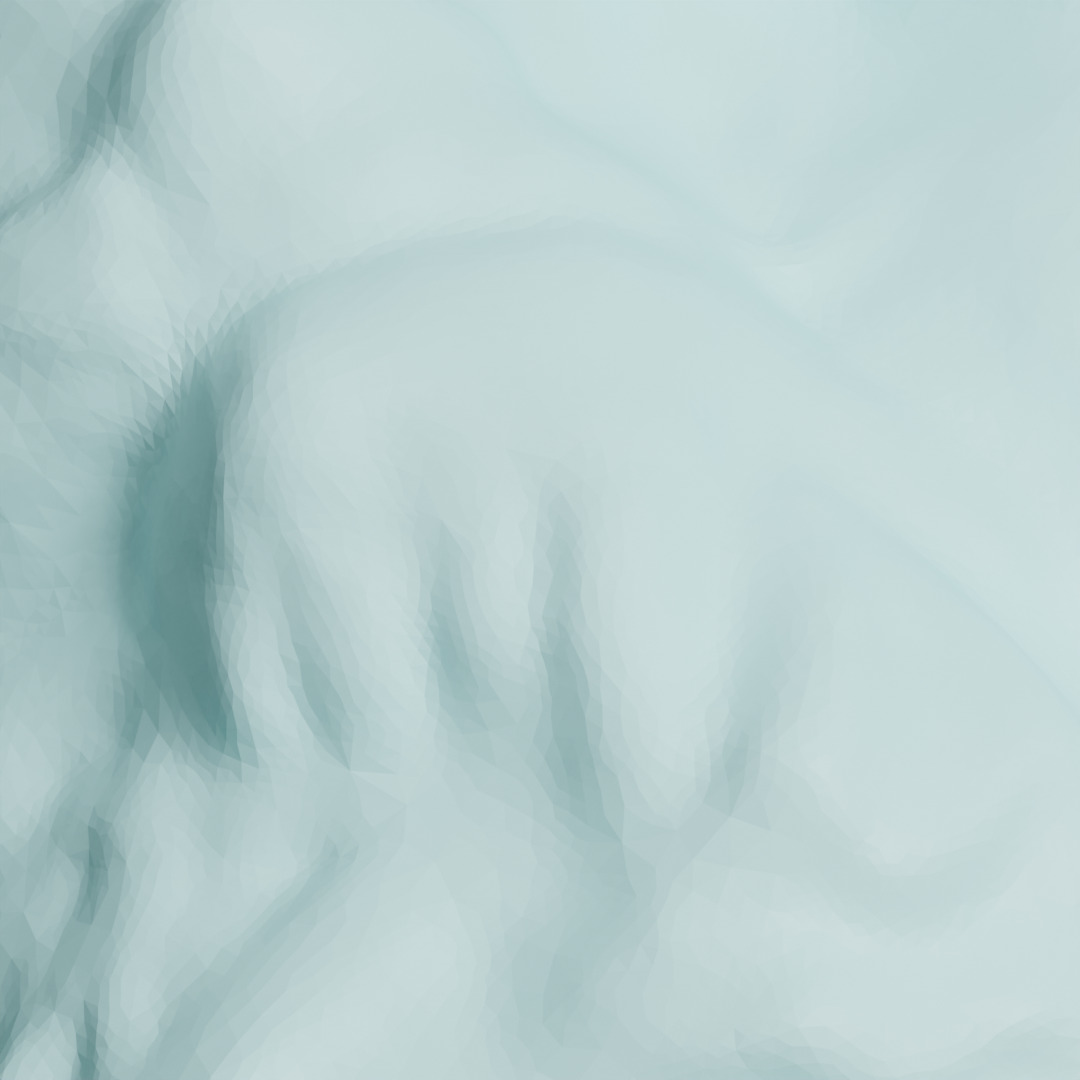 }\\ 
    \includegraphics[width=1.2in]{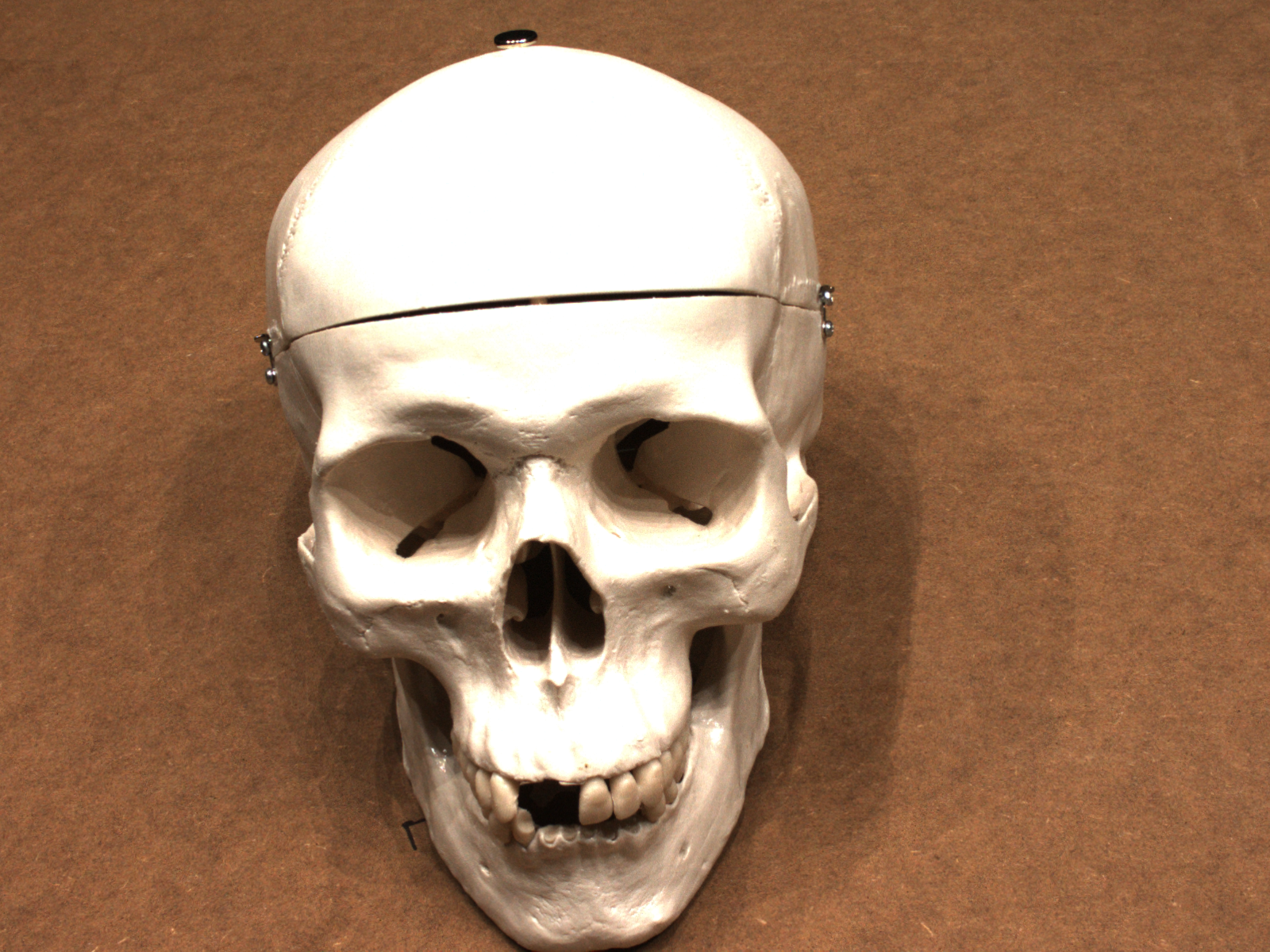 }
  \includegraphics[width=1.2in]{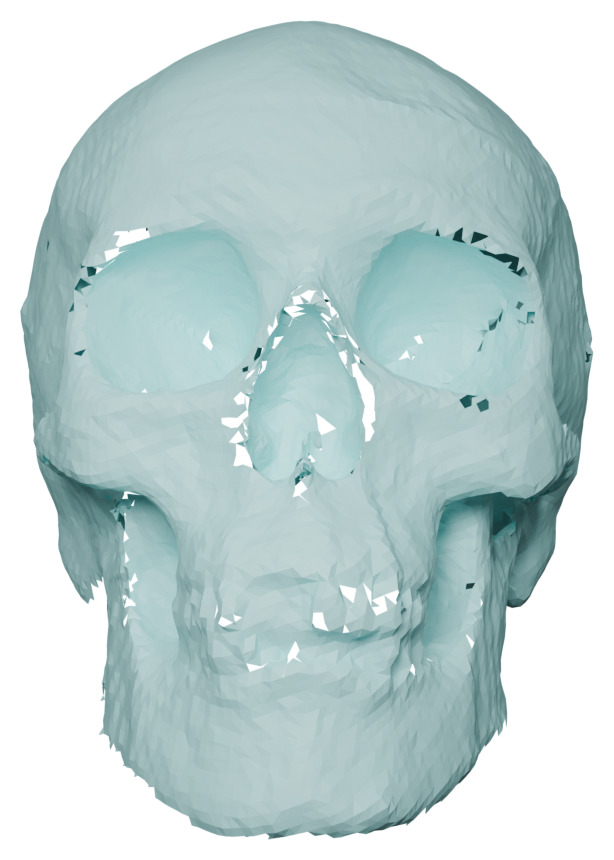 }
    \includegraphics[width=1.2in]{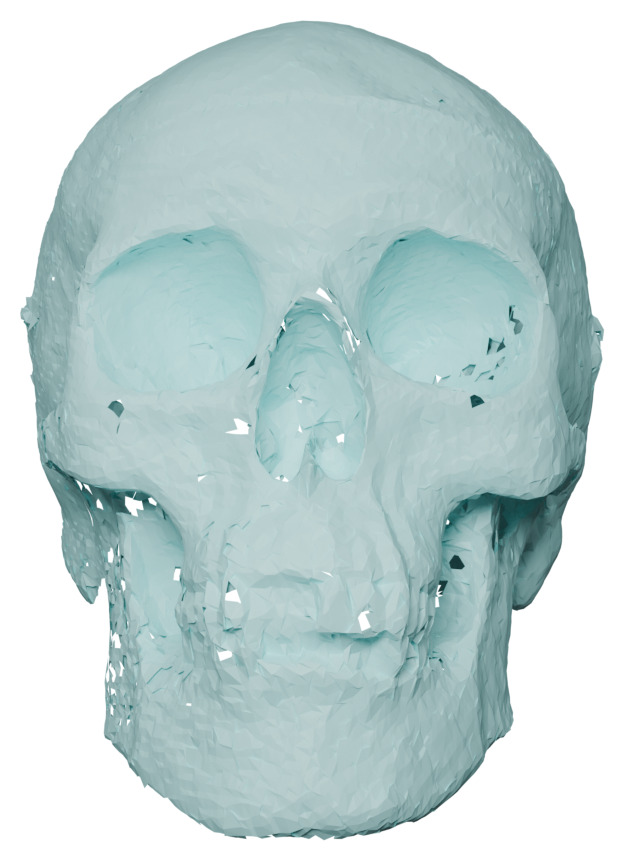 }
    \includegraphics[width=1.2in]{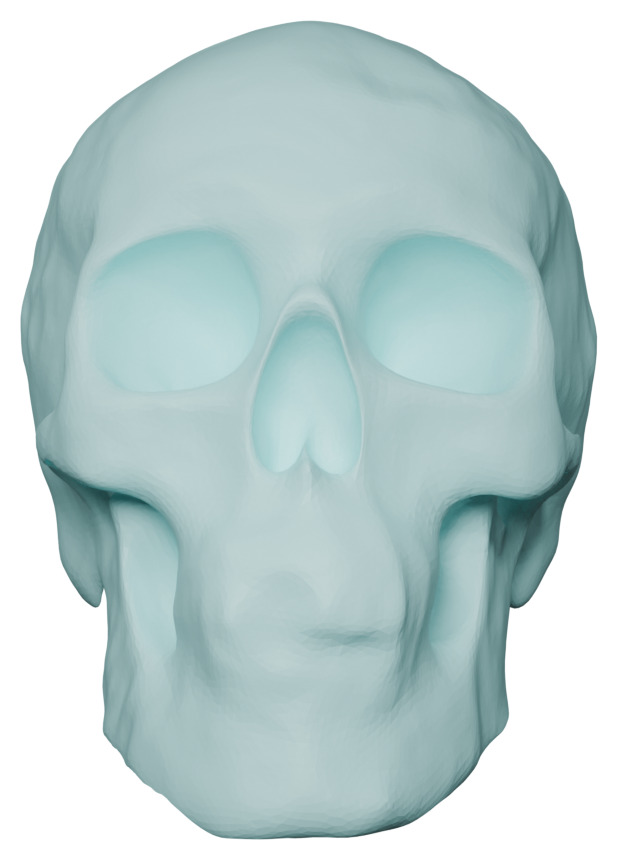 }
    \includegraphics[width=1.2in]{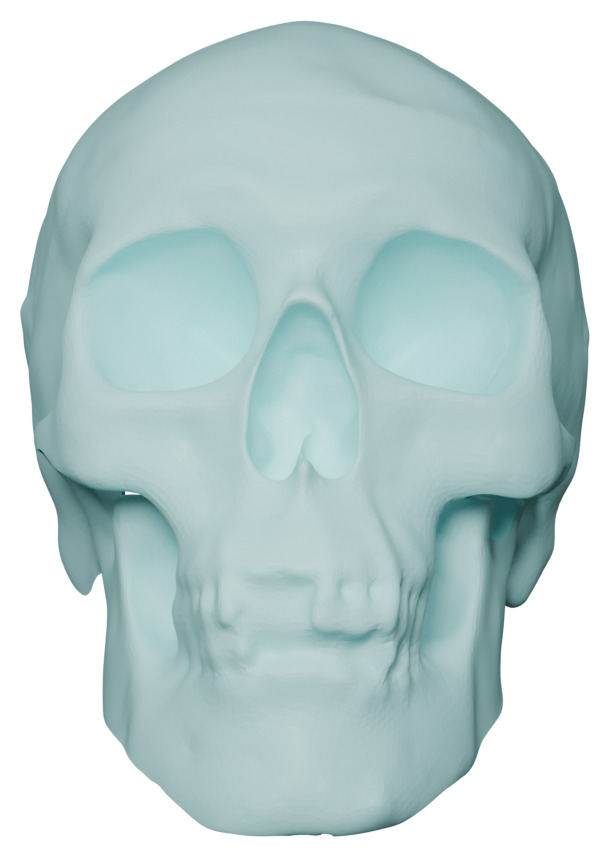 }\\
    \includegraphics[width=1.2in]{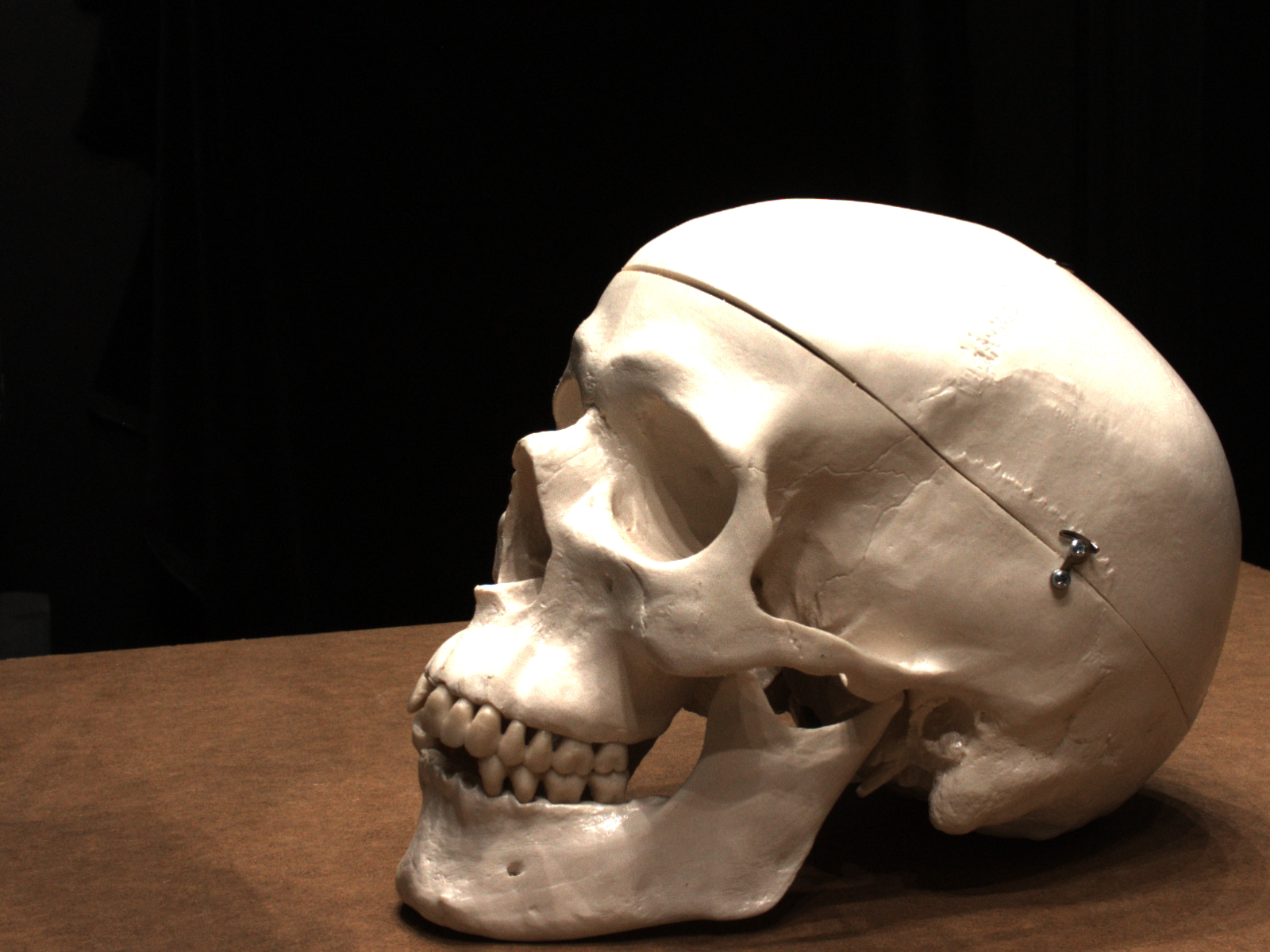 }
    \includegraphics[width=1.2in]{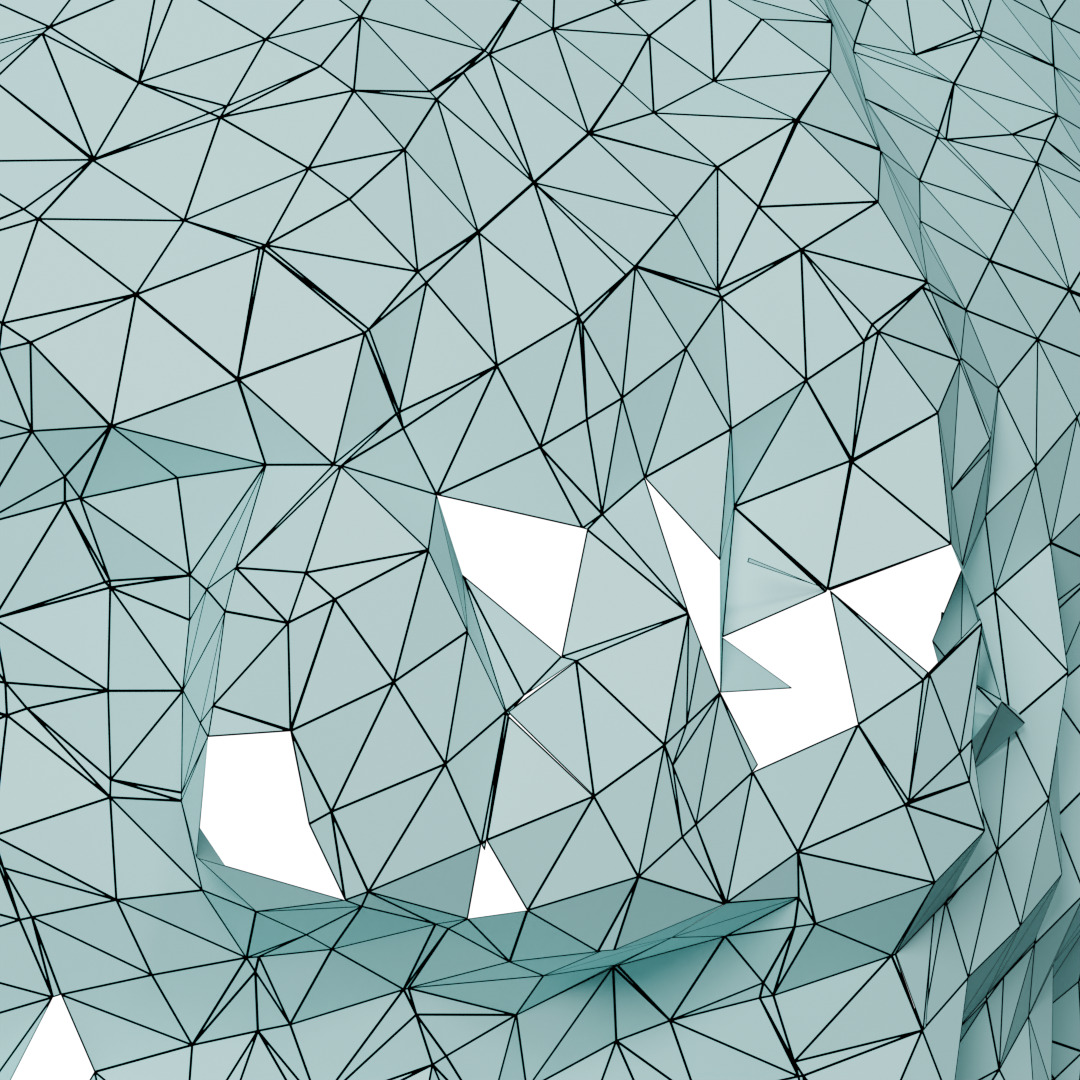 }
    \includegraphics[width=1.2in]{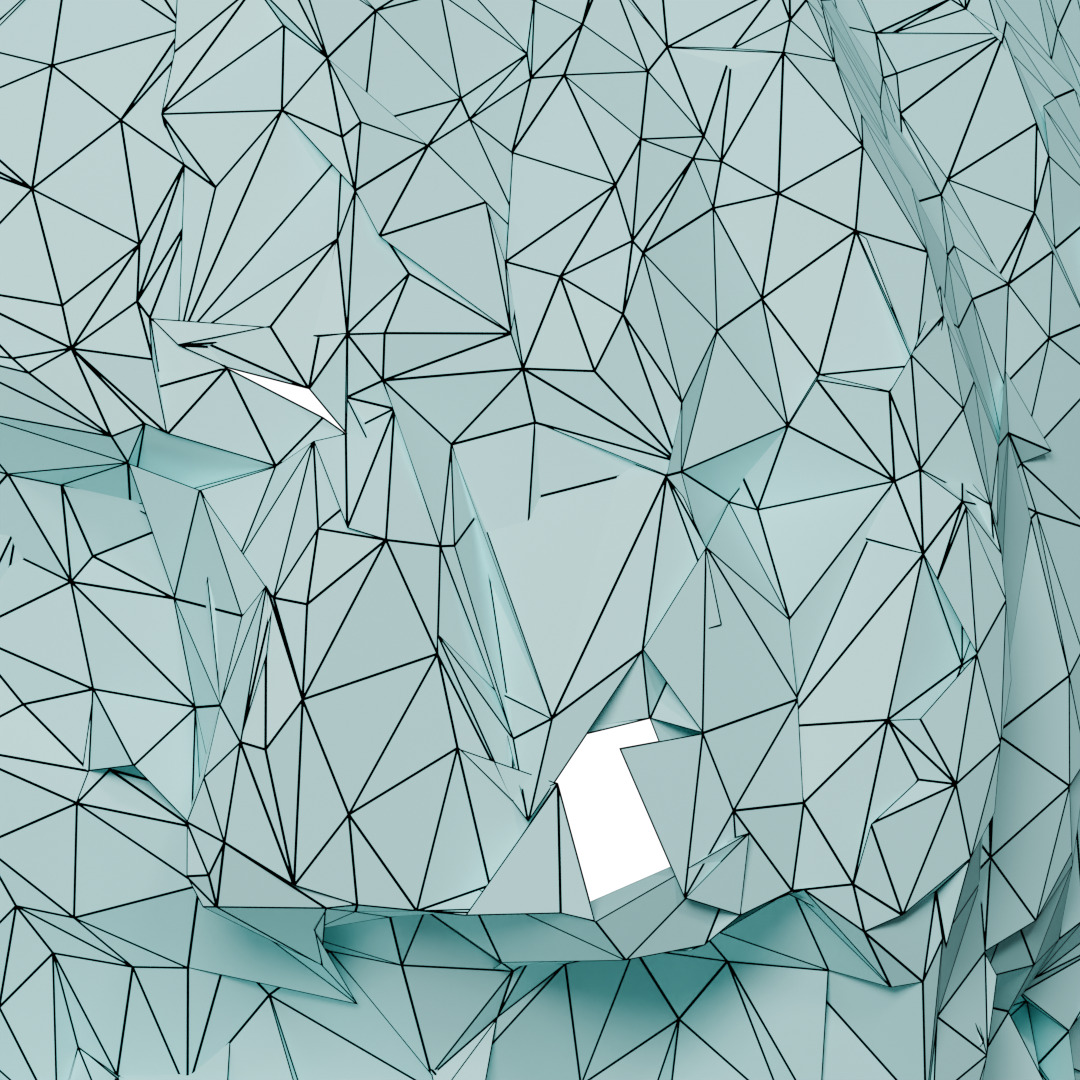 }
    \includegraphics[width=1.2in]{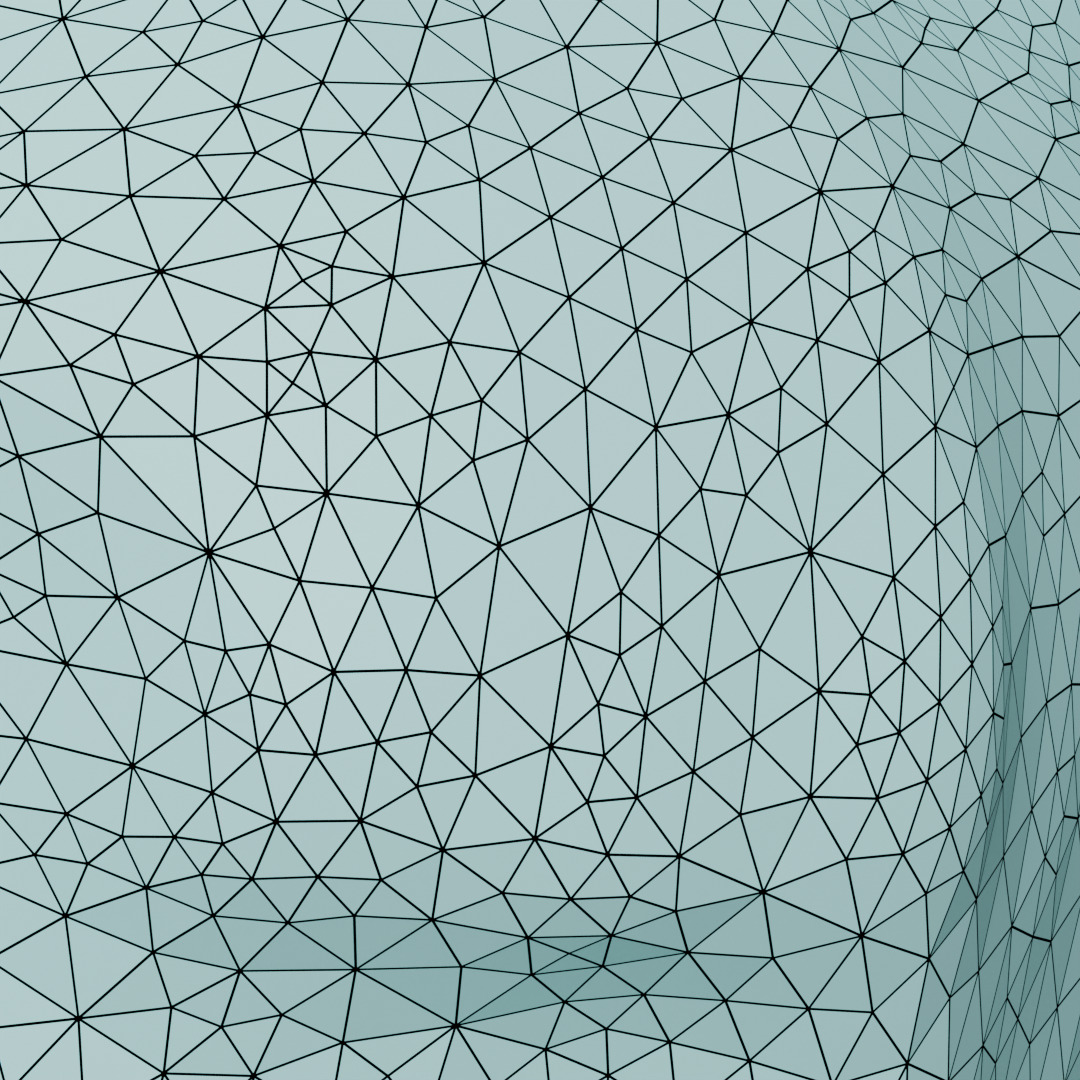 }
    \includegraphics[width=1.2in]{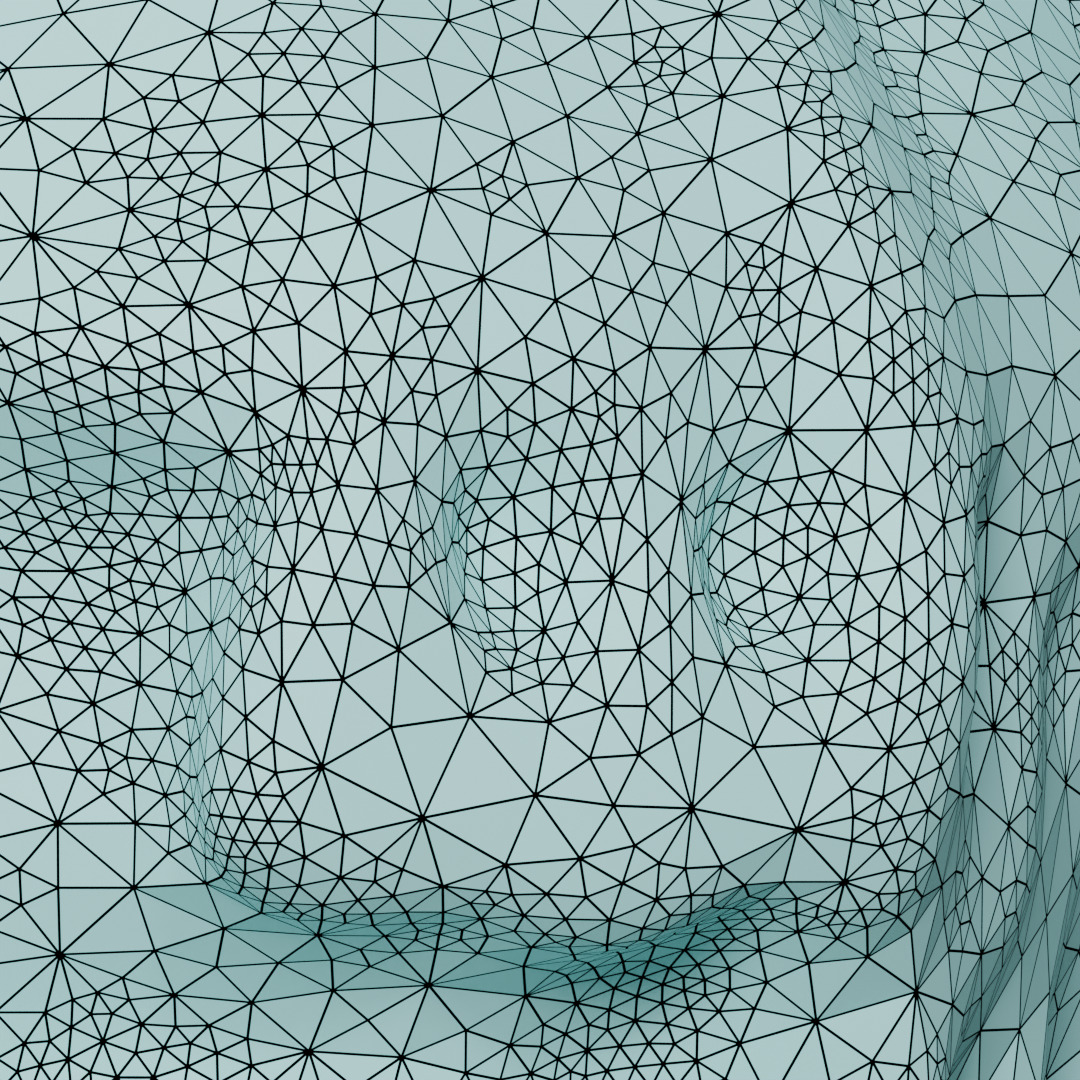 }\\
    \includegraphics[width=1.2in]{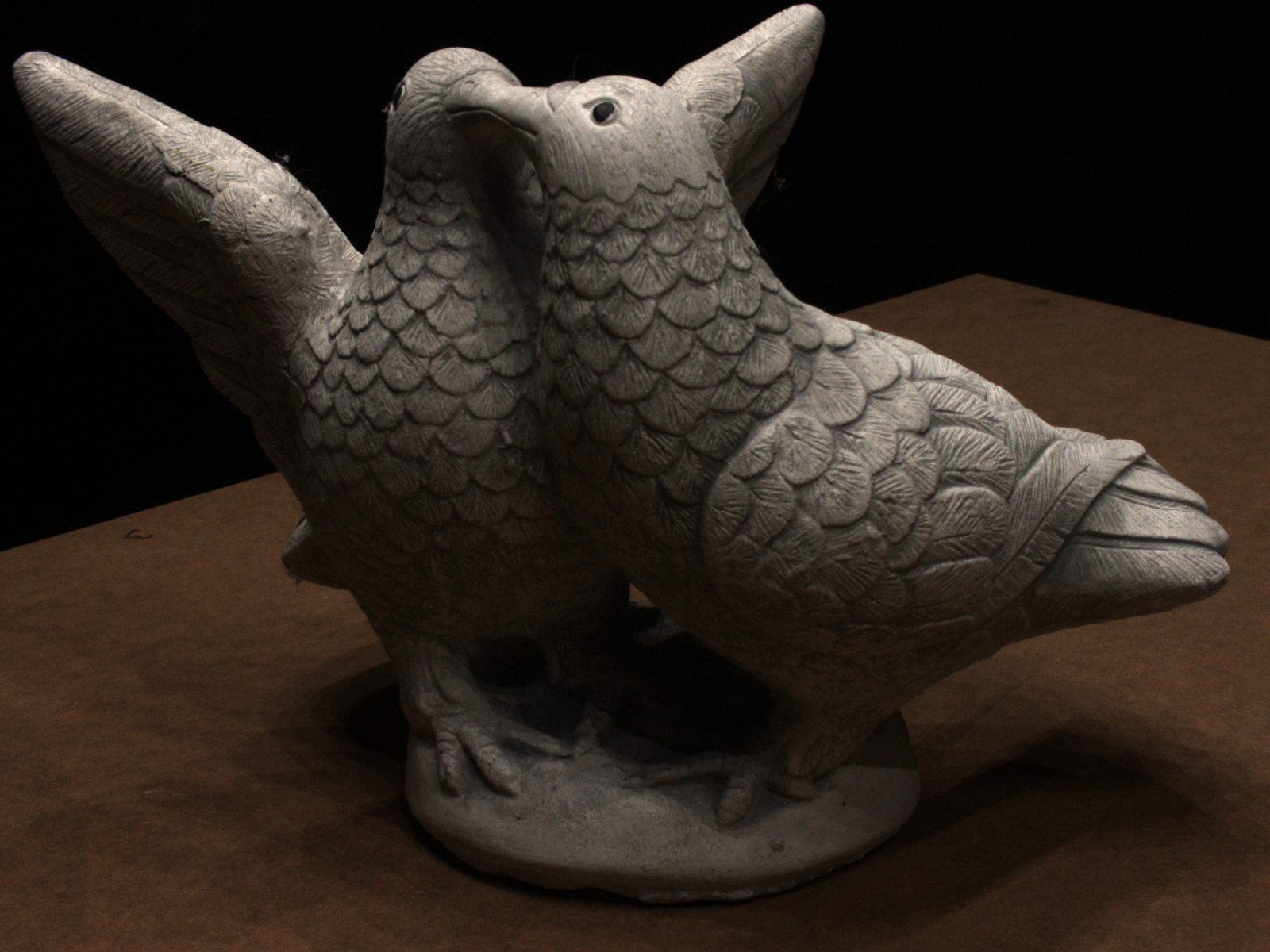 }
  \includegraphics[width=1.2in]{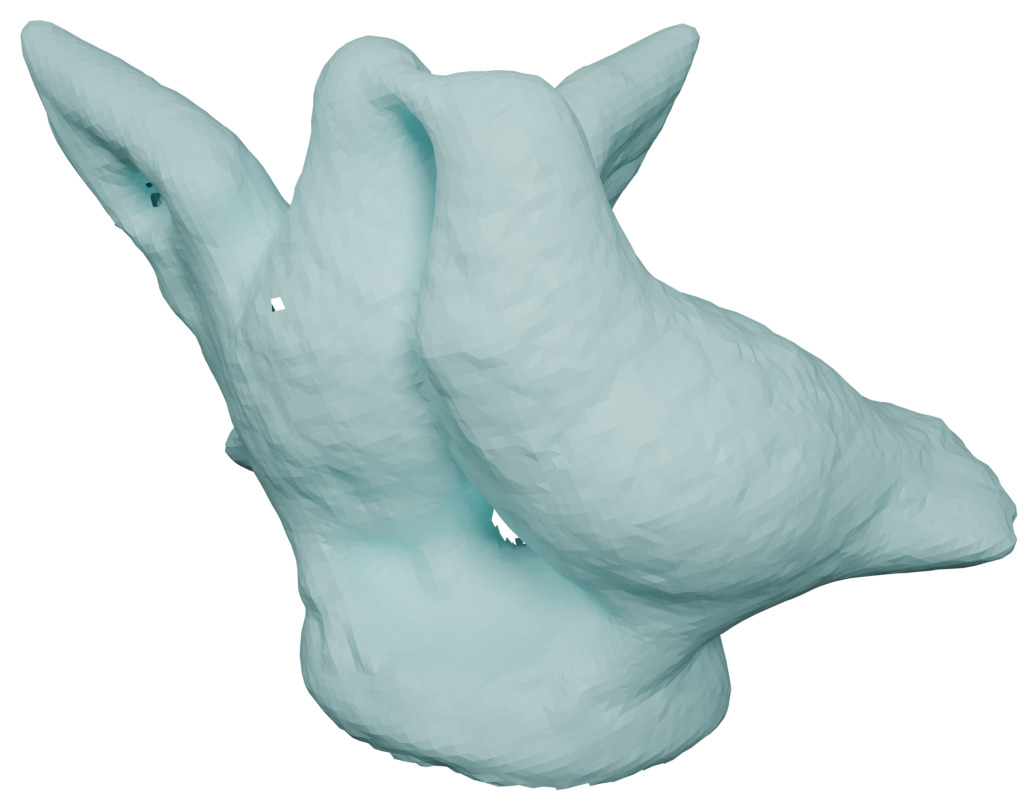 }
    \includegraphics[width=1.2in]{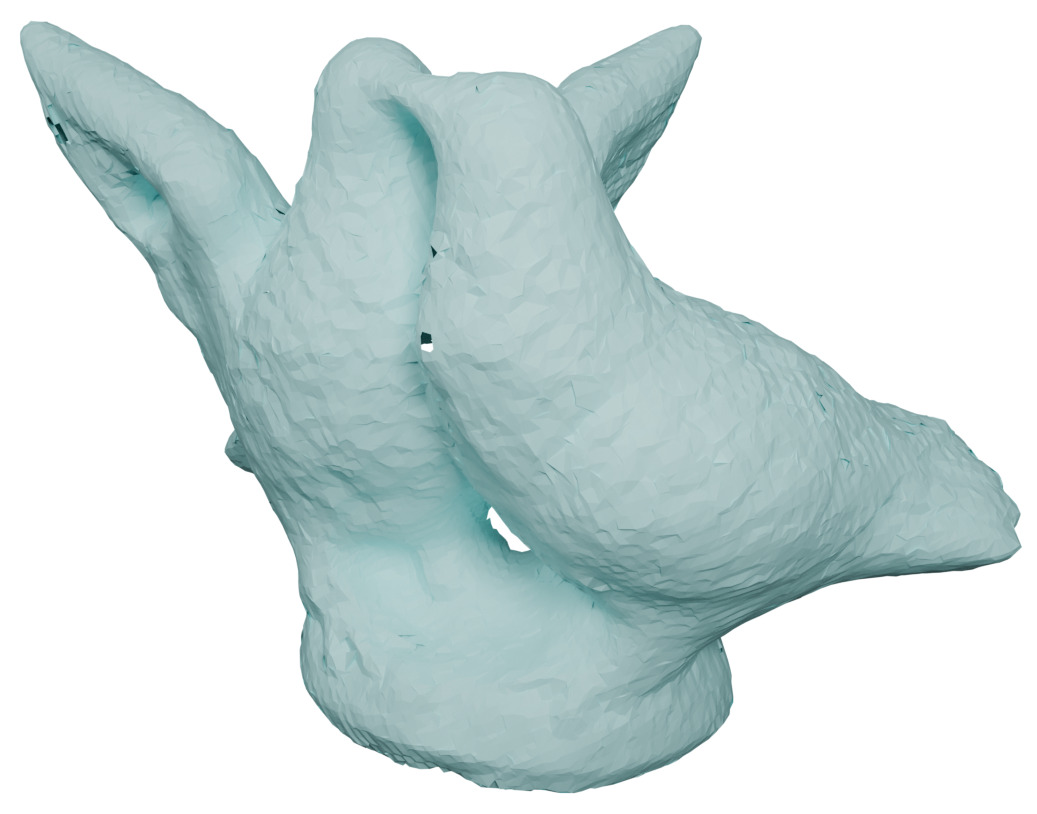 }
    \includegraphics[width=1.2in]{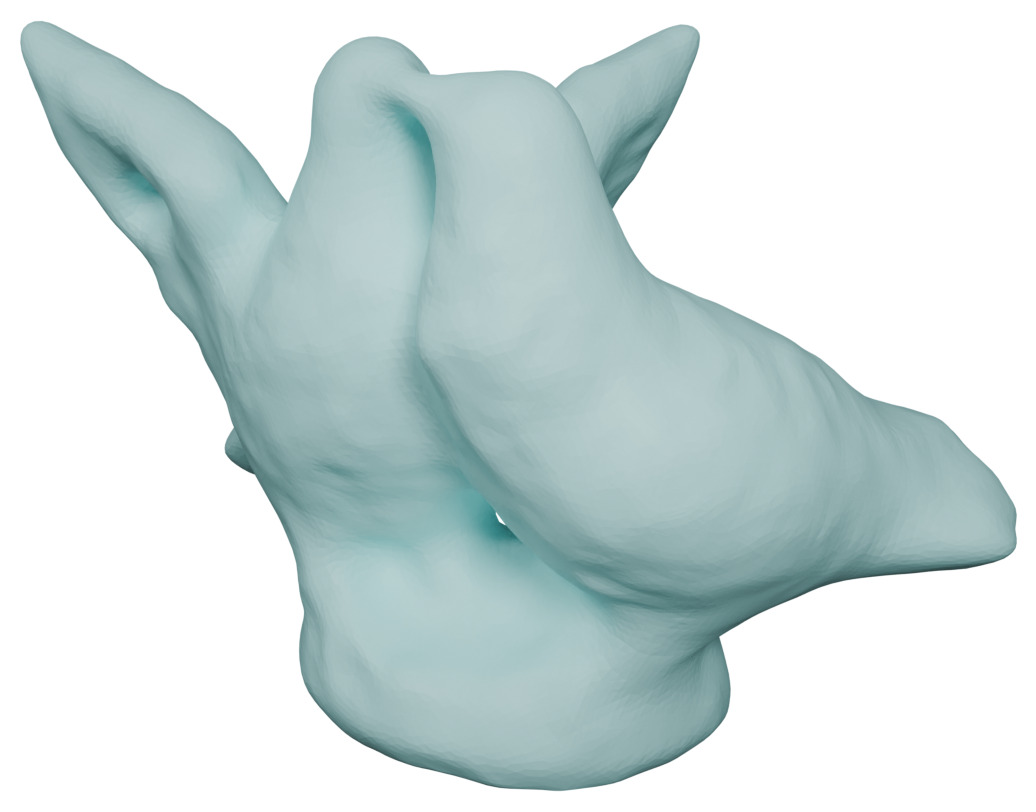 }
    \includegraphics[width=1.2in]{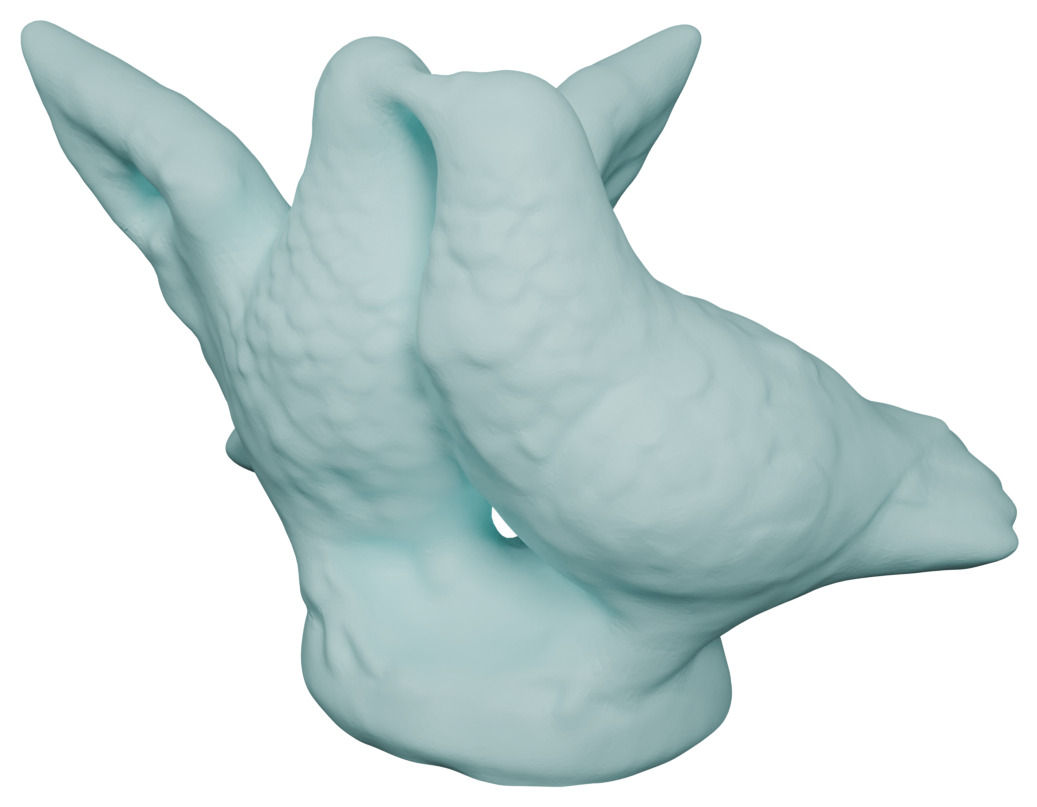 }\\
    \includegraphics[width=1.2in]{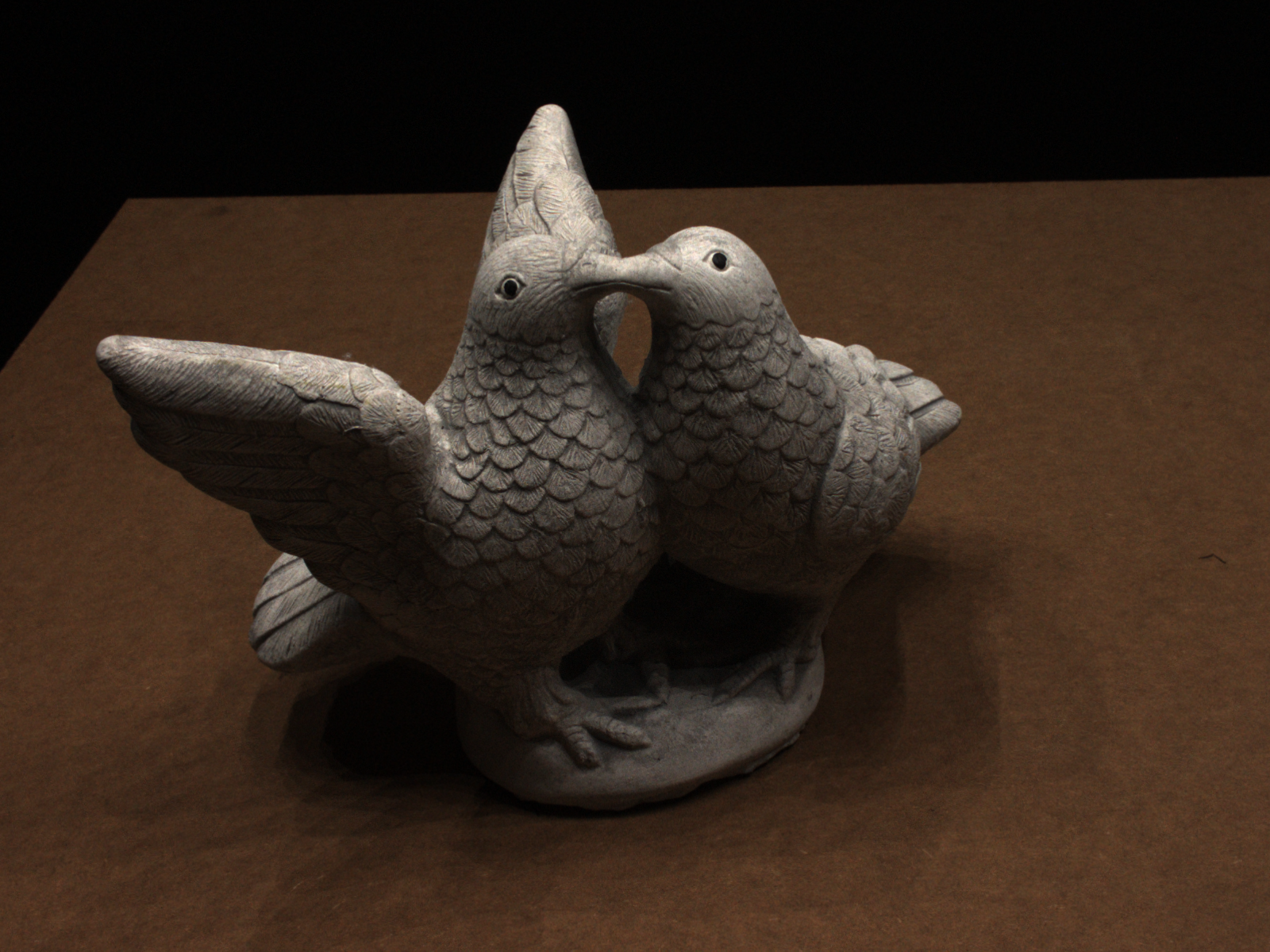 }
    \includegraphics[width=1.2in]{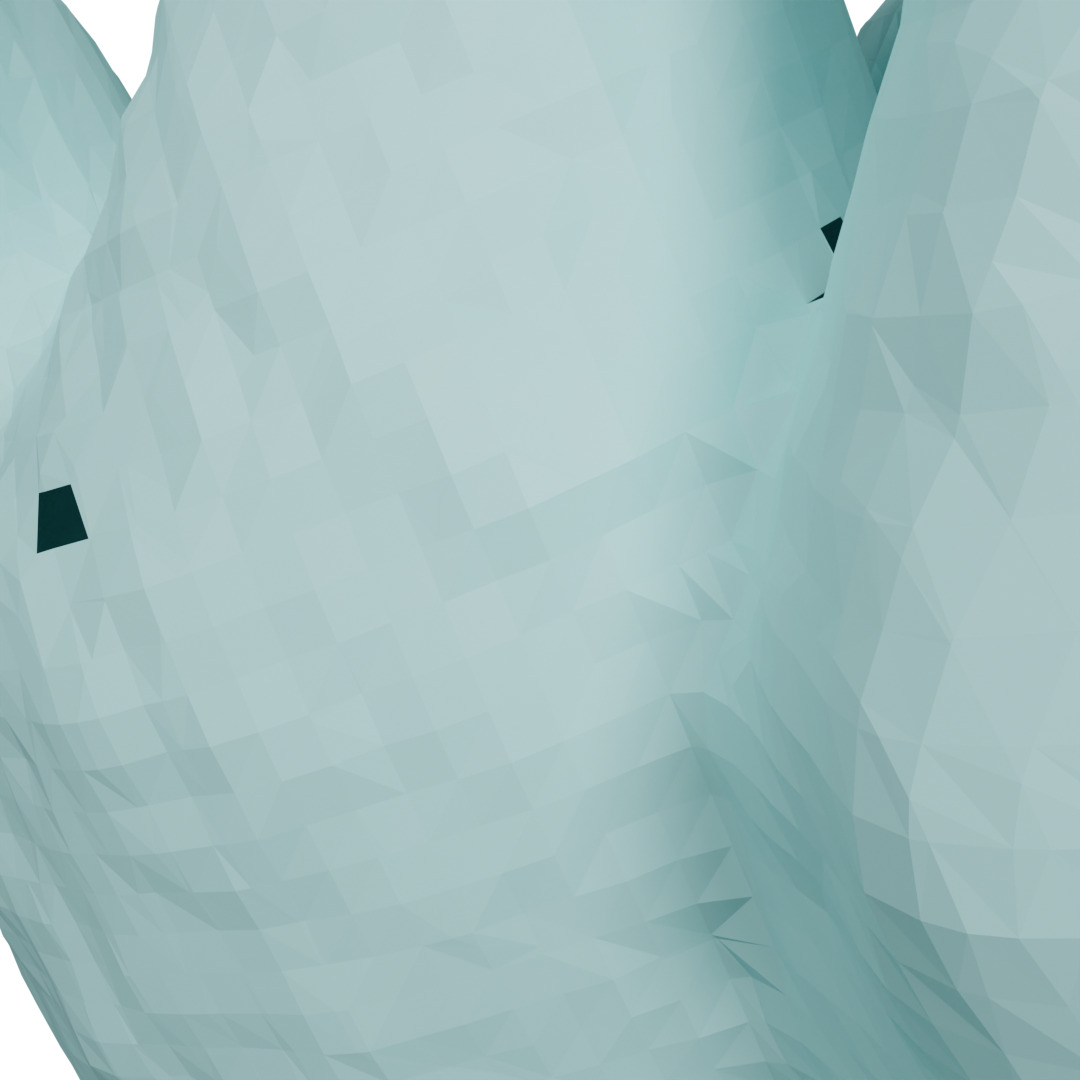 }
    \includegraphics[width=1.2in]{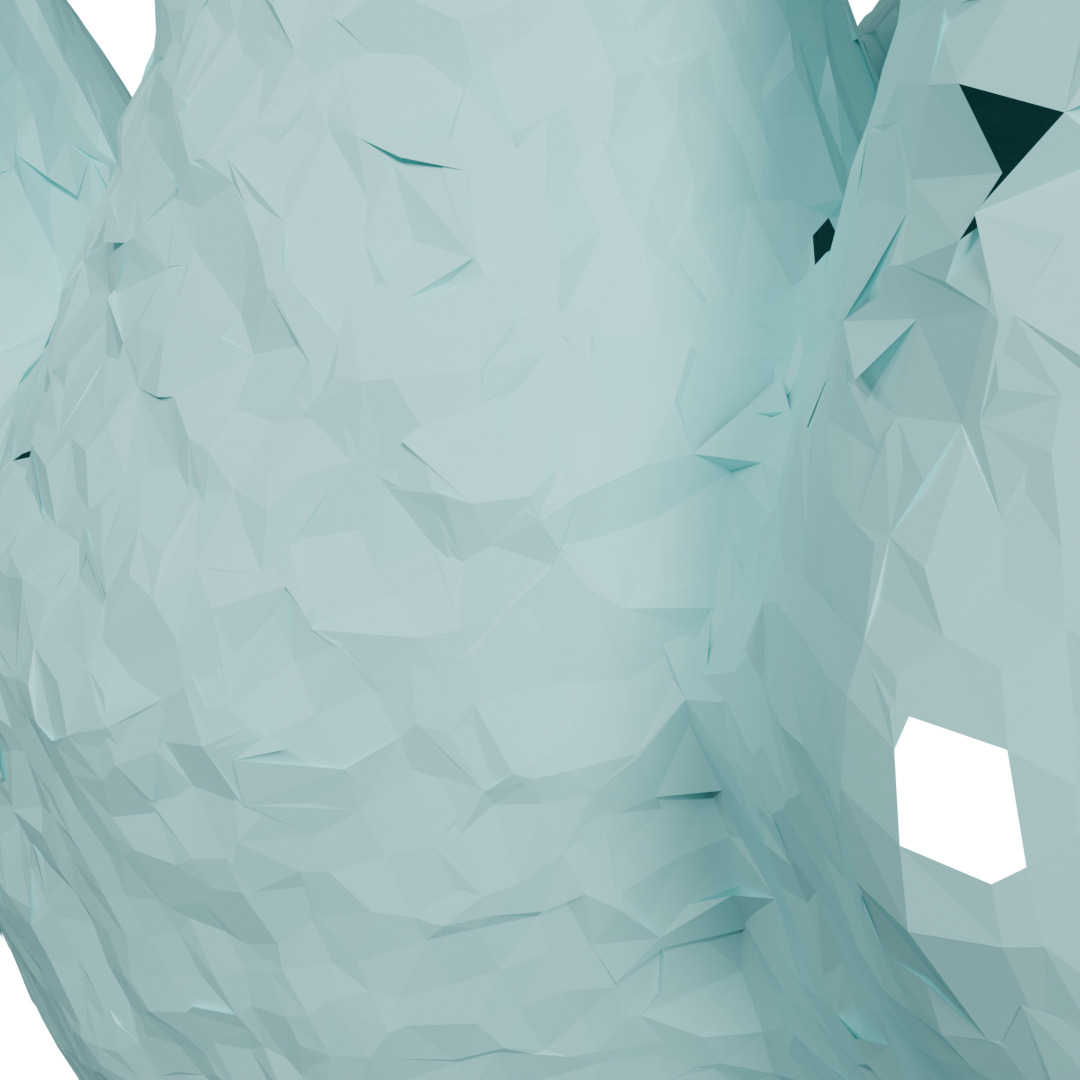 }
    \includegraphics[width=1.2in]{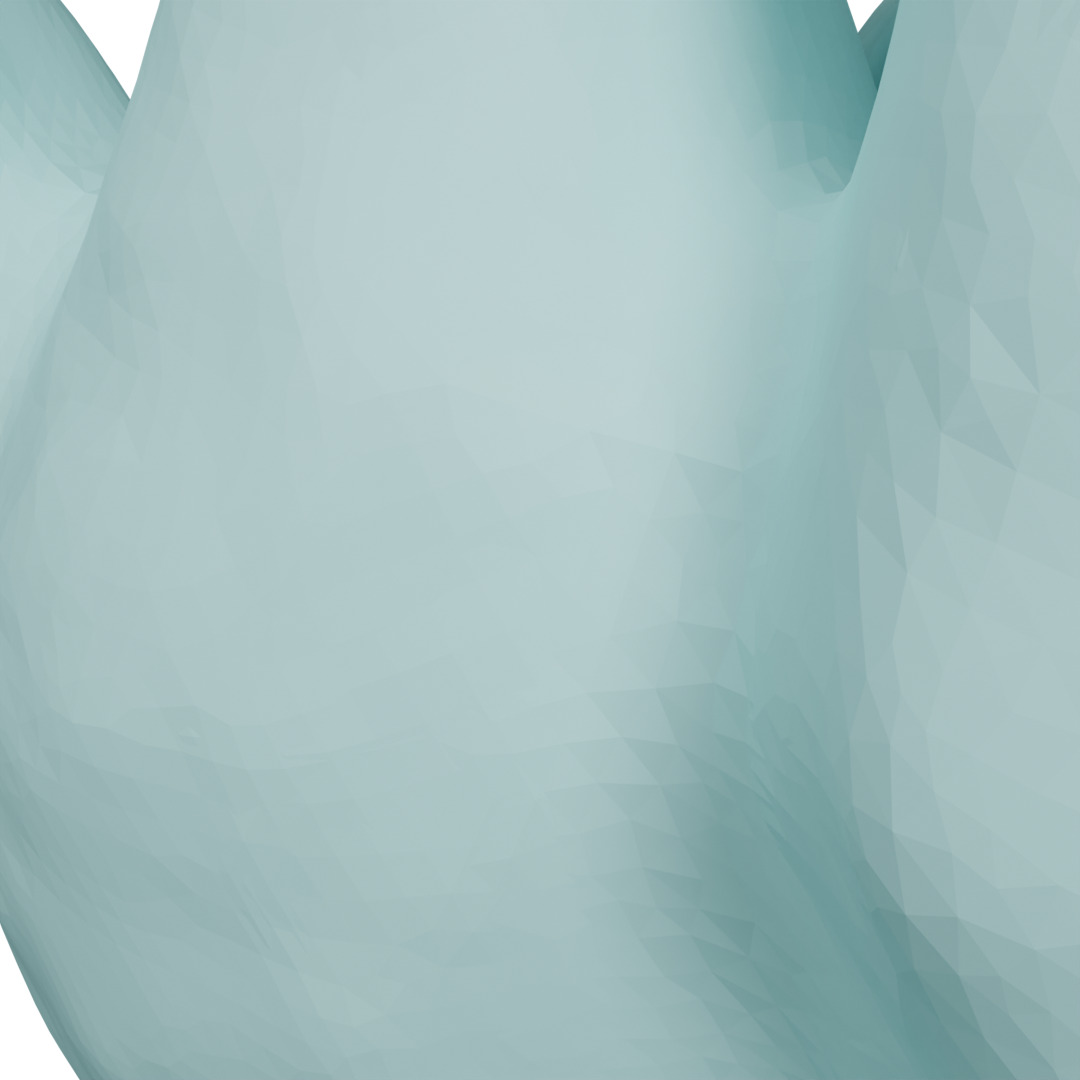 }
    \includegraphics[width=1.2in]{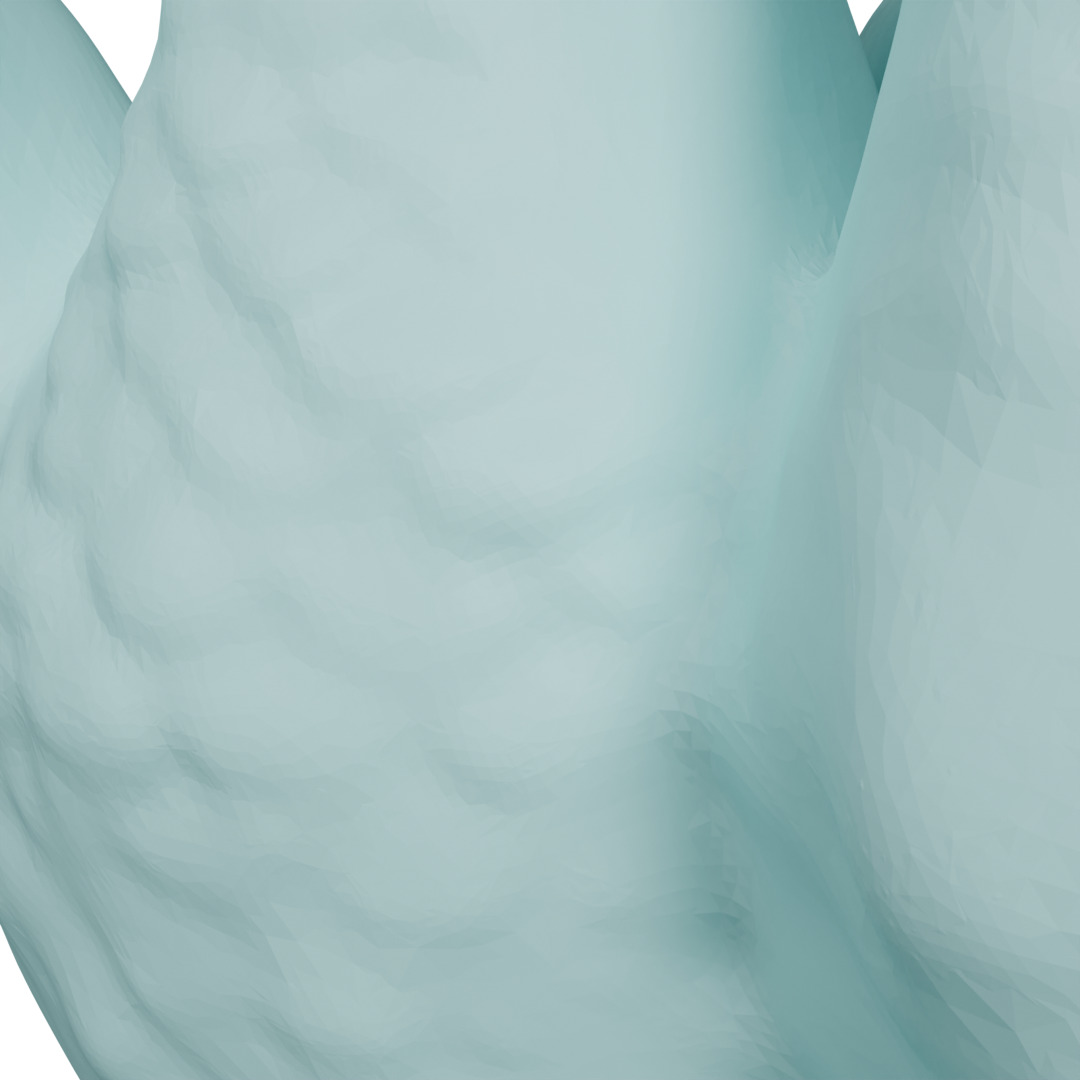 }\\
    \caption{Visual results on the DTU dataset, where the input UDFs were learned from multi-view images using NeUDF. Both DCUDF and DCUDF2 use the same iso-value $r=0.005$. }
    \label{fig:dtu}
\end{figure*}

\subsection{Ablation Studies}
\label{subsec:ablation}
In our DCUDF2 framework, several components collectively enhance the optimization's effectiveness. The accuracy-aware loss plays a critical role by dynamically steering the optimization toward regions requiring more attention, thus preventing over-smoothing. The optimization direction correction avoids potential traps in local minima, ensuring smooth progressions of the optimization process. Additionally, the incorporation of topology editing significantly improves the method's robustness, making it less sensitive to the user-specified iso-value $r$ for computing the initial double covering. Furthermore, the activation mask progressively reduces the number of vertices involved in the optimization, thereby improving the time efficiency. To quantitatively evaluate the impact of these individual components, we conducted an ablation study, the results of which are detailed in Table~\ref{tab:ablation},\ref{tab:w} and visually illustrated in Figures~\ref{fig:U-shape}, ~\ref{fig:rate} and ~\ref{fig:subdivision}.

\begin{table}[tb]
  \caption{Ablation study on the ShapeNet-Cars dataset to test the effect of our three additional components.
  }
  \label{tab:ablation}
  \centering
  \setlength\tabcolsep{2pt}
  \begin{tabular}{l|c|c|c}
    \hline
                              & CD(gt to pred)     & CD(pred to gt) &    CD(average) \\ \hline\hline
    
  Ours(full)       &   3.195      &   3.437         &   3.316   \\  
  Ours(w/o three components) & 3.280 & 3.474 & 3.377  \\
  DCUDF  & 3.359& 3.535& 3.447 \\
\hline
  \bottomrule
  \end{tabular}
\end{table}

\subsection{Runtime Efficiency}
\label{subsec:runtime}

Our enhancements in DCUDF2 have improved its runtime efficiency, achieving a threefold speed increase at a resolution of $128^3$ on the DeepFashion3D dataset.
However, despite these improvements, DCUDF2 still does not match the speed of non-optimization methods such as CAP-UDF and MeshUDF, and also lags behind DMUDF, as detailed in Table~\ref{tab:runtime}. Further advancements in reducing computational time remain a crucial area for future research.

\begin{table}[tb]
\caption{Average runtime comparison on the DeepFashion3D dataset at resolutions of $128^3$, $256^3$, and $512^3$. DMUDF encounters memory constraints and is unable to process the dataset at resolutions of $256^3$ and higher.}
  
  \label{tab:runtime}
  \centering
  \begin{tabular}{l|c|c|c|c}
    \hline
                              & MeshUDF     & DMUDF &    DCUDF &    DCUDF2   \\ \hline\hline
    
  $128^3$      &  2.7s       &   4.8s      &   90s   & 34s   \\  
  $256^3$ & 9.3s & - & 223s & 84s \\
  $512^3$ & 40s & - & 1013s & 403s \\
  \bottomrule
  \end{tabular}
\end{table} 

\begin{table}[h]
  \caption{Results  in the Shapenet-Car dataset with different $w_s$.
  }
  \label{tab:w}
  \centering
  \begin{tabular}{l|c|c|c}
    \hline
          DCUDF2           & $w_s=2$     & $w_s=3$ &    $w_s=4$   \\ \hline\hline
    
  CD (average, $10^{-3}$)       &  3.30       &   3.31      &   3.32     \\  

  \bottomrule
  \end{tabular}
\end{table}

\section{Conclusion}
\label{sec:conclusion}

This paper presents DCUDF2, a significant improvement on the original DCUDF for extracting zero level sets from unsigned distance fields. Our enhancements incorporate self-adaptive weights that adjust to local  geometry and an accuracy-aware loss function that prevents over-smoothing while preserving fine details. Moreover, we propose a topology correction strategy to reduce dependency on the hyper-parameter $r$, and introduce a suite of new operations aimed at improving runtime efficiency. Experimental results confirm that DCUDF2 exceeds existing methods in terms of geometric fidelity and topological accuracy. With its robustness and the superior results it achieves, DCUDF2 has potential to expand the use of UDFs across various application domains.

\bibliographystyle{IEEEtran}
\bibliography{egbib}

\end{document}